%% file: main.tex
\documentclass[12pt,oneside]{book}
\usepackage[utf8]{inputenc}
\usepackage[T1]{fontenc}
\usepackage{tikz}
\usetikzlibrary{arrows.meta}   % for arrow styles

\usepackage{lmodern}
\usepackage{amsmath,amssymb,amsthm,mathtools}
\usepackage{bm}        % bold math symbols
\usepackage[most]{tcolorbox}
\usepackage{graphicx}
\usepackage{float}      % controlled placement of pedagogical figures
\usepackage{hyperref}
\usepackage[normalem]{ulem}
\usepackage{enumitem}  % custom lists
\usepackage{geometry}
\usepackage{fancyhdr}
\usepackage{geometry}
\usepackage{textgreek} % allows direct use of Greek letters like β
\usepackage{tcolorbox}
\usepackage{xparse}    % for custom environments
\usepackage{listings}  % for code snippets
\usepackage{xcolor}    % syntax highlighting

\usepackage[
    backend=biber,
    style=numeric,
    sorting=nyt
]{biblatex}

\tcbset{
  mybox/.style={
    enhanced,
    boxrule=0.8pt,
    arc=2mm,
    left=2mm, right=2mm, top=1mm, bottom=1mm,
    coltitle=black,
    fonttitle=\bfseries,
    title={#1},
    attach title to upper,
  },
  conceptbox/.style={
    mybox={#1},
    colback=blue!6,
    colframe=blue!55!black,
  },
  derivbox/.style={
    mybox={#1},
    colback=black!3,
    colframe=black!45,
  },
  insightbox/.style={
    mybox={#1},
    colback=green!6,
    colframe=green!50!black,
  },
}

\newtcolorbox{ConceptBox}[1]{conceptbox={#1}}
\newtcolorbox{DerivBox}[1]{derivbox={#1}}
\newtcolorbox{InsightBox}[1]{insightbox={#1}}

\NewDocumentEnvironment{faq}{+b}{
  \section*{Frequently Asked Questions}
  \begin{description}[leftmargin=2em,labelsep=1em]
  #1
  \end{description}
}{}

\newcommand{\bookdate}{8 September 2026}

\title{The Little Book of Generative AI Foundations: An Intuitive Mathematical Primer}
\author{
Tianhua Chen\\[0.3em]
School of Computing and Engineering\\
University of Huddersfield\\[0.3em]
\texttt{t.chen@hud.ac.uk}\\[1.5em]
Preprint version\\
\bookdate
}
\date{}

\hypersetup{
  hidelinks,
  pdftitle={The Little Book of Generative AI Foundations: An Intuitive Mathematical Primer},
  pdfauthor={Tianhua Chen}
}

\begin{document}

\frontmatter
\maketitle

\include{chapters/prefer_new_short}

\tableofcontents

\mainmatter

%================================================
% Chapters are included here
%================================================
\include{chapters/cht_1_2_merged}
\include{chapters/cht3}
\include{chapters/cht4_new}
\include{chapters/cht5_new}
\include{chapters/cht_diff_fund}
\include{chapters/cht_score_diffusion}
\include{chapters/cht_8_9_merged_new}
\include{chapters/cht10_new}
\include{chapters/cht_closing_remarks}
\appendix
\include{chapters/cht_appendix1}
\include{chapters/cht_appendix2}

\backmatter
\printbibliography[heading=bibintoc,title={References}]
\end{document}

%% file: chapters/prefer_new_short.tex
\chapter*{Preface}
\addcontentsline{toc}{chapter}{Preface}

Generative artificial intelligence has developed rapidly in recent years, with
new models, systems, and applications appearing at remarkable speed. Beneath
this fast-moving surface, however, many of the central ideas rest on a smaller
set of mathematical principles: latent variables, likelihoods, variational
bounds, invertible transformations, stochastic noising processes, score fields,
adversarial comparison, and energy landscapes.

This book, \emph{The Little Book of Generative AI Foundations: An Intuitive
Mathematical Primer}, was written to provide a compact but rigorous pathway
through these ideas. It does not attempt to catalogue every modern architecture,
implementation detail, or state-of-the-art refinement. Instead, it focuses on
the mathematical foundations that make the major families of generative models
understandable. The aim is to make these foundations more accessible without
removing their mathematical substance. It is written as a foundation-building primer for
mathematically curious researchers, practitioners, and students who want to
understand generative modelling from first
principles.

The word ``little'' in the title should therefore be understood in terms of
scope rather than depth. The book is deliberately selective in what it covers,
but within that chosen scope the treatment is intentionally careful. Core ideas
are developed step by step, with the aim of making the underlying mathematical
structure visible rather than treating models as black-box recipes. The aim is
foundational rather than frontier-oriented: readers already working at the
research frontier may find many topics familiar, but the book is designed to
make the mathematical logic behind these topics explicit, intuitive, and
coherent.

The references are also kept deliberately focused. Their role is to support the
main historical and technical connections behind the exposition, rather than to
turn the book into a large survey-style bibliography.

\section*{Approach and Organisation}
%\addcontentsline{toc}{section}{Approach and Organisation}

The central purpose of the book is to connect the main families of generative
models through a coherent mathematical narrative. The ordering is chosen to make
the mathematics unfold as naturally and compactly as possible, rather than to
follow a strict historical chronology. Models are introduced in a sequence that
allows later ideas to build on earlier ones, so that the reader can see how
latent variables, variational objectives, diffusion processes, score fields,
exact density models, adversarial learning, and energy-based modelling relate
to one another.

The exposition follows a simple principle: mathematical tools are introduced
when they become necessary for understanding a modelling idea. The required
linear algebra, probability, calculus, Gaussian algebra, and density
transformation tools are therefore embedded within the modelling narrative
rather than separated into a long preliminary review.

The book contains more derivation than is strictly necessary for a high-level
overview. This is intentional: many ideas in generative modelling become clearer
when their mathematical objectives are unpacked. The derivations are developed
step by step, with necessary concepts introduced or reviewed when needed, so
that the mathematics remains connected to the modelling intuition rather than
appearing as isolated technical blocks.

The book begins with PCA and autoencoders, using them to introduce linear
transformations, projection, reconstruction, and latent structure.
Probabilistic PCA then turns these ideas into the first latent-variable
generative model in the book.

The next stage develops variational generative modelling. The Evidence Lower
Bound, the Expectation--Maximisation algorithm, and variational inference are
introduced through probabilistic latent variables, before being extended to the
Variational Autoencoder.

The book then turns to diffusion models. Denoising Diffusion Probabilistic
Models are first introduced in discrete time as sequential latent-variable
models. This is followed by the calculus needed for continuous-time generative
modelling, and then by the score-based view of diffusion, where generation is
guided by learned score fields and reverse-time stochastic dynamics.

The following chapter studies exact density models, including normalising flows
and autoregressive factorisations. These models show how likelihoods can remain
tractable by design, either through invertible transformations or through the
chain rule of probability.

The final chapter looks beyond tractable likelihoods. It introduces GANs,
Wasserstein GANs, and Energy-Based Models, showing how generative learning can
proceed through comparison, geometric discrepancy, or scalar energy landscapes
rather than explicit likelihood evaluation.

% Two appendices collect supporting derivations on Gaussian algebra and diffusion
% reverse posteriors.

\section*{Origin and Version}
%\addcontentsline{toc}{section}{Origin and Version}

This book was developed as an independently authored scholarly project,
motivated by a long-standing interest in making the mathematical foundations of
modern generative AI more transparent and accessible. It was not developed as
part of a funded research grant, commissioned institutional output, formal
module-development activity, or official course documentation for any specific
module or programme.

The manuscript grew out of a broader independent project on the probabilistic
and mathematical foundations of modern generative modelling. Earlier short-form
versions of some of these ideas appeared in preprint form, first as an SSRN
preprint and later in revised form as an arXiv
preprint~\cite{chen2025plvm,chen2025unified}. The present book expands that
line of thought into a fuller mathematical primer, with a more systematic
treatment of the foundations leading from classical latent-variable models to
modern generative AI.

The present manuscript is the second preprint version of this book. Compared
with the first version, it adds visual explanations throughout, alongside minor
clarifications to the surrounding text. It remains a working preprint and may
receive further revisions, corrections, clarifications, or additional material.
A revised version may be submitted for formal book publication.

\section*{Reuse}
%\addcontentsline{toc}{section}{Reuse}

This manuscript is made available as a preprint for scholarly reading,
citation, research reference, and study. Readers are welcome to cite the book
and to refer to it as a supporting resource for research, learning, and teaching,
with appropriate attribution.

Substantial reproduction, adaptation, redistribution, or conversion of the
material into derivative teaching, training, or course resources requires
permission from the author, unless otherwise permitted by applicable copyright
exceptions or by a separate licence attached to a future version of this work.

\section*{Acknowledgements}
%\addcontentsline{toc}{section}{Acknowledgements}

This book grew out of my continuing efforts to understand and explain the
mathematical foundations of modern generative AI from first principles. Many of
the explanations became clearer through repeated attempts to connect ideas from
linear algebra, probability, variational inference, stochastic modelling, and
modern deep generative models into a coherent narrative.

I am grateful to the wider academic community whose papers, books, lectures,
and discussions have shaped the broader context in which this manuscript was
developed. Any remaining errors, omissions, or unclear explanations are my own.

\vspace{1cm}

\begin{flushright}
Tianhua Chen\\
\textit{\bookdate}
\end{flushright}

%% file: chapters/cht_1_2_merged.tex
\chapter{Linear Algebra Foundations: From PCA to Autoencoders}
\section*{Overview}
\addcontentsline{toc}{section}{Overview}

Principal Component Analysis (PCA) is an accessible and natural entry point for
this book, for two main reasons:
\begin{itemize}
   \item \textbf{Building a bridge to probabilistic and generative models.}  
    As a widely known dimensionality reduction technique, PCA provides a natural
    stepping stone to several ideas developed later in the book. Its
    reconstruction formulation leads naturally to autoencoders, where
    compression and reconstruction are learned directly from data. Its
    probabilistic extension, Probabilistic PCA (PPCA), then reframes this
    familiar deterministic method from a generative modelling perspective. This
    progression—from algebraic projection, to reconstruction, to probabilistic
    latent-variable modelling—foreshadows the techniques used in modern
    frameworks such as Variational Autoencoders (VAEs) and diffusion models,
    where similar principles reappear in more flexible and expressive forms.

    \item \textbf{Revisiting core linear algebra concepts.}  
    PCA naturally illustrates key ideas from linear algebra—projections, basis
    transformations, eigenvalues, and eigenvectors—which underpin many
    constructions in the rest of the book. Generative models operate in
    high-dimensional spaces, where data are represented and transformed using
    matrices. Concepts such as covariance matrices, variance analysis, matrix
    decompositions, and changes of basis repeatedly appear as tools for
    understanding the structure of data.
\end{itemize}

For these reasons, PCA is an ideal starting point for the book.

\section*{Concept Map}
\addcontentsline{toc}{section}{Concept Map}

This chapter unfolds as a guided journey from basic linear transformations to PCA,
and then from PCA to autoencoders.
Each step is chosen to build intuition, connect naturally to what follows, and
prepare for the probabilistic generative models developed in later chapters.

\begin{itemize}

    \item[\(\rightarrow\)] \textbf{We begin by understanding matrices as transformations.}  
    Matrix--vector multiplication is introduced not as a mechanical rule, but as a way of understanding how vectors and coordinate systems are reshaped.  
    This geometric viewpoint provides the intuition needed for PCA and, more broadly, for the transformation-based perspective that recurs throughout later generative models.

    \item[\(\rightarrow\)] \textbf{We then develop two complementary views of matrix multiplication and connect them to PCA.}  
    The \emph{column-combination} view shows how a matrix defines new directions from which outputs are constructed, while the \emph{row--dot--product} view interprets multiplication in terms of projection and alignment.  
    Together, these ideas foreshadow the two central roles of PCA: changing basis and measuring variation through projection.

    \item[\(\rightarrow\)] \textbf{Next, PCA is formulated as the search for directions of maximal variance, and eigenvalues and eigenvectors enter as the key tools for solving this problem.}  
    By projecting data onto candidate directions and measuring their spread, PCA becomes an optimisation problem.  
    Because the covariance matrix is symmetric, its eigenvectors form an orthonormal basis, and rewriting the objective in this basis shows that the direction of greatest variance is the eigenvector associated with the largest eigenvalue.

    \item[\(\rightarrow\)] \textbf{We then extend this argument to full dimensionality reduction and reconstruction.}  
    The leading eigenvectors define an ordered set of orthogonal principal directions, each capturing as much remaining variance as possible.  
    This leads naturally to projection, reconstruction, and the explained variance ratio, giving both a geometric and practical understanding of how PCA compresses data.

    \item[\(\rightarrow\)] \textbf{The chapter then reveals the deeper structure of PCA as a change of basis.}  
    Diagonalising the covariance matrix is interpreted as a similarity transformation that preserves total variance while reorganising it into decorrelated coordinates aligned with the principal directions.  
    This viewpoint also provides a natural entry point to broader matrix decompositions such as SVD.

    \item[\(\rightarrow\)] \textbf{Finally, the emphasis shifts from variance to reconstruction, creating a bridge from PCA to autoencoders.}  
    PCA is reinterpreted as an encoder--decoder model that minimises reconstruction error, after which linear autoencoders recover the same optimal subspace and nonlinear autoencoders extend this idea from linear subspaces to curved manifolds.  
    This progression reveals both the power and the limitation of deterministic reconstruction models, motivating the probabilistic latent-variable models developed in the next chapter.

\end{itemize}

The chapter trajectory is summarised visually below, from basic linear transformations through PCA and reconstruction to the autoencoder bridge used later in the book.

\input{figures/ch01/chapter1_foundations_at_a_glance.tex}

\medskip
\noindent
\section{Matrices as Linear Transformations: Basis Directions, Projections, and Composition}

Matrices are best understood not merely as tables of numbers, but as representations of transformations acting on vectors.
A matrix specifies a \emph{linear transformation}: a rule that moves vectors in a way that preserves addition and scaling.
Understanding how matrices act on vectors is a foundational skill, not only for linear methods such as PCA, but also for more complex generative models that are built from sequences of linear and nonlinear transformations.

\subsection{Matrices as Transformations of Basis Directions}

\paragraph{The standard basis and the identity transformation.}

We begin with the simplest possible setting.
In two dimensions, the standard (or identity) basis vectors are
\[
\hat{e}_1 =
\begin{bmatrix}1 \\ 0\end{bmatrix},
\qquad
\hat{e}_2 =
\begin{bmatrix}0 \\ 1\end{bmatrix}.
\]
Every vector in $\mathbb{R}^2$ can be written uniquely as a linear combination of these two directions.

Placing these basis vectors as the columns of a matrix produces the identity matrix
\[
I =
\begin{bmatrix}
1 & 0 \\
0 & 1
\end{bmatrix},
\]
which represents the transformation that leaves all vectors unchanged.
Geometrically, the identity transformation leaves every vector unchanged, and therefore preserves directions, lengths, and angles.

\paragraph{Example: a $90^\circ$ rotation.}

To see how a matrix can actively transform space, consider
\[
R =
\begin{bmatrix}
0 & -1 \\
1 & 0
\end{bmatrix}.
\]
The columns of this matrix show how the basis vectors are transformed:
\[
\hat{e}_1 \mapsto
\begin{bmatrix}0 \\ 1\end{bmatrix},
\qquad
\hat{e}_2 \mapsto
\begin{bmatrix}-1 \\ 0\end{bmatrix}.
\]
Each basis direction is rotated by $90^\circ$ counterclockwise.
Since every vector is built from these basis vectors, the same rotation is applied uniformly to the entire plane.

This example illustrates a central principle of linear algebra:
\begin{quote}
\emph{A linear transformation is completely determined by where it sends the basis vectors.}
\end{quote}

\paragraph{Beyond rotations: general linear transformations.}

A matrix does not need to describe a rotation.
Its columns do not have to be orthogonal, nor do they need to have unit length.

For example, consider
\[
A =
\begin{bmatrix}
1 & 1 \\
0 & 1
\end{bmatrix}.
\]
Here, the first basis vector remains unchanged, while the second basis vector is sent from
$\begin{bmatrix}0 \\ 1\end{bmatrix}$ to $\begin{bmatrix}1 \\ 1\end{bmatrix}$.
This transformation corresponds to a shear: one axis is tilted while the other remains fixed.

Despite their differences, both rotation and shear transformations are described in exactly the same way:
by specifying how the basis vectors are mapped.
Once this is known, the effect of the transformation on every other vector is fully determined.

Figure~\ref{fig:ch1-matrix-transformations} makes this basis-vector viewpoint geometric: the transformed basis vectors determine how the entire coordinate grid is carried by the matrix.

\input{figures/ch01/fig_ch01_01_matrix_transformations.tex}

We now turn to how these transformations act on arbitrary vectors through matrix multiplication.

\subsection{Two Views of Matrix Multiplication}

\paragraph{Matrix--vector multiplication: the column-combination view.}

Let $A$ be a matrix with columns $v_1$ and $v_2$.
For a vector $x = \begin{bmatrix}a \\ b\end{bmatrix}$, matrix--vector multiplication gives
\[
Ax = a\,v_1 + b\,v_2.
\]

This expression reveals the first fundamental interpretation of matrix multiplication.
The columns of the matrix define a new set of directions, and the entries of the vector specify how to combine them.
Multiplying a matrix by a vector therefore constructs the output as a weighted combination of the matrix’s column directions.

This interpretation connects directly to how vectors themselves are constructed.
Any vector can be written as
\[
x = a\,\hat{e}_1 + b\,\hat{e}_2,
\]
a combination of the standard basis vectors.
Matrix--vector multiplication simply replaces these standard basis directions with the transformed coordinate directions encoded by the columns of $A$.

In a general case, the same idea applies:
\[
Ax = \sum_{j=1}^n x_j\,a_j,
\]
where $a_j$ denotes the $j$-th column of $A$.

\paragraph{Matrix--vector multiplication: the row-by-column view.}

There is a second, equally important way to understand matrix--vector multiplication.
Rather than focusing on columns, we can examine individual entries of the output vector.

In this view, the $i$-th entry of $Ax$ is obtained by taking the dot product between the $i$-th row of $A$ and the input vector.
This row-by-column computation is not an arbitrary numerical rule—it has a clear geometric meaning.

\begin{tcolorbox}[
  colback=blue!6,
  colframe=blue!55!black,
  title={Why the Dot Product Encodes Projection},
  boxrule=0.8pt,
  arc=2mm,
  left=2mm,
  right=2mm,
  top=1mm,
  bottom=1mm
]

\textbf{Cosine similarity.}
The cosine similarity between two vectors $v_1$ and $v_2$ is
\[
\cos\theta = \frac{v_1^\top v_2}{\|v_1\|\,\|v_2\|},
\]
where $\theta$ is the angle between them.

\textbf{Scalar projection.}
Rearranging this expression leads to the scalar projection of $v_1$ onto the direction of $v_2$:
\[
\text{proj}_{v_2}(v_1)
=
\frac{v_1^\top v_2}{\|v_2\|}.
\]
This quantity measures how much of $v_1$ lies along $v_2$.

\textbf{Dot product as a scaled projection.}
The dot product can therefore be written as
\[
v_1^\top v_2
=
\|v_2\| \times \text{(projection of $v_1$ onto $v_2$)}.
\]
If $v_2$ has unit length, the dot product equals the scalar projection.

\textbf{Vector projection.}
Multiplying the scalar projection by the unit direction of $v_2$ yields the corresponding \emph{vector projection}:
\[
\mathrm{proj}_{v_2}(v_1)
=
\frac{v_1^\top v_2}{\|v_2\|^2}\, v_2.
\]
giving the vector representation of the component of $v_1$ lying along the direction of $v_2$.

\end{tcolorbox}
The discussion in the box shows that the dot product appearing in the row-by-column computation has a geometric interpretation as a measure of projection or alignment. From this perspective, each entry of $Ax$ records how strongly the input vector aligns with a particular row direction of 
$A$.

\paragraph{Extending to matrix--matrix multiplication.}

The same ideas extend naturally to matrix--matrix multiplication.
Multiplying two matrices corresponds to composing two linear transformations.

The product $AB$ can be understood column by column:
\[
AB =
\big[\,A b_1 \;\; A b_2 \;\; \cdots \;\; A b_p\,\big],
\]
where each column of $B$ is transformed individually by $A$. Equivalently, when applied to a vector \(v\), the product \(AB\) represents two successive transformations:
\[
(AB)v = A(Bv).
\]
The vector is first transformed by \(B\), and the result is then transformed by \(A\). Each stage may in turn be understood through the same column-combination and row-by-column perspectives introduced above.

\medskip
\noindent
Together, these perspectives form the linear-algebraic foundation that directly informs the formulation of PCA and will reappear repeatedly as we build towards more expressive generative models:
\begin{itemize}
\item columns describe how a transformation reshapes space,
\item column combinations explain how transformed vectors are constructed,
\item row-wise dot products reveal how projections are extracted,
\item and matrix--matrix multiplication chains transformations together.
\end{itemize}

\section{Eigenvectors, Eigenvalues, and a Matrix’s Natural Coordinate System}
Before delving into PCA, we introduce another fundamental tool for understanding how matrices act on space.
In the previous section, we interpreted matrices as linear transformations that reshape vectors by mixing, stretching, and projecting along different directions.
To reason more precisely about this action, we now seek directions along which the transformation becomes as simple as possible.

Eigenvectors and eigenvalues provide exactly this language.
They identify special directions of a matrix—directions that are not mixed with others under the transformation, but are instead mapped to scalar multiples of themselves.

\paragraph{Eigenvalues and eigenvectors.}
Given a square matrix $A \in \mathbb{R}^{d \times d}$, a nonzero vector $v \in \mathbb{R}^d$ is called an eigenvector of $A$ if
\[
A v = \lambda v
\]
for some scalar $\lambda \in \mathbb{R}$, called the corresponding eigenvalue.
This equation tells us that when $A$ acts on $v$, the output remains along the same line spanned by $v$; its magnitude is scaled by $|\lambda|$, and if $\lambda<0$ the orientation is reversed.

For a general vector, multiplication by $A$ may rotate, stretch, shear, or otherwise distort the direction.
Eigenvectors therefore identify directions that are not mixed with others under the transformation: the output remains along the same line, and differs only by a scalar factor $\lambda$.
Eigenvectors can therefore be viewed as directions of \emph{pure scaling} under the transformation defined by $A$.

\paragraph{Eigen-decomposition: expressing a matrix in its natural coordinate system.}
If a matrix $A \in \mathbb{R}^{d \times d}$ has $d$ linearly independent eigenvectors, we may assemble them as the columns of a matrix $V$, and place the corresponding eigenvalues along the diagonal of a matrix $\Lambda$.

Concretely, if $v_1, v_2, \dots, v_d$ are eigenvectors of $A$ with corresponding eigenvalues $\lambda_1, \lambda_2, \dots, \lambda_d$, then
\[
A [\, v_1 \; v_2 \; \cdots \; v_d \,]
=
[\, A v_1 \; A v_2 \; \cdots \; A v_d \,]
=
[\, \lambda_1 v_1 \; \lambda_2 v_2 \; \cdots \; \lambda_d v_d \,].
\]

In compact matrix form, this relationship can be written as
\[
A V = V \Lambda,
\]
where
\[
V = [\, v_1 \; v_2 \; \cdots \; v_d \,],
\qquad
\Lambda =
\begin{bmatrix}
\lambda_1 & 0 & \cdots & 0 \\
0 & \lambda_2 & \cdots & 0 \\
\vdots & \vdots & \ddots & \vdots \\
0 & 0 & \cdots & \lambda_d
\end{bmatrix}.
\]

Rearranging this expression gives the eigen-decomposition
\[
A = V \Lambda V^{-1}.
\]

\begin{tcolorbox}[
  colback=blue!6,
  colframe=blue!55!black,
  title={Intuition: Eigen-decomposition as a Change of Basis},
  boxrule=0.8pt,
  arc=2mm,
  left=2mm,
  right=2mm,
  top=1mm,
  bottom=1mm
]

A central idea in linear algebra is that the same vector can be described using
different coordinate systems.
Eigen-decomposition makes this explicit by introducing a \emph{new basis}
in which the action of a matrix becomes especially simple.

The matrix $V$ collects the eigenvectors of $A$ as its columns.
These eigenvectors define a new set of coordinate axes—an eigenvector basis for the space.
Writing a vector in this basis means describing it by its coordinates along these
preferred directions.

\medskip
\textbf{Changing coordinates.}
Given a vector $x$ expressed in the standard basis, we change coordinates by applying
\[
x_{\text{eig}} = V^{-1} x.
\]
This operation does not change the vector itself; it only changes how the vector is
\emph{represented}, converting it into eigenvector coordinates.

\medskip
\textbf{Simple action in the eigenbasis.}
Once expressed in the eigenvector basis, the action of $A$ becomes particularly simple:
each coordinate is scaled independently, with the scaling factor given by the
corresponding eigenvalue, as captured by diagonal matrix $\Lambda$:
\[
x_{\text{eig}} \;\mapsto\; \Lambda x_{\text{eig}}.
\]

\medskip
\textbf{Returning to the original basis.}
After this scaling, we convert the result back to the original coordinate system via
\[
x' = V(\Lambda x_{\text{eig}}),
\]
which reconstructs the vector in the standard basis.

\medskip
Putting everything together, the decomposition $A = V \Lambda V^{-1}$ can be read as
\[
x
\;\xrightarrow{\;V^{-1}\;}
\text{change of basis}
\;\xrightarrow{\;\Lambda\;}
\text{independent scaling}
\;\xrightarrow{\;V\;}
\text{return to original basis}.
\]

Eigen-decomposition therefore shows that a matrix acts as simple scaling
\emph{once the right coordinate system is chosen}.
\end{tcolorbox}

\paragraph{The special case of symmetric matrices.}
For a general square matrix, an eigen-decomposition may fail to exist over the reals, or may involve complex eigenvalues or an insufficient set of linearly independent eigenvectors. However, when $A$ is a real symmetric matrix ($A = A^\top$), the situation becomes much simpler.

In this case, several important properties hold:
\begin{itemize}
    \item All eigenvalues of $A$ are real.
    \item There exists an \emph{orthogonal} matrix $U$, whose columns are eigenvectors of $A$, such that
    \[
    A = U \Lambda U^\top,
    \]
    where the columns of $U$ have unit length and are mutually orthogonal, with $U^\top U = I$.
\end{itemize}

This special structure means that symmetric matrices can always be diagonalised using an orthonormal change of basis, so the change of coordinates itself preserves lengths and angles.
As we will see shortly, this property is crucial for PCA, since the empirical covariance matrix is always symmetric.

With these tools in place, we are now ready to formulate and solve the PCA problem.

\section{Principal Component Analysis via Variance Maximisation}

Principal Component Analysis (PCA) \cite{pearson1901,hotelling1933} provides a systematic way to reduce the
dimensionality of data by introducing a new coordinate system adapted to the
structure of the data.
The core idea is to identify new coordinate axes along which the data vary the
most.
By aligning these axes with directions of maximal spread, PCA reveals where the
data are most informative and enables lower-dimensional representations with
minimal loss of structure.

The intuition behind PCA can be summarised by two key principles:
\begin{enumerate}
    \item \textbf{Variance as informativeness.}  
    Directions along which the data exhibit high variance tend to separate data
    points more clearly and therefore carry more useful structural information.
    Directions with little variance contribute less to distinguishing the data.

    \item \textbf{Sequential orthogonal maximisation.}  
    PCA first identifies the single direction that captures the largest possible
    variance.
    Subsequent directions are chosen to capture as much remaining variance as
    possible while remaining orthogonal to those already selected.
\end{enumerate}

\subsection{Formulating PCA}

\paragraph{Centring the data and defining projected variance.}
To formulate PCA conveniently, we first centre the data so that the data cloud
is positioned at the origin.
Given data points $\{x^{(i)}\}_{i=1}^n \subset \mathbb{R}^d$, we assume
\[
\frac{1}{n}\sum_{i=1}^n x^{(i)} = 0.
\]
This centring step ensures that variability is measured relative to the centre
of the data, rather than being influenced by an arbitrary offset.

As discussed earlier when introducing projections and dot products, variability
along a direction can be understood by examining how strongly the data project
onto that direction.
Let $v \in \mathbb{R}^d$ be a unit vector representing a candidate direction onto
which the data will be projected, so that $\|v\| = 1$.
For a data point $x^{(i)}$, the scalar projection onto $v$ is
\[
v^\top x^{(i)},
\]
since $v$ has unit length.
The corresponding vector projection is
\[
(v^\top x^{(i)})\,v.
\]
This again shows that, when $v$ is a unit vector, the dot product gives the
scalar projection onto that direction.

Because the data are centred, the mean of the projected scalar values is also
zero:
\[
\frac{1}{n}\sum_{i=1}^n v^\top x^{(i)}
=
v^\top \left(\frac{1}{n}\sum_{i=1}^n x^{(i)}\right)
=
0.
\]
To quantify how widely these projected values are spread, we measure their
variance about this mean.
The variance of the data projected along direction $v$ is therefore
\[
\mathrm{Var}(v^\top X)
=
\frac{1}{n}\sum_{i=1}^n \bigl(v^\top x^{(i)}\bigr)^2.
\]
Geometrically, directions with larger and more widely dispersed projections
correspond to higher variance.

\paragraph{Covariance matrix form.}
The projected variance can be written compactly in matrix notation:
\[
\mathrm{Var}(v^\top X)
=
\frac{1}{n}\sum_{i=1}^n v^\top x^{(i)} (x^{(i)})^\top v
=
v^\top \left( \frac{1}{n}\sum_{i=1}^n x^{(i)} (x^{(i)})^\top \right) v.
\]
The matrix inside the parentheses is the empirical covariance matrix,
\[
\Sigma = \frac{1}{n} \sum_{i=1}^n x^{(i)} (x^{(i)})^\top.
\]
Thus, the variance of the data projected along direction $v$ is
\[
\mathrm{Var}(v^\top X) = v^\top \Sigma\, v.
\]

The PCA problem is therefore to find the direction that maximises this quantity
under a unit-length constraint:
\[
\max_{v \in \mathbb{R}^d} \; v^\top \Sigma v
\quad \text{subject to} \quad v^\top v = 1.
\]

\subsection{Solving the PCA optimisation problem}

The covariance matrix $\Sigma$ is symmetric by construction.
As discussed in the previous section, symmetric matrices admit an orthogonal
eigen-decomposition:
\[
\Sigma = U \Lambda U^\top,
\qquad
\Lambda = \mathrm{diag}(\lambda_1,\ldots,\lambda_d),
\]
where the columns of $U$ are orthonormal eigenvectors and the eigenvalues are
ordered as
\[
\lambda_1 \ge \lambda_2 \ge \cdots \ge \lambda_d \ge 0.
\]

To solve the optimisation problem, we express the objective in this eigenvector
coordinate system.

\paragraph{Step 1: Change of basis to the eigenvector coordinates.}
Following the basis-change interpretation introduced earlier, we express the
direction vector $v$ in the eigenvector basis as
\[
\tilde{v} = U^\top v.
\]
Since $U$ is orthogonal, $U^\top = U^{-1}$, so this operation simply changes the
coordinates of $v$ from the standard basis to the eigenvector basis.

Orthogonal transformations also preserve vector length:
\[
\tilde{v}^\top \tilde{v}
=
(U^\top v)^\top (U^\top v)
=
v^\top U U^\top v
=
v^\top v.
\]
Consequently, the unit-length constraint $v^\top v = 1$ becomes
\[
\tilde{v}^\top \tilde{v} = 1.
\]

\paragraph{Step 2: Diagonalising the objective.}
Substituting the eigen-decomposition into the objective yields
\[
v^\top \Sigma v
=
v^\top U \Lambda U^\top v
=
(U^\top v)^\top \Lambda (U^\top v)
=
\tilde{v}^\top \Lambda \tilde{v}.
\]
Because $\Lambda$ is diagonal, this expression simplifies to
\[
\tilde{v}^\top \Lambda \tilde{v}
=
\sum_{i=1}^d \lambda_i \tilde{v}_i^2.
\]
This shows explicitly that the variance captured along direction $v$ is a
weighted sum of the eigenvalues, with weights determined by how strongly $v$
aligns with each eigenvector.

\paragraph{Step 3: Bounding the variance.}
Since $\tilde{v}_i^2 \ge 0$ and $\sum_{i=1}^d \tilde{v}_i^2 = 1$, the objective
is a weighted average of the eigenvalues:
\[
\sum_{i=1}^d \lambda_i \tilde{v}_i^2
\le
\lambda_1 \sum_{i=1}^d \tilde{v}_i^2
=
\lambda_1.
\]
No combination of smaller eigenvalues can exceed the largest one.

\paragraph{Step 4: Optimal direction.}
The upper bound is achieved only when all the weight is placed on the largest
eigenvalue:
\[
\tilde{v} = [1,\,0,\,\ldots,\,0]^\top.
\]
Transforming back to the original coordinates gives
\[
v = U \tilde{v} = u_1,
\]
the eigenvector associated with the largest eigenvalue $\lambda_1$.

Thus, the direction that maximises the projected variance is precisely the
leading eigenvector of the covariance matrix, and the maximum variance equals
$\lambda_1$.

\paragraph{Subsequent principal components.}
The first principal component is therefore $u_1$.
Additional components are obtained by repeating the optimisation while enforcing
orthogonality to previously selected directions.
The $k$-th principal component is the eigenvector $u_k$ corresponding to the
$k$-th largest eigenvalue, satisfying
\[
u_k^\top u_j = 0 \qquad \text{for all } j < k.
\]

Geometrically, each new principal direction captures as much remaining
variability as possible while remaining orthogonal to the earlier directions.
Because these directions are orthogonal eigenvectors of the covariance matrix,
the resulting principal component coordinates are uncorrelated.
The eigen-decomposition of the covariance matrix naturally provides exactly such
an ordered, orthonormal set of directions.

\subsection{PCA for dimensionality reduction}

Once the principal components have been identified, PCA provides a concrete
procedure for dimensionality reduction by projecting data onto a
low-dimensional subspace and, if desired, reconstructing it back in the original
space.

Let
\[
U_k = [\,u_1\;u_2\;\cdots\;u_k\,] \in \mathbb{R}^{d \times k}
\]
denote the matrix formed by the top $k$ principal components, that is, the
eigenvectors of the covariance matrix $\Sigma$ corresponding to the largest
eigenvalues $\lambda_1 \ge \lambda_2 \ge \cdots \ge \lambda_k$.
PCA reduces the dimensionality of a data point $x \in \mathbb{R}^d$ by
representing it using only these $k$ directions that capture the largest
variance.

\paragraph{Projection (encoding).}
The low-dimensional representation of $x$ is obtained by projecting it onto the
principal-component subspace:
\[
z = U_k^\top x \in \mathbb{R}^k.
\]
The vector $z$ contains the coordinates of $x$ along the principal directions
and constitutes the compressed representation of the data.

\paragraph{Reconstruction (decoding).}
An approximation of $x$ using only these $k$ components is obtained by mapping
the low-dimensional representation back to the original space:
\[
\hat{x} = U_k z = U_k U_k^\top x.
\]
This operation takes the coordinates of $x$ in the principal-component space and
reconstructs a vector in the original feature space by combining the principal
directions in $U_k$.
Because only the top $k$ components are retained, some information is
inevitably lost during this reconstruction.

It is worth noting that when $k = d$, we have $U_k \in \mathbb{R}^{d \times d}$
and
\[
U_k U_k^\top = I,
\]
so the reconstruction is exact.
When $k < d$, however, $U_k U_k^\top$ is no longer the identity matrix, and
$\hat{x}$ is only an approximation to the original data point.

\paragraph{Explained variance ratio.}
In practice, it is important to quantify how much of the total variability is
preserved after dimensionality reduction.
The most commonly used measure for this purpose is the \emph{explained variance
ratio}.

The total variance present in the original dataset is
\[
\sum_{i=1}^d \lambda_i,
\]
while the variance captured by the top $k$ principal components is
\[
\sum_{i=1}^k \lambda_i.
\]
The proportion of variance preserved by the $k$-dimensional representation is
therefore
\[
\text{Explained Variance Ratio}
=
\frac{\sum_{i=1}^k \lambda_i}{\sum_{i=1}^d \lambda_i},
\]
which provides a principled, data-driven criterion for selecting the number of
components and is widely used in practice.

We now move on to examine the structure of PCA more carefully.
This will clarify why the explained variance ratio is a principled way to quantify how much information is preserved by dimensionality reduction, and will reveal the deeper nature of PCA that we explore in the next section.

\section{PCA as a Change of Basis: Similarity Transformations and SVD}
%\section{The Nature of PCA: Similarity Transformations and Change of Basis}
We now revisit PCA from a deeper structural perspective.
In the previous section, PCA emerged as the solution to a variance-maximisation problem.
Here we show that the same procedure can also be understood as a \emph{change of basis} that diagonalises the covariance matrix and reorganises variance along orthogonal directions.
This viewpoint explains why the total variance is preserved, why the eigenvalues naturally quantify variance captured along the principal directions, and why PCA produces uncorrelated coordinates.

This perspective also leads naturally to the Singular Value Decomposition (SVD), which extends the same geometric ideas beyond covariance matrices and beyond square symmetric matrices.
Although SVD is not required for the basic formulation of PCA, it provides a broader linear-algebraic viewpoint that will remain useful throughout this book.

To develop this interpretation, we begin with two key concepts: the trace of a matrix and similarity transformations.

\subsection{Trace, similarity transformations, and variance preservation}

\paragraph{Trace of the covariance matrix.}
For any square matrix $A \in \mathbb{R}^{d \times d}$, the trace is defined as the sum of its diagonal entries:
\[
\mathrm{Tr}(A) = \sum_{i=1}^d a_{ii}.
\]
When $A$ is a covariance matrix $\Sigma$, each diagonal entry $\sigma_{ii}$ is the variance of the $i$-th coordinate.
Thus,
\[
\mathrm{Tr}(\Sigma)
=
\sum_{i=1}^d \Sigma_{ii}
=
\sum_{i=1}^d \mathrm{Var}(X_i),
\]
which is precisely the total variance of the dataset across all coordinates.

This observation is important for PCA.
Although a change of basis may alter how variance is distributed across individual coordinates, the total variance remains unchanged.
The trace provides a compact way to express this total variance.

\paragraph{Similarity transformations.}
Earlier, we saw that expressing vectors in a new basis defined by an invertible
matrix $A$ corresponds to a change of coordinates,
\[
v_A = A^{-1} v.
\]
The same idea applies to linear transformations themselves.
When a linear transformation $T$ is expressed in this new basis, its matrix
representation becomes
\[
T_A = A^{-1} T A,
\]
which is known as a \emph{similarity transformation}. This expression represents the \emph{same} linear transformation, now written in
a different coordinate system determined by the basis $A$, as shown in the
derivation below.
%Although the matrix representation changes, the underlying geometric action of $T$ does not.

\begin{tcolorbox}[
  colback=blue!6,
  colframe=blue!55!black,
  title={Why $T_A = A^{-1} T A$},
  boxrule=0.8pt,
  arc=2mm,
  left=2mm,
  right=2mm,
  top=1mm,
  bottom=1mm
]
If $v_A = A^{-1}v$ denotes the coordinates of a vector $v$ in the new basis defined
by $A$, then the linear transformation $T$ must admit a corresponding
representation $T_A$ in this coordinate system.  
Both matrices describe the \emph{same} geometric transformation, so applying $T_A$
to the coordinate vector $v_A$ must produce the same result as applying $T$ in the
original coordinates and then changing basis:
\[
v'_A = T_A v_A,
\qquad
v'_A = A^{-1}(Tv).
\]
Since these two procedures must yield the same vector in the $A$-coordinates,
substituting $v_A = A^{-1}v$ gives
\[
T_A A^{-1} v = A^{-1} T v
\quad\text{for all } v.
\]
Multiplying both sides on the left by $A$ yields
\[
A T_A A^{-1} = T,
\qquad\Longrightarrow\qquad
T_A = A^{-1} T A.
\]
\end{tcolorbox}

\paragraph{Trace invariance under change of basis.}

A key algebraic property of the trace is its cyclic identity,
\[
\mathrm{Tr}(BC) = \mathrm{Tr}(CB),
\]
which holds whenever the matrix products $BC$ and $CB$ are well defined.

Combining this property with the similarity transformation
$T_A = A^{-1} T A$ yields
\[
\mathrm{Tr}(T_A)
=
\mathrm{Tr}(A^{-1} T A)
=
\mathrm{Tr}(T).
\]

This shows that the trace is invariant under similarity transformations.
As a result, although a change of basis may redistribute how variance is expressed
across individual coordinates, it preserves the total amount of variance.

\subsection{PCA as a similarity transformation}

We now revisit PCA from the perspective of a change of basis.
For an individual data point $x \in \mathbb{R}^d$, the low-dimensional
representation is obtained by projection onto the principal component directions, $z = U_k^\top x.$ At the level of the entire dataset, with centred data points collected row-wise
in a matrix $X$, applying PCA corresponds to the linear transformation $Z = XU,$
where $U$ is the orthogonal matrix whose columns are the eigenvectors of the
covariance matrix $\Sigma$.

\medskip
\noindent
We now examine how this change of basis affects the \emph{covariance structure}
of the data.

\paragraph{Covariance of the transformed data.}
The covariance matrix of the transformed data is
\[
\tilde{\Sigma} = \mathrm{Cov}(Z) = \mathrm{Cov}(XU).
\]

In general, the covariance of a random vector $Y$ can be written as
\[
\mathrm{Cov}(Y)
= \mathbb{E}\big[(Y - \mathbb{E}[Y])^\top (Y - \mathbb{E}[Y])\big].
\]
In our setting, the data have been centred so that $\mathbb{E}[X] = 0$.
Because $U$ is a fixed linear transformation, this implies
\[
\mathbb{E}[Z] = \mathbb{E}[XU] = \mathbb{E}[X]U = 0.
\]
As a result, the covariance of the transformed data simplifies to
\[
\tilde{\Sigma}
= \mathbb{E}\big[(Z - \mathbb{E}[Z])^\top (Z - \mathbb{E}[Z])\big]
= \mathbb{E}[Z^\top Z].
\]

Substituting $Z = XU$ gives
\[
\tilde{\Sigma}
= \mathbb{E}[(XU)^\top (XU)]
= U^\top \mathbb{E}[X^\top X] U
= U^\top \Sigma U.
\]

\medskip
\noindent
\textbf{Before and after PCA.}
We can now state the central structural result explicitly:
\[
\mathrm{Cov}(X) = \Sigma,
\qquad
\mathrm{Cov}(Z) = U^\top \Sigma U.
\]
That is, PCA transforms the covariance matrix by
\[
\Sigma \;\longmapsto\; U^\top \Sigma U,
\]
which is exactly a \emph{similarity transformation} of the original covariance
matrix.

\paragraph{Consequences of the similarity transformation.}
Because similarity transformations preserve the trace, we have
\[
\mathrm{Tr}(\Sigma)
=
\mathrm{Tr}(U^\top \Sigma U)
=
\mathrm{Tr}(\Lambda),
\]
where $\Sigma = U \Lambda U^\top$ is the eigen-decomposition of the covariance
matrix.

This captures the geometric meaning of PCA.
Before PCA, the covariance matrix $\Sigma$ may contain off-diagonal terms,
reflecting correlations between coordinates.
After changing to the eigenvector basis, the covariance becomes diagonal:
\[
U^\top \Sigma U = \Lambda = \mathrm{diag}(\lambda_1,\ldots,\lambda_d).
\]
The transformed coordinates are therefore uncorrelated, and the variance along
each principal direction appears directly on the diagonal as the corresponding
eigenvalue.

PCA does not create or destroy variance.
Rather, it re-expresses the same total variance in a coordinate system aligned
with the natural directions of the data.
The total variance, given by the trace, is preserved, while its distribution is
reorganised from correlated coordinates into orthogonal principal directions.

This also explains why the explained variance ratio has such a natural form.
Retaining the first $k$ principal components preserves variance $\sum_{i=1}^k \lambda_i,$ out of the total $\sum_{i=1}^d \lambda_i.$ Hence
$\frac{\sum_{i=1}^k \lambda_i}{\sum_{i=1}^d \lambda_i}$ 
measures the proportion of total variance retained after discarding the remaining
directions.

Figure~\ref{fig:ch1-pca-change-of-basis} shows this coordinate change directly: the same data cloud becomes axis-aligned in principal-component coordinates while its total variance is preserved.

\input{figures/ch01/fig_ch01_03_pca_change_basis.tex}

\subsection{From eigen-decomposition to singular value decomposition}

The discussion above shows that PCA may be understood as a change of basis that diagonalises a covariance matrix and reorganises variance along orthogonal directions.
This naturally leads to a broader question:
\begin{quote}
What if the matrix we want to understand is not a covariance matrix, not symmetric, or not even square?
\end{quote}
The Singular Value Decomposition (SVD) provides the corresponding general framework \cite{strang2016introduction}.

Eigen-decomposition is based on the relation $A v = \lambda v,$ which identifies directions that are mapped onto themselves up to scaling.
Along such directions, the transformation is especially simple: the output remains on the same line, scaled by $\lambda$.

This viewpoint is powerful, but limited.
It applies only to square matrices, and additional conditions are generally needed to guarantee a full and well-behaved real decomposition.

SVD generalises the idea.
Instead of asking for directions that are preserved by the matrix, it asks a more flexible question:
\emph{which orthogonal input directions are mapped to which orthogonal output directions, and by how much are they stretched?}

\paragraph{The SVD viewpoint.}
For a real matrix $A \in \mathbb{R}^{m \times n}$, the singular value decomposition takes the form
\[
A = U S V^\top,
\]
where $V \in \mathbb{R}^{n \times n}$ contains orthonormal input directions; $U \in \mathbb{R}^{m \times m}$ contains orthonormal output directions; and $S \in \mathbb{R}^{m \times n}$ is diagonal in the rectangular sense, with nonnegative entries $    \sigma_1 \ge \sigma_2 \ge \cdots \ge 0$ along its main diagonal.

These quantities satisfy
\[
A v_i = \sigma_i u_i,
\qquad
A^\top u_i = \sigma_i v_i,
\]
where $v_i$ is a right singular vector, $u_i$ is the corresponding left singular vector, and $\sigma_i$ is the corresponding singular value.

Geometrically, SVD shows that each orthogonal input direction $v_i$ is mapped to an orthogonal output direction $u_i$, scaled by the corresponding singular value $\sigma_i$.
Unlike eigen-decomposition, the input and output directions need not coincide, which is precisely why SVD exists for \emph{any} real matrix, including rectangular and non-symmetric ones.
Eigen-decomposition then appears as a special case: when $A$ is symmetric, the input and output directions coincide ($U = V$), and the singular values become the absolute values of the eigenvalues.

SVD reinforces a central theme that will recur throughout this book:
complex linear transformations often become much easier to understand once the right coordinate system or decomposition is chosen.
PCA achieves this for covariance matrices by rotating into the eigenvector basis.
SVD extends the same philosophy to arbitrary matrices.

Even when later models go beyond linear transformations, this structural habit of thought remains important.
We repeatedly seek representations in which complicated operations decompose into simpler, more interpretable components.
In this sense, SVD is not merely a numerical tool, but a conceptual extension of the same geometric ideas that underlie PCA.

\begin{tcolorbox}[
  colback=blue!6,
  colframe=blue!55!black,
  title={Rank-One Structure and the Geometry of Matrices},
  boxrule=0.8pt,
  arc=2mm,
  left=2mm,
  right=2mm,
  top=1mm,
  bottom=1mm
]

One of the most revealing interpretations of SVD
is its rank-one expansion:
\[
A = \sum_{i=1}^r \sigma_i\, u_i v_i^\top,
\]
where the rank $r$ is the number of independent directions needed to build the matrix.

Each term $\sigma_i\, u_i v_i^\top$ is a \emph{rank-one transformation}, formed as
an outer product of two vectors.
Geometrically, it performs a simple three-step operation:
\[
\text{project onto } v_i
\;\rightarrow\;
\text{scale by } \sigma_i
\;\rightarrow\;
\text{reorient along } u_i.
\]

Because \(v_i^\top x\) is a scalar, the output \(u_i v_i^\top x\) is always a
scalar multiple of the single vector \(u_i\).
All possible outputs therefore lie along one fixed direction, so the
transformation can span only a one-dimensional output space, and hence rank one. It shows that any matrix can be understood as a
superposition of these elementary outer-product transformations, each capturing
one independent mode of action.
%Rather than acting monolithically, a matrix combines several simple
%project–scale–reorient operations and adds their effects together.

\end{tcolorbox}

\section{From PCA to Autoencoders: Reconstruction, Subspaces, and Manifolds}

In the previous sections, PCA was developed primarily through the lens of
\emph{variance maximisation}: we searched for directions along which the data
exhibit the greatest spread, and used these directions as a new coordinate
system.
This viewpoint reveals how PCA reorganises the data so that its most informative
variations become explicit.

There is, however, another natural and deeply intuitive way to understand the
same method.
Instead of asking which directions capture the most variance, we may ask a
reconstruction question:
\begin{quote}
\emph{If we compress the data into a $k$-dimensional representation and then
reconstruct it, which $k$-dimensional representation loses the least
information?}
\end{quote}
This reformulation turns PCA from an eigenvalue problem into an approximation
problem.
The geometry remains the same, but the emphasis shifts from \emph{capturing
variance} to \emph{minimising reconstruction error}.
This shift is important because reconstruction, rather than variance
maximisation, becomes the organising principle behind autoencoders and many
later representation-learning models.

%\subsection{Reconstruction as the Bridge from PCA to Autoencoders}

\subsection{From variance maximisation to reconstruction.}

Let $U_k = [\,u_1,\dots,u_k\,] \in \mathbb{R}^{d\times k}$ denote the matrix whose
columns are the top $k$ eigenvectors of the covariance matrix.
For a centred data point $x_i \in \mathbb{R}^d$, PCA produces the
low-dimensional representation
\[
z_i = U_k^\top x_i,
\]
which records the coordinates of $x_i$ along the principal directions.
Reconstructing from this representation gives
\[
\hat{x}_i = U_k z_i = U_k U_k^\top x_i,
\]
which is precisely the orthogonal projection of $x_i$ onto the
$k$-dimensional subspace spanned by the columns of $U_k$.

Once both the original point $x_i$ and its reconstruction $\hat{x}_i$ are
available, the natural measure of information loss is their difference.
This leads to the squared reconstruction error
\[
\|x_i - \hat{x}_i\|^2
=
\|x_i - U_k U_k^\top x_i\|^2.
\]
Summing across all data points gives the total reconstruction loss
\[
L(U_k)
=
\sum_{i=1}^n \|x_i - U_k U_k^\top x_i\|^2.
\]

From this viewpoint, PCA may be formulated as
\[
\min_{U_k^\top U_k = I_k}
\;
\sum_{i=1}^n \|x_i - U_k U_k^\top x_i\|^2,
\]
which asks:
\begin{quote}
\emph{Which orthonormal $k$-dimensional subspace minimises the distortion caused
by compressing the data and then reconstructing it?}
\end{quote}

This reconstruction formulation is fully equivalent to the earlier
variance-maximisation view of PCA, and is closely related to the classical
best low-rank approximation perspective~\cite{jolliffe2016pca}.
Although the objective is written differently, both formulations recover the
same optimal $k$-dimensional subspace spanned by the leading eigenvectors of the
covariance matrix.

Figure~\ref{fig:ch1-pca-variance-geometry} illustrates the equivalence in two dimensions: the principal direction simultaneously produces the widest projected coordinates and the smallest orthogonal reconstruction residuals.

\input{figures/ch01/fig_ch01_02_pca_variance_geometry.tex}

This perspective is especially valuable because it recasts PCA as a pair of
maps:
\[
x \;\longrightarrow\; z = U_k^\top x
\;\longrightarrow\; \hat{x} = U_k z.
\]
That is, PCA may be viewed as an \emph{encoder} followed by a
\emph{decoder}, both linear and analytically determined.

\paragraph{Autoencoders as learned encoder--decoder models.}

Once PCA is expressed in reconstruction form, a natural generalisation appears.
Instead of deriving the encoder and decoder analytically from eigenvectors, we
may try to \emph{learn} them directly from data by minimising reconstruction
error, leading to the autoencoder viewpoint~\cite{baldi1989neural,hinton2006reducing}.

Formally, an autoencoder introduces two parameterised mappings.
Given an input $x \in \mathbb{R}^d$,
\begin{itemize}
    \item the \textbf{encoder} compresses the input into a latent
    representation
    \[
    z = q_\phi(x) \in \mathbb{R}^k, \qquad k < d,
    \]
    where $q_\phi$ is parameterised by $\phi$;

    \item the \textbf{decoder} reconstructs the input from this latent code:
    \[
    \hat{x} = p_\theta(z),
    \]
    where $p_\theta$ is parameterised by $\theta$.
\end{itemize}

Together they define the familiar encoder--decoder pipeline
\[
x \;\longrightarrow\; z = q_\phi(x)
\;\longrightarrow\; \hat{x} = p_\theta(z).
\]
The parameters $(\phi,\theta)$ are then learned by minimising reconstruction
error, for example
\[
\min_{\phi,\theta}\;
\|p_\theta(q_\phi(x)) - x\|^2,
\]
or, over a dataset,
\[
\min_{\phi,\theta}\;
\sum_{i=1}^n \|p_\theta(q_\phi(x_i)) - x_i\|^2.
\]

This objective directly parallels the reconstruction formulation of PCA:
\[
\min_{U_k^\top U_k = I_k}
\sum_{i=1}^n \|U_k U_k^\top x_i - x_i\|^2.
\]
The essential difference is that, in PCA, the encoder and decoder are fixed
linear maps tied to orthonormal eigenvectors, whereas in an autoencoder they are
treated as learnable functions.

\subsection{When does a linear autoencoder recover PCA?}

The connection between linear autoencoders and PCA is classical
\cite{baldi1989neural}.
By itself, it does not guarantee that PCA will be recovered.
To recover PCA exactly, two structural ingredients matter.

\paragraph{1. Linearity.}
Suppose both encoder and decoder are linear:
\[
q_\phi(x) = W_q x,
\qquad
p_\theta(z) = W_p z,
\]
for matrices
$W_q \in \mathbb{R}^{k\times d}$ and $W_p \in \mathbb{R}^{d\times k}$.
Then the reconstruction becomes
\[
\hat{x} = W_p W_q x.
\]
In this case, the model remains within the geometric setting of PCA:
all reconstructions lie in a $k$-dimensional linear subspace of
$\mathbb{R}^d$.

If nonlinear activation functions are introduced, this restriction disappears.
The model can then represent curved, nonlinear structure rather than a single
flat subspace.
Thus linearity is essential if the goal is to remain within the classical PCA
framework.

\paragraph{2. Subspace versus basis.}
PCA does not merely identify a $k$-dimensional subspace; it identifies it using
an orthonormal basis given by the principal eigenvectors.
A linear autoencoder, by contrast, is not forced to learn an orthonormal basis.
Minimising reconstruction error alone does not impose orthogonality.

This distinction is important, but subtle.
The key geometric object for dimensionality reduction is not a particular basis,
but the \emph{subspace} spanned by that basis.

\paragraph{Subspaces and column spaces.}
Given vectors $\{v_1,\ldots,v_k\} \subset \mathbb{R}^d$, their span is
\[
\mathrm{Span}(v_1,\ldots,v_k)
=
\left\{
\sum_{i=1}^k \alpha_i v_i : \alpha_i \in \mathbb{R}
\right\}.
\]
Different sets of vectors may span exactly the same subspace.
For example,
\[
\mathrm{Span}\!\left(
\begin{bmatrix}1\\0\end{bmatrix},
\begin{bmatrix}0\\1\end{bmatrix}
\right)
=
\mathrm{Span}\!\left(
\begin{bmatrix}1\\1\end{bmatrix},
\begin{bmatrix}1\\-1\end{bmatrix}
\right)
=
\mathbb{R}^2.
\]

When vectors are collected as the columns of a matrix
$A \in \mathbb{R}^{d\times k}$, their span is called the \emph{column space}
(or range) of $A$:
\[
\mathcal{R}(A)
=
\{Az : z \in \mathbb{R}^k\}.
\]

This is the geometric object that matters for dimensionality reduction.
PCA identifies the optimal $k$-dimensional subspace using an orthonormal basis
$U_k = [\,u_1,\dots,u_k\,]$.
A linear autoencoder may identify the same subspace using a different, generally
non-orthogonal basis.

\paragraph{Subspace recovery under reconstruction loss.}
For a linear autoencoder,
\[
\hat{x} = W_p W_q x,
\]
every reconstruction lies in the column space of $W_p$:
\[
\hat{x} \in \mathcal{R}(W_p).
\]
If the model is trained optimally under squared reconstruction error, then this
column space recovers the same optimal $k$-dimensional subspace as PCA, even if
the basis vectors themselves are different.

This distinction is essential:
PCA selects a particular orthonormal basis for the optimal subspace, whereas a
linear autoencoder need only recover the subspace itself.
From the viewpoint of reconstruction, that is sufficient.
Two different bases—orthogonal or not—represent the same geometric object as
long as they span the same set of directions.

If an orthonormal basis is later desired, it can always be obtained by
orthogonalising the learned decoder matrix.
More generally, any full-rank set of vectors spanning a subspace can be converted
into an orthonormal basis for the same subspace using standard procedures such as
Gram--Schmidt orthogonalisation, QR decomposition, or related matrix
factorisations.

For example, a QR decomposition
\[
W_p = QR,
\]
with $Q^\top Q = I_k$, produces a matrix $Q$ whose columns form an orthonormal
basis for exactly the same subspace spanned by $W_p$.
The key point is that orthogonalisation changes the \emph{basis} used to
represent the subspace, but not the subspace itself.

\subsection{From linear subspaces to nonlinear manifolds}

The true power of autoencoders appears once the encoder and decoder are allowed
to be nonlinear.
In the linear case, if the decoder has the form
\[
p_\theta(z) = W_p z,
\]
then all reconstructions lie in the single linear subspace
\[
\{\hat{x} = W_p z : z \in \mathbb{R}^k\}.
\]
No matter how the latent code $z$ varies, the output remains confined to one
flat $k$-dimensional surface inside $\mathbb{R}^d$.

If the decoder becomes nonlinear, this geometry changes fundamentally.
The set of possible reconstructions becomes
\[
\mathcal{M}_\theta
=
\{p_\theta(z) : z \in \mathbb{R}^k\},
\]
which is no longer restricted to a linear subspace.
Instead, it may form a curved low-dimensional surface embedded in the ambient
space.
Such a surface is called a \emph{manifold}.

A nonlinear autoencoder therefore generalises PCA in a profound way.
Rather than learning a single flat subspace, it can learn a curved manifold that
adapts more flexibly to the intrinsic structure of the data.

Figure~\ref{fig:ch1-pca-ae-manifold} separates two ideas that are easy to conflate: a linear autoencoder may use a different basis while spanning the same PCA subspace, whereas a nonlinear decoder changes the geometry itself by allowing a curved reconstruction manifold.

\input{figures/ch01/fig_ch01_04_pca_ae_manifold.tex}

\paragraph{Nonlinear encoder and decoder.}
In practice, nonlinear autoencoders are typically implemented using neural
networks.
The encoder may take the form
\[
z = q_\phi(x),
\]
where $q_\phi$ is a composition of affine maps and nonlinear activations, and
the decoder similarly takes the form
\[
\hat{x} = p_\theta(z).
\]
The overall structure remains
\[
x \;\longrightarrow\; z \;\longrightarrow\; \hat{x},
\]
but the class of admissible transformations is now much richer than in PCA.

This transition—from linear subspaces to nonlinear manifolds—is one of the key
reasons autoencoders became such an important tool in modern representation
learning.
PCA now appears as the simplest member of a much broader family of
reconstruction-based models: linear, orthogonal, and analytically solvable.

\paragraph{Why deterministic autoencoders are not yet generative models.} Despite their greater expressive power, deterministic autoencoders still lack a
crucial ingredient for generative modelling:
\begin{quote}
\emph{a probabilistic description of how latent variables generate data.}
\end{quote}

A deterministic autoencoder maps each input $x$ to a single latent code $z$ and
then to a single reconstruction $\hat{x}$.
This provides no principled way to
\begin{itemize}
    \item assign probabilities to latent representations,
    \item generate new samples by drawing latent variables from a specified
    distribution,
    \item or train the model through likelihood-based principles.
\end{itemize}

To obtain these capabilities, we must move beyond deterministic reconstruction
models and introduce \emph{probabilistic latent-variable models}.
This transition leads directly to probabilistic PCA (PPCA), and later to the
Variational Autoencoder (VAE), where reconstruction is retained but embedded
within a probabilistic framework.

Seen from this perspective, the path is now clear.
PCA introduces linear subspaces through variance maximisation.
Its reconstruction formulation leads naturally to autoencoders.
Linear autoencoders recover the same optimal subspace as PCA, while nonlinear
autoencoders generalise this idea to manifolds.
But to arrive at modern generative modelling, we must go one step further and
equip latent representations with probability distributions.

The next chapter begins this transition with probabilistic PCA.

\section*{Summary and References}
\addcontentsline{toc}{section}{Summary and References}

This chapter introduced the linear-algebraic ideas needed for PCA and used them
to build a bridge from classical dimensionality reduction to autoencoders.

We began by interpreting matrices as linear transformations, emphasising basis
changes, dot products, and projections as the geometric language underlying PCA.
This led naturally to eigenvectors and eigenvalues, which identify the special
directions along which a matrix acts most simply.

Using these tools, PCA was developed first as variance maximisation and then as
a change of basis that diagonalises the covariance matrix.
These two views clarified why the principal components are the leading
eigenvectors, why the eigenvalues measure captured variance, and how PCA achieves
dimensionality reduction through projection and reconstruction.

The chapter then shifted from variance to reconstruction.
Recasting PCA as a linear encoder--decoder model created a direct bridge to
autoencoders.
Linear autoencoders recover the same optimal subspace as PCA, while nonlinear
autoencoders extend this idea from flat subspaces to curved manifolds.

\bigskip

The material in this chapter draws on standard ideas from linear algebra, matrix decomposition and dimensionality reduction. For the underlying linear algebra and matrix-based viewpoint, a useful broad reference is Strang~\cite{strang2016introduction}, while Jolliffe and Cadima~\cite{jolliffe2016pca} provide a modern overview of PCA and its interpretation. For historical context, the classical origins of PCA go back to Pearson~\cite{pearson1901} and Hotelling~\cite{hotelling1933}. The connection between linear autoencoders and PCA is discussed by Baldi and Hornik~\cite{baldi1989neural}, while deep autoencoders were prominently developed by Hinton and Salakhutdinov~\cite{hinton2006reducing}.

%% file: figures/ch01/chapter1_foundations_at_a_glance.tex
\begin{center}
\begin{tikzpicture}[
  >=Latex,
  every node/.style={font=\small},
  box/.style={draw,rounded corners=2pt,align=center,inner sep=5pt,
             text width=3.05cm,minimum height=1.15cm},
  flow/.style={->,thick}
]
\node[font=\normalsize\bfseries] at (6.0,2.75)
  {Linear algebra foundations at a glance};

\node[box] (mat) at (1.7,0.95) {
  \textbf{Matrices as transformations}\\[3pt]
  {\scriptsize basis images determine\\the whole linear map}
};
\node[box] (proj) at (6.0,0.95) {
  \textbf{Dot products and projections}\\[3pt]
  {\scriptsize alignment $\rightarrow$ coordinates\\along chosen directions}
};
\node[box] (eig) at (10.3,0.95) {
  \textbf{Eigenvectors and eigenvalues}\\[3pt]
  {\scriptsize natural directions\\and scaling strengths}
};
\draw[flow] (mat) -- (proj);
\draw[flow] (proj) -- (eig);

\node[box] (pca) at (10.3,-2.30) {
  \textbf{PCA}\\[3pt]
  {\scriptsize maximal variance\\principal subspace}
};
\node[box] (basis) at (6.0,-2.30) {
  \textbf{Basis change and SVD}\\[3pt]
  {\scriptsize diagonalise covariance\\and expose structure}
};
\node[box] (ae) at (1.7,-2.30) {
  \textbf{Reconstruction bridge}\\[3pt]
  {\scriptsize autoencoders\\manifolds\\probability}
};
\draw[flow] (eig) -- (pca);
\draw[flow] (pca) -- (basis);
\draw[flow] (basis) -- (ae);
\end{tikzpicture}
\end{center}

%% file: figures/ch01/fig_ch01_01_matrix_transformations.tex
\begin{figure}[H]
\centering
\begin{minipage}[t]{0.4\textwidth}
\vspace{0pt}\centering
\includegraphics[width=\linewidth]{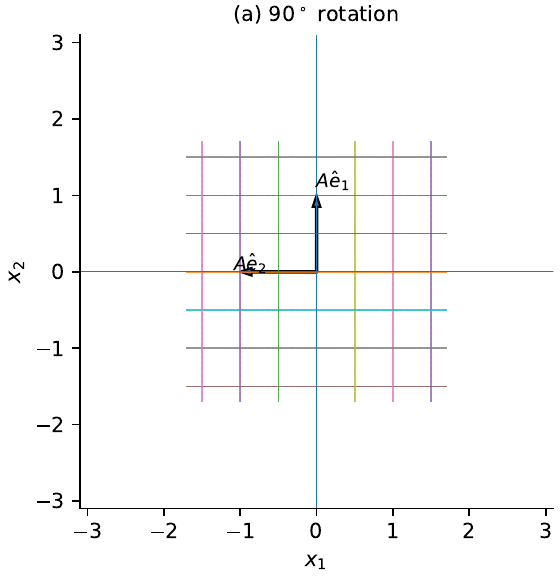}
\end{minipage}\hfill
\begin{minipage}[t]{0.4\textwidth}
\vspace{0pt}\centering
\includegraphics[width=\linewidth]{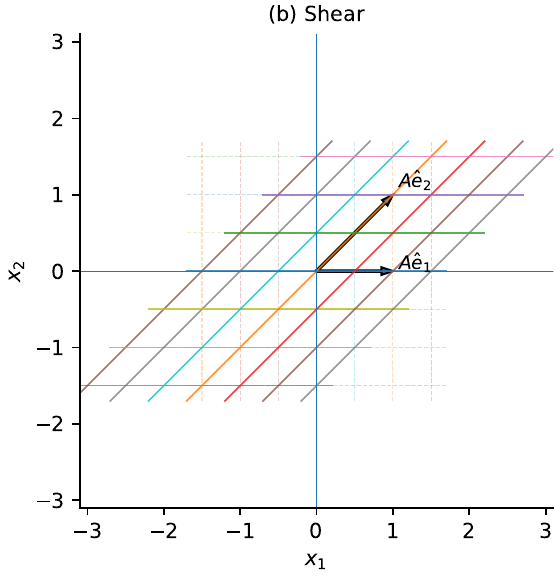}
\end{minipage}
\caption{\textbf{Basis directions determine the transformation.}
The faint dashed lines show the original coordinate grid, while the transformed grid is drawn with solid lines. The arrows $A\hat e_1$ and $A\hat e_2$ are the transformed basis directions and the columns of $A$. A rotation preserves the square grid, whereas a shear tilts one family of grid lines; once the transformed basis directions are fixed, every other vector follows by linear combination.}
\label{fig:ch1-matrix-transformations}
\end{figure}

%% file: figures/ch01/fig_ch01_03_pca_change_basis.tex
\begin{figure}[H]
\centering
\begin{minipage}[t]{0.4\textwidth}
\vspace{0pt}\centering
\includegraphics[width=\linewidth]{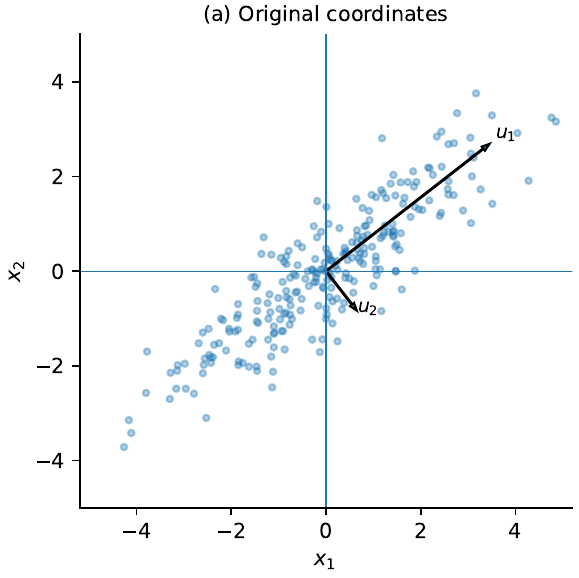}
\end{minipage}\hfill
\begin{minipage}[t]{0.42\textwidth}
\vspace{0pt}\centering
\includegraphics[width=\linewidth]{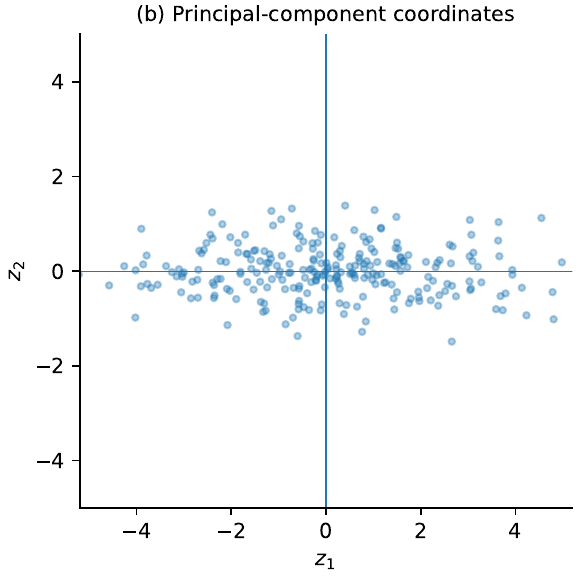}
\end{minipage}
\caption{\textbf{PCA diagonalises covariance by changing basis.}
The same data are shown in the original and principal-component coordinates. In the new basis, the principal directions become the coordinate axes, covariance between the coordinates vanishes, and total variance is preserved.}
\label{fig:ch1-pca-change-of-basis}
\end{figure}

%% file: figures/ch01/fig_ch01_02_pca_variance_geometry.tex
\begin{figure}[H]
\centering
\begin{minipage}[t]{0.42\textwidth}
\vspace{0pt}\centering
\includegraphics[width=\linewidth]{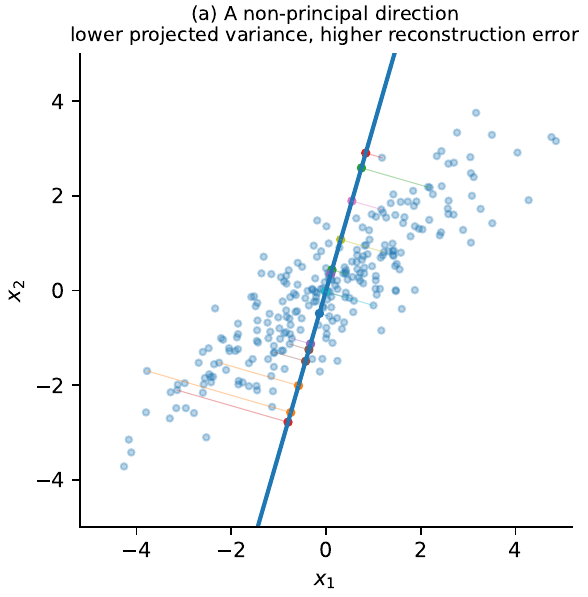}
\end{minipage}\hfill
\begin{minipage}[t]{0.42\textwidth}
\vspace{0pt}\centering
\includegraphics[width=\linewidth]{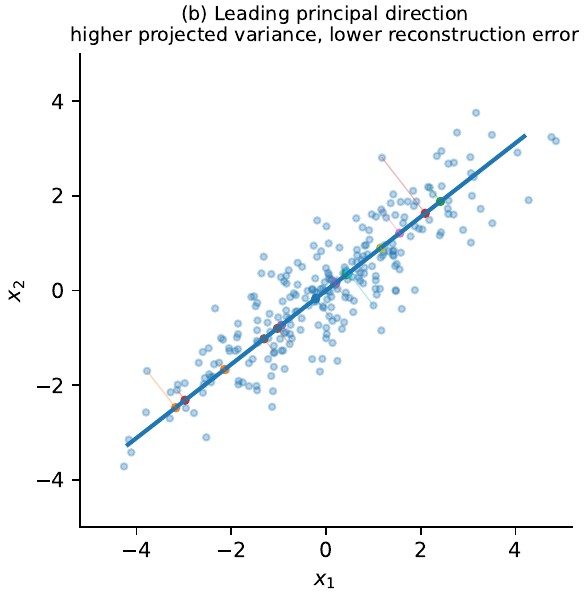}
\end{minipage}
\caption{\textbf{PCA links maximal projected variance and minimal reconstruction error.}
The same centred dataset is projected onto a non-principal direction and the leading principal direction. The principal direction gives greater projected variance and smaller orthogonal reconstruction residuals. Residual segments are shown for only a subset of points for clarity.}
\label{fig:ch1-pca-variance-geometry}
\end{figure}

%% file: figures/ch01/fig_ch01_04_pca_ae_manifold.tex
\begin{figure}[H]
\centering
\begin{minipage}[t]{0.42\textwidth}
  \vspace{0pt}\centering
  \includegraphics[width=\linewidth]{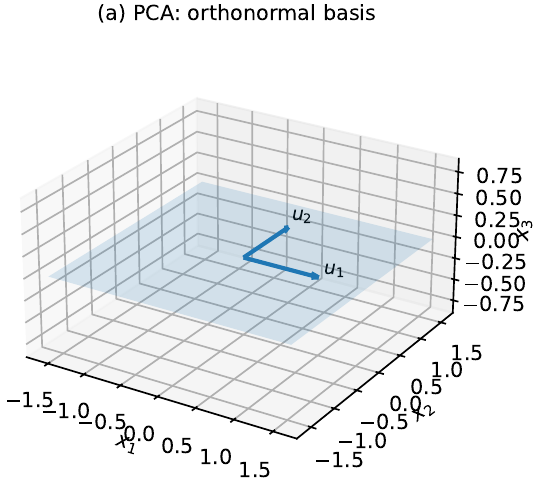}
\end{minipage}\hfill
\begin{minipage}[t]{0.42\textwidth}
  \vspace{0pt}\centering
  \includegraphics[width=\linewidth]{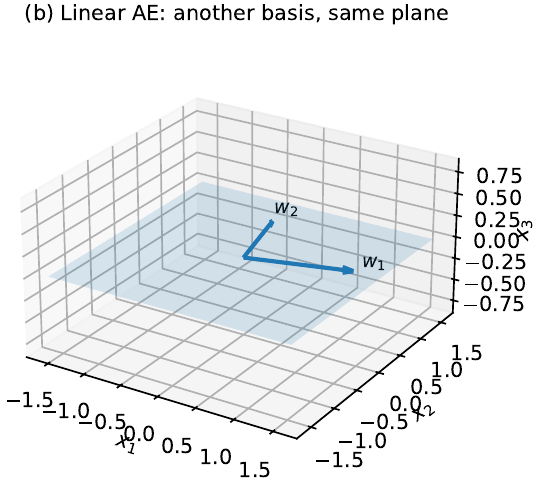}
\end{minipage}

\vspace{0.8em}

\begin{minipage}[t]{0.42\textwidth}
  \vspace{0pt}\centering
  \includegraphics[width=\linewidth]{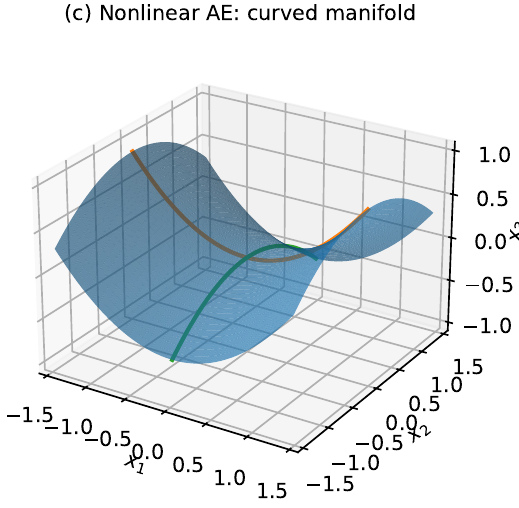}
\end{minipage}\hfill
\begin{minipage}[t]{0.42\textwidth}
  \vspace{0pt}
  \caption{\textbf{From PCA subspaces to learned manifolds.}
  PCA represents a linear subspace using a particular orthonormal basis. A linear autoencoder can recover the same optimal subspace using a different, generally non-orthogonal decoder basis; the basis vectors need not match PCA even when the represented plane does. Once the decoder is nonlinear, the reconstruction set is no longer restricted to one flat subspace and can instead form a curved low-dimensional manifold.}
  \label{fig:ch1-pca-ae-manifold}
\end{minipage}
\end{figure}

%% file: chapters/cht3.tex
\chapter{Probabilistic PCA: A Bridge to Latent-Variable Generative Modelling}
%------------------------------------------------

\section*{Overview}
\addcontentsline{toc}{section}{Overview}

Moving from deterministic representation learning to probabilistic generative modelling, probabilistic Principal Component Analysis (PPCA) is the natural next step after PCA and autoencoders, for two main reasons: 

\begin{itemize}
    \item \textbf{Introducing probabilistic latent-variable modelling.}  
    In the previous chapter, PCA and autoencoders showed how data can be represented through lower-dimensional structure, but they remained fundamentally deterministic.  
    PPCA introduces the key probabilistic ingredients that will recur throughout the rest of the book: a prior over latent variables, a conditional distribution of observations given latent causes, a marginal likelihood for learning, and a posterior distribution for inference.  
    In this sense, PPCA is the first probabilistic latent-variable model in the book.

    \item \textbf{Providing an analytically tractable bridge to modern generative models.}  
    Because PPCA is linear and Gaussian, all of its central quantities can still be derived in closed form.  
    This rare tractability makes it an ideal bridge model: it lets us study likelihood-based learning, posterior inference, Jensen’s inequality, the Evidence Lower Bound (ELBO), and the Expectation Maximisation (EM) algorithm in a setting where every step remains mathematically transparent.  
    These ideas will reappear later in more expressive but less tractable models, such as the Variational Autoencoder.
\end{itemize}

\section*{Concept Map}
\addcontentsline{toc}{section}{Concept Map}

This chapter unfolds as a guided journey from deterministic dimensionality
reduction to probabilistic latent-variable modelling.
\begin{itemize}

    \item[\(\rightarrow\)] \textbf{We begin by formulating PPCA as a probabilistic extension of PCA.}  
    A latent variable is introduced as a hidden lower-dimensional cause of the observed data, and Gaussian noise turns the deterministic PCA reconstruction into a full generative distribution over possible observations.

    \item[\(\rightarrow\)] \textbf{We then work out the marginal likelihood of the observed data.}  
    Because PPCA is linear and Gaussian, the latent variable can be integrated out exactly, showing that the observed data themselves follow a Gaussian distribution with a covariance that separates latent structure from isotropic noise.

    \item[\(\rightarrow\)] \textbf{With the marginal likelihood available explicitly, we consider three optimisation perspectives.}  
    PPCA can be learned through direct analytic maximum likelihood, generic gradient-based optimisation, or the EM algorithm.  
    This comparison shows both why PPCA is unusually tractable and why EM is the most conceptually important route for what follows.

    \item[\(\rightarrow\)] \textbf{The chapter then steps beyond exact PPCA to the general latent-variable setting.}  
    Jensen’s inequality is introduced as the key mathematical tool for replacing an intractable log-likelihood by a tractable lower bound, leading to the ELBO.

    \item[\(\rightarrow\)] \textbf{We next derive EM as an iterative way to maximise this lower bound.}  
    The E-step updates the latent-variable distribution, and the M-step updates the model parameters.  
    In PPCA, this procedure recovers the same solution as the exact analytic derivation, but now understood as part of a much more general inference principle.

    \item[\(\rightarrow\)] \textbf{Finally, we identify a remaining limitation of PPCA that motivates the next chapter.}  
    Although PPCA introduces a prior over latent variables, it does not explicitly regularise the inferred latent space to align with that prior.  
    This limitation points directly to the Variational Autoencoder, where probabilistic inference and latent-space regularisation are combined in a unified learning objective.

\end{itemize}

\medskip
\noindent The main probabilistic relationships of PPCA are summarised visually below.
\input{figures/ch02/chapter2_ppca_at_a_glance}
\medskip

\section{Formulation of Probabilistic PCA}

In the previous chapter, PCA and autoencoders introduced the idea that
high-dimensional data may be represented through a lower-dimensional latent
structure.
However, both remained fundamentally \emph{deterministic}: once a latent
representation is chosen, the reconstruction is fixed.
A central step towards generative modelling is to move beyond this deterministic
picture and allow a latent point to correspond not to a single observation, but
to a \emph{distribution} over possible observations.

Probabilistic Principal Component Analysis (PPCA)~\cite{tipping1999ppca}, is the simplest model in this
book that makes this transition explicit.
It retains the linear latent structure of PCA, but embeds it within a
probabilistic generative process, thus becoming our first concrete example of a
latent-variable generative model.

\paragraph{PPCA as a latent-variable generative model.}
Before introducing the model formally, it is worth pausing on the term
\emph{latent variable}.
The word \emph{latent} simply means \emph{hidden}: it refers to a variable that
is not directly observed, but is assumed to lie behind the data and explain part
of its structure.
In PPCA, the observed vector $\mathbf{x}$ is the data point we see, while
$\mathbf{z}$ is a lower-dimensional hidden representation that we do not observe
directly.
The model assumes that $\mathbf{x}$ is generated from $\mathbf{z}$, rather than
treating $\mathbf{x}$ as an isolated object.
This shift—from observed data alone to observed data generated from hidden
causes—is one of the defining ideas of latent-variable generative modelling.

The PPCA generative process consists of two steps:
\begin{enumerate}
    \item Sample a latent vector
    \[
        \mathbf{z} \sim \mathcal{N}(\mathbf{0}, \mathbf{I}),
    \]
    from a standard Gaussian prior.

    \item Generate an observation by applying a linear transformation, adding a
    mean offset, and then injecting Gaussian noise $\boldsymbol{\epsilon}$:
    \[
        \mathbf{x} = \mathbf{W}\mathbf{z} + \boldsymbol{\mu} + \boldsymbol{\epsilon},
        \qquad
        \boldsymbol{\epsilon} \sim \mathcal{N}(\mathbf{0}, \alpha^2 \mathbf{I}).
    \]
\end{enumerate}

Here,
\begin{itemize}
    \item $\mathbf{z} \in \mathbb{R}^k$ is the latent variable,
    \item $\mathbf{x} \in \mathbb{R}^d$ is the observed data point,
    \item $\mathbf{W} \in \mathbb{R}^{d\times k}$ maps latent coordinates into the observation space,
    \item $\boldsymbol{\mu} \in \mathbb{R}^d$ is the mean of the data,
    \item and $\alpha^2$ controls the isotropic observation noise.
\end{itemize}

This model already contains the essential ingredients of many later generative
models:
a prior over latent variables, a rule that maps latent variables to
observations, and a probability distribution over possible outputs.
The key novelty relative to PCA is that the latent representation no longer
determines a single point exactly.
Instead, it determines the \emph{centre} of a distribution of plausible
observations.

\paragraph{Conditional distribution of the observation.}

Although the generative equation
\[
\mathbf{x} = \mathbf{W}\mathbf{z} + \boldsymbol{\mu} + \boldsymbol{\epsilon}
\]
looks similar to a deterministic mapping, its meaning is fundamentally
probabilistic because of the random noise term.
Conditioned on a latent vector $\mathbf{z}$, the observation is Gaussian:
\[
\mathbf{x}\mid \mathbf{z}
\sim
\mathcal{N}\!\bigl(\mathbf{W}\mathbf{z} + \boldsymbol{\mu},\; \alpha^2 \mathbf{I}\bigr).
\]

This is the central probabilistic shift: once $\mathbf{z}$ is fixed,
$\mathbf{W}\mathbf{z}+\boldsymbol{\mu}$ determines the \emph{mean} of the
observation distribution rather than a single reconstruction.
Figure~\ref{fig:ppca-generative-story} illustrates the geometry for the simple
case $k=1$ and $d=2$. As $z$ varies, the conditional means trace a line in
observation space, while isotropic Gaussian noise generates possible
observations around each mean. The latent variable therefore controls position
along the low-dimensional subspace, whereas the noise allows observations to
deviate from it.

\input{figures/ch02/fig_ch02_01_ppca_generative_story}

\paragraph{Connection with PCA: the zero-noise limit.}

At this point it is natural to ask how PPCA relates back to classical PCA.
The answer is most clearly seen through the zero-noise limit. Recall the PPCA model:
\[
\mathbf{x} = \mathbf{W}\mathbf{z} + \boldsymbol{\mu} + \boldsymbol{\epsilon},
\qquad
\mathbf{z} \sim \mathcal{N}(\mathbf{0}, \mathbf{I}),
\qquad
\boldsymbol{\epsilon} \sim \mathcal{N}(\mathbf{0}, \alpha^2 \mathbf{I}).
\]
As the observation noise tends to zero $\alpha^2 \to 0,$, the conditional distribution
\[
p(\mathbf{x}\mid \mathbf{z})
=
\mathcal{N}\!\bigl(\mathbf{x}\mid \mathbf{W}\mathbf{z}+\boldsymbol{\mu},\; \alpha^2 \mathbf{I}\bigr)
\]
becomes increasingly concentrated around its mean.
In the limit, the randomness disappears and each latent coordinate $\mathbf{z}$
maps to a single deterministic point:
\[
\mathbf{x} = \mathbf{W}\mathbf{z} + \boldsymbol{\mu}.
\]

In this sense, PCA may be understood as the zero-noise limit of PPCA.
The latent subspace remains, but the probabilistic spread around it vanishes.
PPCA therefore contains PCA as a special limiting case while adding something
essentially new: uncertainty around the reconstruction.

In the next section, we will see that PPCA remains unusually tractable:
because the model is linear and Gaussian, the marginal likelihood
\(p(\mathbf{x})\) can still be derived analytically.

\section{Why PPCA Is Analytically Tractable}

Having formulated PPCA as a latent-variable generative model, the next question is how to fit it to data.
In PCA we searched for directions of maximal variance, and in autoencoders we minimised reconstruction error.
In PPCA, by contrast, the model is probabilistic, so the natural fitting principle is to maximise the likelihood of the observed data under the model.

\subsection{From Likelihood to Marginalisation}
Suppose we observe a dataset
\[
\{\mathbf{x}_1,\dots,\mathbf{x}_N\}.
\]
The log-likelihood of the dataset under PPCA is
\[
\sum_{i=1}^N \log p(\mathbf{x}_i).
\]
It is often convenient to divide by \(N\) and write this as an average log-likelihood,
\[
\frac{1}{N}\sum_{i=1}^N \log p(\mathbf{x}_i),
\]
which may also be expressed as an expectation under the empirical data distribution:
\[
\mathbb{E}_{q(\mathbf{x})}\big[\log p(\mathbf{x})\big].
\]
Here, \(q(\mathbf{x})\) simply means the empirical distribution that places equal mass on each observed data point.
Thus, taking an expectation under \(q(\mathbf{x})\) is nothing more than averaging over the training samples, a notation commonly appearing in more general probabilistic setting used later in the book.

The PPCA objective is therefore
\[
\max_{\mathbf{W},\boldsymbol{\mu},\alpha}
\;
\mathbb{E}_{q(\mathbf{x})}\big[\log p(\mathbf{x}; \mathbf{W}, \boldsymbol{\mu}, \alpha)\big].
\]
Here the marginal likelihood depends on the model parameters
\((\mathbf{W}, \boldsymbol{\mu}, \alpha)\).
To avoid overly heavy notation, however, we will often suppress this dependence
and write simply \(p(\mathbf{x})\) when the meaning is clear.

The key quantity here is the marginal likelihood \(p(\mathbf{x})\), which, by the law of total probability, is obtained by integrating out the latent variable:
\[
p(\mathbf{x})
=
\int p(\mathbf{x}\mid \mathbf{z})\,p(\mathbf{z})\,d\mathbf{z}.
\]

\begin{tcolorbox}[
  colback=blue!6,
  colframe=blue!55!black,
  title={Why maximise the log-likelihood?},
  boxrule=0.8pt,
  arc=2mm,
  left=2mm,
  right=2mm,
  top=1mm,
  bottom=1mm
]
We maximise the \emph{log}-likelihood rather than the likelihood itself because it is both numerically safer and algebraically simpler.
For independent data points,
\[
L=\prod_{i=1}^N p(\mathbf{x}_i),
\qquad
\log L=\sum_{i=1}^N \log p(\mathbf{x}_i).
\]
The logarithm turns products of many small probabilities into sums, avoiding numerical underflow and giving the additive form, which is easier to analyse and optimise.
\[
\frac{1}{N}\sum_{i=1}^N \log p(\mathbf{x}_i)
=
\mathbb{E}_{q(\mathbf{x})}\big[\log p(\mathbf{x})\big].
\]
\end{tcolorbox}

At first glance, this marginal likelihood may appear difficult to compute.
Substituting the PPCA model gives
\[
p(\mathbf{x})
=
\int
\mathcal{N}\!\bigl(\mathbf{x}\mid \mathbf{W}\mathbf{z}+\boldsymbol{\mu}, \alpha^2\mathbf{I}\bigr)
\;
\mathcal{N}\!\bigl(\mathbf{z}\mid \mathbf{0}, \mathbf{I}\bigr)
\,d\mathbf{z}.
\]
Since \(\mathbf{z}\) ranges over a continuous latent space, it may seem that evaluating this integral exactly should be intractable.
In most latent-variable models, that intuition is correct.
PPCA is special precisely because its linear-Gaussian structure keeps this marginalisation analytically tractable.

\begin{tcolorbox}[
  colback=blue!5!white,
  colframe=blue!75!black,
  title={Why linear Gaussian models are analytically manageable},
  boxrule=0.8pt,
  arc=2mm,
  left=2mm,
  right=2mm,
  top=1mm,
  bottom=1mm
]
\begin{itemize}
    \item \textbf{Linear transformations preserve Gaussianity.}  
    If a random vector is Gaussian, then any affine transformation of it is also Gaussian.

    \item \textbf{Sums of independent Gaussians are Gaussian.}  
    If two Gaussian random variables are independent, then their sum is again Gaussian.

    \item \textbf{Integrating out Gaussian latent variables remains Gaussian.}  
    If a joint distribution over \((\mathbf{x},\mathbf{z})\) is Gaussian, then marginalising over \(\mathbf{z}\) leaves a Gaussian distribution for \(\mathbf{x}\).
\end{itemize}
\end{tcolorbox}

These facts allow us to identify the distribution of \(\mathbf{x}\) without explicitly computing the integral by brute force.

% \subsection*{A direct derivation from the generative model}
\subsection{Deriving and Interpreting the Marginal Distribution}

Recall the PPCA generative equation:
\[
\mathbf{x} = \mathbf{W}\mathbf{z} + \boldsymbol{\mu} + \boldsymbol{\epsilon},
\qquad
\mathbf{z}\sim\mathcal{N}(\mathbf{0},\mathbf{I}),
\qquad
\boldsymbol{\epsilon}\sim\mathcal{N}(\mathbf{0},\alpha^2\mathbf{I}),
\]
with the latent variable \(\mathbf{z}\) and the noise term \(\boldsymbol{\epsilon}\) assumed to be independent.
This shows that \(\mathbf{x}\) is constructed as an affine transformation of a Gaussian latent variable together with an independent Gaussian noise term.
Because affine transformations of Gaussians remain Gaussian, and sums of independent Gaussians are again Gaussian, it follows that \(\mathbf{x}\) itself must also be Gaussian.

The same marginal can also be derived by expanding the joint density
\[
p(\mathbf{x},\mathbf{z})=p(\mathbf{x}\mid\mathbf{z})p(\mathbf{z})
\]
and then integrating over \(\mathbf{z}\) using the so-called \emph{completing the square} trick.
This algebraic route is reviewed in
Appendix~\ref{app:completing-square}, where the same idea is used to show how
Gaussian joint densities lead to Gaussian marginal and posterior distributions.
For now, we use the simpler distributional argument based on affine
transformations of Gaussian variables.

We can now work out its mean and covariance directly.

\paragraph{Mean.}
Taking expectations gives
\[
\mathbb{E}[\mathbf{x}]
=
\mathbb{E}[\mathbf{W}\mathbf{z} + \boldsymbol{\mu} + \boldsymbol{\epsilon}]
=
\mathbf{W}\,\mathbb{E}[\mathbf{z}] + \boldsymbol{\mu} + \mathbb{E}[\boldsymbol{\epsilon}]
=
\boldsymbol{\mu},
\]
since \(\mathbb{E}[\mathbf{z}]=\mathbf{0}\) and \(\mathbb{E}[\boldsymbol{\epsilon}]=\mathbf{0}\).

\paragraph{Covariance.}
Using
\[
\mathbf{x}=\mathbf{W}\mathbf{z}+\boldsymbol{\mu}+\boldsymbol{\epsilon},
\]
and noting that the constant offset \(\boldsymbol{\mu}\) does not affect covariance, we have
\[
\operatorname{Cov}[\mathbf{x}]
=
\operatorname{Cov}[\mathbf{W}\mathbf{z}+\boldsymbol{\epsilon}].
\]

To expand this more explicitly, recall the general identity
\[
\operatorname{Cov}[a+b]
=
\operatorname{Cov}[a]
+
\operatorname{Cov}[b]
+
\operatorname{Cov}[a,b]
+
\operatorname{Cov}[b,a].
\]
Applying this with \(a=\mathbf{W}\mathbf{z}\) and \(b=\boldsymbol{\epsilon}\) gives
\[
\operatorname{Cov}[\mathbf{x}]
=
\operatorname{Cov}[\mathbf{W}\mathbf{z}]
+
\operatorname{Cov}[\boldsymbol{\epsilon}]
+
\operatorname{Cov}[\mathbf{W}\mathbf{z},\boldsymbol{\epsilon}]
+
\operatorname{Cov}[\boldsymbol{\epsilon},\mathbf{W}\mathbf{z}].
\]

Because \(\mathbf{z}\) and \(\boldsymbol{\epsilon}\) are independent, the cross-covariance terms vanish:
\[
\operatorname{Cov}[\mathbf{W}\mathbf{z},\boldsymbol{\epsilon}] = \mathbf{0},
\qquad
\operatorname{Cov}[\boldsymbol{\epsilon},\mathbf{W}\mathbf{z}] = \mathbf{0}.
\]
Hence
\[
\operatorname{Cov}[\mathbf{x}]
=
\operatorname{Cov}[\mathbf{W}\mathbf{z}]
+
\operatorname{Cov}[\boldsymbol{\epsilon}].
\]

Now we use the standard rule for covariance under a linear transformation:
\[
\operatorname{Cov}[\mathbf{W}\mathbf{z}]
=
\mathbf{W}\operatorname{Cov}[\mathbf{z}]\,\mathbf{W}^\top.
\]
Therefore,
\[
\operatorname{Cov}[\mathbf{x}]
=
\mathbf{W}\operatorname{Cov}[\mathbf{z}]\,\mathbf{W}^\top
+
\operatorname{Cov}[\boldsymbol{\epsilon}].
\]

Finally, substituting
\[
\operatorname{Cov}[\mathbf{z}] = \mathbf{I},
\qquad
\operatorname{Cov}[\boldsymbol{\epsilon}] = \alpha^2\mathbf{I},
\]
yields
\[
\operatorname{Cov}[\mathbf{x}]
=
\mathbf{W}\mathbf{W}^\top + \alpha^2\mathbf{I}.
\]

Putting these together, the marginal distribution of the observation is
\[
p(\mathbf{x})
=
\mathcal{N}\!\bigl(
\mathbf{x}\mid
\boldsymbol{\mu},
\mathbf{W}\mathbf{W}^\top + \alpha^2\mathbf{I}
\bigr).
\]

This is the key analytical result for PPCA: the latent variable has been
integrated out exactly, and the resulting marginal over observations remains
Gaussian.

\paragraph{Interpreting the covariance structure.}
The term $\mathbf{W}\mathbf{W}^\top$ is the covariance contributed in
observation space by the mapped latent variation, because
\[
\operatorname{Cov}(\mathbf{W}\mathbf{z})
=
\mathbf{W}\operatorname{Cov}(\mathbf{z})\mathbf{W}^\top
=
\mathbf{W}\mathbf{W}^\top.
\]
It is low rank,
\[
\operatorname{rank}(\mathbf{W}\mathbf{W}^\top)\le k,
\]
so latent-induced variability is confined to at most a $k$-dimensional
subspace of the $d$-dimensional observation space. If
\[
\mathbf{W}=[\,w_1\;\;w_2\;\;\cdots\;\;w_k\,],
\qquad
\mathbf{W}\mathbf{W}^\top=\sum_{j=1}^k w_jw_j^\top,
\]
the structured covariance can be viewed as a sum of rank-one contributions
associated with the latent directions.

By contrast, $\alpha^2\mathbf{I}$ contributes equal variance in every
observation-space direction. PPCA therefore models variability both
\emph{along} the latent subspace through $\mathbf{W}\mathbf{W}^\top$ and
\emph{around} it through isotropic noise. Figure~\ref{fig:ppca-covariance-decomposition}
summarises this decomposition visually, separating the latent-induced structure
from the isotropic observation noise.

\input{figures/ch02/fig_ch02_02_ppca_covariance_decomposition}

\bigskip

In the next section, we turn from the form of \(p(\mathbf{x})\) to the question
of learning.
Now that the marginal likelihood is available explicitly, we can ask how the
parameters \((\mathbf{W}, \boldsymbol{\mu}, \alpha)\) may be estimated from data.

\section{Learning PPCA: Three Optimisation Perspectives}

Having derived the marginal distribution
\[
p(\mathbf{x})
=
\mathcal{N}\!\bigl(
\mathbf{x}\mid
\boldsymbol{\mu},
\mathbf{W}\mathbf{W}^\top + \alpha^2\mathbf{I}
\bigr),
\]
we can now turn from representation to learning.
The PPCA parameters \((\mathbf{W}, \boldsymbol{\mu}, \alpha)\) are fitted by maximising the log-likelihood of the observed dataset:
\[
\max_{\mathbf{W},\boldsymbol{\mu},\alpha}
\;
\mathbb{E}_{q(\mathbf{x})}\big[\log p(\mathbf{x})\big].
\]

At this point, a natural question arises:
\begin{quote}
\emph{Now that the likelihood is available explicitly, how should we optimise it?}
\end{quote}
There are three useful ways to think about this question.
The first exploits the special analytic structure of PPCA.
The second uses generic gradient-based optimisation.
The third, Expectation--Maximisation (EM), is especially important because it reveals the deeper latent-variable principles that will later lead to variational inference and the VAE.

\subsection{Direct analytic maximum likelihood}

Because the marginal distribution of \(\mathbf{x}\) is Gaussian, PPCA admits an
unusually elegant route to learning.
In principle, one may write down the log-likelihood explicitly, differentiate it
with respect to the parameters, set the resulting derivatives to zero, and solve
for the stationary point:
\[
\nabla_{\mathbf{W}} \log p(\mathbf{x}) = 0,
\qquad
\nabla_{\boldsymbol{\mu}} \log p(\mathbf{x}) = 0,
\qquad
\frac{\partial}{\partial \alpha} \log p(\mathbf{x}) = 0.
\]

This is the classical maximum-likelihood strategy.
Its attraction is clear: if the equations can be solved exactly, then the model
parameters can be expressed directly in terms of the empirical covariance
structure of the data.

For PPCA, this route succeeds because the model remains linear and Gaussian all
the way through.
The final solution can be written in closed form and is closely tied to the
eigenstructure of the sample covariance matrix.
This is one reason PPCA is such an important pedagogical model: it is a genuine
latent-variable generative model that still admits an exact maximum-likelihood
solution.

At the same time, this route also reveals a limitation.
Although elegant, direct analytic optimisation depends strongly on the model
being simple enough for the derivatives to be solved exactly.
As soon as the latent-variable model becomes nonlinear or non-Gaussian, this
closed-form route typically breaks down.
Moreover, even when an analytic solution exists, obtaining it may still require
computationally expensive operations such as matrix inversion or
eigen-decomposition, especially in high dimensions.

\subsection{Gradient-based optimisation}

A second possibility is to treat the log-likelihood simply as an objective
function and optimise it numerically by gradient ascent.
In this view, the parameters are updated iteratively according to
\[
\theta
\leftarrow
\theta + \eta \,\nabla_\theta \mathbb{E}_{q(\mathbf{x})}\big[\log p(\mathbf{x};\theta)\big],
\]
where $\theta = \{\mathbf{W}, \boldsymbol{\mu}, \alpha\},$ and \(\eta\) is a step size.

This is the most general optimisation viewpoint.
Unlike direct analytic maximum likelihood, it does not require that the optimal
parameters themselves be solvable in closed form.
Instead, it treats PPCA like any other differentiable probabilistic model and
uses gradients to climb the likelihood surface iteratively.

In PPCA, the marginal likelihood \(p(\mathbf{x})\) is available explicitly, so
its gradient can also be computed explicitly.
The difference from the analytic route is therefore not that gradients avoid the
likelihood itself, but that they avoid having to solve the optimisation problem
exactly in closed form.
More generally, even when no exact closed-form solution for the parameters
exists, gradient-based methods can still be used whenever the objective or its
gradient can be computed—or at least approximated.

This perspective is worth emphasising because it foreshadows how most modern
generative models are trained.
In later chapters, when closed-form solutions are no longer available,
gradient-based optimisation will become the default computational tool.

However, in the particular case of PPCA, gradient ascent is not the most
illuminating route.
It can certainly be applied, but it does not exploit the special latent-variable
structure of the model.
Nor does it reveal why latent-variable models are often approached through
posterior inference rather than direct optimisation alone.
For that, we need a third viewpoint.

\subsection{Expectation--Maximisation (EM)}

The third route is the Expectation--Maximisation (EM) algorithm~\cite{dempster1977em}. Unlike direct analytic optimisation, EM does not try to solve the likelihood in a
single step.
Unlike generic gradient ascent, it does not treat the latent variables as merely
hidden nuisance variables inside a black-box objective.
Instead, it works \emph{with} the latent-variable structure of the model.

The basic idea is simple.
Although the marginal likelihood
\[
\log p(\mathbf{x})
=
\log \int p(\mathbf{x},\mathbf{z})\,d\mathbf{z}
\]
may be difficult to optimise directly, the joint distribution
\[
p(\mathbf{x},\mathbf{z})
\]
often has a much simpler form.
EM exploits this by alternating between two tasks:
\begin{itemize}
    \item estimating the latent variables under the current parameters;
    \item updating the parameters using those estimates.
\end{itemize}

In PPCA, EM is not strictly necessary, because the model can already be solved
analytically.
But it is still extremely valuable.
It reveals a general optimisation principle for latent-variable models, and that
principle survives even when exact solutions do not.

Seen from a broader perspective, PPCA is especially valuable because it is one of
the few latent-variable models in which all of the central probabilistic
quantities can be written down explicitly:
the prior \(p(\mathbf{z})\), the conditional distribution
\(p(\mathbf{x}\mid\mathbf{z})\), the marginal likelihood \(p(\mathbf{x})\), and
the posterior \(p(\mathbf{z}\mid\mathbf{x})\).
In more complex models, these same quantities remain central, but they no longer
stay analytically tractable.
In particular, learning is driven by the marginal likelihood
\(p(\mathbf{x})\), while inference depends on the posterior
\(p(\mathbf{z}\mid\mathbf{x})\).
This is precisely why EM is so important: it provides a structured way to
alternate between these two aspects of the latent-variable problem.

More importantly for what follows, EM naturally leads to two deeper ideas:
the introduction of an auxiliary distribution over latent variables, and the
replacement of a difficult log-likelihood by a lower bound that can be optimised
iteratively.
These ideas are not unique to PPCA.
They belong to the general theory of latent-variable modelling, and they will
reappear in a broader form when we later study variational inference.

For this reason, we now turn to the mathematical foundation underlying EM:
Jensen’s inequality and the Evidence Lower Bound (ELBO).

\section{Jensen’s Inequality and the Evidence Lower Bound}

To understand why EM is so important, we now need to step back and ask a more
general question.
What do we do when the marginal likelihood
\[
p(\mathbf{x};\theta)=\int p(\mathbf{x},\mathbf{z};\theta)\,d\mathbf{z}
\]
is no longer analytically tractable?
PPCA is special because this integral can still be computed exactly.
Most latent-variable models do not have this luxury.
In more general settings, the latent variable may enter through nonlinear
transformations, non-Gaussian conditionals, or both, and the marginal
likelihood can no longer be worked out in closed form.

This is where Jensen’s inequality enters.
It provides a systematic way to replace the difficult log-likelihood by a
tractable lower bound.
That lower bound is the \emph{Evidence Lower Bound (ELBO)}, and it is
one of the central ideas underlying EM and more 
generative models introduced in later chapters.

\subsection{From Marginal Likelihood to Jensen’s Inequality}

The latent-variable structure of PPCA is not unique to PPCA itself.
More generally, many generative models assume that data are produced in two
stages:
\begin{enumerate}
    \item sample a latent variable \(\mathbf{z}\) from a prior \(p(\mathbf{z})\);
    \item generate an observation \(\mathbf{x}\) from a conditional distribution
    \(p(\mathbf{x}\mid\mathbf{z};\theta)\).
\end{enumerate}
The corresponding marginal likelihood is
\[
p(\mathbf{x};\theta)
=
\int p(\mathbf{x}\mid\mathbf{z};\theta)\,p(\mathbf{z})\,d\mathbf{z}
=
\int p(\mathbf{x},\mathbf{z};\theta)\,d\mathbf{z}.
\]

For a dataset \(\{\mathbf{x}^{(i)}\}\), maximum likelihood learning becomes
\[
\max_\theta \sum_i \log p(\mathbf{x}^{(i)};\theta)
=
\max_\theta \sum_i \log \int p(\mathbf{x}^{(i)},\mathbf{z};\theta)\,d\mathbf{z}.
\]
The difficulty lies in the form of this expression:
the latent variable must be integrated out, and the logarithm sits outside the
integral.
In PPCA, the linear-Gaussian structure allowed us to bypass this obstacle.
In general, we need a more systematic tool.

That tool begins with Jensen’s inequality.
Recall that a \emph{convex} function curves upward, so that the straight line
joining any two points on its graph lies above the graph itself.
A \emph{concave} function curves downward, so that the straight line joining any
two points lies below the graph.
For a random variable \(X\) and a convex function \(f\), Jensen’s inequality states
\[
f(\mathbb{E}[X]) \le \mathbb{E}[f(X)].
\]
If \(f\) is concave, the direction of the inequality reverses:
\[
f(\mathbb{E}[X]) \ge \mathbb{E}[f(X)].
\]

In the present context, the function \(f\) we care about is specifically the
logarithm.
Since \(\log\) is concave,
\[
\log \mathbb{E}[X] \ge \mathbb{E}[\log X].
\]

This is precisely the form we need.
In latent-variable models, the problematic quantity is typically
\[
\log \int (\cdots)\,d\mathbf{z},
\]
or equivalently \(\log \mathbb{E}[\cdot]\), where the logarithm sits outside an
integral or expectation.
Jensen’s inequality allows us to move this logarithm inside the expectation,
trading exactness for a lower bound that is much easier to work with.

\subsection{Deriving the Evidence Lower Bound}

To apply Jensen’s inequality, we first rewrite the marginal likelihood as an
expectation.
Let \(Q(\mathbf{z})\) be any probability distribution over the latent variable,
so that
\[
\int Q(\mathbf{z})\,d\mathbf{z} = 1.
\]
Then
\[
p(\mathbf{x};\theta)
=
\int p(\mathbf{x},\mathbf{z};\theta)\,d\mathbf{z}
=
\int Q(\mathbf{z})
\frac{p(\mathbf{x},\mathbf{z};\theta)}{Q(\mathbf{z})}\,d\mathbf{z}
=
\mathbb{E}_{\mathbf{z}\sim Q}
\!\left[
\frac{p(\mathbf{x},\mathbf{z};\theta)}{Q(\mathbf{z})}
\right].
\]
Taking logarithms gives
\[
\log p(\mathbf{x};\theta)
=
\log
\mathbb{E}_{\mathbf{z}\sim Q}
\!\left[
\frac{p(\mathbf{x},\mathbf{z};\theta)}{Q(\mathbf{z})}
\right].
\]

At first glance this may look like a purely algebraic trick, but it is an
essential step.
The distribution \(Q(\mathbf{z})\) is a freely chosen auxiliary distribution,
introduced so that the marginal likelihood can be expressed as an expectation.
Once this has been done, Jensen’s inequality becomes available.

We are now ready to apply Jensen’s inequality to the rewritten marginal
likelihood:
\[
\log p(\mathbf{x};\theta)
=
\log
\mathbb{E}_{\mathbf{z}\sim Q}
\!\left[
\frac{p(\mathbf{x},\mathbf{z};\theta)}{Q(\mathbf{z})}
\right].
\]
Since the logarithm is concave, Jensen’s inequality gives
\[
\log
\mathbb{E}_{\mathbf{z}\sim Q}
\!\left[
\frac{p(\mathbf{x},\mathbf{z};\theta)}{Q(\mathbf{z})}
\right]
\ge
\mathbb{E}_{\mathbf{z}\sim Q}
\!\left[
\log
\frac{p(\mathbf{x},\mathbf{z};\theta)}{Q(\mathbf{z})}
\right].
\]
Hence,
\[
\log p(\mathbf{x};\theta)
\ge
\mathbb{E}_{\mathbf{z}\sim Q}
\!\left[
\log
\frac{p(\mathbf{x},\mathbf{z};\theta)}{Q(\mathbf{z})}
\right].
\]

This lower bound is called the \emph{Evidence Lower Bound}, or \emph{ELBO}:
\[
\mathcal{L}(Q,\theta;\mathbf{x})
=
\mathbb{E}_{\mathbf{z}\sim Q}
\!\left[
\log
\frac{p(\mathbf{x},\mathbf{z};\theta)}{Q(\mathbf{z})}
\right].
\]
Expanding the logarithm of the fraction gives the equivalent form
\[
\mathcal{L}(Q,\theta;\mathbf{x})
=
\mathbb{E}_{Q}\big[\log p(\mathbf{x},\mathbf{z};\theta)\big]
-
\mathbb{E}_{Q}\big[\log Q(\mathbf{z})\big].
\]

This expression is much easier to work with because the logarithm now acts
directly on the joint density inside an expectation, rather than sitting
outside an intractable integral.
As a result, it becomes much more amenable to iterative optimisation.

This is the central result of the section.
Instead of maximising the generally intractable quantity
\(\log p(\mathbf{x};\theta)\) directly, we may instead maximise a tractable
lower bound that depends on two objects: the model parameters \(\theta\) and
the auxiliary distribution \(Q(\mathbf{z})\).

\subsection{When does the bound become exact?}

A lower bound is useful only if we understand when it is tight.
For Jensen’s inequality, equality holds when the argument of the concave
function—in our case, the logarithm—is constant with respect to the averaging
variable.
Here, that averaging variable is \(\mathbf{z}\), so equality holds precisely
when
\[
\frac{p(\mathbf{x},\mathbf{z};\theta)}{Q(\mathbf{z})}
=
c
\qquad
\text{for all }\mathbf{z},
\]
for some constant \(c\) independent of \(\mathbf{z}\).

Rearranging this expression gives $Q(\mathbf{z})
\propto
p(\mathbf{x},\mathbf{z};\theta).$ To determine the normalising constant, integrate both sides with respect to
\(\mathbf{z}\).
Since \(Q(\mathbf{z})\) is a valid probability distribution, $\int Q(\mathbf{z})\,d\mathbf{z} = 1.$

Therefore,
\[
p(\mathbf{x};\theta)
=
\int p(\mathbf{x},\mathbf{z};\theta)\,d\mathbf{z}
=
c \int Q(\mathbf{z})\,d\mathbf{z}
=
c.
\]
So the constant must be
\[
c = p(\mathbf{x};\theta).
\]
Substituting this back into the equality condition yields
\[
Q(\mathbf{z})
=
\frac{p(\mathbf{x},\mathbf{z};\theta)}{p(\mathbf{x};\theta)}
=
p(\mathbf{z}\mid \mathbf{x};\theta),
\]
which is exactly the true posterior distribution of the latent variable.

Hence the ELBO becomes equal to the true log-likelihood exactly when $Q(\mathbf{z})=p(\mathbf{z}\mid\mathbf{x};\theta),$ that is, when the auxiliary distribution matches the true posterior over the
latent variable. This is a crucial point. The ELBO is not merely an arbitrary surrogate objective.
Its tightness is controlled by how well \(Q(\mathbf{z})\) approximates the true
posterior.

This perspective also clarifies the role of PPCA in the chapter.
PPCA is not special because it avoids the latent-variable problem, but because
it solves it exactly.
Its posterior \(p(\mathbf{z}\mid\mathbf{x})\) is tractable, and its marginal
likelihood \(p(\mathbf{x})\) can also be computed analytically.
For that reason, PPCA does not strictly require the ELBO.

However, the ELBO framework reveals the more general principle that PPCA
exemplifies.
The same latent-variable structure persists in more complex models, but the
exact posterior and exact marginal likelihood are no longer available.
The ELBO is the device that allows us to keep going when exact Gaussian
calculations break down.

\bigskip

We are now in a position to introduce the Expectation--Maximisation algorithm.
The ELBO depends on two coupled ingredients:
the auxiliary distribution \(Q(\mathbf{z})\) and the model parameters
\(\theta\).
This naturally suggests an alternating strategy:
first improve \(Q(\mathbf{z})\) so that the bound becomes as tight as possible,
and then improve \(\theta\) so that the bound itself becomes larger.

That alternating logic is exactly what EM formalises.
In the next section, we will show how EM arises as an iterative procedure for
maximising the ELBO, and how, in the special case of PPCA, it recovers the same
solution that we previously obtained through exact Gaussian calculations.

\section{Expectation--Maximisation: Maximising the Lower Bound}

The fundamental difficulty in latent-variable models is the marginal
log-likelihood
\[
\log p(\mathbf{x};\theta)
=
\log \int p(\mathbf{x},\mathbf{z};\theta)\,d\mathbf{z},
\]
where the logarithm sits outside the marginalisation over the hidden variable
\(\mathbf{z}\).
This makes direct optimisation difficult, even when the joint distribution
\(p(\mathbf{x},\mathbf{z};\theta)\) itself has a relatively simple form.

The ELBO resolves this by introducing a lower bound
\[
\mathcal{L}(Q,\theta;\mathbf{x})
=
\mathbb{E}_{Q}\big[\log p(\mathbf{x},\mathbf{z};\theta)\big]
-
\mathbb{E}_{Q}\big[\log Q(\mathbf{z})\big],
\]
which depends on two coupled objects:
the auxiliary distribution \(Q(\mathbf{z})\) and the model parameters
\(\theta\).
This naturally suggests an alternating optimisation strategy:
\begin{quote}
\emph{if we hold one of these fixed, can we improve the other?}
\end{quote}
That alternating idea is precisely what EM formalises.
At a high level, the reason this is sensible is that once one of the two
quantities is fixed, the ELBO becomes an optimisation problem in the other
alone.
EM exploits this separable structure by alternating between improving the latent
distribution and improving the model parameters.

\subsection{The two EM steps}

Each iteration of EM consists of two stages.

\paragraph{E-step (Expectation step).}
Holding the current parameters \(\theta^{(t)}\) fixed, choose the distribution
\(Q(\mathbf{z})\) that makes the ELBO as tight as possible.
From the previous section, we know that the bound is tightest when
\[
Q(\mathbf{z}) = p(\mathbf{z}\mid \mathbf{x};\theta^{(t)}).
\]
Thus the E-step sets
\[
Q^{(t)}(\mathbf{z})
=
p(\mathbf{z}\mid \mathbf{x};\theta^{(t)}).
\]
In words, the E-step computes the posterior distribution over the latent
variables under the current parameter values.

This also clarifies an important point.
In standard EM, the E-step assumes that the posterior
\(p(\mathbf{z}\mid \mathbf{x};\theta^{(t)})\) is tractable enough to be computed
exactly.
That is the case in PPCA, which is why EM can be applied directly.
More generally, however, the posterior is often not available in closed form.
In those settings, we cannot set \(Q\) equal to the exact posterior and must
instead replace it with an approximation.
That is precisely the step that leads from EM to \emph{variational inference}.

It is worth noting that a tractable posterior does not automatically mean that
the entire optimisation problem can be solved in closed form.
For EM, the crucial requirement is that the posterior
\(p(\mathbf{z}\mid\mathbf{x};\theta)\), or at least the expectations needed under
it, can be computed at the current parameter values.
This is enough to carry out the E-step.
However, the resulting parameter updates may still need to be obtained
iteratively through the M-step rather than by a single direct analytic solution.

PPCA is special because several central quantities, including both the marginal
likelihood and the posterior, are analytically manageable, and the model also
admits a closed-form maximum-likelihood solution.
In many other models, even if the posterior remains tractable enough for EM, the
full parameter optimisation problem may still require an iterative procedure.

\paragraph{M-step (Maximisation step).}
With \(Q^{(t)}\) fixed, update the model parameters by maximising the ELBO with
respect to \(\theta\):
\[
\theta^{(t+1)}
=
\arg\max_{\theta}
\mathcal{L}(Q^{(t)},\theta;\mathbf{x}).
\]
Since the entropy term
\[
-\mathbb{E}_{Q^{(t)}}[\log Q^{(t)}(\mathbf{z})]
\]
does not depend on \(\theta\), this is equivalent to maximising
\[
\theta^{(t+1)}
=
\arg\max_{\theta}
\mathbb{E}_{Q^{(t)}}\big[\log p(\mathbf{x},\mathbf{z};\theta)\big].
\]
This quantity is often called the \emph{expected complete-data log-likelihood}.

Together, the two steps give the EM update pattern:
\[
\boxed{
\begin{aligned}
\text{E-step:}\quad
&Q^{(t)}(\mathbf{z}) = p(\mathbf{z}\mid \mathbf{x};\theta^{(t)}),\\[4pt]
\text{M-step:}\quad
&\theta^{(t+1)}
=
\arg\max_{\theta}
\mathbb{E}_{Q^{(t)}}\big[\log p(\mathbf{x},\mathbf{z};\theta)\big].
\end{aligned}}
\]

\paragraph{Why EM does not decrease the likelihood.}
The monotonicity of EM follows directly from the two optimisation steps and the
lower-bound property of the ELBO. First, after the E-step the bound is tight at
the current parameters:
\[
\mathcal{L}(Q^{(t)},\theta^{(t)};\mathbf{x})
=
\log p(\mathbf{x};\theta^{(t)}).
\]
Second, with $Q^{(t)}$ fixed, the M-step chooses parameters that do not decrease
that bound:
\[
\mathcal{L}(Q^{(t)},\theta^{(t+1)};\mathbf{x})
\ge
\mathcal{L}(Q^{(t)},\theta^{(t)};\mathbf{x}).
\]
Finally, because the ELBO is a lower bound on the log-likelihood,
\[
\log p(\mathbf{x};\theta^{(t+1)})
\ge
\mathcal{L}(Q^{(t)},\theta^{(t+1)};\mathbf{x}).
\]
Combining these relations shows that one full EM iteration cannot decrease the
observed-data log-likelihood. Figure~\ref{fig:em-elbo-geometry} shows the same
argument geometrically as alternating tightening and raising of the bound.

\input{figures/ch02/fig_ch02_04_em_elbo_geometry}

\subsection{Applying EM to PPCA}

In PPCA, the latent posterior is tractable, so EM can be carried out explicitly.
This makes PPCA an especially useful example: it shows how a general ELBO-based
algorithm reduces, in a simple linear-Gaussian model, to closed-form updates.

For PPCA, the posterior distribution of the latent variable is Gaussian. This
follows from the same linear-Gaussian structure used to derive the marginal
distribution, and the completing-the-square derivation is given in
Appendix~\ref{app:completing-square}. The posterior is
\[
p(\mathbf{z}\mid \mathbf{x})
=
\mathcal{N}\!\bigl(
\mathbf{z}\mid
\mathbf{M}^{-1}\mathbf{W}^\top(\mathbf{x}-\boldsymbol{\mu}),
\alpha^2\mathbf{M}^{-1}
\bigr),
\qquad
\mathbf{M}=\mathbf{W}^\top\mathbf{W}+\alpha^2\mathbf{I}.
\]
Figure~\ref{fig:ppca-four-probabilistic-views} brings together the four
probabilistic objects that have now appeared in PPCA: the prior $p(\mathbf{z})$,
the conditional $p(\mathbf{x}\mid\mathbf{z})$, the marginal $p(\mathbf{x})$,
and the posterior $p(\mathbf{z}\mid\mathbf{x})$.

\input{figures/ch02/fig_ch02_03_ppca_probability_views}

From this posterior, the E-step provides the expectations
\[
\mathbb{E}[\mathbf{z}]
=
\mathbf{M}^{-1}\mathbf{W}^\top(\mathbf{x}-\boldsymbol{\mu}),
\]
and
\[
\mathbb{E}[\mathbf{z}\mathbf{z}^\top]
=
\alpha^2\mathbf{M}^{-1}
+
\mathbb{E}[\mathbf{z}]\,\mathbb{E}[\mathbf{z}]^\top.
\]

The M-step then substitutes these expectations into the expected complete-data
log-likelihood and maximises with respect to the parameters
\((\mathbf{W},\boldsymbol{\mu},\alpha)\).
This leads to closed-form update equations for the PPCA parameters.

The important point here is not merely that PPCA can be fitted by EM, but that
the exact PPCA solution may be reinterpreted as the fixed point of an EM
procedure.
In other words, EM does not give a different model; it gives a different
\emph{view} of learning the same model.

More broadly, EM teaches us that learning in latent-variable models can be
understood as an alternation between
\begin{itemize}
    \item \textbf{inference}, where we estimate hidden variables given the data;
    \item \textbf{learning}, where we update the model parameters using those
    inferred latent variables.
\end{itemize}

\subsection{From PPCA to Regularised Latent Representations}
Although PPCA introduces a prior over latent variables, $p(\mathbf{z})=\mathcal{N}(\mathbf{0},\mathbf{I}),$
and defines a valid generative model, it does not yet provide the kind of
flexible and regularised inference framework that will become central in modern
latent-variable models such as the VAE.

For any individual observation \(\mathbf{x}\), the posterior $p(\mathbf{z}\mid \mathbf{x})$
describes which latent coordinates are plausible explanations for that data
point under the model.
If we average these posteriors over the empirical data distribution, we obtain
the \emph{aggregate posterior}: $\mathbb{E}_{q(\mathbf{x})}\!\left[p(\mathbf{z}\mid \mathbf{x})\right].$
This distribution describes how the dataset as a whole is represented in latent
space.

In PPCA, this posterior can be computed exactly because the model is linear and
Gaussian.
This is a strength of PPCA, but it also reveals the source of its limitation:
the clean mathematics depends on strong assumptions.
The latent-to-observed mapping is linear, the observation noise is Gaussian, and
posterior inference can be worked out analytically.

This distinction becomes important when we move beyond the linear-Gaussian world.
Modern generative models aim to keep the same latent-variable idea while making
the mapping from latent variables to observations much more flexible.
Once this flexibility is introduced, the simple closed-form calculations that
made PPCA so transparent are usually no longer available.
We therefore need a new framework that can learn expressive generative mappings,
infer latent representations from data, and keep the latent space organised in a
way that supports generation.

This is the role played by the Variational Autoencoder.
The VAE can be viewed as a nonlinear and variational extension of the PPCA
story: it preserves the idea of latent variables generating observations, but
replaces exact Gaussian calculations with a learnable inference-and-generation
framework.
The details of this framework, including the precise role of the variational
posterior and the latent-space regularisation term, are developed in the next
chapter.

Thus, the transition from PPCA to the VAE is not a move from a non-generative
model to a generative one.
PPCA is already generative.
Rather, the transition is from an analytically tractable linear-Gaussian
latent-variable model to a more flexible nonlinear latent-variable model in
which inference and generation must be learned together.

In the next chapter, the Variational Autoencoder develops this idea fully by
combining reconstruction, probabilistic inference, and latent-space
regularisation within a single variational objective.

\section*{Summary and References}
\addcontentsline{toc}{section}{Summary and References}

This chapter introduced Probabilistic PCA (PPCA) as the first 
probabilistic latent-variable model in the book.
It extended the deterministic picture of PCA by introducing a latent prior, a
probabilistic decoder, a marginal likelihood for learning, and a posterior
distribution for inference.

Because PPCA is linear and Gaussian, all of its central quantities remain
analytically tractable.
This allowed us to derive the marginal likelihood explicitly, interpret its
covariance as the sum of a structured low-rank component and isotropic noise,
and view PPCA as a rare model in which prior, likelihood, marginal, and
posterior can all be written in closed form.

We then considered three perspectives on learning the model: direct analytic
maximum likelihood, gradient-based optimisation, and EM.
Among these, EM was especially important because it exposed the deeper structure
of latent-variable learning as an alternation between inference over hidden
variables and optimisation of model parameters.
This in turn motivated Jensen’s inequality and the ELBO, which generalise the
PPCA story to models where exact marginalisation is no longer possible.

Finally, we identified an important limitation of PPCA: although it introduces a
prior over latent variables, it does not explicitly regularise the aggregate
inferred latent space to align with that prior.
This points directly to the next chapter, where the Variational Autoencoder
(VAE) will combine reconstruction, probabilistic inference, and latent-space
regularisation within a unified variational objective.

\bigskip

The material in this chapter draws on classical work on probabilistic PCA,
latent-variable models, and maximum-likelihood inference.
The original formulation of PPCA was developed by Tipping and
Bishop~\cite{tipping1999ppca}, while the EM algorithm was classically introduced
by Dempster, Laird, and Rubin~\cite{dempster1977em}.
For readers who want a modern and more pedagogical treatment of these ideas,
especially Gaussian latent-variable models, EM, and their probabilistic
interpretation, useful references include Bishop~\cite{bishop2006prml} and
Murphy~\cite{murphy2012ml}.

%% file: figures/ch02/chapter2_ppca_at_a_glance.tex
\begin{center}
\begin{tikzpicture}[
  >=Latex,
  every node/.style={font=\small},
  box/.style={
    draw,
    rounded corners=2pt,
    align=center,
    inner sep=7pt,
    minimum height=1.20cm,
    text width=3.75cm
  },
  lowerbox/.style={
    draw,
    rounded corners=2pt,
    align=center,
    inner sep=7pt,
    minimum height=1.12cm,
    text width=3.75cm
  },
  flow/.style={->,thick}
]

\node[box] (prior) at (0,0) {
  \textbf{Latent prior}\\[2pt]
  $\mathbf{z}\sim\mathcal{N}(\mathbf{0},\mathbf{I})$
};

\node[box] (gen) at (5.15,0) {
  \textbf{Conditional generation}\\[2pt]
  $p(\mathbf{x}\mid\mathbf{z})
  =\mathcal{N}(\mathbf{W}\mathbf{z}+\boldsymbol{\mu},
  \alpha^2\mathbf{I})$
};

\node[box] (marginal) at (10.30,0) {
  \textbf{Marginal data model}\\[2pt]
  $p(\mathbf{x})
  =\mathcal{N}\!\bigl(
  \boldsymbol{\mu},
  \mathbf{W}\mathbf{W}^{\top}+\alpha^2\mathbf{I}
  \bigr)$
};

\node[lowerbox] (posterior) at (5.15,-2.75) {
  \textbf{Posterior inference}\\[2pt]
  $p(\mathbf{z}\mid\mathbf{x})
  \propto p(\mathbf{x}\mid\mathbf{z})p(\mathbf{z})$\\[-1pt]
  exact Gaussian\\in PPCA
};

\node[lowerbox] (learning) at (10.30,-2.75) {
  \textbf{Parameter learning}\\[2pt]
  maximise $\log p(\mathbf{x})$\\[-1pt]
  directly or via ELBO / EM
};

\draw[flow] (prior) -- (gen);
\draw[flow] (gen) -- (marginal);
\draw[flow] (gen) -- node[right] {\scriptsize Bayes} (posterior);
\draw[flow] (marginal) -- node[right] {\scriptsize likelihood} (learning);
\draw[flow] (posterior) -- node[above] {\scriptsize EM} (learning);

\node[font=\normalsize\bfseries] at (5.15,1.45) {PPCA at a glance};

\end{tikzpicture}
\end{center}

%% file: figures/ch02/fig_ch02_01_ppca_generative_story.tex
\begin{figure}[H]
\centering
\begin{minipage}[t]{0.45\textwidth}
  \vspace{0pt}\centering
  \includegraphics[width=\linewidth]{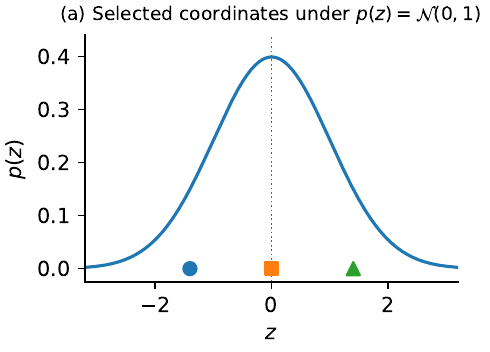}
\end{minipage}\hfill
\begin{minipage}[t]{0.45\textwidth}
  \vspace{0pt}\centering
  \includegraphics[width=\linewidth]{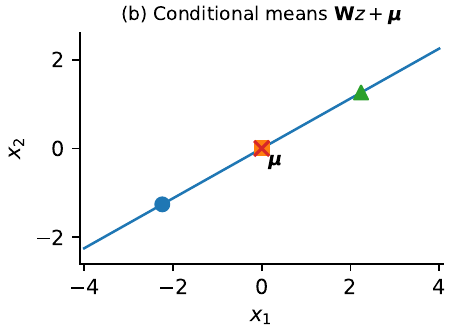}
\end{minipage}

\vspace{0.8em}

\begin{minipage}[t]{0.45\textwidth}
  \vspace{0pt}\centering
  \includegraphics[width=\linewidth]{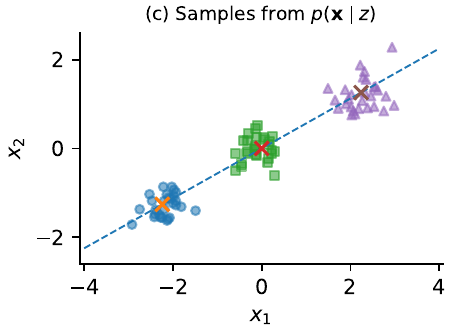}
\end{minipage}\hfill
\begin{minipage}[t]{0.45\textwidth}
  \vspace{0pt}
  \caption{\textbf{From latent coordinates to probabilistic observations.}
  Selected latent coordinates are shown under the prior $p(z)=\mathcal{N}(0,1)$,
  mapped to conditional means $\mathbf{W}z+\boldsymbol{\mu}$ in observation space,
  and then used as the centres of isotropic conditional distributions
  $p(\mathbf{x}\mid z)=\mathcal{N}(\mathbf{W}z+\boldsymbol{\mu},\alpha^2\mathbf{I})$.
  Panel (c) shows samples conditional on the three selected latent values, not
  the marginal $p(\mathbf{x})$.}
  \label{fig:ppca-generative-story}
\end{minipage}
\end{figure}

%% file: figures/ch02/fig_ch02_02_ppca_covariance_decomposition.tex
\begin{figure}[H]
\centering
\begin{minipage}[t]{0.48\textwidth}
  \vspace{0pt}\centering
  \includegraphics[width=\linewidth]{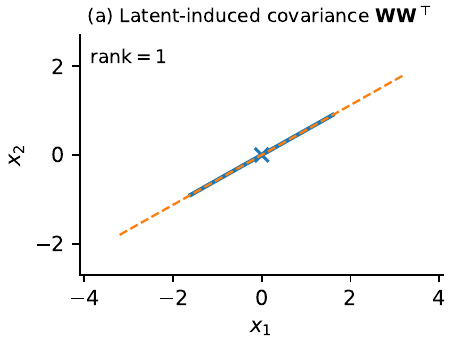}
\end{minipage}\hfill
\begin{minipage}[t]{0.48\textwidth}
  \vspace{0pt}\centering
  \includegraphics[width=\linewidth]{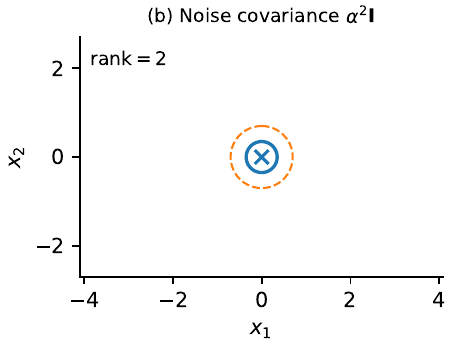}
\end{minipage}

\vspace{0.8em}

\begin{minipage}[t]{0.48\textwidth}
  \vspace{0pt}\centering
  \includegraphics[width=\linewidth]{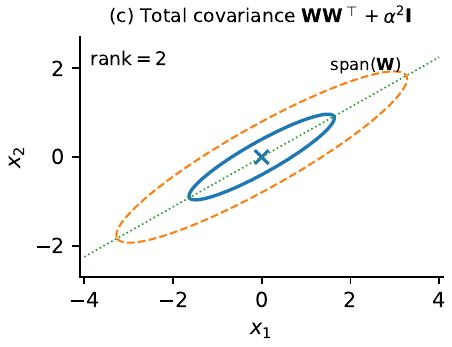}
\end{minipage}\hfill
\begin{minipage}[t]{0.48\textwidth}
  \vspace{0pt}
  \caption{\textbf{Covariance decomposition in PPCA.}
  For the illustrated case $k=1$ and $d=2$, $\mathbf{W}\mathbf{W}^{\top}$ is rank one and contributes
  variance only along $\mathrm{span}(\mathbf{W})$, while $\alpha^2\mathbf{I}$ is
  isotropic and full rank. Their sum is the full-rank marginal covariance,
  combining variation along and around the latent subspace. Solid and dashed
  geometry show one- and two-standard-deviation extents.}
  \label{fig:ppca-covariance-decomposition}
\end{minipage}
\end{figure}

%% file: figures/ch02/fig_ch02_04_em_elbo_geometry.tex
\begin{figure}[H]
\centering
\begin{tikzpicture}[x=1cm,y=1cm,>=Latex,font=\small]
\def\thetacur{2.0}
\def\thetanext{2.714}

% Common schematic functions:
% log likelihood: 3 - 0.25(theta-3)^2
% post-E ELBO: likelihood - 0.10(theta-2)^2
% loose pre-E ELBO: likelihood - 0.60 - 0.15(theta-1.5)^2

% Panel (a): E-step
\begin{scope}[shift={(0,0)}]
  \draw[->] (0,0) -- (5.0,0) node[right] {$\theta$};
  \draw[->] (0,0) -- (0,3.35);

  \draw[thick] plot[domain=0.3:4.7,samples=100]
    (\x,{3 - 0.25*(\x-3)^2});
  \node[anchor=west] at (3.25,3.02) {$\log p(\mathbf{x};\theta)$};

  \draw[dashed] plot[domain=0.3:4.7,samples=100]
    (\x,{3 - 0.25*(\x-3)^2 - 0.60 - 0.15*(\x-1.5)^2});
  \node[anchor=west] at (0.55,0.92) {$\mathcal{L}(Q,\theta;\mathbf{x})$};

  \draw[thick,dash dot] plot[domain=0.3:4.7,samples=100]
    (\x,{3 - 0.25*(\x-3)^2 - 0.10*(\x-2)^2});
  \node[anchor=west] at (3.15,2.12)
    {$\mathcal{L}(Q^{(t)},\theta;\mathbf{x})$};

  \draw[densely dotted] (\thetacur,0) -- (\thetacur,2.75);
  \fill (\thetacur,2.75) circle (1.4pt);
  \node[below] at (\thetacur,0) {$\theta^{(t)}$};

  \draw[->] (1.45,1.28) .. controls (1.35,1.80) and (1.55,2.25) .. (1.90,2.62);
  \node[align=center] at (0.95,1.80) {\scriptsize update\\[-1pt]\scriptsize $Q$};

  \node[font=\small\bfseries] at (2.5,3.72) {(a) E-step: tighten};
\end{scope}

% Panel (b): M-step
\begin{scope}[shift={(7.0,0)}]
  \draw[->] (0,0) -- (5.0,0) node[right] {$\theta$};
  \draw[->] (0,0) -- (0,3.35);

  \draw[thick] plot[domain=0.3:4.7,samples=100]
    (\x,{3 - 0.25*(\x-3)^2});
  \node[anchor=west] at (3.25,3.02) {$\log p(\mathbf{x};\theta)$};

  % Exactly the same post-E-step bound as in panel (a)
  \draw[thick,dash dot] plot[domain=0.3:4.7,samples=100]
    (\x,{3 - 0.25*(\x-3)^2 - 0.10*(\x-2)^2});
  \node[anchor=west] at (3.15,2.12)
    {$\mathcal{L}(Q^{(t)},\theta;\mathbf{x})$};

  \draw[densely dotted] (\thetacur,0) -- (\thetacur,2.75);
  \draw[densely dotted] (\thetanext,0) -- (\thetanext,2.929);
  \fill (\thetacur,2.75) circle (1.4pt);
  \fill (\thetanext,2.929) circle (1.4pt);

  \node[below,xshift=-4pt] at (\thetacur,0) {$\theta^{(t)}$};
  \node[below,xshift=7pt] at (\thetanext,0) {$\theta^{(t+1)}$};

  \draw[->,thick] (2.08,2.78) -- (2.62,2.92);

  \node[font=\small\bfseries] at (2.5,3.72) {(b) M-step: raise};
\end{scope}

\node[align=center] at (6.0,-0.92) {
$\displaystyle
\log p(\mathbf{x};\theta^{(t+1)})
\ge
\mathcal{L}(Q^{(t)},\theta^{(t+1)};\mathbf{x})
\ge
\mathcal{L}(Q^{(t)},\theta^{(t)};\mathbf{x})
=
\log p(\mathbf{x};\theta^{(t)})$};

\end{tikzpicture}
\caption{\textbf{EM as alternating optimisation of the ELBO.}
In the E-step, $Q$ is set to the exact posterior, making the ELBO tight at the
current parameters. Holding this bound fixed, the M-step updates the parameters
to raise it. The inequality below shows why one full iteration cannot decrease
the observed-data log-likelihood. The curves are schematic.}
\label{fig:em-elbo-geometry}
\end{figure}

%% file: figures/ch02/fig_ch02_03_ppca_probability_views.tex
\begin{figure}[H]
\centering
\begin{minipage}[t]{0.45\textwidth}
\centering\includegraphics[width=\linewidth]{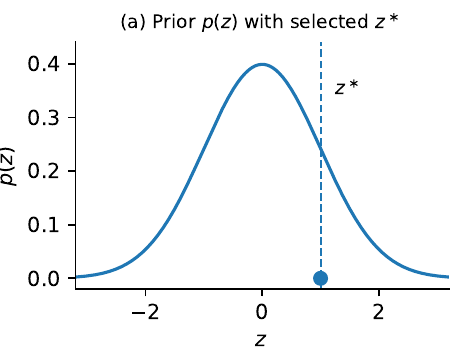}
\end{minipage}\hfill
\begin{minipage}[t]{0.45\textwidth}
\centering\includegraphics[width=\linewidth]{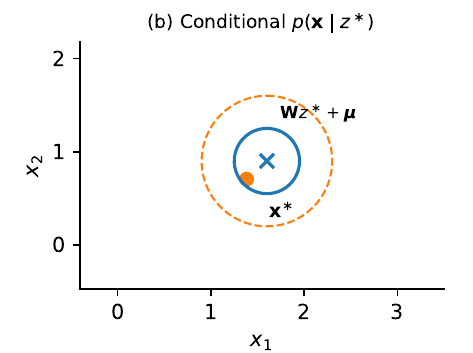}
\end{minipage}

\vspace{0.4em}

\begin{minipage}[t]{0.45\textwidth}
\centering\includegraphics[width=\linewidth]{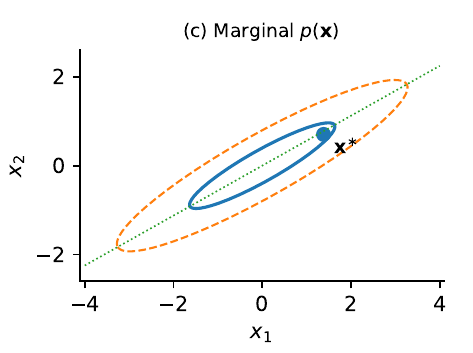}
\end{minipage}\hfill
\begin{minipage}[t]{0.45\textwidth}
\centering\includegraphics[width=\linewidth]{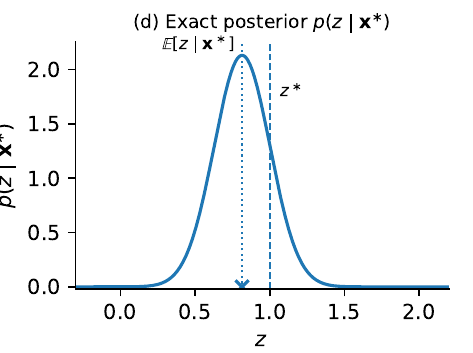}
\end{minipage}
\caption{\textbf{Four probabilistic views of the same PPCA model.}
The prior $p(z)$ describes latent coordinates; fixing $z^\ast$ gives
$p(\mathbf{x}\mid z^\ast)$; integrating over $z$ gives the marginal
$p(\mathbf{x})$; and observing $\mathbf{x}^\ast$ gives the posterior
$p(z\mid\mathbf{x}^\ast)$. The true generating $z^\ast$ is shown in the
posterior panel; observation noise means it need not equal the posterior mean.
In PPCA, all four distributions are Gaussian and available analytically.}
\label{fig:ppca-four-probabilistic-views}
\end{figure}

%% file: chapters/cht4_new.tex
\chapter{The Variational Autoencoder: From Probabilistic Latent Variables to Variational Inference}

\section*{Overview}
\addcontentsline{toc}{section}{Overview}

The Variational Autoencoder (VAE) is one of the foundational models of modern
deep generative learning. It occupies an important position in this book because
it brings together the two strands developed in the previous chapters:
autoencoders as reconstruction-based representation learners, and probabilistic
PCA as a latent-variable generative model.

In a deterministic autoencoder, each input is mapped to a single latent code and
then reconstructed. In PPCA, latent variables are treated probabilistically, but
the model remains linear and Gaussian. The VAE combines these ideas in a more
flexible nonlinear setting: the encoder produces a distribution over latent
variables, the decoder defines a conditional distribution over observations,
and learning proceeds by optimising a variational lower bound rather than an
exact marginal likelihood.

This makes the VAE a natural bridge from classical latent-variable models to
modern deep generative modelling. It retains the probabilistic structure of
PPCA---a prior, a likelihood, a posterior, and a marginal likelihood---but
replaces exact Gaussian inference with a learned approximate posterior. The
Evidence Lower Bound (ELBO), introduced in the previous chapter, now becomes a
practical training objective that balances reconstruction accuracy with
regularisation of the latent space.

Although VAEs are no longer the newest dominant generative paradigm, they remain
conceptually central. They show how probabilistic inference, neural networks,
latent representations, and gradient-based optimisation can be combined in a
single end-to-end generative model. This chapter therefore serves as the first
major step from analytically tractable latent-variable models toward the broader
deep generative frameworks developed later in the book.

\section*{Concept Map}
\addcontentsline{toc}{section}{Concept Map}

This chapter traces the move from deterministic autoencoders to
variational latent-variable generative models.

\begin{itemize}

    \item[\(\rightarrow\)] \textbf{We begin by turning the deterministic autoencoder into a probabilistic model.}  
    Instead of mapping each observation to a single latent code, the encoder now produces a distribution \(q_{\phi}(\mathbf{z}\mid\mathbf{x})\), while the decoder defines a conditional distribution \(p_{\theta}(\mathbf{x}\mid\mathbf{z})\).  
    This gives the model both uncertainty and genuine generative capability.

    \item[\(\rightarrow\)] \textbf{We then derive the learning objective from the latent-variable model itself.}  
    Starting from the intractable marginal likelihood \(p_{\theta}(\mathbf{x})\), we use the ELBO framework developed in the previous chapter and specialise it to the VAE setting.  
    This reveals the familiar decomposition into a reconstruction term and a regularisation term, showing how accurate data fitting is balanced against latent-space structure.

    \item[\(\rightarrow\)] \textbf{Finally, we explain how the VAE is trained and what this achieves.}  
    The variational posterior \(q_{\phi}(\mathbf{z}\mid\mathbf{x})\) plays the role of a learned approximation to the true posterior, and the reparameterisation trick makes this stochastic inference process differentiable and trainable with gradient-based optimisation.

\end{itemize}

\medskip
\noindent The main probabilistic relationships of the VAE are summarised visually below.
\input{figures/ch03/chapter3_vae_at_a_glance}
\medskip

\section{From Autoencoder to Variational Autoencoder}

In the previous chapters, two ideas were developed in parallel.
On the one hand, the autoencoder showed how data can be compressed into a
lower-dimensional latent representation, that is, a compact internal
description or feature code from which the input can be reconstructed.
On the other hand, probabilistic PCA (PPCA) showed how a latent-variable model
can turn such a representation into a genuinely probabilistic generative
process.
The Variational Autoencoder (VAE)~\cite{kingma2014autoencoding,rezende2014stochastic} brings these two threads together.

A standard autoencoder has the familiar deterministic form
\[
\mathbf{x}
\;\xrightarrow{\text{encoder}}\;
\mathbf{z}
\;\xrightarrow{\text{decoder}}\;
\hat{\mathbf{x}}.
\]
Given an input \(\mathbf{x}\), the encoder produces a single latent code
\(\mathbf{z}\), and the decoder maps that code back to a single reconstruction
\(\hat{\mathbf{x}}\).
This structure is useful for compression and representation learning, but it has
an important limitation: the latent code is a fixed point rather than a
probabilistic object.
As a result, the model does not naturally express uncertainty, and it does not
provide a principled latent-variable generative framework.

The VAE addresses this by replacing the deterministic latent code with a
\emph{distribution} over latent variables.
This mirrors the earlier shift from PCA to PPCA: there too, a deterministic
low-dimensional description was turned into a probabilistic latent-variable
model by introducing uncertainty.
Instead of encoding \(\mathbf{x}\) as one precise point in latent space, the
VAE encodes it as a probability distribution that reflects uncertainty about
which latent representation best explains the observation.
This is the key conceptual shift:
\begin{quote}
\emph{the latent representation is no longer a single code, but a distribution
from which codes may be sampled.}
\end{quote}

\subsection{Making the encoder probabilistic}

In a deterministic autoencoder, the encoder is a function
\[
f_{\text{enc}}:\mathbf{x}\mapsto \mathbf{z}.
\]
For each observation \(\mathbf{x}\), it returns exactly one latent vector
\(\mathbf{z}\).

In a VAE, the encoder instead defines a tractable conditional distribution over
latent variables,
\[
q_{\phi}(\mathbf{z}\mid \mathbf{x}),
\]
which serves as a learned approximation to the true posterior.
This distribution is often called the \emph{variational posterior} or
\emph{approximate posterior}.
Its role is to describe which latent variables are plausible given the observed
data point \(\mathbf{x}\).

The word \emph{posterior} is used deliberately.
In the previous PPCA chapter, we saw that in a latent-variable model the true
posterior is
\[
p_{\theta}(\mathbf{z}\mid \mathbf{x}),
\]
the distribution over latent variables after observing \(\mathbf{x}\).
In the VAE, this true posterior is generally intractable.
The main reason is that the encoder and decoder are usually modelled by neural
networks with multiple nonlinear layers, so the posterior no longer retains the
simple linear-Gaussian form that made PPCA analytically solvable.
We therefore introduce the tractable approximation
\(q_{\phi}(\mathbf{z}\mid \mathbf{x})\) instead.
For now, the key structural change is simply this:
the encoder no longer outputs a single vector, but the parameters of a
distribution.

In practice, the encoder is typically implemented by a neural network that maps
\(\mathbf{x}\) to the parameters of a Gaussian distribution.
Most commonly, we assume
\[
q_{\phi}(\mathbf{z}\mid \mathbf{x})
=
\mathcal{N}\!\bigl(
\mathbf{z}\mid
\boldsymbol{\mu}_{\phi}(\mathbf{x}),
\operatorname{diag}(\boldsymbol{\sigma}_{\phi}^2(\mathbf{x}))
\bigr),
\]
or, in a simplified isotropic presentation,
\[
q_{\phi}(\mathbf{z}\mid \mathbf{x})
=
\mathcal{N}\!\bigl(
\mathbf{z}\mid
\boldsymbol{\mu}_{\phi}(\mathbf{x}),
\boldsymbol{\sigma}_{\phi}^2(\mathbf{x})\mathbf{I}
\bigr).
\]
Thus the encoder network does not produce \(\mathbf{z}\) directly.
Instead, it produces the mean and variance parameters that specify a Gaussian
distribution in latent space.

Conceptually, the encoder now has the form
\[
\mathbf{x}
\;\longrightarrow\;
\bigl(\boldsymbol{\mu}_{\phi}(\mathbf{x}),
\boldsymbol{\sigma}_{\phi}(\mathbf{x})\bigr)
\;\longrightarrow\;
q_{\phi}(\mathbf{z}\mid \mathbf{x}).
\]
A latent code is then obtained by sampling
\[
\mathbf{z}\sim q_{\phi}(\mathbf{z}\mid \mathbf{x}).
\]

For a fixed input \(\mathbf{x}\), the contrast with a deterministic
autoencoder is therefore direct: the autoencoder returns one latent point,
whereas the VAE encoder defines a distribution of plausible latent values from
which different codes may be sampled. Figure~\ref{fig:vae-ae-versus-posterior}
shows this point-versus-distribution shift geometrically.

\input{figures/ch03/fig_ch03_01_ae_vs_vae_latent_representation}

\begin{tcolorbox}[
  colback=blue!6,
  colframe=blue!55!black,
  title={Why choose a tractable Gaussian approximate posterior?},
  boxrule=0.8pt,
  arc=2mm,
  left=2mm,
  right=2mm,
  top=1mm,
  bottom=1mm
]

The true posterior \(p_{\theta}(\mathbf{z}\mid\mathbf{x})\) in a VAE is usually
too complicated to compute exactly, so we replace it with a simpler family
\(q_{\phi}(\mathbf{z}\mid\mathbf{x})\) that can be represented and learned in
practice. A Gaussian choice is especially convenient for two main reasons.

\begin{itemize}
    \item It is simple to represent.
    A neural network only needs to output a mean and a variance, giving a
    compact and differentiable way to describe the latent distribution.

    \item It is easy to work with computationally.  
    We can sample from it easily, and many of the mathematical quantities that
    will appear later in the learning objective become much easier to evaluate
    when Gaussian distributions are used.
\end{itemize}
What tractability means here is that the approximate posterior is chosen
to be simple enough to handle during learning.
It does not mean that the entire VAE becomes analytically solvable again.
Unlike PPCA, the decoder is still nonlinear, the exact data likelihood remains
intractable, and we will still need a surrogate objective and gradient-based
optimisation to train the model.

\end{tcolorbox}

\subsection{Making the decoder probabilistic}

A similar shift occurs on the decoding side.
In a deterministic autoencoder, the decoder is a function
\[
f_{\text{dec}}:\mathbf{z}\mapsto \hat{\mathbf{x}},
\]
which produces a single reconstruction \(\hat{\mathbf{x}}\) for each latent
input.

In a VAE, the decoder instead defines a conditional distribution
\[
p_{\theta}(\mathbf{x}\mid \mathbf{z}),
\]
which describes how an observation is generated from a latent variable.
This is the generative part of the model.

Again, the decoder is typically parameterised by a neural network.
Given \(\mathbf{z}\), the decoder outputs the parameters of a distribution in
data space.
For continuous data, a common choice is a Gaussian distribution such as
\[
p_{\theta}(\mathbf{x}\mid \mathbf{z})
=
\mathcal{N}\!\bigl(
\mathbf{x}\mid
\boldsymbol{\mu}_{\theta}(\mathbf{z}),
\sigma^2 \mathbf{I}
\bigr),
\]
while for binary data one often uses a Bernoulli likelihood.
The exact choice depends on the type of data being modelled, but the underlying
principle is always the same:
\begin{quote}
\emph{the decoder does not output a single deterministic reconstruction, but a
distribution over possible observations given the latent variable.}
\end{quote}

In practice, one often uses the mean or mode of this distribution as the
model's representative reconstruction, but conceptually the decoder defines a
full conditional probability law rather than a single point prediction.
So, whereas the encoder maps data to a distribution in latent space, the
decoder maps latent variables to a distribution in data space.

\subsection{Prior and Learning Difficulty}

\paragraph{The latent prior.} To complete the probabilistic latent-variable model, we also introduce a prior
distribution over latent variables:
\[
p(\mathbf{z}).
\]
As in PPCA, the standard VAE usually begins with a simple standard Gaussian
prior,
\[
p(\mathbf{z})=\mathcal{N}(\mathbf{0},\mathbf{I}).
\]

This prior plays two roles.

First, it provides the starting point for generation:
after training, we may sample
\[
\mathbf{z}\sim p(\mathbf{z}),
\]
and then generate an observation by sampling from the decoder distribution
\[
\mathbf{x}\sim p_{\theta}(\mathbf{x}\mid \mathbf{z}).
\]

Second, it provides the reference distribution against which the inferred latent
distribution will later be regularised.
This is an important step beyond PPCA: there, the prior was present, but the
model did not explicitly encourage the inferred latent representations to align
with it.
In the VAE, that alignment will become part of the learning objective itself.

In this way, the prior gives the latent space a global structure, while the
decoder turns points in that space into data.

\paragraph{Why this creates a new difficulty.}

Putting these pieces together, the VAE consists of three probabilistic
components:
\[
p(\mathbf{z}),
\qquad
q_{\phi}(\mathbf{z}\mid \mathbf{x}),
\qquad
p_{\theta}(\mathbf{x}\mid \mathbf{z}),
\]
namely a prior over latent variables, an approximate posterior for inference,
and a decoder likelihood for generation.

This gives the VAE a structure that resembles both the autoencoder and PPCA,
but is richer than either one.
Like the autoencoder, it has an encoder--decoder architecture.
Like PPCA, it is a probabilistic latent-variable model.
But unlike PPCA, the encoder and decoder are now typically nonlinear neural
networks, which greatly increase modelling power while also destroying the
analytic tractability we previously relied on.

In PPCA, the linear-Gaussian structure allowed quantities such as the marginal
likelihood \(p(\mathbf{x})\) and the posterior \(p(\mathbf{z}\mid \mathbf{x})\)
to be derived explicitly.
In the VAE, these quantities are generally no longer available in closed form. So although the VAE has a clean probabilistic structure, its learning problem is
no longer analytically solvable in the way PPCA was.
This brings us directly to the central question of the chapter:
\begin{quote}
\emph{If the true marginal likelihood and posterior are intractable, what
objective should we optimise instead?}
\end{quote}

The answer is the same general principle introduced in the previous chapter:
the Evidence Lower Bound (ELBO).
In the next section, we specialise that principle to the VAE and derive the
objective that makes the model trainable.

\section{From Intractable Likelihood to the ELBO}

With the probabilistic structure of the VAE in place, we can now ask how the
model should be learned.
As in PPCA, the natural principle is maximum likelihood: we would like to choose
the parameters so that the model assigns high probability to the observed data.

For a single observation \(\mathbf{x}\), the relevant quantity is the marginal
likelihood
\[
p_{\theta}(\mathbf{x})
=
\int p_{\theta}(\mathbf{x},\mathbf{z})\,d\mathbf{z}
=
\int p_{\theta}(\mathbf{x}\mid \mathbf{z})\,p(\mathbf{z})\,d\mathbf{z},
\]
and over a dataset \(\{\mathbf{x}^{(i)}\}\) the learning objective becomes
\[
\max_{\theta}\sum_i \log p_{\theta}(\mathbf{x}^{(i)}).
\]

This has the same formal appearance as the PPCA objective from the previous
chapter, but there is now a crucial difference.
In PPCA, the linear-Gaussian structure allowed the integral over
\(\mathbf{z}\) to be computed exactly.
In the VAE, the decoder \(p_{\theta}(\mathbf{x}\mid\mathbf{z})\) is typically
parameterised by a nonlinear neural network, which destroys the simple
linear-Gaussian form that previously allowed exact marginalisation.
As a result, the marginal likelihood
\[
\log p_{\theta}(\mathbf{x})
=
\log \int p_{\theta}(\mathbf{x}\mid \mathbf{z})\,p(\mathbf{z})\,d\mathbf{z}
\]
is generally intractable.

The same difficulty immediately appears in the posterior:
\[
p_{\theta}(\mathbf{z}\mid \mathbf{x})
=
\frac{p_{\theta}(\mathbf{x}\mid \mathbf{z})\,p(\mathbf{z})}
{p_{\theta}(\mathbf{x})}.
\]
The numerator can be written down, but normalising it requires the same
intractable marginal likelihood \(p_{\theta}(\mathbf{x})\).
So in the VAE, both the marginal likelihood and the true posterior are
typically unavailable in closed form.

This is precisely the point at which the variational machinery from the
previous chapter becomes essential.
Rather than optimising the intractable log-likelihood directly, we optimise a
tractable lower bound: the Evidence Lower Bound (ELBO).

\subsection{Introducing the variational posterior}

In Chapter 2, the ELBO was derived by introducing an auxiliary distribution
\(Q(\mathbf{z})\) over the latent variable.
In the VAE, that auxiliary distribution now becomes a learned, parameterised
approximation to the true posterior:
\[
Q(\mathbf{z})
\;\longrightarrow\;
q_{\phi}(\mathbf{z}\mid \mathbf{x}),
\]
where \(q_{\phi}(\mathbf{z}\mid \mathbf{x})\) is the encoder network introduced
in the previous section.

This substitution is the key step that turns the general ELBO framework into
the VAE objective.
Instead of trying to compute the intractable true posterior
\(p_{\theta}(\mathbf{z}\mid \mathbf{x})\), we use the tractable approximation
\(q_{\phi}(\mathbf{z}\mid \mathbf{x})\) and optimise a lower bound on
\(\log p_{\theta}(\mathbf{x})\).

Applying the ELBO construction gives
\[
\log p_{\theta}(\mathbf{x})
\geq
\mathbb{E}_{q_{\phi}(\mathbf{z}\mid \mathbf{x})}
\!\left[
\log \frac{p_{\theta}(\mathbf{x},\mathbf{z})}
{q_{\phi}(\mathbf{z}\mid \mathbf{x})}
\right].
\]
Thus the VAE objective for a single data point is
\[
\mathcal{L}(\mathbf{x};\theta,\phi)
=
\mathbb{E}_{q_{\phi}(\mathbf{z}\mid \mathbf{x})}
\!\left[
\log \frac{p_{\theta}(\mathbf{x},\mathbf{z})}
{q_{\phi}(\mathbf{z}\mid \mathbf{x})}
\right].
\]

This is the ELBO specialised to the VAE setting.
It depends on both sets of parameters:
\begin{itemize}
    \item \(\theta\), the parameters of the decoder network \(p_{\theta}(\mathbf{x}\mid \mathbf{z})\);
    \item \(\phi\), the parameters of the encoder network \(q_{\phi}(\mathbf{z}\mid \mathbf{x})\).
\end{itemize}
Training the VAE means maximising this lower bound jointly with respect to both
parts of the model.

We will now rewrite the same objective in three equivalent forms.
Each form highlights a different aspect of the learning problem:
the latent-variable structure, the gap to the exact likelihood, and the
practical balance between reconstruction and regularisation.

\subsection{Three Equivalent Forms of the ELBO}

\paragraph{First form of the ELBO: joint-density form.} The most compact form of the ELBO is
\[
\mathcal{L}(\mathbf{x};\theta,\phi)
=
\mathbb{E}_{q_{\phi}(\mathbf{z}\mid \mathbf{x})}
\!\left[
\log \frac{p_{\theta}(\mathbf{x},\mathbf{z})}
{q_{\phi}(\mathbf{z}\mid \mathbf{x})}
\right].
\]
Expanding the logarithm makes its structure more explicit:
\[
\mathcal{L}(\mathbf{x};\theta,\phi)
=
\mathbb{E}_{q_{\phi}(\mathbf{z}\mid \mathbf{x})}
[\log p_{\theta}(\mathbf{x},\mathbf{z})]
-
\mathbb{E}_{q_{\phi}(\mathbf{z}\mid \mathbf{x})}
[\log q_{\phi}(\mathbf{z}\mid \mathbf{x})].
\]

This form is conceptually useful because it displays exactly the same latent-variable
structure as in the general ELBO derivation from the previous chapter.
The first term asks the sampled latent variable to be compatible with the joint
model \(p_{\theta}(\mathbf{x},\mathbf{z})\), while the second term comes from
the variational distribution itself.

It also makes clear what the VAE is doing.
The model wants to explain the observation \(\mathbf{x}\) through a latent
variable \(\mathbf{z}\), but because the exact posterior is unavailable, it
optimises a lower bound that depends on the approximate posterior
\(q_{\phi}(\mathbf{z}\mid \mathbf{x})\) instead.

\paragraph{Second form of the ELBO: exact likelihood minus posterior mismatch.}

Using Bayes' rule,
\[
p_{\theta}(\mathbf{x},\mathbf{z})
=
p_{\theta}(\mathbf{z}\mid \mathbf{x})\,p_{\theta}(\mathbf{x}),
\]
we can rewrite the ELBO as
\[
\begin{aligned}
\mathcal{L}(\mathbf{x};\theta,\phi)
&=
\mathbb{E}_{q_{\phi}(\mathbf{z}\mid \mathbf{x})}
\!\left[
\log \frac{p_{\theta}(\mathbf{z}\mid \mathbf{x})\,p_{\theta}(\mathbf{x})}
{q_{\phi}(\mathbf{z}\mid \mathbf{x})}
\right] \\[4pt]
&=
\log p_{\theta}(\mathbf{x})
-
\mathbb{E}_{q_{\phi}(\mathbf{z}\mid \mathbf{x})}
\!\left[
\log \frac{q_{\phi}(\mathbf{z}\mid \mathbf{x})}
{p_{\theta}(\mathbf{z}\mid \mathbf{x})}
\right].
\end{aligned}
\]

\begin{tcolorbox}[
  colback=blue!6,
  colframe=blue!55!black,
  title={KL divergence},
  boxrule=0.8pt,
  arc=2mm,
  left=2mm,
  right=2mm,
  top=1mm,
  bottom=1mm
]
For two probability distributions \(q\) and \(p\), the
\emph{Kullback--Leibler (KL) divergence} from \(q\) to \(p\) is defined as
\[
\mathrm{KL}(q\,\|\,p)
=
\mathbb{E}_{q}\!\left[\log \frac{q}{p}\right].
\]
It measures how different \(q\) is from \(p\) when expectations are taken under
\(q\).

Two basic facts are especially important here:
\begin{itemize}
    \item \(\mathrm{KL}(q\,\|\,p)\ge 0\);
    \item \(\mathrm{KL}(q\,\|\,p)=0\) if and only if \(q=p\) almost everywhere.
\end{itemize}

So KL divergence behaves like a non-negative measure of mismatch between two
distributions.
It is not a true distance in the geometric sense, because it is asymmetric, but
it is extremely useful in probabilistic modelling.
\end{tcolorbox}

Recognising the second term as a KL divergence gives
\[
\boxed{
\mathcal{L}(\mathbf{x};\theta,\phi)
=
\log p_{\theta}(\mathbf{x})
-
\mathrm{KL}\!\bigl(
q_{\phi}(\mathbf{z}\mid \mathbf{x})
\,\|\,\,
p_{\theta}(\mathbf{z}\mid \mathbf{x})
\bigr).
}
\]

This form shows that the ELBO differs from the true log-likelihood only by the
divergence between the approximate posterior and the true posterior.
Since KL divergence is always non-negative,
\[
\mathrm{KL}\!\bigl(
q_{\phi}(\mathbf{z}\mid \mathbf{x})
\,\|\,\,
p_{\theta}(\mathbf{z}\mid \mathbf{x})
\bigr)\ge 0,
\]
it follows immediately that
\[
\mathcal{L}(\mathbf{x};\theta,\phi)\le \log p_{\theta}(\mathbf{x}),
\]
which is why the ELBO is indeed a lower bound.

This form also makes clear what ELBO maximisation is trying to achieve:
to increase a lower bound on the model evidence while simultaneously reducing
the mismatch between the approximate posterior and the true posterior.

\paragraph{Third form of the ELBO: reconstruction plus regularisation.}

To obtain the form most commonly used in practice, we factor the joint
distribution as
\[
p_{\theta}(\mathbf{x},\mathbf{z})
=
p_{\theta}(\mathbf{x}\mid \mathbf{z})\,p(\mathbf{z}).
\]
Substituting this into the ELBO gives
\[
\begin{aligned}
\mathcal{L}(\mathbf{x};\theta,\phi)
&=
\mathbb{E}_{q_{\phi}(\mathbf{z}\mid \mathbf{x})}
\!\left[
\log \frac{p_{\theta}(\mathbf{x}\mid \mathbf{z})\,p(\mathbf{z})}
{q_{\phi}(\mathbf{z}\mid \mathbf{x})}
\right] \\[4pt]
&=
\mathbb{E}_{q_{\phi}(\mathbf{z}\mid \mathbf{x})}
\!\left[
\log p_{\theta}(\mathbf{x}\mid \mathbf{z})
\right]
+
\mathbb{E}_{q_{\phi}(\mathbf{z}\mid \mathbf{x})}
\!\left[
\log p(\mathbf{z})-\log q_{\phi}(\mathbf{z}\mid \mathbf{x})
\right].
\end{aligned}
\]
The second expectation is again a negative KL divergence, so we obtain
\[
\boxed{
\mathcal{L}(\mathbf{x};\theta,\phi)
=
\mathbb{E}_{q_{\phi}(\mathbf{z}\mid \mathbf{x})}
\!\left[
\log p_{\theta}(\mathbf{x}\mid \mathbf{z})
\right]
-
\mathrm{KL}\!\bigl(
q_{\phi}(\mathbf{z}\mid \mathbf{x})
\,\|\,\,
p(\mathbf{z})
\bigr).
}
\]

This is the most familiar VAE objective.
It has two terms with distinct roles:
\begin{itemize}
    \item the \textbf{reconstruction term}
    \[
    \mathbb{E}_{q_{\phi}(\mathbf{z}\mid \mathbf{x})}
    [\log p_{\theta}(\mathbf{x}\mid \mathbf{z})],
    \]
    which encourages the decoder to assign high probability to the observed data
    given sampled latent codes and therefore acts as a reconstruction objective;

    \item the \textbf{regularisation term}
    \[
    \mathrm{KL}\!\bigl(
    q_{\phi}(\mathbf{z}\mid \mathbf{x})
    \,\|\,\,
    p(\mathbf{z})
    \bigr),
    \]
    which encourages the approximate posterior to remain close to the prior.
\end{itemize}

The first term promotes accurate reconstruction through the decoder.
The second term prevents the encoder from placing each data point arbitrarily
in latent space.
Instead, it encourages the inferred latent distributions to remain compatible
with the assumed prior, thereby giving the latent space a more coherent global
structure.

This regularisation is what makes generative sampling meaningful.
After training, we may draw
\[
\mathbf{z}\sim p(\mathbf{z}),
\qquad
\mathbf{x}\sim p_{\theta}(\mathbf{x}\mid \mathbf{z}),
\]
and expect the resulting samples to resemble the data distribution.

This also addresses the limitation highlighted at the end of the PPCA chapter.
In PPCA, the model introduced a prior, but did not explicitly regularise the
inferred latent space to align with it.
In the VAE, that alignment is built directly into the learning objective.

The three forms derived above are algebraically equivalent but emphasise
different aspects of the same objective. Figure~\ref{fig:vae-elbo-three-views}
brings them together before we summarise what ELBO maximisation achieves.

\input{figures/ch03/fig_ch03_02_vae_elbo_three_views}

\paragraph{What the ELBO achieves.}

We can now see the VAE objective from two complementary viewpoints.

From the inference viewpoint, the ELBO says:
\begin{quote}
\emph{learn an approximate posterior that is close to the true posterior while
still explaining the data well.}
\end{quote}

From the representation-learning viewpoint, it says:
\begin{quote}
\emph{learn latent codes that reconstruct the data accurately, but do so in a
way that keeps the latent space organised and compatible with a simple prior.}
\end{quote}

This is why the VAE is more than an autoencoder with noise.
Its objective is not simply to reconstruct.
It is to perform approximate probabilistic inference inside a latent-variable
generative model.

At the same time, this objective introduces a new computational challenge.
The reconstruction term is an expectation over a stochastic latent variable
sampled from \(q_{\phi}(\mathbf{z}\mid \mathbf{x})\).
To optimise the ELBO with respect to the encoder parameters \(\phi\), and hence
train the approximate posterior jointly with the decoder, we need a way to pass
gradients through this sampling process.

That is the purpose of the reparameterisation trick, which will be the
starting point of the next section.

\section{Learning the Variational Autoencoder: Reparameterisation and Optimisation}

The previous section showed that the VAE is trained by maximising the ELBO,
\[
\mathcal{L}(\mathbf{x};\theta,\phi)
=
\mathbb{E}_{q_{\phi}(\mathbf{z}\mid \mathbf{x})}
\!\left[
\log p_{\theta}(\mathbf{x}\mid \mathbf{z})
\right]
-
\mathrm{KL}\!\bigl(
q_{\phi}(\mathbf{z}\mid \mathbf{x})
\,\|\,\,
p(\mathbf{z})
\bigr).
\]
The remaining question is how this objective can be optimised in practice, given
that it contains an expectation over stochastic latent samples drawn from the
encoder distribution.

%\paragraph{Why naive sampling blocks gradients}

The reconstruction term in the ELBO contains an expectation over latent samples:
\[
\mathbb{E}_{q_{\phi}(\mathbf{z}\mid \mathbf{x})}
\!\left[
\log p_{\theta}(\mathbf{x}\mid \mathbf{z})
\right].
\]
This means that, during training, we must sample \(\mathbf{z}\) from the encoder
distribution and then pass that sample through the decoder.

At first sight, this seems straightforward:
sample
\[
\mathbf{z}\sim q_{\phi}(\mathbf{z}\mid \mathbf{x}),
\]
evaluate the reconstruction term, and optimise.
The difficulty is that the sampling distribution itself depends on the encoder
parameters \(\phi\).
If we treat $\mathbf{z}\sim q_{\phi}(\mathbf{z}\mid \mathbf{x})$
as a black-box stochastic operation, then the dependence of the sample
\(\mathbf{z}\) on \(\phi\) is hidden inside the sampling step, preventing
ordinary backpropagation through the encoder.

In other words, the decoder parameters \(\theta\) are easy to differentiate
through once \(\mathbf{z}\) is given, but the encoder parameters \(\phi\) are
more problematic because they determine \emph{how} \(\mathbf{z}\) is sampled.
To train the VAE efficiently by gradient-based optimisation, we therefore need a
way to rewrite this stochastic sampling step as a differentiable computation.

\paragraph{The reparameterisation trick.}
The key idea is to separate the source of randomness from the dependence on the
encoder parameters.
For the common case of a Gaussian approximate posterior with diagonal
covariance,
\[
q_{\phi}(\mathbf{z}\mid \mathbf{x})
=
\mathcal{N}\!\bigl(
\mathbf{z}\mid
\boldsymbol{\mu}_{\phi}(\mathbf{x}),
\operatorname{diag}(\boldsymbol{\sigma}_{\phi}^2(\mathbf{x}))
\bigr),
\]
we can generate a latent sample by first drawing noise from a standard normal
distribution
\[
\boldsymbol{\epsilon}\sim\mathcal{N}(\mathbf{0},\mathbf{I}),
\]
and then defining
\[
\mathbf{z}
=
\boldsymbol{\mu}_{\phi}(\mathbf{x})
+
\boldsymbol{\sigma}_{\phi}(\mathbf{x})\odot \boldsymbol{\epsilon}.
\]
This is the reparameterisation trick~\cite{kingma2014autoencoding,rezende2014stochastic}.
Figure~\ref{fig:vae-reparameterisation} contrasts the direct stochastic view
with the reparameterised computation.

\input{figures/ch03/fig_ch03_03_vae_reparameterisation}

The trick does not remove stochasticity; it relocates it to the auxiliary noise
\(\boldsymbol{\epsilon}\), which is independent of the encoder parameters.
For a fixed noise draw, the latent variable is a deterministic differentiable
function
\[
\mathbf{z}=\mathbf{z}(\mathbf{x},\phi,\boldsymbol{\epsilon}),
\]
so ordinary pathwise gradients can propagate through the encoder using standard
backpropagation.

\paragraph{Joint optimisation of encoder and decoder.} With the reparameterisation trick in place, the VAE can be trained end-to-end by
maximising the ELBO jointly with respect to both parameter sets:
\[
(\theta,\phi)
\quad\longmapsto\quad
\max \; \mathcal{L}(\mathbf{x};\theta,\phi).
\]
In practice, for each input the encoder produces
\(\boldsymbol{\mu}_{\phi}(\mathbf{x})\) and
\(\boldsymbol{\sigma}_{\phi}(\mathbf{x})\), a latent sample is formed via
reparameterisation, the ELBO is evaluated, and gradients are then
backpropagated through both encoder and decoder.
This optimisation is typically carried out using minibatch stochastic gradient
methods or one of their modern variants.

This is one of the most important conceptual achievements of the VAE.
It turns approximate posterior inference and generative learning into a single
differentiable computation graph.
From this point onward, the encoder and decoder are no longer trained as
separate objects: they are coupled through the ELBO and learned together.

In this sense, training a VAE is an instance of variational inference.
Rather than computing the true posterior \(p_{\theta}(\mathbf{z}\mid\mathbf{x})\)
exactly, we learn a tractable approximation \(q_{\phi}(\mathbf{z}\mid\mathbf{x})\)
and optimise the ELBO jointly with respect to both the decoder parameters
\(\theta\) and the encoder parameters \(\phi\).

\paragraph{A variant: \(\beta\)-VAE.}

The balance between reconstruction and regularisation can be adjusted in
different ways, and one important extension is the \(\beta\)-VAE~\cite{higgins2017beta}, which
introduces a weighting factor on the KL term:
\[
\mathcal{L}_{\beta}
=
\mathbb{E}_{q_{\phi}(\mathbf{z}\mid \mathbf{x})}
\!\left[
\log p_{\theta}(\mathbf{x}\mid \mathbf{z})
\right]
-
\beta\,
\mathrm{KL}\!\bigl(
q_{\phi}(\mathbf{z}\mid \mathbf{x})
\,\|\,\,
p(\mathbf{z})
\bigr).
\]
When \(\beta>1\), the latent space is more strongly regularised, often leading
to more structured or disentangled representations, but usually at the cost of
reconstruction quality.

This extension makes especially clear a central tension in variational
latent-variable models:
\begin{quote}
\emph{the latent space must be regular enough to support generation, but
informative enough to support useful representation and reconstruction.}
\end{quote}

Seen from this perspective, the VAE marks an important step in the development
of modern generative modelling.
PPCA introduced a tractable probabilistic latent-variable model.
The VAE carries that same latent-variable logic into the nonlinear neural-network
setting by replacing exact inference with a learned approximation and by turning
the ELBO into a practical training objective.

These ideas will continue to matter in later chapters.
In particular, although diffusion models are often introduced from other
viewpoints, ELBO-based reasoning and latent-variable perspectives also provide a
useful framework for approaching them.
In the next chapter, we will see how these same probabilistic ideas can be
extended further in one of the most important modern generative AI frameworks.

\section*{Summary and References}
\addcontentsline{toc}{section}{Summary and References}

This chapter introduced the Variational Autoencoder (VAE) as a probabilistic
extension of the autoencoder and as a nonlinear continuation of the
latent-variable framework developed through PPCA.

Because the model is nonlinear, neither the marginal likelihood
\(p_{\theta}(\mathbf{x})\) nor the true posterior
\(p_{\theta}(\mathbf{z}\mid\mathbf{x})\) is generally available in closed form.
We then saw that the ELBO admits several equivalent forms, the most familiar of
which decomposes into a reconstruction term and a regularisation term.
This made clear that the VAE is not merely reconstructing inputs, but learning
a latent representation that remains compatible with a simple prior and hence
supports meaningful generation.

Finally, the chapter showed how  the reparameterisation trick rewrites stochastic latent sampling as a
differentiable transformation of parameter-free noise, allowing encoder and
decoder to be trained jointly by gradient-based optimisation

The original VAE formulation was introduced by Kingma and
Welling~\cite{kingma2014autoencoding}, with closely related stochastic-gradient
variational methods developed by Rezende, Mohamed, and
Wierstra~\cite{rezende2014stochastic}.
For broader probabilistic modelling background, see Bishop~\cite{bishop2006prml} and
Murphy~\cite{murphy2012ml}.
For a comprehensive introduction to VAEs and an overview of their more recent developments, see Kingma and Welling~\cite{kingma2019introduction}.

%% file: figures/ch03/chapter3_vae_at_a_glance.tex
\begin{center}
\begin{tikzpicture}[
  >=Latex,
  every node/.style={font=\small},
  box/.style={
    draw,
    rounded corners=2pt,
    align=center,
    inner sep=6pt,
    minimum height=1.02cm,
    text width=3.0cm
  },
  objective/.style={
    draw,
    rounded corners=2pt,
    align=center,
    inner sep=6pt,
    minimum height=1.08cm,
    text width=4.2cm
  },
  smallbox/.style={
    draw,
    rounded corners=2pt,
    align=center,
    inner sep=5pt,
    minimum height=0.88cm,
    text width=2.4cm
  },
  flow/.style={->,thick},
  secondary/.style={->,densely dashed}
]

\node[smallbox] (x) at (0,0) {
  \textbf{Observed data}\\[1pt]
  $\mathbf{x}$
};

\node[box] (enc) at (4.1,0) {
  \textbf{Variational encoder}\\[1pt]
  $q_{\phi}(\mathbf{z}\mid\mathbf{x})$
};

\node[smallbox] (z) at (8.1,0) {
  \textbf{Latent sample}\\[1pt]
  $\mathbf{z}$
};

\node[box] (dec) at (12.2,0) {
  \textbf{Probabilistic decoder}\\[1pt]
  $p_{\theta}(\mathbf{x}\mid\mathbf{z})$
};

\node[smallbox] (prior) at (8.1,-2.45) {
  \textbf{Latent prior}\\[1pt]
  $p(\mathbf{z})=\mathcal{N}(\mathbf{0},\mathbf{I})$
};

\node[objective] (elbo) at (3.0,-2.45) {
  \textbf{ELBO objective}\\[2pt]
  $\mathbb{E}_{q_{\phi}(\mathbf{z}\mid\mathbf{x})}
   [\log p_{\theta}(\mathbf{x}\mid\mathbf{z})]$
  $-\;
  \mathrm{KL}\!\left(
  q_{\phi}(\mathbf{z}\mid\mathbf{x})
  \,\|\,p(\mathbf{z})\right)$
};

\draw[flow] (x) -- (enc);
\draw[flow] (enc) -- (z);
\draw[flow] (z) -- (dec);

% Prior supplies latent samples when generating new observations.
\draw[flow] (prior) -- node[right] {\scriptsize generation} (z);

% Dashed links indicate distributions entering the ELBO terms.
\draw[secondary] (prior) -- (elbo);
\draw[secondary] (enc) -- (elbo);
\draw[secondary] (dec.south west) to[out=-120,in=0] (elbo.east);

\node[font=\normalsize\bfseries] at (6.1,1.45) {VAE at a glance};

\end{tikzpicture}
\end{center}

%% file: figures/ch03/fig_ch03_01_ae_vs_vae_latent_representation.tex
\begin{figure}[H]
\centering
\includegraphics[width=0.94\textwidth]{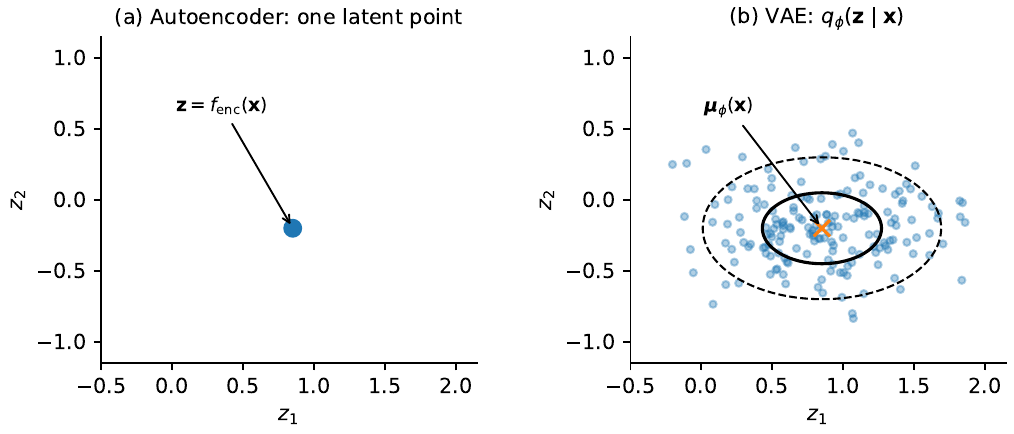}
\caption{\textbf{From a deterministic latent code to a variational posterior.}
For a fixed input $\mathbf{x}$, a standard autoencoder produces one latent
point, whereas a VAE encoder specifies a distribution
$q_{\phi}(\mathbf{z}\mid\mathbf{x})$ of plausible latent values. Here the common
diagonal-Gaussian choice is shown through samples, its mean, and elliptical
contours. The shared centre is chosen only to make the point-versus-distribution
contrast visually direct.}
\label{fig:vae-ae-versus-posterior}
\end{figure}

%% file: figures/ch03/fig_ch03_02_vae_elbo_three_views.tex
\begin{figure}[H]
\centering
\begin{tikzpicture}[
  >=Latex,
  every node/.style={font=\small},
  elbo/.style={
    draw,
    rounded corners=2pt,
    align=center,
    inner sep=7pt,
    text width=12.2cm
  },
  view/.style={
    draw,
    rounded corners=2pt,
    align=center,
    inner sep=7pt,
    text width=5.75cm,
    minimum height=2.35cm
  },
  flow/.style={->,thick}
]

\node[elbo] (joint) at (0,2.25) {
  \textbf{Joint-density form}\\[4pt]
  $\displaystyle
  \mathcal{L}(\mathbf{x};\theta,\phi)
  =
  \mathbb{E}_{q_{\phi}(\mathbf{z}\mid\mathbf{x})}
  \!\left[
  \log
  \frac{p_{\theta}(\mathbf{x},\mathbf{z})}
       {q_{\phi}(\mathbf{z}\mid\mathbf{x})}
  \right]$
};

\node[view] (inf) at (-3.45,-1.2) {
  \textbf{Inference view}\\[4pt]
  $\displaystyle
  \begin{aligned}
  \mathcal{L}
  &=
  \log p_{\theta}(\mathbf{x})\\[-1pt]
  &\quad-
  \mathrm{KL}\!\left(
  q_{\phi}(\mathbf{z}\mid\mathbf{x})
  \,\|\,p_{\theta}(\mathbf{z}\mid\mathbf{x})
  \right)
  \end{aligned}$\\[5pt]
  The gap to the exact log-likelihood is the
  approximate-posterior mismatch.
};

\node[view] (train) at (3.45,-1.2) {
  \textbf{Training view}\\[4pt]
  $\displaystyle
  \begin{aligned}
  \mathcal{L}
  &=
  \mathbb{E}_{q_{\phi}(\mathbf{z}\mid\mathbf{x})}
  [\log p_{\theta}(\mathbf{x}\mid\mathbf{z})]\\[-1pt]
  &\quad-
  \mathrm{KL}\!\left(
  q_{\phi}(\mathbf{z}\mid\mathbf{x})
  \,\|\,p(\mathbf{z})
  \right)
  \end{aligned}$\\[5pt]
  Expected reconstruction log-likelihood is balanced
  against regularisation toward the prior.
};

\draw[flow] (joint) -- node[left] {\scriptsize Bayes' rule} (inf);
\draw[flow] (joint) -- node[right] {\scriptsize factor the joint} (train);

\node[font=\normalsize\bfseries] at (0,3.45)
  {The same ELBO, viewed three ways};

\end{tikzpicture}
\caption{\textbf{Three equivalent views of the VAE ELBO.}
The joint-density form is the generic variational objective. Bayes' rule gives
the inference view as the exact log-likelihood minus posterior mismatch, while
factoring the joint gives the training view as expected reconstruction
log-likelihood minus regularisation toward the latent prior. All three are
algebraically equivalent.}
\label{fig:vae-elbo-three-views}
\end{figure}

%% file: figures/ch03/fig_ch03_03_vae_reparameterisation.tex
\begin{figure}[H]
\centering
\begin{tikzpicture}[
  >=Latex,
  every node/.style={font=\small},
  box/.style={
    draw,
    rounded corners=2pt,
    align=center,
    inner sep=5pt,
    minimum height=0.92cm,
    text width=2.55cm
  },
  tinybox/.style={
    draw,
    rounded corners=2pt,
    align=center,
    inner sep=4pt,
    minimum height=0.82cm,
    text width=2.0cm
  },
  flow/.style={->,thick},
  stochastic/.style={->,thick,dashed}
]

% ---------------- Direct sampling ----------------
\begin{scope}[shift={(-6.15,0)}]
\node[font=\normalsize\bfseries] at (4.75,2.05)
  {(a) Direct parameter-dependent sampling};

\node[tinybox] (x1) at (0,0.55) {$\mathbf{x}$};
\node[box] (enc1) at (3.0,0.55) {
  encoder\\[-1pt]
  $\boldsymbol{\mu}_{\phi}(\mathbf{x}),
   \boldsymbol{\sigma}_{\phi}(\mathbf{x})$
};
\node[box] (draw1) at (6.25,0.55) {
  stochastic draw\\[-1pt]
  $\mathbf{z}\sim q_{\phi}(\mathbf{z}\mid\mathbf{x})$
};
\node[box] (dec1) at (9.5,0.55) {
  decoder\\[-1pt]
  $p_{\theta}(\mathbf{x}\mid\mathbf{z})$
};

\draw[flow] (x1) -- (enc1);
\draw[stochastic] (enc1) -- (draw1);
\draw[flow] (draw1) -- (dec1);

\node[align=center,text width=8.8cm] at (4.75,-0.95) {
Treating the draw as a black-box stochastic operation hides the ordinary
pathwise backpropagation route to the encoder parameters $\phi$.
};
\end{scope}

% ---------------- Reparameterised sampling ----------------
\begin{scope}[shift={(-6.15,-4.55)}]
\node[font=\normalsize\bfseries] at (4.75,2.05)
  {(b) Reparameterised sampling};

\node[tinybox] (x2) at (0,0.55) {$\mathbf{x}$};
\node[box] (enc2) at (3.0,0.55) {
  encoder\\[-1pt]
  $\boldsymbol{\mu}_{\phi}(\mathbf{x}),
   \boldsymbol{\sigma}_{\phi}(\mathbf{x})$
};
\node[tinybox] (eps) at (3.0,-1.05) {
  $\boldsymbol{\epsilon}\sim
  \mathcal{N}(\mathbf{0},\mathbf{I})$
};
\node[box] (combine) at (6.25,0.55) {
  deterministic transform\\[-1pt]
  $\mathbf{z}
  =
  \boldsymbol{\mu}_{\phi}
  +
  \boldsymbol{\sigma}_{\phi}
  \odot\boldsymbol{\epsilon}$
};
\node[box] (dec2) at (9.5,0.55) {
  decoder\\[-1pt]
  $p_{\theta}(\mathbf{x}\mid\mathbf{z})$
};

\draw[flow] (x2) -- (enc2);
\draw[flow] (enc2) -- (combine);
\draw[flow] (eps) -- (combine);
\draw[flow] (combine) -- (dec2);

\node[align=center,text width=8.8cm] at (4.75,-2.15) {
For fixed $\boldsymbol{\epsilon}$, $\mathbf{z}$ is a deterministic
differentiable function of the encoder outputs, exposing a pathwise gradient
to $\phi$.
};
\end{scope}

\end{tikzpicture}
\caption{\textbf{The reparameterisation trick separates randomness from the
encoder parameters.}
Directly sampling $\mathbf{z}\sim q_{\phi}(\mathbf{z}\mid\mathbf{x})$ obscures
the ordinary pathwise gradient through the sampling step. Instead, sampling
parameter-free $\boldsymbol{\epsilon}\sim\mathcal{N}(\mathbf{0},\mathbf{I})$ and
setting $\mathbf{z}=\boldsymbol{\mu}_{\phi}(\mathbf{x})
+\boldsymbol{\sigma}_{\phi}(\mathbf{x})\odot\boldsymbol{\epsilon}$ preserves
the same Gaussian distribution while making $\mathbf{z}$ differentiable in the
encoder outputs for a fixed noise draw.}
\label{fig:vae-reparameterisation}
\end{figure}

%% file: chapters/cht5_new.tex
\chapter{Denoising Diffusion Probabilistic Models}
\section*{Overview}
\addcontentsline{toc}{section}{Overview}

Diffusion models represent one of the most important modern families of
generative models, achieving strong performance in a wide range of
applications, particularly in image synthesis.
At their core is a simple but powerful idea: instead of generating a sample in a
single step, learn to transform noise gradually into structured data through a
sequence of denoising operations.

This chapter focuses on the discrete-time \emph{Denoising Diffusion
Probabilistic Model} (DDPM) framework~\cite{ho2020denoising}, which provides one of the clearest
probabilistic formulations of this idea.
Historically, diffusion-based generative modelling was developed by
Sohl-Dickstein et al.~\cite{sohl2015deep}, drawing inspiration from ideas in
physics, particularly processes in which randomness accumulates gradually over
time.
In the physical world, for example, a drop of ink released into water slowly
disperses until its structure becomes indistinguishable from its surroundings.
Diffusion models take this intuition and reverse it:
instead of watching order dissolve into noise, we learn how to start from noise
and gradually reconstruct structure.

In this book, however, we first approach diffusion models from the probabilistic
and latent-variable viewpoint developed through PPCA and the VAE.
This provides a natural entry point before introducing the continuous-time
dynamical perspective.
Later, in Chapter~5, we return to the relevant calculus, dynamical, and
density-evolution viewpoints and build toward a more continuous interpretation in Chapter~6.

From a methodological perspective, diffusion models still share important
structural connections with earlier latent-variable models.
The key difference is that, instead of using a single latent variable or a
single stochastic encoding step, diffusion introduces a whole sequence of latent
states connected through time.
In practice, this gives diffusion models two linked phases: a fixed forward
process that gradually corrupts data into noise, and a learned reverse
denoising process that gradually reconstructs structure from noise.
This temporal design gives the model much greater flexibility, but also changes
how inference and learning must be formulated.

\section*{Concept Map}
\addcontentsline{toc}{section}{Concept Map}

The chapter extends the modelling logic of earlier chapters to explain sequential noising and denoising through latent variables, posterior inference, and variational learning.

\begin{itemize}

    \item[\(\rightarrow\)] \textbf{We begin with the forward diffusion process.}  
    A clean sample \(\mathbf{x}_0\) is gradually corrupted by Gaussian noise
    over \(T\) steps, producing a latent trajectory
    \(\mathbf{x}_1,\dots,\mathbf{x}_T\).
    This defines a fixed Markov chain together with a closed-form relation
    between any noisy state \(\mathbf{x}_t\) and the original sample
    \(\mathbf{x}_0\).

    \item[\(\rightarrow\)] \textbf{We then introduce the reverse process as the learned generative model.}  
    Starting from a simple Gaussian prior at time \(T\), the model learns a
    reverse denoising chain that aims to reconstruct the data distribution.
    This reverse chain is the unknown part of diffusion and the main object of
    learning.

    \item[\(\rightarrow\)] \textbf{Next, we recast diffusion in ELBO form.}  
    The observed variable is the clean sample \(\mathbf{x}_0\), while the latent
    variable is the whole trajectory \(\mathbf{x}_{1:T}\).
    The known forward process plays the role of the auxiliary distribution in
    the ELBO, turning an intractable likelihood into a tractable lower bound.

    \item[\(\rightarrow\)] \textbf{We decompose the ELBO into local denoising terms.}  
    This shows that the key trainable quantity at each time step is the
    mismatch between the learned reverse transition
    \(p_\theta(\mathbf{x}_{t-1}\mid \mathbf{x}_t)\) and the true reverse
    posterior \(q(\mathbf{x}_{t-1}\mid \mathbf{x}_t,\mathbf{x}_0)\).

    \item[\(\rightarrow\)] \textbf{We then derive the true reverse posterior and rewrite it in the noise domain.}  
    Because the forward process is Gaussian, the true reverse posterior is also
    Gaussian.
    Reparameterising its mean through the noise variable removes the explicit
    dependence on \(\mathbf{x}_0\) and leads to the standard DDPM
    parameterisation.

    \item[\(\rightarrow\)] \textbf{Finally, we obtain the practical DDPM objective and generation procedure.}  
    The learning problem simplifies to noise prediction, so training becomes a
    regression task on Gaussian noise, while sampling becomes an iterative
    denoising process that gradually transforms pure noise into a data sample.

\end{itemize}

\medskip
\noindent The main probabilistic structure of DDPMs is summarised visually below.
\input{figures/ch04/chapter4_ddpm_at_a_glance}
\medskip

\section{The Forward Diffusion Process: From Data to Noise}

Diffusion models depart from the generative frameworks we have studied so far in
one important way.
In PPCA and the VAE, each observation is described through a comparatively
direct latent representation.
In diffusion models, by contrast, a data point is associated with a
\emph{sequence} of latent states that records a gradual transformation through
time.
Generation is therefore no longer a one-off decoding step, but a trajectory.

At a high level, DDPMs consist of two complementary procedures.
The \textbf{forward process} gradually corrupts a clean data sample
\(\mathbf{x}_0\) until it becomes nearly pure Gaussian noise.
The \textbf{reverse process}, or \textbf{denoising process}, aims to undo this
corruption step by step, starting from noise and gradually recovering a
structured sample.

We begin with the simpler half of the model: the forward process.
In the diffusion literature, this process is conventionally denoted by \(q\).
It is fixed and known, rather than learned from data.
Its role is to define how a clean sample \(\mathbf{x}_0\) is progressively
corrupted over \(T\) discrete time steps, producing a sequence of increasingly
noisy latent states
\[
\mathbf{x}_1,\mathbf{x}_2,\dots,\mathbf{x}_T.
\]

\subsection{Gaussian Noising and the Closed-Form Forward Process}

\paragraph{A step-by-step Gaussian noising process.} Let \(\mathbf{x}_0\) denote an observed data sample.
The forward diffusion process produces the sequence above by adding a small
amount of Gaussian noise at each step.
Formally, the transition from \(\mathbf{x}_{t-1}\) to \(\mathbf{x}_t\) is
defined by
\[
q(\mathbf{x}_t\mid \mathbf{x}_{t-1})
=
\mathcal{N}\!\bigl(
\mathbf{x}_t \mid
\sqrt{1-\beta_t}\,\mathbf{x}_{t-1},
\beta_t\mathbf{I}
\bigr),
\]
where \(\beta_t\in(0,1)\) controls how much noise is injected at step \(t\).

Equivalently, we may write the same step as
\[
\mathbf{x}_t
=
\sqrt{1-\beta_t}\,\mathbf{x}_{t-1}
+
\sqrt{\beta_t}\,\mathbf{z}_t,
\qquad
\mathbf{z}_t\sim\mathcal{N}(\mathbf{0},\mathbf{I}).
\]
This update has a simple interpretation:
\(\sqrt{1-\beta_t}\) preserves part of the previous signal, while
\(\sqrt{\beta_t}\) scales the fresh Gaussian perturbation added at the current
step.
Thus each new state \(\mathbf{x}_t\) is slightly noisier than the one before it.

Here \(\beta_t\) controls how much fresh Gaussian noise is added at step \(t\),
and therefore how quickly the original structure of the data is destroyed as
the process evolves.

\paragraph{A closed-form link between \texorpdfstring{$\mathbf{x}_t$}{x\_t} and \texorpdfstring{$\mathbf{x}_0$.}{x\_0}} The one-step rule above tells us how \(\mathbf{x}_t\) depends on
\(\mathbf{x}_{t-1}\), but it is often more useful to relate \(\mathbf{x}_t\)
directly to the original clean sample \(\mathbf{x}_0\).
This turns out to be one of the most useful features of the forward diffusion
process, because it lets us jump directly to any time step \(t\) without having
to simulate the entire chain.

Starting from
\[
\mathbf{x}_t
=
\sqrt{1-\beta_t}\,\mathbf{x}_{t-1}
+
\sqrt{\beta_t}\,\mathbf{z}_t,
\qquad
\mathbf{z}_t\sim\mathcal{N}(\mathbf{0},\mathbf{I}),
\]
we can recursively substitute earlier steps into later ones.

For example,
\[
\begin{aligned}
\mathbf{x}_1
&=
\sqrt{1-\beta_1}\,\mathbf{x}_0
+
\sqrt{\beta_1}\,\mathbf{z}_1,\\[4pt]
\mathbf{x}_2
&=
\sqrt{1-\beta_2}\,\mathbf{x}_1
+
\sqrt{\beta_2}\,\mathbf{z}_2.
\end{aligned}
\]
Substituting the first into the second gives
\[
\mathbf{x}_2
=
\sqrt{(1-\beta_2)(1-\beta_1)}\,\mathbf{x}_0
+
\sqrt{1-\beta_2}\sqrt{\beta_1}\,\mathbf{z}_1
+
\sqrt{\beta_2}\,\mathbf{z}_2.
\]

At this point it is convenient to define
\[
\alpha_t = 1-\beta_t,
\qquad
\bar{\alpha}_t = \prod_{s=1}^{t}\alpha_s.
\]
Here \(\alpha_t\) is the signal-retention factor at step \(t\), while
\(\bar{\alpha}_t\) is the cumulative signal-retention factor up to time \(t\).

The pattern is now clear.
At each new step, the remaining signal is multiplied by another factor
\(\sqrt{1-\beta_t}=\sqrt{\alpha_t}\), while a fresh Gaussian perturbation is
added.
After \(t\) steps, the signal inherited from \(\mathbf{x}_0\) is therefore
scaled by
\[
\prod_{s=1}^{t}\sqrt{1-\beta_s}
=
\prod_{s=1}^{t}\sqrt{\alpha_s}
=
\sqrt{\bar{\alpha}_t}.
\]

Meanwhile, the many independent Gaussian noise terms can be combined into a
single overall Gaussian variable.
Since a weighted sum of independent Gaussian variables is still Gaussian, the
accumulated noise can be written as one standard Gaussian scaled by the total
variance factor \(1-\bar{\alpha}_t\).
This gives the closed-form relation
\[
\mathbf{x}_t
=
\sqrt{\bar{\alpha}_t}\,\mathbf{x}_0
+
\sqrt{1-\bar{\alpha}_t}\,\boldsymbol{\epsilon},
\qquad
\boldsymbol{\epsilon}\sim\mathcal{N}(\mathbf{0},\mathbf{I}).
\]

In probabilistic form, this is
\[
q(\mathbf{x}_t\mid \mathbf{x}_0)
=
\mathcal{N}\!\left(
\mathbf{x}_t \mid
\sqrt{\bar{\alpha}_t}\,\mathbf{x}_0,\,
(1-\bar{\alpha}_t)\mathbf{I}
\right).
\]

This notation is important:
\begin{itemize}
    \item \(\mathbf{z}_t\) denotes the \emph{local} Gaussian noise newly added at
    step \(t\);
    \item \(\boldsymbol{\epsilon}\) denotes the \emph{overall accumulated} noise
    that summarises the effect of all noising steps up to time \(t\).
\end{itemize}

The closed-form expression above is extremely useful.
It tells us that after \(t\) steps, the sample \(\mathbf{x}_t\) still contains a
fraction \(\sqrt{\bar{\alpha}_t}\) of the original signal, while the remaining
part has been replaced by Gaussian noise.
When \(t=0\), we have \(\bar{\alpha}_0=1\), so \(\mathbf{x}_t=\mathbf{x}_0\).
When \(t=T\) is large enough that \(\bar{\alpha}_T\approx 0\), repeated
Gaussian corruption has washed out the original structure and the sample is
driven close to isotropic Gaussian noise.

This direct relationship between \(\mathbf{x}_t\) and \(\mathbf{x}_0\) will
later make training much more efficient, because it lets us sample noisy states
at arbitrary time steps without simulating the whole trajectory.

Figure~\ref{fig:ddpm-forward-closed-form} summarises both effects visually:
the signal coefficient decreases as the noise coefficient increases, while the
marginal distribution progressively loses its original structure.

\input{figures/ch04/fig_ch04_01_forward_diffusion}

\subsection{The forward chain and the Markov assumption}

Because the diffusion process introduces a whole sequence of latent states, the
full forward trajectory is described by the joint distribution
\[
q(\mathbf{x}_{0:T}) = q(\mathbf{x}_0)\,q(\mathbf{x}_{1:T}\mid \mathbf{x}_0).
\]
By the chain rule of probability,
\[
q(\mathbf{x}_{0:T})
=
q(\mathbf{x}_0)
\prod_{t=1}^{T}
q(\mathbf{x}_t\mid \mathbf{x}_{t-1},\mathbf{x}_{t-2},\dots,\mathbf{x}_0).
\]

The crucial simplifying assumption is that the forward process is
\textbf{Markovian}.
This means that each new state depends only on the immediately preceding one:
\[
q(\mathbf{x}_t\mid \mathbf{x}_{t-1},\mathbf{x}_{t-2},\dots,\mathbf{x}_0)
=
q(\mathbf{x}_t\mid \mathbf{x}_{t-1}).
\]
Hence the full forward distribution factorises as
\[
q(\mathbf{x}_{0:T})
=
q(\mathbf{x}_0)\prod_{t=1}^{T}q(\mathbf{x}_t\mid \mathbf{x}_{t-1}).
\]

This makes the forward process mathematically tractable and easy to analyse.
But it is not the main object of learning.
Once the data sample \(\mathbf{x}_0\) and the noise schedule
\(\{\beta_t\}_{t=1}^{T}\) are fixed, the forward process \(q\) is fully known.

\begin{tcolorbox}[
  colback=blue!6,
  colframe=blue!55!black,
  title={Forward-process summary},
  boxrule=0.8pt,
  arc=2mm,
  left=2mm,
  right=2mm,
  top=1mm,
  bottom=1mm
]
\[
q(\mathbf{x}_t\mid \mathbf{x}_{t-1})
=
\mathcal{N}\!\bigl(
\mathbf{x}_t\mid
\sqrt{1-\beta_t}\,\mathbf{x}_{t-1},
\beta_t\mathbf{I}
\bigr),
\qquad
\alpha_t=1-\beta_t,
\qquad
\bar{\alpha}_t=\prod_{s=1}^{t}\alpha_s.
\]

\[
\mathbf{x}_t
=
\sqrt{\bar{\alpha}_t}\,\mathbf{x}_0
+
\sqrt{1-\bar{\alpha}_t}\,\boldsymbol{\epsilon},
\qquad
\boldsymbol{\epsilon}\sim\mathcal{N}(\mathbf{0},\mathbf{I}),
\]
so equivalently
\[
q(\mathbf{x}_t\mid \mathbf{x}_0)
=
\mathcal{N}\!\left(
\mathbf{x}_t\mid
\sqrt{\bar{\alpha}_t}\,\mathbf{x}_0,\,
(1-\bar{\alpha}_t)\mathbf{I}
\right).
\]

\[
q(\mathbf{x}_{0:T})
=
q(\mathbf{x}_0)\prod_{t=1}^{T}q(\mathbf{x}_t\mid \mathbf{x}_{t-1}).
\]
\end{tcolorbox}

What ultimately matters for generation, however, is the \textbf{reverse} or
\textbf{denoising} process:
starting from a random Gaussian point and gradually reconstructing a data-like
sample.
That reverse process is not known analytically and must be learned, typically by
a neural network.
To understand it, we will now bring back the tools of ELBO and variational
inference developed earlier and adapt them to this time-dependent latent
sequence, so that the reverse process can be expressed, analysed, and eventually
trained.

\section{The Reverse Process and the ELBO Formulation}

The forward process gave us a fixed and tractable way to transform a clean data
sample \(\mathbf{x}_0\) into a sequence of increasingly noisy latent states
\(\mathbf{x}_1,\dots,\mathbf{x}_T\), eventually approaching Gaussian noise.
That process, however, is not the part we wish to learn.
For generation, what matters is the opposite direction:
starting from noise and gradually recovering structure.
This is the role of the \textbf{reverse process}, conventionally denoted by
\(p\).

At a high level, the reverse process tries to undo the corruption introduced by
the forward chain.
Starting from a simple Gaussian sample \(\mathbf{x}_T\), it generates
\(\mathbf{x}_{T-1},\mathbf{x}_{T-2},\dots,\mathbf{x}_0\) step by step, with the
goal that the final state \(\mathbf{x}_0\) should look like a genuine data
sample.
In other words, if the forward process moves
\[
\mathbf{x}_0 \longrightarrow \mathbf{x}_1 \longrightarrow \cdots \longrightarrow \mathbf{x}_T,
\]
then the reverse process aims to model the opposite direction
\[
\mathbf{x}_T \longrightarrow \mathbf{x}_{T-1} \longrightarrow \cdots \longrightarrow \mathbf{x}_0.
\]

\subsection{The learned reverse chain}

To describe generation probabilistically, we introduce a parameterised reverse
process.
By the general chain rule of probability, the joint distribution over the whole
reverse trajectory could in principle be written as
\[
\begin{aligned}
p_\theta(\mathbf{x}_{0:T})
&= p_\theta(\mathbf{x}_0,\mathbf{x}_{1:T}) \\[4pt]
&= p_\theta(\mathbf{x}_0\mid \mathbf{x}_{1:T})\,p_\theta(\mathbf{x}_{1:T}) \\[4pt]
&= p_\theta(\mathbf{x}_0\mid \mathbf{x}_{1:T})\,
   p_\theta(\mathbf{x}_1\mid \mathbf{x}_{2:T})\,
   p_\theta(\mathbf{x}_{2:T}) \\[4pt]
&= p_\theta(\mathbf{x}_0\mid \mathbf{x}_{1:T})\,
   p_\theta(\mathbf{x}_1\mid \mathbf{x}_{2:T})\,
   p_\theta(\mathbf{x}_2\mid \mathbf{x}_{3:T})\,
   \cdots\,
   p_\theta(\mathbf{x}_{T-1}\mid \mathbf{x}_T)\,
   p(\mathbf{x}_T).
\end{aligned}
\]
This is the fully general expansion: each earlier state may depend on all later
states.

In diffusion models, however, we make the same simplifying choice as in the
forward process and assume that the reverse chain is also \textbf{Markovian}.
That is,
\[
p_\theta(\mathbf{x}_{t-1}\mid \mathbf{x}_{t:T})
=
p_\theta(\mathbf{x}_{t-1}\mid \mathbf{x}_t).
\]
Under this first-order Markov assumption, the reverse joint simplifies to
\[
p_\theta(\mathbf{x}_{0:T})
=
p(\mathbf{x}_T)\prod_{t=1}^{T} p_\theta(\mathbf{x}_{t-1}\mid \mathbf{x}_t).
\]
This expression should be read from right to left.
We begin from a simple prior at the final time step,
typically
\[
p(\mathbf{x}_T)=\mathcal{N}(\mathbf{0},\mathbf{I}),
\]
and then apply a sequence of learned reverse transitions
\(p_\theta(\mathbf{x}_{t-1}\mid \mathbf{x}_t)\) until we eventually obtain a
sample at time \(0\).

This is the key unknown object in the model.
Unlike the forward process \(q\), which was fully specified once the noise
schedule \(\{\beta_t\}\) was chosen, the reverse transitions are not known in
closed form and must therefore be learned from data.
In practice, the parameters \(\theta\) belong to a neural network that will be
trained to model these reverse denoising steps.

\subsection{Why the reverse process is difficult}

At first sight, one might imagine simply reversing the forward rule.
But this is not straightforward.
Although the forward transition
\[
q(\mathbf{x}_t\mid \mathbf{x}_{t-1})
\]
is Gaussian and analytically defined, the corresponding reverse transition
\[
q(\mathbf{x}_{t-1}\mid \mathbf{x}_t)
\]
is not directly available in a usable form.

To see why, recall Bayes' rule:
\[
q(\mathbf{x}_{t-1}\mid \mathbf{x}_t)
=
\frac{
q(\mathbf{x}_t\mid \mathbf{x}_{t-1})\,q(\mathbf{x}_{t-1})
}{
q(\mathbf{x}_t)
}.
\]
The forward conditional \(q(\mathbf{x}_t\mid \mathbf{x}_{t-1})\) is known by
construction.
The difficulty lies in the marginal distributions
\(q(\mathbf{x}_{t-1})\) and \(q(\mathbf{x}_t)\).
These are not simple fixed Gaussians chosen by hand; rather, they are induced by
taking the unknown data distribution \(q(\mathbf{x}_0)\) and pushing it through
the forward noising chain.
So, although the forward step itself is simple, the true reverse conditional
depends implicitly on the underlying data distribution and is therefore not
available as a simple rule that we can directly evaluate or sample from.

A second difficulty appears when we consider likelihood-based learning.
The model defines a joint distribution over the whole reverse trajectory
\(\mathbf{x}_{0:T}\), but what we ultimately care about is the marginal
likelihood of the observed clean data:
\[
p_\theta(\mathbf{x}_0)
=
\int p_\theta(\mathbf{x}_{0:T})\,d\mathbf{x}_{1:T}.
\]
This is the diffusion-model analogue of the marginal likelihoods we encountered
in earlier latent-variable models.
The observed variable is the clean sample \(\mathbf{x}_0\), while the whole
trajectory \(\mathbf{x}_{1:T}\) plays the role of latent variables and must be
integrated out.

This quantity \(p_\theta(\mathbf{x}_0)\) is the key object we would like to
learn and optimise, because it measures how much probability the model assigns
to a real data sample.
If a generative model assigns high probability to the observed data, it means it
has learned a good distribution over the data domain.
But in diffusion, this marginal likelihood is defined only indirectly through
the reverse chain
\[
p_\theta(\mathbf{x}_{0:T})
=
p(\mathbf{x}_T)\prod_{t=1}^{T} p_\theta(\mathbf{x}_{t-1}\mid \mathbf{x}_t).
\]
So learning the reverse transitions \(p_\theta(\mathbf{x}_{t-1}\mid \mathbf{x}_t)\)
well is precisely how we hope to build a model that assigns high likelihood to
\(\mathbf{x}_0\).

The difficulty, however, is that computing \(p_\theta(\mathbf{x}_0)\) requires
integrating over the entire latent trajectory \(\mathbf{x}_{1:T}\).
Since this trajectory is high-dimensional and time-indexed, the integral is
generally intractable.

So, just as in the VAE, the central learning problem is:
\begin{quote}
\emph{how can we optimise the model when the exact marginal likelihood is not
available in closed form?}
\end{quote}
The answer is again to use an \textbf{Evidence Lower Bound (ELBO)}.

\subsection{The Diffusion ELBO}
\paragraph{Recasting the ELBO in the diffusion setting} At this point, it is helpful to connect the diffusion model back to the general
ELBO framework developed in earlier chapters.

In PPCA and the VAE, we wrote the marginal likelihood in the generic latent-variable form
\[
p(\mathbf{x}) = \int p(\mathbf{x},\mathbf{z})\,d\mathbf{z},
\]
where \(\mathbf{x}\) was the observed variable and \(\mathbf{z}\) was the latent
variable.
In diffusion models, the same structure is still present, but the notation is
now specialised as follows:
\[
\mathbf{x}\;\longrightarrow\;\mathbf{x}_0,
\qquad
\mathbf{z}\;\longrightarrow\;\mathbf{x}_{1:T}.
\]
That is, the observed variable is the clean data sample \(\mathbf{x}_0\), while
the latent variable is no longer a single code but the whole noisy trajectory
\(\mathbf{x}_{1:T}\).

Accordingly, the marginal likelihood now takes the form
\[
p_\theta(\mathbf{x}_0)
=
\int p_\theta(\mathbf{x}_{0:T})\,d\mathbf{x}_{1:T}.
\]

In the general ELBO derivation, we also introduced an auxiliary distribution
\(Q(\mathbf{z})\) over the latent variables.
In the diffusion setting, this role is naturally played by
\[
q(\mathbf{x}_{1:T}\mid \mathbf{x}_0),
\]
namely the \emph{known forward process conditioned on the observed data}.

This point deserves emphasis.
In the VAE, the auxiliary distribution had to be introduced as a learned
approximation \(q_\phi(\mathbf{z}\mid \mathbf{x})\), because the true posterior
was intractable and no closed-form encoder was available.
In diffusion models, however, the situation is different:
the forward noising process has already given us a complete conditional
distribution over the latent trajectory once \(\mathbf{x}_0\) is known.
That is, for a given clean data point \(\mathbf{x}_0\), the model already tells
us exactly how to sample its noisy versions
\(\mathbf{x}_1,\mathbf{x}_2,\dots,\mathbf{x}_T\).
So the auxiliary distribution required by the ELBO is not an extra learned
construction added on top of the model; it is already built into the model as
the forward diffusion process itself.

This is one of the elegant features of diffusion models.
The “variational distribution” is already available:
\[
Q(\mathbf{Z})
\;\longrightarrow\;
q(\mathbf{x}_{1:T}\mid \mathbf{x}_0).
\]
In other words, diffusion does not need to learn a separate encoder network for
the latent variables in the way the VAE does.
The known forward process already provides the conditional distribution over
latent states that the ELBO requires.

\begin{tcolorbox}[
  colback=blue!6,
  colframe=blue!55!black,
  title={Diffusion-model quantities in ELBO form},
  boxrule=0.8pt,
  arc=2mm,
  left=2mm,
  right=2mm,
  top=1mm,
  bottom=1mm
]
\[
\text{observed variable: }\mathbf{x}\rightarrow \mathbf{x}_0,
\qquad
\text{latent variable: }\mathbf{z}\rightarrow \mathbf{x}_{1:T}.
\]

\[
\text{learned generative joint: }\;
p(\mathbf{x},\mathbf{z})
\rightarrow
p_\theta(\mathbf{x}_{0:T})
=
p(\mathbf{x}_T)\prod_{t=1}^{T} p_\theta(\mathbf{x}_{t-1}\mid \mathbf{x}_t).
\]

\[
\text{auxiliary / variational distribution: }\;
Q(\mathbf{z})
\rightarrow
q(\mathbf{x}_{1:T}\mid \mathbf{x}_0),
\]
which is already known from the forward diffusion process.

\[
\text{marginal likelihood: }\;
p_\theta(\mathbf{x}_0)
=
\int p_\theta(\mathbf{x}_{0:T})\,d\mathbf{x}_{1:T},
\]
which is generally intractable because it requires marginalising out the whole
latent trajectory.
\end{tcolorbox}

\paragraph{The diffusion ELBO.} With these identifications in place, the standard ELBO construction gives
\[
\log p_\theta(\mathbf{x}_0)
\ge
\mathbb{E}_{q(\mathbf{x}_{1:T}\mid \mathbf{x}_0)}
\!\left[
\log
\frac{
p_\theta(\mathbf{x}_{0:T})
}{
q(\mathbf{x}_{1:T}\mid \mathbf{x}_0)
}
\right].
\]
This is the ELBO specialised to the diffusion setting.

To train the model over an entire dataset, we then take an outer expectation
over the empirical data distribution \(q(\mathbf{x}_0)\), that is, over the
clean training samples themselves.
This gives the overall objective
\[
\mathbb{E}_{q(\mathbf{x}_0)}
\left[
\log p_\theta(\mathbf{x}_0)
\right]
\ge
\mathbb{E}_{q(\mathbf{x}_0)}
\mathbb{E}_{q(\mathbf{x}_{1:T}\mid \mathbf{x}_0)}
\!\left[
\log
\frac{
p_\theta(\mathbf{x}_{0:T})
}{
q(\mathbf{x}_{1:T}\mid \mathbf{x}_0)
}
\right].
\]

So, from the latent-variable viewpoint developed earlier in the book, diffusion
models fit naturally into the same general probabilistic framework:
they are still generative latent-variable models, but now with a
time-indexed latent sequence rather than a single latent code.
Rather than optimising the intractable marginal likelihood directly, we optimise
a tractable lower bound built from two ingredients:
the learned reverse generative model \(p_\theta(\mathbf{x}_{0:T})\) and the
known forward process \(q(\mathbf{x}_{1:T}\mid \mathbf{x}_0)\).

The crucial next step is therefore to expand and simplify this ELBO so that we
can see how the learning problem decomposes across diffusion steps.
That will be the goal of the next section, where we unpack the reverse process
more explicitly and derive the step-wise objective that eventually leads to the
practical DDPM training loss.

\section{From the Diffusion ELBO to a Step-wise Objective}

We now take the diffusion ELBO derived in the previous section and rewrite it
into a form that exposes what the model is actually being trained to learn.
The key idea is that, although the likelihood involves the whole latent
trajectory \(\mathbf{x}_{1:T}\), the resulting objective can be decomposed into
local denoising steps.

\subsection{Expanding the Diffusion ELBO}

\paragraph{Substituting the forward and reverse chains.}

Recall the two factorizations already established:
\[
p_\theta(\mathbf{x}_{0:T})
=
p(\mathbf{x}_T)\prod_{t=1}^{T} p_\theta(\mathbf{x}_{t-1}\mid \mathbf{x}_t),
\]
and
\[
q(\mathbf{x}_{1:T}\mid \mathbf{x}_0)
=
\prod_{t=1}^{T} q(\mathbf{x}_t\mid \mathbf{x}_{t-1}).
\]
As noted earlier, because the forward process is conditionally defined from the
clean data sample \(\mathbf{x}_0\), we may equivalently keep this conditioning
explicit and write
\[
q(\mathbf{x}_{1:T}\mid \mathbf{x}_0)
=
\prod_{t=1}^{T} q(\mathbf{x}_t\mid \mathbf{x}_{t-1},\mathbf{x}_0).
\]
By the Markov property, the conditioning on \(\mathbf{x}_0\) is redundant in the
one-step forward transitions, but keeping it visible will soon be useful when we
rewrite the ELBO using Bayes' rule.

Substituting the forward and reverse chains into the ELBO gives
\[
\begin{aligned}
\mathcal{L}_{\mathrm{ELBO}}
&=
\mathbb{E}_{q(\mathbf{x}_0)}
\mathbb{E}_{q(\mathbf{x}_{1:T}\mid \mathbf{x}_0)}
\!\left[
\log
\frac{
p(\mathbf{x}_T)\prod_{t=1}^{T} p_\theta(\mathbf{x}_{t-1}\mid \mathbf{x}_t)
}{
\prod_{t=1}^{T} q(\mathbf{x}_t\mid \mathbf{x}_{t-1},\mathbf{x}_0)
}
\right].
\end{aligned}
\]
Using the logarithm to turn products into sums, this becomes
\[
\begin{aligned}
\mathcal{L}_{\mathrm{ELBO}}
&=
\mathbb{E}_{q(\mathbf{x}_0)}
\mathbb{E}_{q(\mathbf{x}_{1:T}\mid \mathbf{x}_0)}
\!\left[
\log p(\mathbf{x}_T)
+
\sum_{t=1}^{T}
\log
\frac{
p_\theta(\mathbf{x}_{t-1}\mid \mathbf{x}_t)
}{
q(\mathbf{x}_t\mid \mathbf{x}_{t-1},\mathbf{x}_0)
}
\right].
\end{aligned}
\]

At first sight, this already looks like a step-wise objective.
However, it is not yet written in the most useful form.
The denominator here is still the \emph{forward} noising distribution, whereas
the model is trying to learn a \emph{reverse} denoising process.

\paragraph{What should the learned denoising process approximate?}

The learned reverse transition
\[
p_\theta(\mathbf{x}_{t-1}\mid \mathbf{x}_t)
\]
is the model's denoising rule: given a noisy state \(\mathbf{x}_t\), it should
produce a distribution over the slightly cleaner state \(\mathbf{x}_{t-1}\).
So the natural learning target is not the forward corruption rule
\(q(\mathbf{x}_t\mid \mathbf{x}_{t-1})\), but the corresponding
\emph{true reverse posterior} induced by the forward process.

It is important here to distinguish two closely related objects:

\begin{itemize}
    \item the \textbf{learned reverse process} or \textbf{learned denoising process}
    \[
    p_\theta(\mathbf{x}_{t-1}\mid \mathbf{x}_t),
    \]
    which is the model we parameterise and train;

    \item the \textbf{true reverse posterior induced by the forward process}
    \[
    q(\mathbf{x}_{t-1}\mid \mathbf{x}_t,\mathbf{x}_0),
    \]
    which is the exact one-step denoising distribution implied by the known
    noising mechanism, once the clean sample \(\mathbf{x}_0\) is given.
\end{itemize}

Why do we want the learned denoising process to match this posterior?
Because the forward process tells us exactly how real data are corrupted.
So, during training, it also tells us what the correct one-step reversal of that
corruption should be.
If the learned reverse transition imitates the true reverse posterior at every
time step, then the whole learned reverse chain will become capable of undoing
the forward noising process and recovering the data distribution.

A subtle but important point is that we do \emph{not} work directly with
\[
q(\mathbf{x}_{t-1}\mid \mathbf{x}_t).
\]
That object is the unconditional reverse distribution induced by the full data
distribution and is not directly tractable.
During training, however, the clean data sample \(\mathbf{x}_0\) is observed.
This means we can instead work with
\[
q(\mathbf{x}_{t-1}\mid \mathbf{x}_t,\mathbf{x}_0),
\]
which is the conditional posterior induced by the known forward process for that
specific observed sample.
This is precisely why we keep \(\mathbf{x}_0\) explicit: conditioning on the
observed data makes the relevant reverse distribution analytically tractable.

So the task of learning the reverse process can be stated more precisely as:
\begin{quote}
\emph{train the modelled denoising transition
\(p_\theta(\mathbf{x}_{t-1}\mid \mathbf{x}_t)\)
to match the true reverse posterior
\(q(\mathbf{x}_{t-1}\mid \mathbf{x}_t,\mathbf{x}_0)\)
at every diffusion step.}
\end{quote}

\subsection{Rewriting the ELBO with Bayes' rule}\begin{tcolorbox}[
  colback=blue!6,
  colframe=blue!55!black,
  title={A useful trick for exposing the true reverse posterior},
  boxrule=0.8pt,
  arc=2mm,
  left=2mm,
  right=2mm,
  top=1mm,
  bottom=1mm
]
The trick is to keep every term conditioned on the observed clean sample x0, and then
apply Bayes’ rule. Starting from the forward transition $q(\mathbf{x}_t\mid \mathbf{x}_{t-1},\mathbf{x}_0),$ conditioned on \(\mathbf{x}_0\), Bayes' rule gives:
\[
\begin{aligned}
q(\mathbf{x}_t\mid \mathbf{x}_{t-1},\mathbf{x}_0)
&=
\frac{
q(\mathbf{x}_{t-1},\mathbf{x}_t\mid \mathbf{x}_0)
}{
q(\mathbf{x}_{t-1}\mid \mathbf{x}_0)
}=
\frac{
q(\mathbf{x}_{t-1}\mid \mathbf{x}_t,\mathbf{x}_0)\,
q(\mathbf{x}_t\mid \mathbf{x}_0)
}{
q(\mathbf{x}_{t-1}\mid \mathbf{x}_0)
}.
\end{aligned}
\]
Rearranging gives the true reverse posterior:
\[
q(\mathbf{x}_{t-1}\mid \mathbf{x}_t,\mathbf{x}_0)
=
\frac{
q(\mathbf{x}_t\mid \mathbf{x}_{t-1},\mathbf{x}_0)\,
q(\mathbf{x}_{t-1}\mid \mathbf{x}_0)
}{
q(\mathbf{x}_t\mid \mathbf{x}_0)
}.
\]
\end{tcolorbox}

To bring this true reverse posterior into the ELBO, we apply Bayes' rule to the
forward transition:
\[
q(\mathbf{x}_t\mid \mathbf{x}_{t-1},\mathbf{x}_0)
=
\frac{
q(\mathbf{x}_{t-1}\mid \mathbf{x}_t,\mathbf{x}_0)\,
q(\mathbf{x}_t\mid \mathbf{x}_0)
}{
q(\mathbf{x}_{t-1}\mid \mathbf{x}_0)
}.
\]
Substituting this into the denominator of the ELBO gives
\[
\begin{aligned}
\mathcal{L}_{\mathrm{ELBO}}
&=
\mathbb{E}_{q(\mathbf{x}_0)}
\mathbb{E}_{q(\mathbf{x}_{1:T}\mid \mathbf{x}_0)}
\!\left[
\log p(\mathbf{x}_T)
+
\sum_{t=1}^{T}
\log
\frac{
p_\theta(\mathbf{x}_{t-1}\mid \mathbf{x}_t)\,
q(\mathbf{x}_{t-1}\mid \mathbf{x}_0)
}{
q(\mathbf{x}_{t-1}\mid \mathbf{x}_t,\mathbf{x}_0)\,
q(\mathbf{x}_t\mid \mathbf{x}_0)
}
\right].
\end{aligned}
\]
Now split the logarithm of the product into a sum:
\[
\begin{aligned}
\mathcal{L}_{\mathrm{ELBO}}
&=
\mathbb{E}_{q(\mathbf{x}_0)}
\mathbb{E}_{q(\mathbf{x}_{1:T}\mid \mathbf{x}_0)}
\!\left[
\log p(\mathbf{x}_T)
+
\sum_{t=1}^{T}
\log
\frac{
p_\theta(\mathbf{x}_{t-1}\mid \mathbf{x}_t)
}{
q(\mathbf{x}_{t-1}\mid \mathbf{x}_t,\mathbf{x}_0)
}
+
\sum_{t=1}^{T}
\log
\frac{
q(\mathbf{x}_{t-1}\mid \mathbf{x}_0)
}{
q(\mathbf{x}_t\mid \mathbf{x}_0)
}
\right].
\end{aligned}
\]

The second sum telescopes.
To see this explicitly,
\[
\begin{aligned}
\sum_{t=1}^{T}
\log
\frac{
q(\mathbf{x}_{t-1}\mid \mathbf{x}_0)
}{
q(\mathbf{x}_t\mid \mathbf{x}_0)
}
&=
\log
\frac{q(\mathbf{x}_0\mid \mathbf{x}_0)}{q(\mathbf{x}_1\mid \mathbf{x}_0)}
+
\log
\frac{q(\mathbf{x}_1\mid \mathbf{x}_0)}{q(\mathbf{x}_2\mid \mathbf{x}_0)}
+\cdots+
\log
\frac{q(\mathbf{x}_{T-1}\mid \mathbf{x}_0)}{q(\mathbf{x}_T\mid \mathbf{x}_0)}
\\[6pt]
&=
\log\!\left[
\frac{q(\mathbf{x}_0\mid \mathbf{x}_0)}{q(\mathbf{x}_1\mid \mathbf{x}_0)}
\cdot
\frac{q(\mathbf{x}_1\mid \mathbf{x}_0)}{q(\mathbf{x}_2\mid \mathbf{x}_0)}
\cdots
\frac{q(\mathbf{x}_{T-1}\mid \mathbf{x}_0)}{q(\mathbf{x}_T\mid \mathbf{x}_0)}
\right]
\\[6pt]
&=
\log\!\left[
\frac{q(\mathbf{x}_0\mid \mathbf{x}_0)}{q(\mathbf{x}_T\mid \mathbf{x}_0)}
\right] =
\log q(\mathbf{x}_0\mid \mathbf{x}_0)
-\log q(\mathbf{x}_T\mid \mathbf{x}_0).
\end{aligned}
\]
because all intermediate terms cancel pairwise.
% Since \(q(\mathbf{x}_0\mid \mathbf{x}_0)\) is a degenerate certainty term, it is
% constant and may be ignored for optimisation purposes.
The term \(q(\mathbf{x}_0\mid \mathbf{x}_0)\) is formally a degenerate
distribution concentrated at the observed value \(\mathbf{x}_0\), corresponding
to a Dirac delta rather than an ordinary density. Since it contains no learnable
parameters, it contributes only an additive constant to the objective and may be
ignored for optimisation purposes.
So the ELBO becomes
\[
\begin{aligned}
\mathcal{L}_{\mathrm{ELBO}}
&=
\mathbb{E}_{q(\mathbf{x}_0)}
\mathbb{E}_{q(\mathbf{x}_{1:T}\mid \mathbf{x}_0)}
\!\left[
\log p(\mathbf{x}_T)
-
\log q(\mathbf{x}_T\mid \mathbf{x}_0)
+
\sum_{t=1}^{T}
\log
\frac{
p_\theta(\mathbf{x}_{t-1}\mid \mathbf{x}_t)
}{
q(\mathbf{x}_{t-1}\mid \mathbf{x}_t,\mathbf{x}_0)
}
\right]
+\text{const}.
\end{aligned}
\]

\subsection{From the global ELBO to local learning terms}

We now reorganise the ELBO into the standard step-wise decomposition.
There are two main pieces to handle:
\begin{itemize}
    \item the \textbf{boundary term} involving the endpoint \(\mathbf{x}_T\);
    \item the \textbf{time-step terms} involving the learned denoising transitions
    \(p_\theta(\mathbf{x}_{t-1}\mid \mathbf{x}_t)\).
\end{itemize}

Starting from the expression obtained above, we first isolate the part involving only the endpoint \(\mathbf{x}_T\):
\[
\mathbb{E}_{q(\mathbf{x}_{1:T}\mid \mathbf{x}_0)}
\!\big[
\log p(\mathbf{x}_T)-\log q(\mathbf{x}_T\mid \mathbf{x}_0)
\big].
\]
Since this integrand depends only on \(\mathbf{x}_T\), all the other latent
variables \(\mathbf{x}_1,\dots,\mathbf{x}_{T-1}\) can be marginalised out under
the forward chain, leaving only the marginal \(q(\mathbf{x}_T\mid \mathbf{x}_0)\):
\[
\begin{aligned}
&\mathbb{E}_{q(\mathbf{x}_{1:T}\mid \mathbf{x}_0)}
\!\big[
\log p(\mathbf{x}_T)-\log q(\mathbf{x}_T\mid \mathbf{x}_0)
\big] \\[4pt]
&\qquad=
\mathbb{E}_{q(\mathbf{x}_T\mid \mathbf{x}_0)}
\!\big[
\log p(\mathbf{x}_T)-\log q(\mathbf{x}_T\mid \mathbf{x}_0)
\big] \\[4pt]
&\qquad=
-\mathbb{E}_{q(\mathbf{x}_T\mid \mathbf{x}_0)}
\!\left[
\log \frac{q(\mathbf{x}_T\mid \mathbf{x}_0)}{p(\mathbf{x}_T)}
\right] \\[4pt]
&\qquad=
-\mathrm{KL}\!\bigl(
q(\mathbf{x}_T\mid \mathbf{x}_0)\,\|\,p(\mathbf{x}_T)
\bigr).
\end{aligned}
\]
This is the \textbf{prior-matching term}: it compares the distribution reached
by the forward process at its final time step with the simple Gaussian prior
from which the learned reverse process begins.

We now turn to the sum over \(t\).
Here one must handle the inner expectation
\(\mathbb{E}_{q(\mathbf{x}_{1:T}\mid \mathbf{x}_0)}\) carefully.
The whole sum sits inside the expectation, but each summand
\[
\log
\frac{
p_\theta(\mathbf{x}_{t-1}\mid \mathbf{x}_t)
}{
q(\mathbf{x}_{t-1}\mid \mathbf{x}_t,\mathbf{x}_0)
}
\]
depends only on the local variables \(\mathbf{x}_{t-1}\) and \(\mathbf{x}_t\),
together with the observed sample \(\mathbf{x}_0\).
So for each fixed \(t\), all the other latent variables can again be
marginalised out.

For the special case \(t=1\), this gives
\[
\begin{aligned}
&\mathbb{E}_{q(\mathbf{x}_{1:T}\mid \mathbf{x}_0)}
\!\left[
\log
\frac{
p_\theta(\mathbf{x}_0\mid \mathbf{x}_1)
}{
q(\mathbf{x}_0\mid \mathbf{x}_1,\mathbf{x}_0)
}
\right] =
\mathbb{E}_{q(\mathbf{x}_1\mid \mathbf{x}_0)}
\!\left[
\log
\frac{
p_\theta(\mathbf{x}_0\mid \mathbf{x}_1)
}{
q(\mathbf{x}_0\mid \mathbf{x}_1,\mathbf{x}_0)
}
\right].
\end{aligned}
\]
Since \(q(\mathbf{x}_0\mid \mathbf{x}_1,\mathbf{x}_0)\) is degenerate once
\(\mathbf{x}_0\) is conditioned on, it corresponds to a point mass at the
observed value and contains no learnable parameters. It therefore contributes
only an additive constant, leaving the \textbf{reconstruction term}
\[
\mathbb{E}_{q(\mathbf{x}_1\mid \mathbf{x}_0)}
\big[
\log p_\theta(\mathbf{x}_0\mid \mathbf{x}_1)
\big].
\]

For the remaining terms \(t=2,\dots,T\), we do the same localisation of the
inner expectation:\[
\begin{aligned}
&\mathbb{E}_{q(\mathbf{x}_{1:T}\mid \mathbf{x}_0)}
\!\left[
\log
\frac{
p_\theta(\mathbf{x}_{t-1}\mid \mathbf{x}_t)
}{
q(\mathbf{x}_{t-1}\mid \mathbf{x}_t,\mathbf{x}_0)
}
\right] \\[6pt]
&\qquad=
\mathbb{E}_{q(\mathbf{x}_{t-1},\mathbf{x}_t\mid \mathbf{x}_0)}
\!\left[
\log
\frac{
p_\theta(\mathbf{x}_{t-1}\mid \mathbf{x}_t)
}{
q(\mathbf{x}_{t-1}\mid \mathbf{x}_t,\mathbf{x}_0)
}
\right] \\[6pt]
&\qquad=
\mathbb{E}_{q(\mathbf{x}_t\mid \mathbf{x}_0)}
\mathbb{E}_{q(\mathbf{x}_{t-1}\mid \mathbf{x}_t,\mathbf{x}_0)}
\!\left[
\log
\frac{
p_\theta(\mathbf{x}_{t-1}\mid \mathbf{x}_t)
}{
q(\mathbf{x}_{t-1}\mid \mathbf{x}_t,\mathbf{x}_0)
}
\right].
\end{aligned}
\]
Now recognise the inner expectation as the negative of a KL divergence:
\[
\begin{aligned}
&\mathbb{E}_{q(\mathbf{x}_{t-1}\mid \mathbf{x}_t,\mathbf{x}_0)}
\!\left[
\log
\frac{
p_\theta(\mathbf{x}_{t-1}\mid \mathbf{x}_t)
}{
q(\mathbf{x}_{t-1}\mid \mathbf{x}_t,\mathbf{x}_0)
}
\right] \\[4pt]
&\qquad=
-\mathbb{E}_{q(\mathbf{x}_{t-1}\mid \mathbf{x}_t,\mathbf{x}_0)}
\!\left[
\log
\frac{
q(\mathbf{x}_{t-1}\mid \mathbf{x}_t,\mathbf{x}_0)
}{
p_\theta(\mathbf{x}_{t-1}\mid \mathbf{x}_t)
}
\right] \\[4pt]
&\qquad=
-\mathrm{KL}\!\big(
q(\mathbf{x}_{t-1}\mid \mathbf{x}_t,\mathbf{x}_0)
\,\|\,\,
p_\theta(\mathbf{x}_{t-1}\mid \mathbf{x}_t)
\big).
\end{aligned}
\]
Therefore each intermediate time-step contribution becomes
\[
-
\mathbb{E}_{q(\mathbf{x}_t\mid \mathbf{x}_0)}
\Big[
\mathrm{KL}\!\big(
q(\mathbf{x}_{t-1}\mid \mathbf{x}_t,\mathbf{x}_0)
\,\|\,\,
p_\theta(\mathbf{x}_{t-1}\mid \mathbf{x}_t)
\big)
\Big].
\]

Putting the boundary term and the time-step terms together, the negative ELBO
can be written as
\[
\begin{aligned}
-\mathcal{L}_{\mathrm{ELBO}}
=
\mathbb{E}_{q(\mathbf{x}_0)}
\Big[
&\underbrace{
\mathrm{KL}\!\bigl(
q(\mathbf{x}_T\mid \mathbf{x}_0)\,\|\,p(\mathbf{x}_T)
\bigr)
}_{\text{prior-matching term}}
\\[4pt]
&+
\sum_{t=2}^{T}
\underbrace{
\mathbb{E}_{q(\mathbf{x}_t\mid \mathbf{x}_0)}
\Big[
\mathrm{KL}\!\big(
q(\mathbf{x}_{t-1}\mid \mathbf{x}_t,\mathbf{x}_0)
\,\|\,\,
p_\theta(\mathbf{x}_{t-1}\mid \mathbf{x}_t)
\big)
\Big]
}_{\text{step-wise denoising terms}}
\\[4pt]
&-
\underbrace{
\mathbb{E}_{q(\mathbf{x}_1\mid \mathbf{x}_0)}
\big[
\log p_\theta(\mathbf{x}_0\mid \mathbf{x}_1)
\big]
}_{\text{reconstruction term}}
\Big].
\end{aligned}
\]

This decomposition separates three roles: prior matching at the noisy endpoint,
local reverse-transition matching at intermediate steps, and reconstruction at
the data end. Figure~\ref{fig:ddpm-elbo-local-decomposition} shows how these
terms sit along the forward and reverse chains.

\input{figures/ch04/fig_ch04_02_elbo_local_decomposition}

\noindent
The key consequence is that learning reduces to matching, at each time step,
the learned denoising transition
$p_\theta(\mathbf{x}_{t-1}\mid \mathbf{x}_t)$ to the true reverse posterior
$q(\mathbf{x}_{t-1}\mid \mathbf{x}_t,\mathbf{x}_0)$. The next step is
therefore to work out the analytical form of this posterior.

\section{From the True Reverse Posterior to DDPM Learning}

The previous sections showed that the central trainable terms in the diffusion
ELBO compare the learned denoising transition
\[
p_\theta(\mathbf{x}_{t-1}\mid \mathbf{x}_t)
\]
with the true reverse posterior
\[
q(\mathbf{x}_{t-1}\mid \mathbf{x}_t,\mathbf{x}_0).
\]
The purpose of this section is to connect these two views:
first by identifying the exact Gaussian reverse posterior implied by the forward
process, and then by showing how that posterior leads to the practical DDPM
learning objective and sampling procedure.

\subsection{The true reverse posterior from known forward quantities}

From the previous section, we have the identity
\[
q(\mathbf{x}_{t-1}\mid \mathbf{x}_t,\mathbf{x}_0)
=
\frac{
q(\mathbf{x}_t\mid \mathbf{x}_{t-1},\mathbf{x}_0)\,
q(\mathbf{x}_{t-1}\mid \mathbf{x}_0)
}{
q(\mathbf{x}_t\mid \mathbf{x}_0)
}.
\]
This is useful because every term on the right-hand side is determined by the
forward process, which is already known. In particular,
\[
q(\mathbf{x}_t\mid \mathbf{x}_{t-1},\mathbf{x}_0)
=
q(\mathbf{x}_t\mid \mathbf{x}_{t-1})
=
\mathcal{N}\!\bigl(
\mathbf{x}_t \mid
\sqrt{\alpha_t}\,\mathbf{x}_{t-1},
(1-\alpha_t)\mathbf{I}
\bigr),
\]
where \(\alpha_t=1-\beta_t\). From the earlier closed-form forward relation, we
also know
\[
q(\mathbf{x}_{t-1}\mid \mathbf{x}_0)
=
\mathcal{N}\!\bigl(
\mathbf{x}_{t-1}\mid
\sqrt{\bar{\alpha}_{t-1}}\,\mathbf{x}_0,
(1-\bar{\alpha}_{t-1})\mathbf{I}
\bigr),
\]
and
\[
q(\mathbf{x}_t\mid \mathbf{x}_0)
=
\mathcal{N}\!\bigl(
\mathbf{x}_t\mid
\sqrt{\bar{\alpha}_t}\,\mathbf{x}_0,
(1-\bar{\alpha}_t)\mathbf{I}
\bigr).
\]

So the true reverse posterior is expressed entirely in terms of Gaussian
quantities from the forward diffusion process. Because the terms above are Gaussian, the posterior in
\(\mathbf{x}_{t-1}\) must also be Gaussian.
So we may write
\[
q(\mathbf{x}_{t-1}\mid \mathbf{x}_t,\mathbf{x}_0)
=
\mathcal{N}\!\bigl(
\mathbf{x}_{t-1}\mid
\tilde{\boldsymbol{\mu}}_t(\mathbf{x}_t,\mathbf{x}_0),
\tilde{\beta}_t\mathbf{I}
\bigr),
\]
for some mean \(\tilde{\boldsymbol{\mu}}_t(\mathbf{x}_t,\mathbf{x}_0)\) and
variance coefficient \(\tilde{\beta}_t\).

Using the same completing-the-square idea introduced earlier, and derived in
Appendix~\ref{app:diffusion-posterior}, the true reverse posterior takes the
following Gaussian form:
% \[
% q(\mathbf{x}_{t-1}\mid \mathbf{x}_t,\mathbf{x}_0)
% =
% \mathcal{N}\!\Bigl(
% \mathbf{x}_{t-1}\mid
% \tilde{\boldsymbol{\mu}}_t(\mathbf{x}_t,\mathbf{x}_0),
% \tilde{\beta}_t\mathbf{I}
% \Bigr),
% \]
with
\[
\tilde{\boldsymbol{\mu}}_t(\mathbf{x}_t,\mathbf{x}_0)
=
\frac{
\sqrt{\bar{\alpha}_{t-1}}\,\beta_t
}{
1-\bar{\alpha}_t
}\,\mathbf{x}_0
+
\frac{
\sqrt{\alpha_t}(1-\bar{\alpha}_{t-1})
}{
1-\bar{\alpha}_t
}\,\mathbf{x}_t,
\]
and
\[
\tilde{\beta}_t
=
\frac{
1-\bar{\alpha}_{t-1}
}{
1-\bar{\alpha}_t
}\,\beta_t.
\]

This result is worth interpreting.
The posterior mean
\(\tilde{\boldsymbol{\mu}}_t(\mathbf{x}_t,\mathbf{x}_0)\)
is a weighted combination of two pieces of information:
the clean sample \(\mathbf{x}_0\), which tells us where the trajectory began,
and the current noisy state \(\mathbf{x}_t\), which tells us where we are at
time \(t\).
So the true reverse posterior identifies the most plausible slightly cleaner
state by combining knowledge of both the clean origin and the current noisy
observation.

The posterior variance \(\tilde{\beta}_t\mathbf{I}\) is also fixed by the
forward diffusion schedule.
That is, once the forward noising process is chosen, the uncertainty of the
corresponding reverse step is determined as well.
So, from the ELBO viewpoint, the reverse model is not being asked to match an
arbitrary denoising distribution, but a specific Gaussian target induced by the
known forward process.

Figure~\ref{fig:ddpm-true-reverse-posterior} illustrates this target as the
Gaussian fusion of the two known forward-process factors.

\input{figures/ch04/fig_ch04_03_true_reverse_posterior}

Although this posterior is exact, it still depends explicitly on the clean data
sample \(\mathbf{x}_0\).
That is acceptable during training, when \(\mathbf{x}_0\) is known, but it is
not usable directly at generation time, when \(\mathbf{x}_0\) is precisely the
unknown object we wish to create.

\subsection{Reparameterising the posterior through the noise variable}

The key observation is that the forward process already provides a direct
relationship between the clean sample \(\mathbf{x}_0\), the noisy sample
\(\mathbf{x}_t\), and a standard Gaussian noise variable:
\[
\mathbf{x}_t
=
\sqrt{\bar{\alpha}_t}\,\mathbf{x}_0
+
\sqrt{1-\bar{\alpha}_t}\,\boldsymbol{\epsilon},
\qquad
\boldsymbol{\epsilon}\sim \mathcal{N}(\mathbf{0},\mathbf{I}).
\]

Solving this for \(\mathbf{x}_0\) gives
\[
\mathbf{x}_0
=
\frac{1}{\sqrt{\bar{\alpha}_t}}
\Bigl(
\mathbf{x}_t-\sqrt{1-\bar{\alpha}_t}\,\boldsymbol{\epsilon}
\Bigr).
\]
We now substitute this into the posterior mean
\(\tilde{\boldsymbol{\mu}}_t(\mathbf{x}_t,\mathbf{x}_0)\) in order to remove
its explicit dependence on \(\mathbf{x}_0\).

Starting from
\[
\tilde{\boldsymbol{\mu}}_t(\mathbf{x}_t,\mathbf{x}_0)
=
\frac{
\sqrt{\bar{\alpha}_{t-1}}\,\beta_t
}{
1-\bar{\alpha}_t
}\,\mathbf{x}_0
+
\frac{
\sqrt{\alpha_t}(1-\bar{\alpha}_{t-1})
}{
1-\bar{\alpha}_t
}\,\mathbf{x}_t,
\]
substituting the expression for \(\mathbf{x}_0\) yields
\[
\begin{aligned}
\tilde{\boldsymbol{\mu}}_t(\mathbf{x}_t,\boldsymbol{\epsilon})
&=
\frac{
\sqrt{\bar{\alpha}_{t-1}}\,\beta_t
}{
1-\bar{\alpha}_t
}
\cdot
\frac{1}{\sqrt{\bar{\alpha}_t}}
\Bigl(
\mathbf{x}_t-\sqrt{1-\bar{\alpha}_t}\,\boldsymbol{\epsilon}
\Bigr)
+
\frac{
\sqrt{\alpha_t}(1-\bar{\alpha}_{t-1})
}{
1-\bar{\alpha}_t
}\,\mathbf{x}_t.
\end{aligned}
\]

After simplifying the coefficients, this becomes
\[
\tilde{\boldsymbol{\mu}}_t(\mathbf{x}_t,\boldsymbol{\epsilon})
=
\frac{1}{\sqrt{\alpha_t}}
\left(
\mathbf{x}_t
-
\frac{\beta_t}{\sqrt{1-\bar{\alpha}_t}}\,
\boldsymbol{\epsilon}
\right).
\]

\begin{tcolorbox}[
  colback=blue!6,
  colframe=blue!55!black,
  title={Algebra behind the coefficient simplification},
  boxrule=0.8pt,
  arc=2mm,
  left=2mm,
  right=2mm,
  top=1mm,
  bottom=1mm
]
Using
\[
\bar{\alpha}_t=\alpha_t\bar{\alpha}_{t-1},
\qquad
\frac{\sqrt{\bar{\alpha}_{t-1}}}{\sqrt{\bar{\alpha}_t}}
=
\frac{1}{\sqrt{\alpha_t}},
\]
the coefficient of \(\mathbf{x}_t\) becomes
\[
\frac{\beta_t}{\sqrt{\alpha_t}(1-\bar{\alpha}_t)}
+
\frac{\sqrt{\alpha_t}(1-\bar{\alpha}_{t-1})}{1-\bar{\alpha}_t}
=
\frac{
\beta_t+\alpha_t(1-\bar{\alpha}_{t-1})
}{
\sqrt{\alpha_t}(1-\bar{\alpha}_t)
}.
\]
Since
\[
\beta_t=1-\alpha_t,
\qquad
\bar{\alpha}_t=\alpha_t\bar{\alpha}_{t-1},
\]
we have
\[
\beta_t+\alpha_t(1-\bar{\alpha}_{t-1})
=
(1-\alpha_t)+\alpha_t-\alpha_t\bar{\alpha}_{t-1}
=
1-\bar{\alpha}_t,
\]
so this simplifies to
\[
\frac{1-\bar{\alpha}_t}{\sqrt{\alpha_t}(1-\bar{\alpha}_t)}
=
\frac{1}{\sqrt{\alpha_t}}.
\]
Similarly, the coefficient of \(\boldsymbol{\epsilon}\) becomes
\[
\frac{
\sqrt{\bar{\alpha}_{t-1}}\,\beta_t
}{
\sqrt{\bar{\alpha}_t}\sqrt{1-\bar{\alpha}_t}
}
=
\frac{\beta_t}{\sqrt{\alpha_t}\sqrt{1-\bar{\alpha}_t}}.
\]
\end{tcolorbox}

So the true reverse posterior may be rewritten as
\[
\boxed{
\begin{aligned}
q(\mathbf{x}_{t-1}\mid \mathbf{x}_t,\boldsymbol{\epsilon})
&=
\mathcal{N}\!\left(
\mathbf{x}_{t-1}\mid
\frac{1}{\sqrt{\alpha_t}}
\left(
\mathbf{x}_t
-
\frac{\beta_t}{\sqrt{1-\bar{\alpha}_t}}\,
\boldsymbol{\epsilon}
\right),
\tilde{\beta}_t\mathbf{I}
\right).
\end{aligned}
}
\]

This form is extremely important.
It shows that the reverse step can be characterised through the current noisy
sample \(\mathbf{x}_t\) and the aggregate noise \(\boldsymbol{\epsilon}\) that
produced it.
So instead of trying to recover \(\mathbf{x}_0\) directly, the model can aim to
predict the underlying noise.

\subsection{Parameterising the learned reverse process and deriving the practical DDPM objective}

This leads to the standard DDPM parameterisation.
Instead of using the true noise \(\boldsymbol{\epsilon}\), which is unknown to
the model, we introduce a neural network
\[
\boldsymbol{\epsilon}_\theta(\mathbf{x}_t,t)
\]
to predict it from the noisy sample \(\mathbf{x}_t\) and the time step \(t\).

The learned reverse transition is then defined as
\[
p_\theta(\mathbf{x}_{t-1}\mid \mathbf{x}_t)
=
\mathcal{N}\!\bigl(
\mathbf{x}_{t-1}\mid
\boldsymbol{\mu}_\theta(\mathbf{x}_t,t),
\tilde{\beta}_t\mathbf{I}
\bigr),
\]
with mean
\[
\boldsymbol{\mu}_\theta(\mathbf{x}_t,t)
=
\frac{1}{\sqrt{\alpha_t}}
\left(
\mathbf{x}_t
-
\frac{\beta_t}{\sqrt{1-\bar{\alpha}_t}}\,
\boldsymbol{\epsilon}_\theta(\mathbf{x}_t,t)
\right).
\]

This has a clear interpretation:
if the network predicts the true noise accurately, then the learned reverse mean
matches the exact reverse posterior mean, and so the learned denoising process
matches the correct local reverse step.

\bigskip

Now recall that the central trainable terms in the ELBO are
\[
\mathbb{E}_{q(\mathbf{x}_t\mid \mathbf{x}_0)}
\Big[
\mathrm{KL}\!\big(
q(\mathbf{x}_{t-1}\mid \mathbf{x}_t,\mathbf{x}_0)
\,\|\,\,
p_\theta(\mathbf{x}_{t-1}\mid \mathbf{x}_t)
\big)
\Big].
\]
Now both distributions are Gaussian with the same variance
\(\tilde{\beta}_t\mathbf{I}\).
So their KL divergence reduces to a squared difference between means:
\[
\mathrm{KL}\!\big(
q(\mathbf{x}_{t-1}\mid \mathbf{x}_t,\mathbf{x}_0)
\,\|\,\,
p_\theta(\mathbf{x}_{t-1}\mid \mathbf{x}_t)
\big)
\propto
\bigl\|
\tilde{\boldsymbol{\mu}}_t(\mathbf{x}_t,\boldsymbol{\epsilon})
-
\boldsymbol{\mu}_\theta(\mathbf{x}_t,t)
\bigr\|^2.
\]
Substituting the two mean expressions gives
\[
\tilde{\boldsymbol{\mu}}_t(\mathbf{x}_t,\boldsymbol{\epsilon})
-
\boldsymbol{\mu}_\theta(\mathbf{x}_t,t)
=
\frac{\beta_t}{\sqrt{\alpha_t}\sqrt{1-\bar{\alpha}_t}}
\Bigl(
\boldsymbol{\epsilon}
-
\boldsymbol{\epsilon}_\theta(\mathbf{x}_t,t)
\Bigr).
\]
So, up to a time-dependent scaling factor, the KL term becomes a mean-squared
error in noise prediction:
\[
\mathrm{KL}\!\big(
q(\mathbf{x}_{t-1}\mid \mathbf{x}_t,\mathbf{x}_0)
\,\|\,\,
p_\theta(\mathbf{x}_{t-1}\mid \mathbf{x}_t)
\big)
\propto
\bigl\|
\boldsymbol{\epsilon}
-
\boldsymbol{\epsilon}_\theta(\mathbf{x}_t,t)
\bigr\|^2.
\]

This is the central simplification in DDPMs: the local variational terms reduce,
up to a time-dependent coefficient, to a regression problem in the added
Gaussian noise. The exact ELBO therefore induces a weighted noise-prediction
mean-squared error. In practice, the commonly used simplified objective of
Ho et al.~\cite{ho2020denoising} drops this time-dependent weighting and uses
\[
\boxed{
\mathcal{L}_{\mathrm{simple}}
=
\mathbb{E}_{t,\mathbf{x}_0,\boldsymbol{\epsilon}}
\Bigl[
\bigl\|
\boldsymbol{\epsilon}
-
\boldsymbol{\epsilon}_\theta(\mathbf{x}_t,t)
\bigr\|^2
\Bigr]
}
\]
where $\mathbf{x}_t
=
\sqrt{\bar{\alpha}_t}\,\mathbf{x}_0
+
\sqrt{1-\bar{\alpha}_t}\,\boldsymbol{\epsilon}.$
Figure~\ref{fig:ddpm-noise-prediction-bridge} summarises the algebraic bridge
from local posterior matching to this practical noise-prediction loss.

\input{figures/ch04/fig_ch04_04_noise_prediction_bridge}

\subsection{How training and sampling work}

The training loop now becomes conceptually simple.
For each clean data sample \(\mathbf{x}_0\), we:
\begin{enumerate}
    \item sample a time step \(t\);
    \item sample Gaussian noise \(\boldsymbol{\epsilon}\sim\mathcal{N}(\mathbf{0},\mathbf{I})\);
    \item construct the noisy state
    \[
    \mathbf{x}_t
    =
    \sqrt{\bar{\alpha}_t}\,\mathbf{x}_0
    +
    \sqrt{1-\bar{\alpha}_t}\,\boldsymbol{\epsilon};
    \]
    \item train the network to predict \(\boldsymbol{\epsilon}\) from
    \((\mathbf{x}_t,t)\).
\end{enumerate}
Because \(\mathbf{x}_t\) can be sampled directly from \(\mathbf{x}_0\), training
does not require simulating the full forward chain step by step.

Once trained, the model generates data by running the learned reverse process
iteratively.
We begin from
\[
\mathbf{x}_T\sim\mathcal{N}(\mathbf{0},\mathbf{I}),
\]
and then for \(t=T,T-1,\dots,1\) sample
\[
\mathbf{x}_{t-1}\sim p_\theta(\mathbf{x}_{t-1}\mid \mathbf{x}_t).
\]
At each step, the model uses its predicted noise
\(\boldsymbol{\epsilon}_\theta(\mathbf{x}_t,t)\) to compute the reverse mean and
remove a small amount of corruption.
After repeating this over all time steps, the final state \(\mathbf{x}_0\) is
obtained as the generated sample.

\bigskip
\noindent
This completes the main probabilistic story of DDPMs in discrete time.
The forward process is fixed, the reverse process is learned, the ELBO reduces
to local denoising targets, and the final practical objective becomes noise
prediction.

\section*{Summary and References}
\addcontentsline{toc}{section}{Summary and References}

This chapter introduced Denoising Diffusion Probabilistic Models (DDPMs) from a
probabilistic latent-variable perspective.
Unlike PPCA and the VAE, where a single latent variable plays the central role,
diffusion models work with a full time-indexed latent trajectory
\(\mathbf{x}_{1:T}\), generated by progressively corrupting data through a fixed
Gaussian Markov process.

We first showed how the forward process transforms a clean sample
\(\mathbf{x}_0\) into increasingly noisy states, and derived the closed-form
relation between \(\mathbf{x}_t\) and \(\mathbf{x}_0\).
We then turned to the reverse process, which is the part that must be learned,
and used the ELBO framework to express diffusion as a latent-variable
generative model with an intractable marginal likelihood.

The ELBO was then decomposed into local denoising terms, revealing that the key
trainable quantity at each time step is the mismatch between the learned reverse
transition \(p_\theta(\mathbf{x}_{t-1}\mid \mathbf{x}_t)\) and the true reverse
posterior \(q(\mathbf{x}_{t-1}\mid \mathbf{x}_t,\mathbf{x}_0)\).
Because the forward process is Gaussian, this posterior is also Gaussian and
can be derived in closed form.
However, since it depends explicitly on the clean sample \(\mathbf{x}_0\), it
cannot be used directly at generation time.

Finally, by reparameterising the posterior through the noise variable, we showed
how diffusion learning reduces to noise prediction.
This yields the practical DDPM objective: train a neural network to predict the
Gaussian noise that produced a noisy sample at a given time step, and then use
that learned predictor to iteratively denoise pure Gaussian noise into a data
sample.
In this way, DDPMs extend the probabilistic latent-variable line developed in
earlier chapters into a sequential generative framework built on gradual
corruption and reconstruction.

\medskip
\noindent

The diffusion-based generative modelling framework was introduced by
Sohl-Dickstein et al.~\cite{sohl2015deep}, who framed generation as the learned
reversal of a gradual noising process inspired by nonequilibrium
thermodynamics. This work established several core ingredients that remain
central to DDPMs: a fixed forward corruption process, a learned reverse process,
and a sequence of intermediate latent states linking data to noise.

The modern DDPM formulation was developed by Ho, Jain, and
Abbeel~\cite{ho2020denoising}, who showed how this sequential latent-variable
model could be trained effectively through a variational objective that
simplifies to noise prediction. Later refinements, such as those by Nichol and
Dhariwal~\cite{nichol2021improved}, improved the practical training and
sampling behaviour of DDPMs. The score-based and continuous-time viewpoints
developed later in the book provide a complementary interpretation of the same
denoising principle, but the focus of this chapter is the discrete
latent-variable construction of DDPMs.

%% file: figures/ch04/chapter4_ddpm_at_a_glance.tex
\begin{center}
\begin{tikzpicture}[
  >=Latex,
  every node/.style={font=\small},
  state/.style={draw,rounded corners=2pt,minimum width=1.65cm,minimum height=0.82cm,align=center},
  wide/.style={draw,rounded corners=2pt,text width=4.2cm,minimum height=1.02cm,align=center,inner sep=6pt},
  flow/.style={->,thick},
  rev/.style={->,thick,dashed}
]
\node[font=\normalsize\bfseries] at (6.3,2.0) {DDPM at a glance};

\node[state] (x0) at (0,0.65) {$\mathbf{x}_0$\\clean data};
\node[state] (x1) at (3.0,0.65) {$\mathbf{x}_1$};
\node at (5.0,0.65) {$\cdots$};
\node[state] (xt) at (7.0,0.65) {$\mathbf{x}_t$};
\node at (9.0,0.65) {$\cdots$};
\node[state] (xT) at (12.0,0.65) {$\mathbf{x}_T$\\Gaussian noise};

\draw[flow] (x0) -- node[above] {\scriptsize fixed $q$} (x1);
\draw[flow] (x1) -- (4.45,0.65);
\draw[flow] (5.55,0.65) -- (xt);
\draw[flow] (xt) -- (8.45,0.65);
\draw[flow] (9.55,0.65) -- node[above] {\scriptsize add noise} (xT);

\draw[rev] (xT.south) to[out=-135,in=-45] node[midway,above] {\scriptsize learned reverse $p_\theta$} (xt.south);
\draw[rev] (xt.south) to[out=-135,in=-45] (x1.south);
\draw[rev] (x1.south) to[out=-135,in=-45] node[below] {\scriptsize denoise} (x0.south);

\node[wide] (shortcut) at (3.25,-2.15) {
\textbf{Training shortcut}\\[2pt]
sample $t,\boldsymbol{\epsilon}$ and jump directly to
$\displaystyle
\mathbf{x}_t=
\sqrt{\bar{\alpha}_t}\mathbf{x}_0+
\sqrt{1-\bar{\alpha}_t}\boldsymbol{\epsilon}$
};

\node[wide] (learn) at (9.2,-2.15) {
\textbf{Learning}\\[2pt]
predict $\boldsymbol{\epsilon}$ from $(\mathbf{x}_t,t)$ so that
$p_\theta(\mathbf{x}_{t-1}\mid\mathbf{x}_t)$ matches the local reverse target
};

\draw[flow] (shortcut) -- (learn);
\end{tikzpicture}
\end{center}

%% file: figures/ch04/fig_ch04_01_forward_diffusion.tex
\begin{figure}[H]
\centering
\includegraphics[width=0.98\textwidth]{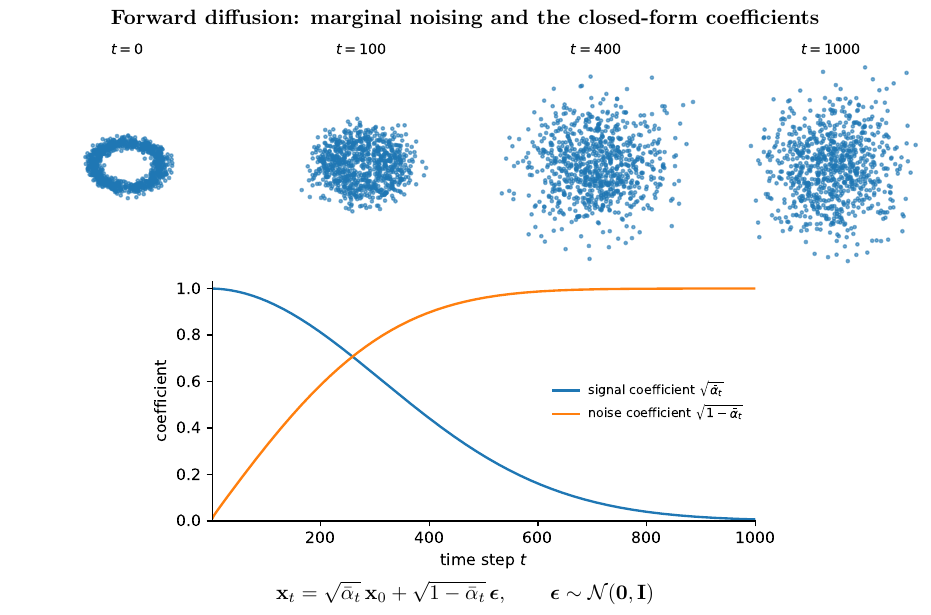}
\caption{\textbf{Forward diffusion and its closed-form noising relation.}
The upper row shows marginal samples at selected diffusion times under the
illustrative linear noise schedule. As $t$ increases, the data structure is
progressively overwhelmed by Gaussian noise, while the lower panel shows the
signal coefficient decreasing as the noise coefficient increases. The
snapshots are sampled directly from $q(\mathbf{x}_t\mid\mathbf{x}_0)$, so they
represent distributions at different times rather than one realised Markov
trajectory.}
\label{fig:ddpm-forward-closed-form}
\end{figure}

%% file: figures/ch04/fig_ch04_02_elbo_local_decomposition.tex
\begin{figure}[H]
\centering
\begin{tikzpicture}[
  >=Latex,
  every node/.style={font=\small},
  state/.style={draw,circle,minimum size=0.76cm,inner sep=1pt},
  term/.style={draw,rounded corners=2pt,align=center,inner sep=5pt},
  flow/.style={->,thick},
  reverse/.style={->,thick,dashed}
]
\node[font=\normalsize\bfseries] at (6.1,3.0)
  {The diffusion ELBO decomposes along the chain};

\node[state] (x0) at (0,1.35) {$x_0$};
\node[state] (x1) at (2.15,1.35) {$x_1$};
\node at (3.75,1.35) {$\cdots$};
\node[state] (xtm1) at (5.7,1.35) {$x_{t-1}$};
\node[state] (xt) at (8.15,1.35) {$x_t$};
\node at (9.8,1.35) {$\cdots$};
\node[state] (xT) at (12.0,1.35) {$x_T$};

% Forward chain
\draw[flow] (x0) -- node[above] {\scriptsize fixed $q$} (x1);
\draw[flow] (x1) -- (3.2,1.35);
\draw[flow] (4.3,1.35) -- (xtm1);
\draw[flow] (xtm1) -- node[above] {\scriptsize $q(x_t\mid x_{t-1})$} (xt);
\draw[flow] (xt) -- (9.25,1.35);
\draw[flow] (10.35,1.35) -- (xT);

% Learned reverse chain
\draw[reverse] (x1) -- node[below] {\scriptsize $p_\theta$} (x0);
\draw[reverse] (xt) -- node[below] {\scriptsize $p_\theta(x_{t-1}\mid x_t)$} (xtm1);
\draw[reverse] (xT) -- node[below] {\scriptsize learned reverse} (10.35,1.35);

\node[term,text width=3.45cm] (recon) at (1.1,-1.15) {
\textbf{Reconstruction}\\[2pt]
$\displaystyle
-\mathbb E_{q(x_1\mid x_0)}
[\log p_\theta(x_0\mid x_1)]$
};

\node[term,text width=5.15cm] (local) at (7.0,-1.35) {
\textbf{Intermediate steps, $t=2,\ldots,T$}\\[2pt]
$\displaystyle
\begin{aligned}
&\mathbb E_{q(x_t\mid x_0)}\Big[
\mathrm{KL}\!\big(\\[-1pt]
&q(x_{t-1}\mid x_t,x_0)
\,\|\,p_\theta(x_{t-1}\mid x_t)
\big)\Big]
\end{aligned}$
};

\node[term,text width=3.85cm] (prior) at (10.55,-3.65) {
\textbf{Endpoint / prior matching}\\[2pt]
$\displaystyle
\mathrm{KL}\!\big(
q(x_T\mid x_0)\,\|\,p(x_T)
\big)$
};

\draw[flow] (x0.south) to[out=-90,in=90] (recon.north west);
\draw[flow] (xtm1.south) to[out=-90,in=90] (local.north west);
\draw[flow] (xt.south) to[out=-90,in=90] (local.north east);
\draw[flow] (xT.south) to[out=-90,in=90] (prior.north);

\node[font=\scriptsize,align=center,text width=11.5cm] at (5.8,-5.0) {
The global latent-trajectory objective becomes a boundary term,
a reconstruction term, and a sum of local reverse-transition matching terms.
};

\end{tikzpicture}
\caption{\textbf{From the diffusion ELBO to local denoising terms.}
The fixed forward chain $q$ creates the latent trajectory, while $p_\theta$
models transitions in the reverse direction. Reorganising the negative ELBO
gives endpoint prior matching, data reconstruction, and intermediate KL terms
comparing each learned reverse transition with the exact posterior
$q(\mathbf{x}_{t-1}\mid\mathbf{x}_t,\mathbf{x}_0)$.}
\label{fig:ddpm-elbo-local-decomposition}
\end{figure}

%% file: figures/ch04/fig_ch04_03_true_reverse_posterior.tex
\begin{figure}[H]
\centering
\includegraphics[width=0.8\textwidth]{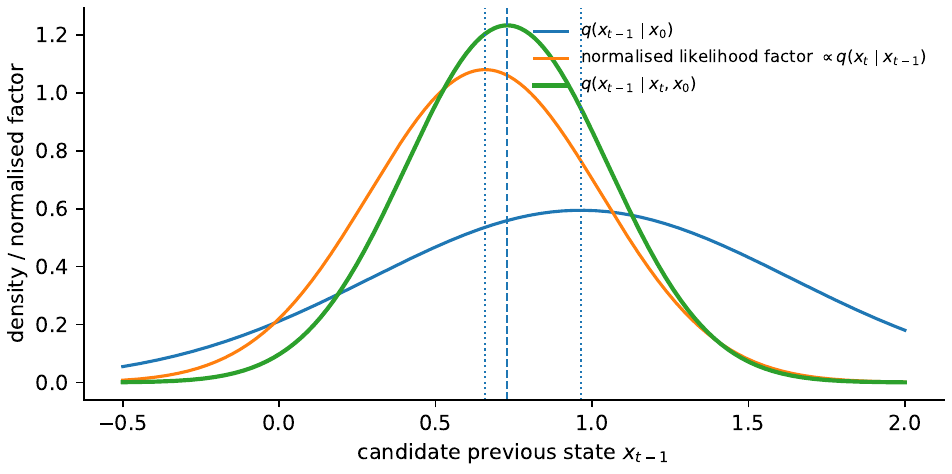}
\caption{\textbf{The exact reverse posterior as a Gaussian fusion.}
For fixed $\mathbf{x}_0$ and $\mathbf{x}_t$, the forward marginal
$q(\mathbf{x}_{t-1}\mid\mathbf{x}_0)$ and the transition likelihood
$q(\mathbf{x}_t\mid\mathbf{x}_{t-1})$, viewed as a function of
$\mathbf{x}_{t-1}$, provide two Gaussian factors. Their product gives
$q(\mathbf{x}_{t-1}\mid\mathbf{x}_t,\mathbf{x}_0)$. The one-dimensional plot
visualises the same componentwise calculation used in the isotropic
multivariate case.}
\label{fig:ddpm-true-reverse-posterior}
\end{figure}

%% file: figures/ch04/fig_ch04_04_noise_prediction_bridge.tex
\begin{figure}[H]
\centering
\begin{tikzpicture}[
  >=Latex,
  every node/.style={font=\small},
  box/.style={
    draw,
    rounded corners=2pt,
    align=center,
    inner sep=6pt,
    text width=5.45cm,
    minimum height=1.38cm
  },
  narrow/.style={
    draw,
    rounded corners=2pt,
    align=center,
    inner sep=6pt,
    text width=4.15cm,
    minimum height=1.15cm
  },
  flow/.style={->,thick}
]

\node[box] (true) at (2.9,1.45) {
\textbf{Exact reverse-posterior mean}\\[3pt]
$\displaystyle
\tilde{\boldsymbol{\mu}}_t
=
\frac{1}{\sqrt{\alpha_t}}
\left(
\mathbf{x}_t-
\frac{\beta_t}{\sqrt{1-\bar{\alpha}_t}}
\,\boldsymbol{\epsilon}
\right)$
};

\node[box] (learned) at (9.1,1.45) {
\textbf{Learned reverse mean}\\[3pt]
$\displaystyle
\boldsymbol{\mu}_{\theta}
=
\frac{1}{\sqrt{\alpha_t}}
\left(
\mathbf{x}_t-
\frac{\beta_t}{\sqrt{1-\bar{\alpha}_t}}
\,\boldsymbol{\epsilon}_{\theta}(\mathbf{x}_t,t)
\right)$
};

\draw[flow] (true) -- (learned);

\node[narrow] (samevar) at (2.9,-0.95) {
\textbf{Shared fixed variance}\\[2pt]
$q:\ \tilde{\beta}_t\mathbf I
\qquad
p_\theta:\ \tilde{\beta}_t\mathbf I$
};

\node[narrow] (meanmse) at (9.1,-0.95) {
\textbf{Gaussian KL becomes\\mean matching}\\[2pt]
$\displaystyle
\mathrm{KL}(q\,\|\,p_\theta)
\propto
\|\tilde{\boldsymbol{\mu}}_t-\boldsymbol{\mu}_\theta\|^2$
};

\draw[flow] (true.south) -- (samevar.north);
\draw[flow] (learned.south) -- (meanmse.north);
\draw[flow] (samevar) -- (meanmse);

\node[box] (noise) at (6.0,-3.45) {
\textbf{Shared coefficients leave a time-dependent scale}\\[3pt]
$\displaystyle
\|\tilde{\boldsymbol{\mu}}_t-\boldsymbol{\mu}_\theta\|^2
\propto
\|\boldsymbol{\epsilon}
-\boldsymbol{\epsilon}_{\theta}(\mathbf{x}_t,t)\|^2$
};

\draw[flow] (meanmse) -- (noise);

\node[narrow] (simple) at (6.0,-5.95) {
\textbf{Practical DDPM objective}\\[3pt]
$\displaystyle
\mathcal L_{\mathrm{simple}}
=
\mathbb E
\left[
\|\boldsymbol{\epsilon}
-\boldsymbol{\epsilon}_{\theta}(\mathbf{x}_t,t)\|^2
\right]$
};

\draw[flow] (noise) -- (simple);

\end{tikzpicture}

\caption{\textbf{From exact posterior matching to noise prediction.}
Writing the exact and learned reverse means in the same form, with the same
fixed variance, turns the local Gaussian KL into squared mean matching. Their
shared coefficients reduce this to a time-weighted squared error between the
true noise $\boldsymbol{\epsilon}$ and its prediction
$\boldsymbol{\epsilon}_\theta(\mathbf{x}_t,t)$. The commonly used
$\mathcal{L}_{\mathrm{simple}}$ objective drops this time-dependent weighting.}
\label{fig:ddpm-noise-prediction-bridge}
\end{figure}

%% file: chapters/cht_diff_fund.tex
\chapter{Calculus Foundations for Continuous Time Generative Modelling}

\section*{Overview}
\addcontentsline{toc}{section}{Overview}

In the previous chapter, we studied denoising diffusion probabilistic models
(DDPMs) in discrete time, where generation is described as a sequence of finite
noising and denoising steps. That perspective is powerful, but it also raises a
natural next question: what happens when the step size becomes smaller and
smaller, so that the evolution is better viewed as continuous in time?

This question is not unique to diffusion models. More generally, many methods in machine learning and generative modelling can be
viewed through repeated local updates of the form
\[
x_{t+\Delta t}\approx x_t+\text{small update}.
\]
A more concrete example is
\[
x_{t+\Delta t}=x_t+\alpha f(x_t)\Delta t+\beta\,\text{noise},
\]
where the next state is obtained from the current one by adding a structured
change, and in some settings also a random perturbation. In discrete
time, such rules are written step by step. In continuous time, the same idea is
expressed instead through infinitesimal motion: how the state changes over an
arbitrarily small time interval, and how randomness enters that motion. This is
why the language of calculus becomes essential.

For that reason, the chapter begins with \emph{motion and trajectories}. Before
studying full probability distributions, it is helpful to understand how an
individual state moves under a local update rule, how such motion becomes a
differential equation in continuous time, and how deterministic and stochastic
motion can be described in a unified way.

But motion alone is not enough for generative modelling. Ultimately, what
matters is not just where one particle moves, but how a whole cloud of particles
evolves so that its probability density takes the form we want. We therefore
move next to \emph{density evolution}: how particle motion reshapes probability
mass, first under deterministic transport and then under drift-plus-noise
dynamics, ultimately leading to the Fokker--Planck equation.

So the chapter develops the progression
\[
\text{discrete updates}
\;\longrightarrow\;
\text{continuous motion}
\;\longrightarrow\;
\text{density evolution}.
\]
This provides the calculus foundation for the next chapter, where these ideas
are connected to score-based generative modelling and then to fully continuous
diffusion.

\section*{Concept Map}
\addcontentsline{toc}{section}{Concept Map}

The chapter builds the calculus foundation needed to move from discrete update
rules to continuous-time generative dynamics, and then from particle motion to
the evolution of whole probability densities.

\begin{itemize}
    \item[\(\rightarrow\)] \textbf{We first study deterministic continuous-time motion through contraction.}  
    In one dimension, contraction shows how a differential equation determines a
    trajectory explicitly. In multiple dimensions, the same idea extends through
    matrices and the matrix exponential, revealing how different directions may
    contract at different rates.

    \item[\(\rightarrow\)] \textbf{We then introduce stochastic motion through Brownian motion.}  
    Brownian motion provides the canonical model of continuous-time random
    fluctuation, with Gaussian increments whose variance grows linearly in time.
    This gives the basic random ingredient that later underlies diffusion.

    \item[\(\rightarrow\)] \textbf{Next, we combine drift and noise in the Ornstein--Uhlenbeck process.}  
    The OU process shows how deterministic contraction and stochastic
    perturbation can coexist in a single system, producing a mean-reverting
    drift-plus-noise dynamics that foreshadows later continuous-time generative
    models.

    \item[\(\rightarrow\)] \textbf{We then move from particle motion to deterministic density evolution.}  
    Starting from probability conservation under a flow map, we derive the
    finite-time change-of-variable formula, then pass to infinitesimal time to
    obtain the one-dimensional Liouville equation and the corresponding
    log-density change along a trajectory.

    \item[\(\rightarrow\)] \textbf{We extend deterministic density evolution to multiple dimensions.}  
    This introduces the Jacobian as the local linearisation of the flow, the
    determinant as the measure of local volume change, and the divergence as the
    quantity that captures net local outflow. These lead to the
    multidimensional Liouville equation and its trajectory-wise form.

    \item[\(\rightarrow\)] \textbf{Finally, we add Brownian diffusion and arrive at the Fokker--Planck equation.}  
    Pure Brownian motion is seen to smooth density through Gaussian averaging,
    which produces the diffusion equation. Adding deterministic drift then
    yields the full Fokker--Planck equation, showing how transport and diffusion
    combine to govern stochastic density evolution in continuous time.

\end{itemize}

\medskip
\noindent The main relationships developed in this chapter are summarised visually below.
\input{figures/ch05/chapter5_continuous_time_at_a_glance}
\medskip

\section{From Discrete Updates to Continuous Motion}

This section introduces the basic language of continuous-time motion that will
be used throughout the rest of the chapter. We begin by setting up the core
ideas of state, velocity, differential equations, and local approximation,
which allow discrete update rules to be reinterpreted as continuous motion
laws. We then study three fundamental examples: deterministic contraction,
which pulls states toward the origin in a structured way; Brownian motion,
which introduces continuous-time random fluctuation; and the
Ornstein--Uhlenbeck process, which combines drift and noise in a single
stochastic system. Together, these examples provide the basic motion patterns
that will later underlie density evolution and continuous-time generative
dynamics.

\subsection{Basic Language for Continuous-Time Motion}

We begin by introducing the basic language used to describe motion in
continuous time. The aim is to make precise what it means for a state to evolve
smoothly, how its local change is measured, and how discrete update rules can
be reinterpreted in infinitesimal form.

Let \(x(t)\) denote the position of a quantity at time \(t\). In one
dimension, \(x(t)\) is simply a scalar-valued function of time. In multiple
dimensions, it becomes a vector whose coordinates all evolve with time. The
first derivative, \(\frac{dx}{dt}\), measures the instantaneous rate of change
of \(x\) with respect to time, and is commonly interpreted as velocity. It is
also often written in dot notation as \(\dot{x}(t)\). Similarly, the second
derivative, \(\frac{d^2x}{dt^2}=\ddot{x}(t)\), measures the rate of change of
velocity, and is commonly interpreted as acceleration.

The symbols \(dx\) and \(dt\) are often read informally as an infinitesimal
change in position and an infinitesimal change in time. The basic intuition is
that if \(\frac{dx}{dt}\) is the rate of change, then over an extremely small
time interval, the corresponding change in position is represented by
\[
dx=\frac{dx}{dt}\,dt.
\]
So, in infinitesimal form, the local change is again understood as
\[
\text{change}=\text{rate}\times\text{duration}.
\]

Once we write equations describing how the position changes over time, we arrive
at \emph{differential equations}. In the simplest case, such an equation
specifies the velocity of the state at each possible location as a function of
its current position, and sometimes of time as well. In higher dimensions, this
local motion rule can be viewed geometrically as assigning to each point in
space a direction of motion. One may therefore think of a differential equation
as a rule of motion, and of a vector field as the picture formed by all the
local arrows indicating how a particle would move if placed at different
locations.

These ideas are especially useful here because they allow us to reinterpret
iterative update rules as continuous motion laws. Over a small time interval
\(\Delta t\), a first-order approximation gives
\[
x(t+\Delta t)\approx x(t)+\frac{dx}{dt}\,\Delta t.
\]
This says that the value at a slightly later time is approximately equal to the
current value plus velocity times a small time increment. It is therefore the
continuous-time analogue of a discrete update rule.

\begin{tcolorbox}[
  colback=blue!4,
  colframe=blue!50!black,
  title={First-order Taylor expansion},
  boxrule=0.6pt,
  arc=2mm,
  left=2mm,
  right=2mm,
  top=1mm,
  bottom=1mm
]
Taylor expansion approximates a smooth function near a chosen point by using
its derivatives at that point. In local form, for a small displacement
\(\Delta x\), we may write
\[
f(x+\Delta x)
=
\sum_{k=0}^{\infty}
\frac{f^{(k)}(x)}{k!}(\Delta x)^k.
\]
Writing out the first few terms gives
\[
f(x+\Delta x)
=
f(x)
+
f'(x)\Delta x
+
\frac{f''(x)}{2!}(\Delta x)^2
+
\frac{f^{(3)}(x)}{3!}(\Delta x)^3
+\cdots.
\]

The \emph{first-order} Taylor approximation keeps only the constant term and
the linear term:
\[
f(x+\Delta x)
\approx
f(x)+f'(x)\Delta x.
\]
The discarded terms involve powers such as \((\Delta x)^2\), \((\Delta x)^3\),
and so on. When \(\Delta x\) is very small, these higher-order terms become
much smaller than the linear term.

In the present setting, the function is the trajectory \(x(t)\), the expansion
variable is time, and the small displacement is \(\Delta t\). Thus,
\[
x(t+\Delta t)
=
x(t)
+
\left.\frac{dx}{dt}\right|_{t}\Delta t
+
O(\Delta t^2).
\]
Keeping only the first-order term gives $x(t+\Delta t)
\approx
x(t)
+
\left.\frac{dx}{dt}\right|_{t}\Delta t.$
\end{tcolorbox}

\subsection{Deterministic Contraction}

We first consider deterministic motion, where the future evolution is fully
specified once the current state is known. The simplest example is
\emph{contraction}: a motion in which the state is continuously pulled back
toward the origin. This begins in one dimension and then extends naturally to
multiple dimensions, where different coordinates may contract at different
rates.

\paragraph{One-dimensional contraction.}
In one dimension, contraction toward the origin is described by
\[
\frac{dx}{dt}=-\alpha x,
\qquad \alpha>0.
\]
This says that the velocity is always proportional to the current position, but
with the opposite sign. If \(x>0\), then \(\dot{x}<0\), so the state moves left
toward the origin. If \(x<0\), then \(\dot{x}>0\), so the state moves right
toward the origin. In both cases, the motion is directed inward, and the
further the state is from the origin, the stronger the pull back.

This equation can be solved directly:
\[
\begin{aligned}
\frac{dx}{dt}=-\alpha x
&\;\Longrightarrow\;
\frac{1}{x}\frac{dx}{dt}=-\alpha
\;\Longrightarrow\;
\frac{d}{dt}\log|x|=-\alpha \\[4pt]
&\;\Longrightarrow\;
\log|x(t)|=-\alpha t + C
\;\Longrightarrow\;
x(t)=Ce^{-\alpha t}.
\end{aligned}
\]
Using the initial condition \(x(0)=x_0\), we obtain \(C=x_0\), so
\[
\boxed{
x(t)=x_0e^{-\alpha t}.
}
\]

This makes the behaviour of the system very clear. At time \(t=0\), the state
is \(x_0\). As time increases, the exponential factor \(e^{-\alpha t}\) shrinks
toward zero, so the magnitude of \(x(t)\) decays toward the origin. The
parameter \(\alpha\) controls the speed of contraction: larger \(\alpha\) gives
faster decay, while smaller \(\alpha\) gives slower decay.

So even this simple example already illustrates a basic principle of
continuous-time motion: the differential equation specifies the local direction
and speed of motion, while its solution reveals the resulting trajectory over
time.

\paragraph{Extension to multiple dimensions.}
The same idea extends naturally to higher dimensions. Suppose now that
\(x(t)\in\mathbb{R}^2\) is a two-dimensional state vector,
\[
x(t)=
\begin{bmatrix}
x_1(t)\\
x_2(t)
\end{bmatrix}.
\]
A simple multidimensional contraction model is
\[
\frac{d}{dt}x(t)=Ax(t),
\]
where \(A\) is a constant matrix. To represent independent contraction along
each coordinate axis, we may choose
\[
A=
\begin{bmatrix}
-\alpha_1 & 0\\
0 & -\alpha_2
\end{bmatrix},
\qquad \alpha_1,\alpha_2>0.
\]
Then the system becomes
\[
\frac{d}{dt}
\begin{bmatrix}
x_1(t)\\
x_2(t)
\end{bmatrix}
=
\begin{bmatrix}
-\alpha_1 & 0\\
0 & -\alpha_2
\end{bmatrix}
\begin{bmatrix}
x_1(t)\\
x_2(t)
\end{bmatrix},
\]
which is equivalent to the pair of scalar equations
\[
\dot{x}_1(t)=-\alpha_1x_1(t),
\qquad
\dot{x}_2(t)=-\alpha_2x_2(t).
\]
Each coordinate therefore contracts independently according to its own rate:
\[
x_1(t)=x_1(0)e^{-\alpha_1 t},
\qquad
x_2(t)=x_2(0)e^{-\alpha_2 t}.
\]
In compact form, this may be written as
\[
x(t)=e^{At}x(0),
\]
where \(e^{At}\) is the \emph{matrix exponential}. In the present example,
\(A\) is diagonal, so the matrix exponential simply applies the usual scalar
exponential to each diagonal entry:
\[
e^{At}
=
\begin{bmatrix}
e^{-\alpha_1 t} & 0\\
0 & e^{-\alpha_2 t}
\end{bmatrix}.
\]

\begin{tcolorbox}[
  colback=blue!4,
  colframe=blue!50!black,
  title={Matrix exponential through the Maclaurin series},
  boxrule=0.6pt,
  arc=2mm,
  left=2mm,
  right=2mm,
  top=1mm,
  bottom=1mm
]
The Taylor expansion of a smooth function around zero is called the
\emph{Maclaurin series}. For a scalar function \(f\), it has the form
\[
f(u)
=
\sum_{k=0}^{\infty}\frac{f^{(k)}(0)}{k!}u^k
=
f(0)
+
f'(0)u
+
\frac{f''(0)}{2!}u^2
+
\frac{f^{(3)}(0)}{3!}u^3
+\cdots.
\]
For the exponential function, all derivatives are again \(e^u\), so at
\(u=0\) they are all equal to \(1\). Therefore
\[
e^u
=
1+u+\frac{u^2}{2!}+\frac{u^3}{3!}+\cdots
=
\sum_{k=0}^{\infty}\frac{u^k}{k!}.
\]

The matrix exponential is defined by using the same series, but replacing the
scalar \(u\) by the matrix \(At\):
\[
\boxed{
e^{At}
=
I
+
At
+
\frac{(At)^2}{2!}
+
\frac{(At)^3}{3!}
+\cdots.
}
\]
The scalar \(1\) is replaced by the identity matrix \(I\), as \(I\) plays
the role of the multiplicative identity for matrices. For the diagonal matrix used here $A=
\begin{bmatrix}
-\alpha_1 & 0\\
0 & -\alpha_2
\end{bmatrix},$
% \[

% \]
we have
\[
At=
\begin{bmatrix}
-\alpha_1 t & 0\\
0 & -\alpha_2 t
\end{bmatrix}.
\]
Powers of a diagonal matrix remain diagonal:
\[
(At)^k
=
\begin{bmatrix}
(-\alpha_1 t)^k & 0\\
0 & (-\alpha_2 t)^k
\end{bmatrix}.
\]
Substituting these powers into the series gives
\[
e^{At}
=
\begin{bmatrix}
\sum_{k=0}^{\infty}\frac{(-\alpha_1 t)^k}{k!} & 0\\
0 & \sum_{k=0}^{\infty}\frac{(-\alpha_2 t)^k}{k!}
\end{bmatrix}
=
\begin{bmatrix}
e^{-\alpha_1 t} & 0\\
0 & e^{-\alpha_2 t}
\end{bmatrix}.
\]
\end{tcolorbox}

This multidimensional example introduces an important new feature: the motion
may now shrink at different speeds in different directions. If
\(\alpha_1>\alpha_2\), then the first coordinate decays more quickly than the
second, so the state contracts faster in the \(x_1\)-direction than in the
\(x_2\)-direction. Thus the motion still heads toward the origin, but the
contraction need not be uniform in all directions.

The diagonal form also shows that there is still no interaction between
coordinates: each component evolves on its own. Although the example is written
in two dimensions for clarity, the same idea extends directly to
\(\mathbb{R}^d\). A diagonal matrix
\[
A=\mathrm{diag}(-\alpha_1,-\alpha_2,\dots,-\alpha_d)
\]
produces independent exponential contraction along each coordinate direction.

This makes diagonal contraction a useful first multidimensional example. It
introduces matrix-based continuous motion in the simplest possible way, and also
provides a natural bridge from scalar differential equations to vector-valued
continuous dynamics.

\subsection{Brownian Motion}

The discussion so far has been entirely deterministic: once the initial state is
fixed, the future trajectory is completely determined by the governing
differential equation. Many continuous-time generative models, however, also
involve randomness. At each small time interval, the state may receive a random
perturbation, and the resulting motion is no longer fully predictable from its
initial condition alone.

The basic model of such random continuous-time fluctuation is \emph{Brownian
motion}, also called \emph{Wiener motion}~\cite{oksendal2003sde}. It is usually denoted by
\(\{W_t\}_{t\ge 0}\), and may be characterised by three basic properties:
\begin{enumerate}
    \item \(W_0=0\);
    \item increments over disjoint time intervals are independent;
    \item for any \(\Delta t>0\),
    \[
    W_{t+\Delta t}-W_t \sim \mathcal{N}(0,\Delta t).
    \]
\end{enumerate}

The third property is especially important. It says that over a short time
interval of length \(\Delta t\), the random increment has mean zero and variance
\(\Delta t\). So Brownian motion introduces fluctuations that are unbiased on
average, but whose uncertainty grows with elapsed time.

The Gaussian assumption is also natural. If we imagine the motion over a short
interval as the accumulated effect of many tiny, approximately independent
random perturbations, then their total is well modelled by a Gaussian random
variable. This is closely related to the intuition behind the central limit
theorem: when many small independent effects are added together, their sum tends
toward a Gaussian form.

A useful way to understand this is to divide the interval \([0,t]\) into many
small pieces and write \(W_t\) as a sum of independent Gaussian increments. If
we divide \([0,t]\) into \(n\) equal pieces of length $\Delta t=\frac{t}{n},$ then 
\[
W_t-W_0
=
\sum_{k=0}^{n-1}\bigl(W_{(k+1)\Delta t}-W_{k\Delta t}\bigr)
=
\sum_{k=0}^{n-1}\Delta W_k,
\qquad
\Delta W_k:=W_{(k+1)\Delta t}-W_{k\Delta t}.
\]

Since \(W_0=0\), this becomes
\[
W_t=\sum_{k=0}^{n-1}\Delta W_k.
\]

\begin{tcolorbox}[
  colback=blue!4,
  colframe=blue!50!black,
  title={Mean and variance of Brownian motion at time \(t\)},
  boxrule=0.6pt,
  arc=2mm,
  left=2mm,
  right=2mm,
  top=1mm,
  bottom=1mm
]
Each increment satisfies
\[
\Delta W_k\sim \mathcal{N}(0,\Delta t),
\qquad
\mathbb{E}[\Delta W_k]=0,
\qquad
\mathrm{Var}(\Delta W_k)=\Delta t=\frac{t}{n}.
\]
Therefore
\[
\mathbb{E}[W_t]
=
\sum_{k=0}^{n-1}\mathbb{E}[\Delta W_k]
=
0,
\]
and, using independence,
\[
\mathrm{Var}(W_t)
=
\sum_{k=0}^{n-1}\mathrm{Var}(\Delta W_k)
=
\sum_{k=0}^{n-1}\frac{t}{n}
=
t.
\]
Hence
\[
\mathbb{E}[W_t]=0,
\qquad
\mathrm{Var}(W_t)=t.
\]
\end{tcolorbox}

So at each fixed time \(t\), Brownian motion is centred at zero, but its spread
grows linearly with time. This means that Brownian motion does not pull the
state toward any preferred location; instead, it describes random wandering
whose uncertainty increases as time passes.

A useful heuristic way to think about Brownian motion is through the
approximation
\[
W_{t+\Delta t}-W_t \approx \sqrt{\Delta t}\,\varepsilon_t,
\qquad
\varepsilon_t\sim\mathcal{N}(0,1).
\]
This makes the scaling especially clear. Because the increment has variance
\(\Delta t\), its typical size is of order \(\sqrt{\Delta t}\), not
\(\Delta t\). This square-root scaling is one of the most characteristic
features of stochastic continuous-time motion and will later reappear in
diffusion models.

Brownian motion is often written informally in differential form as $dW_t,$
which represents an infinitesimal random increment. Although \(dW_t\) is not an
ordinary differential in the deterministic sense, this notation is extremely
useful because it allows stochastic motion to be written in a form parallel to
ordinary differential equations.
\subsection{The Ornstein--Uhlenbeck Process}

A particularly important example of stochastic continuous-time motion is the
Ornstein--Uhlenbeck (OU) process. It combines deterministic contraction with
Brownian noise:
\[
dX_t=-\alpha X_t\,dt+\sigma\,dW_t,
\qquad
\alpha>0,\ \sigma>0.
\]

This equation contains two parts. The drift term $-\alpha X_t\,dt$ pulls the state back toward the origin, just as in the one-dimensional
contraction model introduced earlier. The noise term $\sigma\,dW_t$ injects random perturbations driven by Brownian motion. So the OU process
balances two competing tendencies: deterministic pull toward the origin and
stochastic fluctuations away from it.

Over a small time interval \(\Delta t\), one may write the corresponding update
heuristically as
\[
X_{t+\Delta t}
\approx
X_t-\alpha X_t\,\Delta t+\sigma\sqrt{\Delta t}\,\varepsilon_t,
\qquad
\varepsilon_t\sim\mathcal{N}(0,1).
\]
This discrete-time view is very useful. It shows that the next state is obtained
from the current one by adding two pieces:
\begin{itemize}
    \item a deterministic correction of size \(\Delta t\), and
    \item a random perturbation of size \(\sqrt{\Delta t}\).
\end{itemize}
So the OU process already has the basic drift-plus-noise structure that will
later appear in Langevin dynamics and diffusion models.

The OU process also has an explicit solution:
\[
X_t
=
X_0e^{-\alpha t}
+
\sigma \int_0^t e^{-\alpha(t-s)}\,dW_s.
\]
This formula reveals its structure clearly. The first term, $X_0e^{-\alpha t},$ is exactly the deterministic exponential contraction of the initial state. The
second term accumulates all past random perturbations, but each perturbation is
weighted by the decay factor \(e^{-\alpha(t-s)}\). So older noise contributions
gradually lose influence, while more recent ones matter more.

This is why the OU process is often described as a \emph{mean-reverting}
stochastic process: it remains random while retaining a stable restoring
tendency. Figure~\ref{fig:continuous-three-motion-laws} places this behaviour
beside deterministic contraction and Brownian motion, making clear how the OU
process combines the two basic effects.

So far, however, we have still described motion only at the level of individual
particles. The next step is to understand how such motion reshapes an entire
probability density over time.

\input{figures/ch05/fig_ch05_01_three_motion_laws}

%\section{From Particle Motion to Deterministic Density Evolution}
\section{Deterministic Density Evolution}

The previous section focused on the motion of individual particles. That
viewpoint is already useful, because many continuous-time models are indeed
described by differential equations acting on states or trajectories. For
generative modelling, however, the motion of one particle is only part of the
story. What we ultimately care about is how a whole collection of such particles
changes its \emph{probability density} over time.

This section therefore makes an important transition: from \emph{trajectory
dynamics} to \emph{density dynamics}. We begin with the simplest deterministic
setting, where particles are transported by a map and probability mass is
preserved. This leads first to a finite-time change-of-variable formula, then
to an infinitesimal density evolution law, and finally to its multidimensional
form through divergence. Throughout, the underlying idea is the same:
\[
\text{particle motion}
\;\longrightarrow\;
\text{transport of probability mass}
\;\longrightarrow\;
\text{deterministic density evolution}.
\]

\subsection{Finite-Time Density Transformation}

We begin with the simplest density principle: \emph{probability is conserved
under deterministic transport}. Here this means that particles are moved by the
dynamics from one location to another in a fully determined way, while the
associated probability mass is simply carried with them.

Suppose that at time \(0\), a particle has position \(x_0\), distributed
according to density \(p_0(x_0)\). Suppose further that the deterministic
dynamics transport this initial point to a later position
\[
x_t=\phi_t(x_0),
\]
where \(\phi_t\) denotes the trajectory or flow map that moves the particle from time \(0\) to time \(t\).

To understand how densities change, consider a very small interval around
\(x_0\), of length \(dx_0\). The probability mass contained in that tiny
interval is approximately
\[
p_0(x_0)\,dx_0.
\]
After evolving to time \(t\), that same small piece of probability mass is
transported to a new interval around \(x_t=\phi_t(x_0)\). Let the length of
that transported interval be \(dx_t\). Since deterministic motion merely
transports particles and does not create or destroy probability mass, the total
probability must remain the same:
\[
p_0(x_0)\,dx_0=p_t(x_t)\,dx_t.
\]

The key question is therefore how the small interval \(dx_0\) is transformed
under the flow map. In one dimension, this local stretching or compression is
measured by the derivative of the map. The derivation is as follows.

\begin{tcolorbox}[
  colback=blue!4,
  colframe=blue!50!black,
  title={Deriving the one-dimensional change-of-variable formula},
  boxrule=0.6pt,
  arc=2mm,
  left=2mm,
  right=2mm,
  top=1mm,
  bottom=1mm
]
Track not only the point \(x_0\), but also a nearby point \(x_0+dx_0\). Under
the same flow map,
\[
x_t=\phi_t(x_0), \qquad x_t+dx_t=\phi_t(x_0+dx_0),
\]
so $dx_t=\phi_t(x_0+dx_0)-\phi_t(x_0).$
Because \(dx_0\) is very small, a first-order Taylor expansion of \(\phi_t\)
around \(x_0\) gives
\[
\phi_t(x_0+dx_0)\approx \phi_t(x_0)+\frac{d\phi_t(x_0)}{dx_0}\,dx_0.
\]
Hence $dx_t\approx \frac{d\phi_t(x_0)}{dx_0}\,dx_0
= \frac{dx_t}{dx_0}\,dx_0$.
Substituting into probability conservation,
\[
p_0(x_0)\,dx_0
=
p_t(x_t)\left|\frac{dx_t}{dx_0}\right|dx_0,
\]
and cancelling \(dx_0\) yields
\[
\boxed{
p_t(x_t)=p_0(x_0)\left|\frac{dx_t}{dx_0}\right|^{-1}.
}
\]
\end{tcolorbox}

This is the one-dimensional change-of-variable formula induced by a
deterministic flow. Its interpretation is very intuitive. If the transformation
stretches an interval, then the same amount of probability is spread over a
larger region, so the density decreases. If it compresses an interval, then the
same probability mass is squeezed into a smaller region, so the density
increases. Thus the derivative \(\frac{dx_t}{dx_0}\) measures how much local
length has changed, and its reciprocal tells us how the density rescales.

The same idea extends naturally to multiple dimensions. In one dimension, we
tracked how a small \emph{length} changes under the map. In two dimensions, we
must track how a small \emph{area} changes; in three dimensions, a small
\emph{volume}; and in general, a small \(d\)-dimensional volume element. The
quantity that measures this local scaling is no longer an ordinary derivative,
but the determinant of the Jacobian matrix of the transformation. Thus the
multidimensional change-of-variable formula becomes
\[
p_t(x_t)
=
p_0(x_0)
\left|
\det\!\left(\frac{\partial x_t}{\partial x_0}\right)
\right|^{-1}.
\]
So the determinant plays, in higher dimensions, the role that
\(\frac{dx_t}{dx_0}\) plays in one dimension, measuring how much a tiny local region has been stretched or compressed by the flow.

It is important to note that this formula describes a \emph{finite-time}
deterministic transformation. It tells us how density changes between time
\(0\) and time \(t\), but it does not yet describe the infinitesimal rate of
density change. For that, we next derive a local-in-time continuity equation.

\subsection{Infinitesimal Density Evolution in One Dimension: The Liouville Equation}

The finite-time change-of-variable formula tells us how a density transforms
between time \(0\) and time \(t\). In continuous-time modelling, however, it is
often even more important to understand how density changes over an
\emph{infinitesimal} time interval, because continuous dynamics is built from
such local updates. This leads to the \emph{Liouville equation}, also called the
continuity equation in the deterministic setting.

Suppose particles move according to the one-dimensional ordinary differential
equation
\[
\frac{dx}{dt}=f(x,t),
\]
where \(f(x,t)\) is the local velocity field. To understand how the density
changes, we now adopt a fixed-space viewpoint: rather than following one
particle, we stand at a small interval in space and watch probability mass flow
through it.

Consider a short interval \([x,x+\Delta x]\). The total probability mass inside
this interval at time \(t\) is approximately
\[
p(x,t)\,\Delta x.
\]
Over an infinitesimal time interval, this local mass can change only because
probability flows in through the left boundary or flows out through the right
boundary.

The relevant flow quantity is the \emph{probability flux}, namely density times
velocity:
\[
p(x,t)f(x,t).
\]
This tells us how much probability mass crosses a location per unit time. Thus
the inflow at the left boundary is
\[
p(x,t)f(x,t),
\]
while the outflow at the right boundary is
\[
p(x+\Delta x,t)f(x+\Delta x,t).
\]

Conservation of probability therefore says that the rate of change of
probability mass inside the interval must equal inflow minus outflow:
\[
\frac{\partial}{\partial t}p(x,t)\,\Delta x
=
p(x,t)f(x,t)-p(x+\Delta x,t)f(x+\Delta x,t).
\]
Dividing both sides by \(\Delta x\) gives
\[
\frac{\partial}{\partial t}p(x,t)
=
\frac{p(x,t)f(x,t)-p(x+\Delta x,t)f(x+\Delta x,t)}{\Delta x}.
\]
Rearranging,
\[
\frac{\partial}{\partial t}p(x,t)
=
-
\frac{p(x+\Delta x,t)f(x+\Delta x,t)-p(x,t)f(x,t)}{\Delta x}.
\]
Now let \(\Delta x\to 0\). The right-hand side becomes a spatial derivative, so
we obtain
\[
\boxed{
\partial_t p(x,t)
=
-\partial_x\bigl(p(x,t)f(x,t)\bigr).
}
\]
This is the one-dimensional Liouville equation.

The equation says that density changes according to the net outflow of
probability flux. If more probability flows out of a region than into it, the
density decreases there. If more flows in than out, the density increases.
It is often useful to expand the right-hand side using the product rule:
\[
\partial_t p
=
-\bigl(f\,\partial_x p + p\,\partial_x f\bigr).
\]
This reveals two distinct contributions.

The term $-f\,\partial_x p$ depends on the spatial slope of the density itself. It therefore describes
\emph{transport}: an existing density profile is being carried through space by
the velocity field. If the flow simply moves particles from one place to
another without changing their relative spacing, then the main effect is that
the whole density profile shifts.

By contrast, the term $-p\,\partial_x f$ depends on the spatial slope of the velocity field. It therefore describes
\emph{deformation}: nearby particles no longer move at exactly the same speed,
so they either spread apart or compress together. If \(\partial_x f>0\), the
velocity increases with position, so nearby particles separate and the density
decreases. If \(\partial_x f<0\), the velocity decreases with position, so
nearby particles compress and the density increases.

\subsection{Density Change Along a Trajectory}

The Liouville equation describes how density changes at a fixed location in
space. But there is another viewpoint that is also very useful: how does the
density change \emph{along the trajectory of a moving particle}? This viewpoint
shifts attention from what happens at a fixed position to what is experienced by
a particle as it moves with the flow.

Suppose a particle follows the deterministic dynamics
\[
\frac{dx(t)}{dt}=f(x(t),t).
\]
Now consider the density evaluated along that particle path:
\[
p(x(t),t).
\]
Because both the position and time are changing, we differentiate this quantity
using the chain rule. Equivalently, this is the \emph{total derivative}: it
combines the explicit change with respect to time and the implicit change coming
from the fact that the particle itself is moving through space. Thus
\[
\frac{d}{dt}p(x(t),t)
=
\partial_t p(x(t),t)
+
\frac{dx(t)}{dt}\,\partial_x p(x(t),t).
\]
Since \(\dot{x}(t)=f(x(t),t)\), this becomes
\[
\frac{d}{dt}p(x(t),t)
=
\partial_t p + f\,\partial_x p.
\]

We now substitute the Liouville equation
\[
\partial_t p
=
-\partial_x(pf)
=
-(f\,\partial_x p + p\,\partial_x f).
\]
This gives
\[
\frac{d}{dt}p(x(t),t)
=
-(f\,\partial_x p + p\,\partial_x f)+f\,\partial_x p.
\]
The transport terms cancel, leaving
\[
\frac{d}{dt}p(x(t),t)
=
-p(x(t),t)\,\partial_x f(x(t),t).
\]
Dividing both sides by \(p(x(t),t)\) gives
\[
\frac{d}{dt}\log p(x(t),t)
=
-\partial_x f(x(t),t).
\]
Hence
\[
\boxed{
\frac{d}{dt}\log p(x(t),t)
=
-\partial_x f(x(t),t).
}
\]

This is an important result. In the previous subsection, the density evolution
was decomposed into a transport part and a deformation part. Here we see that,
once we move with the particle itself, the transport part disappears: from the
particle’s own point of view, it is no longer observing the profile being
carried through space, because it is travelling with that flow. What remains is
only the local expansion or compression effect.

Thus, if \(\partial_x f>0\), nearby particles tend to separate and the density
along the trajectory decreases. If \(\partial_x f<0\), nearby particles tend to
compress and the density increases. So the fixed-space and moving-particle viewpoints emphasise different aspects of
the same dynamics. The Liouville equation describes how the whole density field
changes across space, whereas the formula above tells us how density changes
along an individual trajectory.

\subsection{Divergence and the Multidimensional Liouville Equation}

We now extend density evolution from one dimension to a full multidimensional
state space. The main new feature is that both the motion and the probability
flux become vector-valued. In one dimension, the scalar derivative
\(\partial_x\) was enough to describe how probability flowed in or out of a
small interval. In multiple dimensions, the corresponding quantity is the
\emph{divergence}, which measures the net outflow of a vector field from a tiny
region of space. This section shows how divergence arises naturally from local
volume change, and how it leads to the multidimensional Liouville equation.

Let \(x(t)\in\mathbb{R}^d\) and suppose particles evolve according to
\[
\frac{dx}{dt}=v(x,t),
\]
where the velocity field is
\[
v(x,t)=
\begin{bmatrix}
v_1(x,t)\\
v_2(x,t)\\
\vdots\\
v_d(x,t)
\end{bmatrix}.
\]

As in the one-dimensional case, the basic idea is still local conservation of
probability. But before we can write down the density equation, we first need to
understand how a tiny region of space changes under the flow. In one dimension,
this was measured by the derivative of the map, which told us how a small
interval was stretched or compressed. In multiple dimensions, the corresponding
objects are the \emph{Jacobian matrix} and its determinant.

Over a very small time step \(t\to t+dt\), the motion is approximated by
\[
x(t+dt)=x(t)+v(x(t),t)\,dt.
\]
So the infinitesimal transformation is
\[
x\mapsto x+v(x,t)\,dt.
\]

To understand how a tiny volume changes, we ask how the updated point depends on
its starting location. That dependence is captured by differentiating the map
with respect to \(x\). This gives the \emph{Jacobian matrix} of the
transformation:
\[
\frac{\partial x(t+dt)}{\partial x}
=
I+\frac{\partial v}{\partial x}dt.
\]
If we denote the Jacobian matrix of the velocity field by
\[
J_v(x,t)=\frac{\partial v}{\partial x},
\]
then
\[
\frac{\partial x(t+dt)}{\partial x}=I+J_v(x,t)\,dt.
\]

The Jacobian tells us how a small displacement changes under the flow. To pass
from displacements to volumes, we take its determinant.

\begin{tcolorbox}[
  colback=blue!4,
  colframe=blue!50!black,
  title={Local volume scaling and divergence},
  boxrule=0.6pt,
  arc=2mm,
  left=2mm,
  right=2mm,
  top=1mm,
  bottom=1mm
]
The local volume scaling is
\[
\det\!\bigl(I+J_v(x,t)\,dt\bigr).
\]
For small \(dt\), the first-order expansion gives
\[
\det(I+A\,dt)=1+\operatorname{tr}(A)\,dt+O(dt^2).
\]
Applying this with \(A=J_v(x,t)\) yields
\[
\det\!\bigl(I+J_v(x,t)\,dt\bigr)
=
1+\operatorname{tr}(J_v(x,t))\,dt+O(dt^2).
\]
The trace of the Jacobian has a special name:
\[
\operatorname{tr}(J_v(x,t))
=
\nabla\cdot v(x,t)
=
\sum_{i=1}^d \frac{\partial v_i}{\partial x_i}(x,t).
\]
Hence
\[
\det\!\bigl(I+J_v(x,t)\,dt\bigr)
=
1+(\nabla\cdot v(x,t))\,dt+O(dt^2),
\]
so a small volume element evolves as
\[
dV(t+dt)=\bigl(1+(\nabla\cdot v(x,t))\,dt\bigr)dV(t),
\]
up to first order.
\end{tcolorbox}

This gives the key geometric interpretation of divergence: it is the
instantaneous rate at which a tiny volume expands or contracts under the flow.
If \(\nabla\cdot v>0\), the flow locally expands volume. If \(\nabla\cdot v<0\),
it locally compresses volume.

With that geometric picture in place, we can now derive the density-evolution
equation. The probability mass inside a small moving volume must remain the
same:
\[
p(x,t)\,dV(t)
=
p(x+v(x,t)dt,t+dt)\,dV(t+dt).
\]
Substituting the volume scaling relation,
\[
dV(t+dt)=\bigl(1+(\nabla\cdot v)\,dt\bigr)dV(t),
\]
we obtain
\[
p(x,t)
=
p(x+vdt,t+dt)\bigl(1+(\nabla\cdot v)\,dt\bigr).
\]

Next, apply a multivariable first-order Taylor expansion to the density.
\begin{tcolorbox}[
  colback=blue!4,
  colframe=blue!50!black,
  title={Deriving the multidimensional Liouville equation},
  boxrule=0.6pt,
  arc=2mm,
  left=2mm,
  right=2mm,
  top=1mm,
  bottom=1mm
]
For a function \(p(x,t)\) depending on position and time, the first-order
multivariable Taylor expansion is
\[
p(x+\Delta x,t+\Delta t)
\approx
p(x,t)+\nabla p(x,t)\cdot \Delta x+\partial_t p(x,t)\Delta t.
\]
In the flow considered here,
\(\Delta x=v(x,t)dt\) and \(\Delta t=dt\), hence
\[
\begin{aligned}
p(x+vdt,t+dt)
&=
p(x,t)+\nabla p(x,t)\cdot(vdt)+\partial_t p(x,t)dt+O(dt^2)\\
&=
p(x,t)+(v\cdot\nabla p)dt+(\partial_t p)dt+O(dt^2).
\end{aligned}
\]

Substituting this into probability conservation,
\[
p(x,t)
=
p(x+vdt,t+dt)\bigl(1+(\nabla\cdot v)dt\bigr),
\]
gives
\[
\begin{aligned}
p
&=
\left[p+(\partial_t p)dt+(v\cdot\nabla p)dt\right]
\left[1+(\nabla\cdot v)dt\right]
+O(dt^2)\\
&=
p+(\partial_t p)dt+(v\cdot\nabla p)dt+p(\nabla\cdot v)dt+O(dt^2).
\end{aligned}
\]
After subtracting \(p\), dividing by \(dt\), and letting \(dt\to0\),
\[
0=\partial_t p+v\cdot\nabla p+p(\nabla\cdot v).
\]
Using \(\nabla\cdot(pv)=v\cdot\nabla p+p(\nabla\cdot v)\), this becomes
\[
0=\partial_t p+\nabla\cdot(pv),
\]
or equivalently,
\[
\boxed{
\partial_t p(x,t)=-\nabla\cdot\bigl(p(x,t)v(x,t)\bigr).
}
\]
\end{tcolorbox}

This is the multidimensional Liouville equation. It is the natural
generalisation of the one-dimensional continuity equation. Instead of the scalar
derivative \(\partial_x(pf)\), we now have the divergence of the vector-valued
probability flux \(pv\). So the density changes according to the net outflow of
probability from a small region of space.

The trajectory-wise viewpoint also extends naturally. Along a path satisfying
\[
\dot{x}(t)=v(x(t),t),
\]
the log-density evolves according to
\[
\boxed{
\frac{d}{dt}\log p(x(t),t)
=
-\nabla\cdot v(x(t),t).
}
\]
So in multiple dimensions, just as in one dimension, once we move with the
particle itself, the transport part disappears and the remaining density change
is governed by local expansion or compression of the flow.
Figure~\ref{fig:continuous-flow-density-jacobian} summarises the geometric
connection between finite-time volume change, density rescaling, and the
infinitesimal divergence relation.

\input{figures/ch05/fig_ch05_02_flow_density_jacobian}

% \section{From Deterministic Transport to Stochastic Density Evolution}
\section{Stochastic Density Evolution and the Fokker--Planck Equation}

So far, the density evolution has been entirely deterministic: probability mass
is transported by the flow, and its local change is governed by divergence.
However, many continuous-time generative models also include random
perturbations. Once such diffusion is added, density no longer changes only
through transport and local expansion or compression; it also spreads through
space. This leads from the Liouville equation to the \emph{Fokker--Planck
equation}~\cite{risken1996fokker}.

\subsection{Pure Brownian Diffusion as Gaussian Smoothing}

To build intuition, consider first pure Brownian diffusion in one dimension:
\[
dX_t=\sigma\,dW_t.
\]
Over a small interval \(\Delta t\), the increment is Gaussian:
\[
X_{t+\Delta t}-X_t \sim \mathcal{N}(0,\sigma^2\Delta t).
\]
Thus a particle previously at location \(y\) has some probability of arriving
near \(x\) after this short time step, and that probability is described by a
Gaussian kernel centred at \(y\). Therefore the updated density is obtained by
collecting contributions from all nearby previous locations:
\[
p(x,t+\Delta t)
=
\int p(y,t)\,
\frac{1}{\sqrt{2\pi \sigma^2\Delta t}}
\exp\!\left(
-\frac{(x-y)^2}{2\sigma^2\Delta t}
\right)dy.
\]

In other words, the density at \(x\) after a short time is a weighted average of
the previous density around \(x\), where the weights are given by the
short-time Gaussian transition kernel. This is a convolution with a narrow
Gaussian kernel. Intuitively, probability mass diffuses into \(x\) from nearby
locations, while some of the probability previously near \(x\) diffuses away.
The net result is smoothing and spreading of the density.

To make this more convenient to analyse, let the Gaussian increment be denoted
by \(\eta\), where
\[
\eta\sim \mathcal{N}(0,\sigma^2\Delta t).
\]
Introduce the change of variable
\[
\eta=x-y,
\qquad\text{so that}\qquad
y=x-\eta.
\]
Then the density update may be written as
\[
p(x,t+\Delta t)
=
\int p(x-\eta,t)\,G_{\Delta t}(\eta)\,d\eta,
\]
where
\[
G_{\Delta t}(\eta)
=
\frac{1}{\sqrt{2\pi \sigma^2\Delta t}}
\exp\!\left(
-\frac{\eta^2}{2\sigma^2\Delta t}
\right)
\]
is the Gaussian density of the increment. Since
\[
\int G_{\Delta t}(\eta)\,d\eta=1,
\]
this can also be viewed as an expectation:
\[
p(x,t+\Delta t)=\mathbb{E}_{\eta}\bigl[p(x-\eta,t)\bigr].
\]

\subsection{From Brownian Diffusion to the Fokker--Planck Equation}

We now expand \(p(x-\eta,t)\) around \(x\) using Taylor series. In general,
for a sufficiently smooth function of one variable, $f(x+h)=\sum_{k=0}^{\infty}\frac{f^{(k)}(x)}{k!}h^k.$ So, taking \(f(\cdot)=p(\cdot,t)\) and \(h=-\eta\), we obtain
\[
\begin{aligned}
p(x-\eta,t)
&=
\sum_{k=0}^{\infty}
\frac{\partial_x^{\,k}p(x,t)}{k!}(-\eta)^k \\[4pt]
&=
p(x,t)
-\eta\,\partial_x p(x,t)
+\frac{(-\eta)^2}{2!}\partial_{xx}p(x,t)
+\frac{(-\eta)^3}{3!}\partial_{xxx}p(x,t)
+\cdots \\[4pt]
&=
p(x,t)-\eta\,\partial_x p(x,t)
+\frac{\eta^2}{2}\partial_{xx}p(x,t)
+O(\eta^3).
\end{aligned}
\]
Taking expectation with respect to \(\eta\) gives
\[
p(x,t+\Delta t)
=
\mathbb{E}_{\eta}\bigl[p(x-\eta,t)\bigr]
=
p(x,t)
-
\mathbb{E}[\eta]\partial_x p(x,t)
+
\frac{1}{2}\mathbb{E}[\eta^2]\partial_{xx}p(x,t)
+\cdots.
\]
Because the Gaussian increment has mean zero,
\[
\mathbb{E}[\eta]=0,
\]
the first-order term vanishes. Its variance is
\[
\mathbb{E}[\eta^2]=\sigma^2\Delta t,
\]
so the second-order term survives:
\[
p(x,t+\Delta t)
=
p(x,t)+\frac{\sigma^2\Delta t}{2}\partial_{xx}p(x,t)+o(\Delta t).
\]
Subtracting \(p(x,t)\) from both sides, dividing by \(\Delta t\), and taking
\(\Delta t\to 0\), we obtain
\[
\boxed{
\partial_t p(x,t)=\frac{\sigma^2}{2}\partial_{xx}p(x,t).
}
\]
Thus pure Brownian motion gives the pure diffusion equation.

This is the density-level signature of random spreading. Unlike the Liouville
equation, which moves density through deterministic transport, the diffusion
equation smooths and spreads the density through space.

Now suppose we also include deterministic drift:
\[
dX_t=f(X_t,t)\,dt+\sigma\,dW_t.
\]
Over a short interval, the particle undergoes two effects:
\begin{itemize}
    \item a deterministic displacement of order \(\Delta t\) due to the drift
    \(f\),
    \item a random Gaussian displacement of order \(\sqrt{\Delta t}\) due to the
    Brownian term.
\end{itemize}
The resulting density evolution combines the deterministic Liouville transport
term with the diffusion term derived above:
\[
\boxed{
\partial_t p(x,t)
=
-\partial_x\bigl(f(x,t)p(x,t)\bigr)
+
\frac{\sigma^2}{2}\partial_{xx}p(x,t).
}
\]
This is the one-dimensional \textbf{Fokker--Planck equation} for constant
diffusion coefficient \(\sigma\). Figure~\ref{fig:continuous-liouville-fokker-planck}
compares the corresponding density evolution under drift alone, diffusion
alone, and their combination.

\input{figures/ch05/fig_ch05_03_liouville_fokker_planck}

\subsection{Multidimensional Form and Interpretation}

In multiple dimensions, the deterministic part becomes
\[
-\nabla\cdot\bigl(p(x,t)v(x,t)\bigr),
\]
while the diffusion term becomes
\[
\frac{\sigma^2}{2}\Delta p(x,t),
\]
where the Laplacian is
\[
\Delta p
=
\nabla\cdot(\nabla p)
=
\sum_{i=1}^d \frac{\partial^2 p}{\partial x_i^2}.
\]
Thus, for constant isotropic diffusion coefficient \(\sigma\), the
multidimensional Fokker--Planck equation takes the form
\[
\boxed{
\partial_t p(x,t)
=
-\nabla\cdot\bigl(p(x,t)v(x,t)\bigr)
+
\frac{\sigma^2}{2}\Delta p(x,t).
}
\]

This equation is one of the most important density-evolution laws in
continuous-time probabilistic modelling. The Liouville equation tells us how
densities move under deterministic flow. The Fokker--Planck equation extends
that picture by adding stochastic diffusion. Together, they provide the main
bridge from elementary continuous-time motion to the density dynamics that
underlie fully continuous diffusion models.

In the next chapter, we will build directly on this foundation. The key next
step is to understand how density evolution can be linked to the gradient of
log-density, and how that viewpoint leads naturally to score-based learning and
fully continuous diffusion.

\section*{Summary and References}
\addcontentsline{toc}{section}{Summary and References}

This chapter developed the calculus foundation needed to move from discrete
updates to continuous-time generative dynamics. We began by introducing the
basic language of continuous motion through derivatives, differential
equations, infinitesimal change, and first-order Taylor approximation.

We then studied three core motion patterns. Deterministic contraction showed how
a differential equation determines a trajectory, first in one dimension and then
in multiple dimensions through matrices and the matrix exponential. Brownian
motion introduced continuous-time random fluctuation through Gaussian
increments, and the Ornstein--Uhlenbeck process combined drift and noise in a
single mean-reverting system.

The chapter then shifted from particle motion to density evolution. For
deterministic flows, we derived the finite-time change-of-variable formula and
then the Liouville equation, together with its trajectory-wise log-density form.
Extending this to multiple dimensions introduced the Jacobian, determinant, and
divergence as the key tools for describing local volume change and density flow.

Finally, we added Brownian diffusion and arrived at the Fokker--Planck
equation, which combines deterministic transport with stochastic spreading.
This provides the main bridge to the next chapter, where density evolution will
be linked to gradients of log-density, score-based learning, and fully
continuous diffusion models.

\bigskip

The material in this chapter draws on standard ideas from calculus,
differential equations, stochastic processes, and continuous-time probability.
For broad machine learning background and probabilistic modelling perspectives,
useful general references include Bishop~\cite{bishop2006prml} and
Murphy~\cite{murphy2012ml}. For stochastic differential equations and Brownian
motion, a standard introduction is Øksendal~\cite{oksendal2003sde}. For the
continuity equation, transport, and related partial-differential-equation
viewpoints, a useful reference is Evans~\cite{evans2010pde}. For the
Fokker--Planck equation and stochastic diffusion more specifically, a classical
reference is Risken~\cite{risken1996fokker}.

%% file: figures/ch05/chapter5_continuous_time_at_a_glance.tex
\begin{center}
\resizebox{0.98\textwidth}{!}{%
\begin{tikzpicture}[
  >=Latex,
  every node/.style={font=\small},
  stage/.style={
    draw, rounded corners=2pt, align=center, inner sep=6pt,
    text width=3.0cm, minimum height=1.35cm
  },
  next/.style={
    draw, rounded corners=2pt, align=center, inner sep=6pt,
    text width=2.65cm, minimum height=1.35cm
  },
  flow/.style={->,thick}
]
\node[font=\normalsize\bfseries] at (6.15,1.95)
  {Continuous-time foundations at a glance};

\node[stage] (disc) at (0,0) {
  \textbf{Small discrete}\\[-1pt]\textbf{update}\\[3pt]
  $x_{t+\Delta t}\approx x_t+$ local change
};

\node[stage] (motion) at (4.1,0) {
  \textbf{Particle-level}\\[-1pt]\textbf{motion}\\[3pt]
  {\scriptsize ODE: $dX_t=f(X_t,t)\,dt$}\\
  {\scriptsize SDE:}\\[-1pt]{\scriptsize $dX_t=f\,dt+\sigma\,dW_t$}
};

\node[stage] (density) at (8.2,0) {
  \textbf{Density-level}\\[-1pt]\textbf{evolution}\\[3pt]
  Liouville: transport\\
  Fokker--Planck: transport $+$ diffusion
};

\node[next] (score) at (12.3,0) {
  \textbf{Next chapter}\\[3pt]
  density evolution\\
  $\longrightarrow$ score\\
  $\nabla_x\log p_t(x)$
};

\draw[flow] (disc) -- node[above] {\scriptsize limit} (motion);
\draw[flow] (motion) -- node[above] {\scriptsize ensemble} (density);
\draw[flow] (density) -- (score);
\end{tikzpicture}%
}
\end{center}

%% file: figures/ch05/fig_ch05_01_three_motion_laws.tex
\begin{figure}[H]
\centering
\begin{minipage}[t]{0.45\textwidth}
  \vspace{0pt}\centering
  \includegraphics[width=\linewidth]{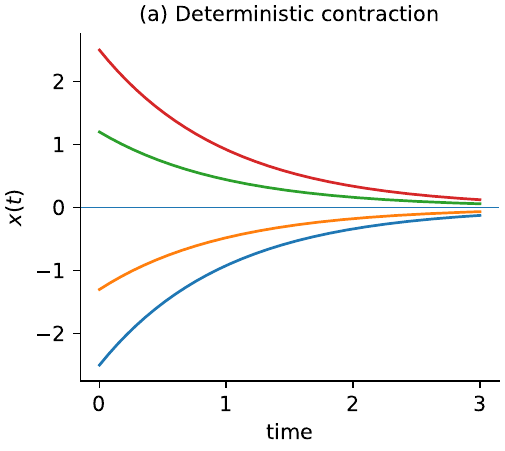}\\[-0.2em]
  {\small $\displaystyle \frac{dX_t}{dt}=-\alpha X_t$}
\end{minipage}\hfill
\begin{minipage}[t]{0.45\textwidth}
  \vspace{0pt}\centering
  \includegraphics[width=\linewidth]{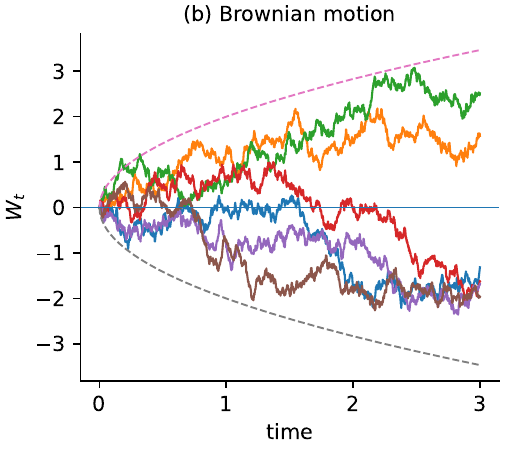}\\[-0.2em]
  {\small $\displaystyle W_{t+\Delta t}-W_t\sim\mathcal N(0,\Delta t)$}
\end{minipage}

%\vspace{0.8em}

\begin{minipage}[t]{0.45\textwidth}
  \vspace{0pt}\centering
  \includegraphics[width=\linewidth]{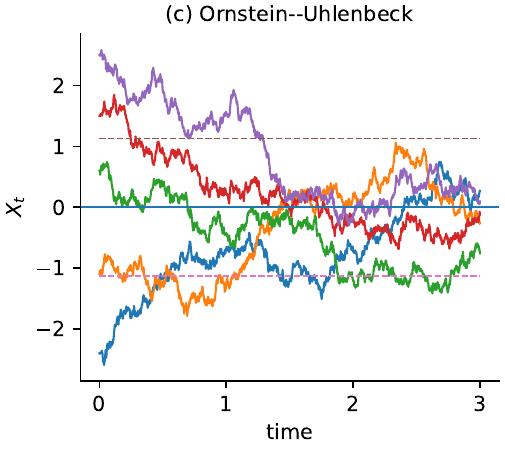}\\[-0.2em]
  {\small $\displaystyle dX_t=-\alpha X_t\,dt+\sigma\,dW_t$}
\end{minipage}\hfill
\begin{minipage}[t]{0.45\textwidth}
  \vspace{0pt}
  \caption{\textbf{Three fundamental continuous-time motion laws.}
  Deterministic contraction produces smooth trajectories that decay toward the
  origin. Brownian motion has zero-mean independent Gaussian increments, so
  paths wander with typical spread proportional to $\sqrt{t}$; the dashed curves
  show $\pm2\sqrt{t}$ for unit Brownian motion. The Ornstein--Uhlenbeck process
  combines random fluctuations with a restoring drift toward the origin; its
  dashed lines show twice the stationary standard deviation.}
  \label{fig:continuous-three-motion-laws}
\end{minipage}
\end{figure}

%% file: figures/ch05/fig_ch05_02_flow_density_jacobian.tex
\begin{figure}[H]
\centering
\resizebox{0.76\textwidth}{!}{%
\begin{tikzpicture}[
  >=Latex,
  every node/.style={font=\small},
  panel/.style={draw,rounded corners=2pt,minimum width=4.0cm,minimum height=3.25cm,align=center},
  eq/.style={draw,rounded corners=2pt,align=center,inner sep=5pt,text width=5.4cm},
  flow/.style={->,thick}
]
\node[font=\normalsize\bfseries] at (6.0,3.05)
  {Deterministic flow: local volume change controls density};

\node[panel] (left) at (2.0,0.65) {};
\node[panel] (right) at (10.0,0.65) {};

\node[font=\bfseries] at (2.0,1.85) {Initial local region};
\node[font=\bfseries] at (10.0,1.85) {After the flow map $F_t$};

\draw[thick] (1.05,0.05) rectangle (2.95,1.10);
\node at (2.0,0.58) {$\Delta V_0$};
\node at (2.0,-0.45) {$p_0(x_0)$};

\draw[thick] (9.25,0.25) rectangle (10.75,0.92);
\node at (10.0,0.58) {$\Delta V_t$};
\node at (10.0,-0.45) {$p_t(x_t)$};

\draw[flow] (4.0,0.75) -- node[above,align=center]
  {$x_t=F_t(x_0)$\\[-1pt]
   $J_t=\partial F_t/\partial x_0$} (8.0,0.75);

\node[eq] (volume) at (3.0,-2.05) {
  \textbf{Local volume scaling}\\[3pt]
  $\displaystyle \Delta V_t\approx |\det J_t|\,\Delta V_0$
};

\node[eq] (density) at (9.0,-2.05) {
  \textbf{Probability conservation}\\[3pt]
  $\displaystyle p_t(F_t(x_0))=\frac{p_0(x_0)}{|\det J_t|}$
};

\draw[flow] (volume) -- (density);

\node[eq,text width=7.0cm] (instant) at (6.0,-4.05) {
  \textbf{Infinitesimal limit along a trajectory}\\[3pt]
  $\displaystyle \frac{d}{dt}\log p(x(t),t)=-\nabla\!\cdot v(x(t),t)$
};

\end{tikzpicture}%
}
\caption{\textbf{Probability conservation links local volume change to
density change.}
A deterministic flow map $F_t$ moves particles and their surrounding volume
elements. The Jacobian determinant measures local volume scaling, so
conservation of probability rescales density by the inverse volume factor.
Infinitesimally, this gives the trajectory-wise Liouville relation: local
expansion lowers density, while compression raises it.}
\label{fig:continuous-flow-density-jacobian}
\end{figure}

%% file: figures/ch05/fig_ch05_03_liouville_fokker_planck.tex
\begin{figure}[H]
\centering
\begin{minipage}[t]{0.48\textwidth}
  \vspace{0pt}\centering
  \includegraphics[width=\linewidth]{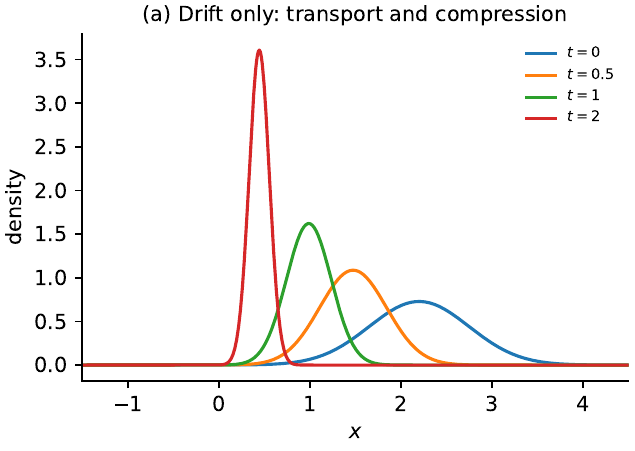}\\[-0.2em]
  {\small $\displaystyle \partial_t p=-\partial_x(fp)$}
\end{minipage}\hfill
\begin{minipage}[t]{0.48\textwidth}
  \vspace{0pt}\centering
  \includegraphics[width=\linewidth]{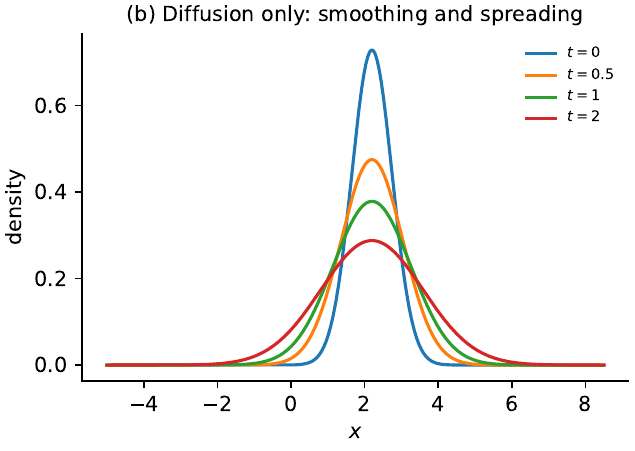}\\[-0.2em]
  {\small $\displaystyle \partial_t p=\frac{\sigma^2}{2}\partial_{xx}p$}
\end{minipage}

\vspace{0.8em}

\begin{minipage}[t]{0.48\textwidth}
  \vspace{0pt}\centering
  \includegraphics[width=\linewidth]{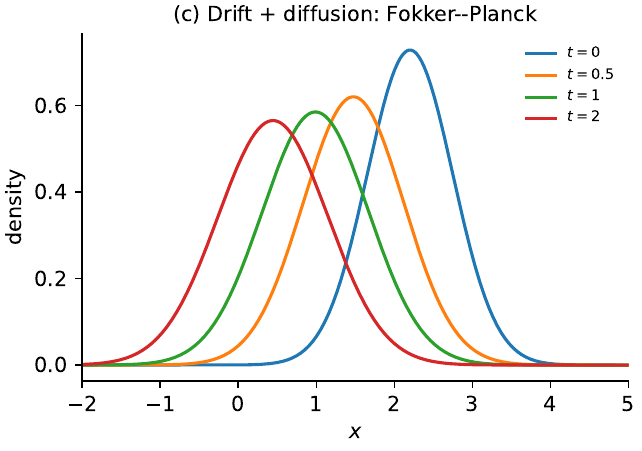}\\[-0.2em]
  {\small $\displaystyle \partial_t p=-\partial_x(fp)+\frac{\sigma^2}{2}\partial_{xx}p$}
\end{minipage}\hfill
\begin{minipage}[t]{0.48\textwidth}
  \vspace{0pt}
  \caption{\textbf{Drift transports probability; diffusion spreads it.}
  Starting from the same Gaussian density, deterministic contraction moves the
  mean toward the origin while shrinking the variance, whereas pure Brownian
  diffusion keeps the mean fixed while increasing the variance. The
  Ornstein--Uhlenbeck process combines both effects: its mean reverts toward zero
  while its variance approaches the stationary value $\sigma^2/(2\alpha)$.
  The curves are exact Gaussian density solutions.}
  \label{fig:continuous-liouville-fokker-planck}
\end{minipage}
\end{figure}

%% file: chapters/cht_score_diffusion.tex
\chapter{Score-Based Generative Modelling and Continuous-Time Diffusion}
\section*{Overview}
\addcontentsline{toc}{section}{Overview}

The earlier DDPM chapter introduced diffusion models as a discrete
latent-variable construction: data are gradually corrupted into noise, and a
learned reverse process gradually reconstructs structure. That viewpoint is
powerful because it connects diffusion models to the probabilistic modelling
ideas developed earlier in the book, including latent variables, Gaussian
transitions, posterior inference, and variational learning.

But diffusion models also admit another interpretation, one that has become
central to modern generative modelling. Instead of asking only how to write down
a reverse Markov chain, we may ask a more local and geometric question: if a
sample is currently located at \(x\), what direction should it move in so that
it becomes more likely under the distribution we want to generate from? The
answer is given by the \emph{score function}, $\nabla_x \log p(x),$
which points in the direction of steepest local increase of log-density.

This change of viewpoint is profound. A probability density describes where
mass is concentrated, but the score describes how to move through the space
defined by that density. It turns a distribution into a vector field. Once this
field is known, it can guide noisy samples back toward regions of high
probability. In this sense, score-based modelling transforms generative
modelling from the problem of explicitly evaluating a density into the problem
of learning the local geometry of that density.

This is one reason score-based generative modelling has become such an important
way of understanding diffusion models. Unlike models that generate data through
a single direct transformation, modern diffusion systems learn how to correct
noise gradually, step by step or continuously, across many levels of corruption.
At each level, the model needs local information about how a noisy sample should
be nudged so that it becomes slightly more data-like. The score provides
precisely this information.

The previous chapter supplied the calculus needed for this perspective. We
studied how particles move under deterministic and stochastic dynamics, and how
whole probability densities evolve through the Liouville and Fokker--Planck
equations. This chapter now uses those tools to explain how score fields,
stochastic sampling, and density evolution come together in score-based
diffusion.

The aim is not to replace the DDPM viewpoint, but to complement it. DDPMs show
how diffusion can be organised as a discrete latent-variable model. The
score-based view shows why denoising can also be understood geometrically and
dynamically: generation is guided by learned local directions along a path of
noisy distributions, gradually carrying samples from randomness back toward
structure.

\section*{Concept Map}
\addcontentsline{toc}{section}{Concept Map}

The chapter develops the score-based interpretation of diffusion models:
from local density geometry, to score learning, to multi-scale denoising, and
finally to continuous-time reverse diffusion.

\begin{itemize}

    \item[\(\rightarrow\)] \textbf{We begin with the score function.}  
    The score \(\nabla_x \log p(x)\) is interpreted as a local probabilistic
    direction field. It points toward regions where the log-density increases
    most rapidly.

    \item[\(\rightarrow\)] \textbf{We connect scores to sampling through Langevin dynamics.}  
    A deterministic score flow moves samples toward high-density regions, while
    Brownian noise prevents collapse and supports exploration. The
    Fokker--Planck equation shows why the target density is stationary under
    this score-plus-noise dynamics.

    \item[\(\rightarrow\)] \textbf{We then ask how the score can be learned.}  
    The Fisher divergence gives a natural objective for matching a learned score
    \(s_\theta(x)\) to the true data score. Score matching rewrites this
    objective using integration by parts, removing the need to evaluate the
    unknown data density.

    \item[\(\rightarrow\)] \textbf{We stabilise score learning through denoising.}  
    Instead of learning the raw data score directly, we add Gaussian noise and
    learn the score of the smoothed distribution \(p_\sigma\). This turns score
    estimation into a regression problem against a denoising direction.

    \item[\(\rightarrow\)] \textbf{We move from one noise level to many.}  
    Large noise levels reveal coarse global structure, while small noise levels
    reveal fine local detail. Multi-scale score learning therefore learns a
    family of score fields across different smoothing scales.

    \item[\(\rightarrow\)] \textbf{We connect multi-scale scores to diffusion.}  
    A family of noisy score fields becomes a diffusion model when the noisy
    distributions are organised as the marginals of one forward noising process.
    In continuous time, this gives
    a path of densities \(p_t(x)\), \(0\leq t\leq T\).%a path of densities $    p_t(x), 0\leq t\leq T.$

    \item[\(\rightarrow\)] \textbf{We reverse diffusion using the score.}  
    The forward process carries data toward noise. The reverse-time dynamics
    uses the time-dependent score $    \nabla_x \log p_t(x)$
    as the local correction that guides samples from noise back toward data.

    \item[\(\rightarrow\)] \textbf{We finally connect back to DDPMs.}  
    DDPMs were previously introduced as discrete latent-variable models. Here we
    reinterpret their noise-prediction objective as carrying score-like
    information about intermediate noisy distributions, showing that the DDPM
    and score-based views are complementary.

\end{itemize}

\medskip
\noindent The chapter's main score-based modelling pathway is summarised visually below.
\input{figures/ch06/chapter6_score_at_a_glance}
\medskip

\section{From the Score Function to Langevin Sampling}

The previous chapter explained how probability densities evolve under
deterministic transport and stochastic diffusion. We saw that ordinary
differential equations induce Liouville-type density evolution, while stochastic
differential equations with Brownian noise lead to the Fokker--Planck equation.
These results are important because they move the discussion beyond the motion
of individual particles and toward the evolution of whole distributions.

A natural next question is then the following: among all possible density
evolutions, which ones are actually useful for generative modelling? In
generative modelling, we do not seek arbitrary motion. Rather, we seek a
dynamics whose behaviour is connected to a probability density of interest,
typically the data distribution or a suitably smoothed version of it. This is
where the score function enters.

The score provides a local probabilistic direction field, and when combined
with noise, it leads to Langevin dynamics, one of the most important bridges
between density modelling and sampling. This section introduces that bridge
carefully. We first explain the score as a probabilistic drift field, then show
how noise is added through Langevin dynamics, and finally explain why the
resulting dynamics is tied to a target density through the notion of
stationarity.

The purpose of this section is therefore not yet to introduce modern diffusion
models in their full form. Rather, it is to establish a simpler and more
fundamental principle: if the score of a target density is available, then it
can be used as a drift field in a stochastic dynamics whose stationary
distribution is the target itself. This idea provides one of the main conceptual
bridges from describing a target density to constructing a dynamics that can
generate samples from it.

\subsection{The Score Function as a Probabilistic Drift Field}

Let \(p(x)\) be a probability density on \(\mathbb{R}^d\). Its \emph{score
function} is defined as
\[
s(x)=\nabla_x \log p(x).
\]
If the density depends on time as well, then the time-dependent score is written
as
\[
s(x,t)=\nabla_x \log p(x,t).
\]

At first sight, it may seem slightly surprising that the gradient of the
\emph{log}-density should be so important. The reason is that the score captures
local directional information in a particularly useful form. The density
\(p(x)\) tells us how much probability mass is concentrated near \(x\), whereas
the score tells us in which direction the log-density increases most rapidly.
It therefore provides a local vector field pointing toward regions of higher
probability.

In one dimension, the score becomes
\[
s(x)
=
\frac{d}{dx}\log p(x)
=
\frac{p'(x)}{p(x)}.
\]
This expression is already revealing. The numerator \(p'(x)\) measures the local
change in density, but dividing by \(p(x)\) normalises that change relative to
the local scale of the density itself. The score therefore measures a relative
slope rather than just an absolute slope. This makes it more meaningful as a
field of probabilistic motion.

A simple Gaussian example makes this interpretation concrete. Consider the
one-dimensional Gaussian density
\[
p(x)
=
\frac{1}{\sqrt{2\pi\sigma^2}}
\exp\left(
-\frac{(x-\mu)^2}{2\sigma^2}
\right).
\]
Taking logarithms gives
\[
\log p(x)
=
-\frac{(x-\mu)^2}{2\sigma^2}
-\frac{1}{2}\log(2\pi\sigma^2).
\]
Differentiating with respect to \(x\), we obtain
\[
\frac{d}{dx}\log p(x)
=
-\frac{x-\mu}{\sigma^2}.
\]
Hence the score is
\[
\boxed{
s(x)
=
-\frac{x-\mu}{\sigma^2}.
}
\]

This Gaussian example immediately shows the geometric meaning of the score. If
\(x>\mu\), then the score is negative, so it points back toward the mean. If
\(x<\mu\), then the score is positive, again pointing back toward the mean. The
further \(x\) is from the mean, the larger the magnitude of the score. Thus, for
a Gaussian density, the score behaves like a restoring field that directs
particles toward the high-density region.

In the multivariate Gaussian case,
\[
p(x)=\mathcal{N}(\mu,\Sigma),
\]
the score becomes
\[
\boxed{
\nabla_x \log p(x)
=
-\Sigma^{-1}(x-\mu).
}
\]
So again, the score is a vector field pointing toward the high-density centre,
but now shaped by the covariance structure. Directions with smaller variance
produce stronger restoring forces, while directions with larger variance produce
weaker ones. Figure~\ref{fig:score-field-geometry} shows how the same idea
extends beyond a single Gaussian: for a multimodal density, the score becomes a
spatially varying field whose local direction depends on the surrounding density
geometry.

\input{figures/ch06/fig_ch06_01_score_field_geometry}

This suggests a natural idea. If we define a deterministic flow by
\[
\frac{dx}{dt}
=
\nabla_x \log p(x),
\]
then particles move in the direction of steepest increase of log-density. Since
the logarithm is monotone, this also moves particles toward regions of higher
density. In this sense, the score may be interpreted as a \emph{probabilistic
drift field}: it tells us how to move if we wish to guide particles toward more
likely regions under the target distribution.

This viewpoint is extremely useful, but it also reveals an important limitation.
A purely deterministic score flow tends to move particles uphill toward modes.
It may therefore collapse trajectories into concentrated high-density regions
rather than explore the full target distribution. This is useful for
optimisation, but not yet sufficient for sampling. To obtain a genuine sampling
dynamics, we must add noise. This leads directly to Langevin dynamics.

\subsection{Langevin Dynamics: Score Plus Noise}

Langevin dynamics combines two ingredients:
\begin{enumerate}
    \item a drift term given by the score, which pulls particles toward
    higher-density regions;
    \item a Brownian noise term, which injects randomness and maintains
    exploration.
\end{enumerate}

In continuous time, the Langevin stochastic differential equation takes the form
\[
\boxed{
dX_t
=
\nabla_x \log p^\ast(X_t)\,dt
+
\sqrt{2}\,dW_t,
}
\]
where \(p^\ast(x)\) denotes the target density of interest. The notation
\(p^\ast\) is used here to emphasise that this is the density we would like to
sample from.

In discrete time, this corresponds heuristically to the update
\[
\boxed{
x_{k+1}
=
x_k
+
\epsilon\,\nabla_x \log p^\ast(x_k)
+
\sqrt{2\epsilon}\,\xi_k,
\qquad
\xi_k\sim\mathcal{N}(0,I),
}
\]
where \(\epsilon>0\) is a small step size.

This update has a direct interpretation. The current state \(x_k\) is first
moved by the score term
\[
\epsilon\,\nabla_x \log p^\ast(x_k),
\]
which nudges the particle toward regions of greater probability under the target
density. Then a Gaussian perturbation
\[
\sqrt{2\epsilon}\,\xi_k
\]
is added. This random term prevents simple collapse into a mode and allows the
particle to keep exploring the state space. In the continuous-time limit, the
deterministic score drift and the Brownian perturbation combine into the
Langevin SDE above.

The structure should look familiar from the previous chapter. Langevin dynamics
is a drift-plus-noise system, and therefore its density evolves according to a
Fokker--Planck equation. For the Langevin dynamics
\[
dX_t
=
\nabla_x \log p^\ast(X_t)\,dt
+
\sqrt{2}\,dW_t,
\]
the drift field is
\[
v(x)=\nabla_x \log p^\ast(x),
\]
and the diffusion coefficient is \(\sigma=\sqrt{2}\). Since the
Fokker--Planck equation for constant isotropic diffusion is
\[
\partial_t p(x,t)
=
-\nabla\cdot\bigl(p(x,t)v(x,t)\bigr)
+
\frac{\sigma^2}{2}\Delta p(x,t),
\]
substituting \(v(x)=\nabla_x\log p^\ast(x)\) and
\(\sigma^2=2\) gives
\[
\boxed{
\partial_t p(x,t)
=
-\nabla\cdot\bigl(p(x,t)\nabla_x \log p^\ast(x)\bigr)
+
\Delta p(x,t).
}
\]

This equation combines two effects. The first term, $-\nabla\cdot\bigl(p\nabla_x \log p^\ast\bigr),$ describes deterministic transport induced by the score field. The second term, $\Delta p,$ describes diffusion caused by Brownian noise. The drift term tends to
concentrate probability toward high-density regions of the target, while the
diffusion term tends to spread probability outward. The important fact is that,
with exactly this combination, the target density becomes stationary. This is
the key reason Langevin dynamics is useful as a sampling mechanism.

It is also useful to compare this with the Ornstein--Uhlenbeck process
introduced earlier. The OU process has a linear drift toward the origin and a
Gaussian stationary distribution. Langevin dynamics generalises the same
principle: instead of using a fixed linear restoring force, it uses the score of
a chosen target density as the restoring field. In this sense, the OU process
may be viewed as a Gaussian special case of a broader score-based sampling
mechanism.

To see this explicitly, suppose the target density is the standard Gaussian
\(p^\ast(x)=\mathcal{N}(0,I)\). Then
\[
\nabla_x\log p^\ast(x)=-x,
\]
and Langevin dynamics becomes
\[
dX_t=-X_t\,dt+\sqrt{2}\,dW_t,
\]
which is precisely an Ornstein--Uhlenbeck process with a particular noise
scaling. Thus the Langevin equation recovers the OU process when the target
density is Gaussian, and extends it to non-Gaussian targets through the score
\(\nabla_x\log p^\ast(x)\).

\subsection{Target and Stationary Distributions}

We now explain why Langevin dynamics is not merely a formal equation, but a
meaningful sampling mechanism.

Suppose a density \(p(x,t)\) evolves according to some density evolution
equation
\[
\partial_t p(x,t)
=
\mathcal{L}[p(x,t)].
\]
A density \(p^\ast(x)\) is called a \emph{stationary distribution} if, once the
system has density \(p^\ast\), it remains there. In PDE language, this means
\[
\mathcal{L}[p^\ast(x)]=0.
\]
Equivalently,
\[
\partial_t p(x,t)=0
\qquad
\text{when}
\qquad
p(x,t)=p^\ast(x).
\]

This concept is important in generative modelling. If the target density is
stationary for a dynamics, then the dynamics is aligned with that target
distribution. If, in addition, trajectories or densities started elsewhere
approach this stationary law over time, then the process becomes a mechanism for
sampling from the target distribution. A full convergence theory requires
additional assumptions and is beyond the present tutorial, but the stationarity
calculation already reveals the central idea.

Let us now verify stationarity for Langevin dynamics. Recall that its
Fokker--Planck equation is
\[
\partial_t p(x,t)
=
-\nabla\cdot\bigl(p(x,t)\nabla_x \log p^\ast(x)\bigr)
+
\Delta p(x,t).
\]
We now substitute
\[
p(x,t)=p^\ast(x).
\]
Then the right-hand side becomes
\[
-\nabla\cdot\bigl(p^\ast(x)\nabla_x \log p^\ast(x)\bigr)
+
\Delta p^\ast(x).
\]
Now use the identity
\[
p^\ast(x)\nabla_x \log p^\ast(x)
=
\nabla_x p^\ast(x),
\]
because
\[
\nabla_x \log p^\ast(x)
=
\frac{\nabla_x p^\ast(x)}{p^\ast(x)}.
\]
Therefore the previous expression simplifies to
\[
-\nabla\cdot(\nabla_x p^\ast(x))
+
\Delta p^\ast(x).
\]
But by definition,
\[
\nabla\cdot(\nabla_x p^\ast)
=
\Delta p^\ast.
\]
Hence
\[
-\Delta p^\ast(x)+\Delta p^\ast(x)=0.
\]
So indeed,
\[
\boxed{
\partial_t p(x,t)=0
\qquad
\text{when}
\qquad
p(x,t)=p^\ast(x).
}
\]

Therefore \(p^\ast\) is a stationary distribution of the Langevin dynamics
\[
\boxed{
dX_t
=
\nabla_x \log p^\ast(X_t)\,dt
+
\sqrt{2}\,dW_t.
}
\]

This is one of the central conceptual results behind score-based generative
modelling. It shows that if we use the score of a target density as the drift
field, and combine it with the appropriate Brownian noise, then the resulting
density evolution preserves that target density. The score is therefore not
merely an interesting derivative of the density. It is precisely the drift field
that makes the desired density stationary under a natural stochastic dynamics.

At the same time, it is important to be precise about what has been shown. The
argument above establishes \emph{stationarity}, not a full proof that every
initial density converges to \(p^\ast\). The latter is a deeper question
involving the long-time behaviour of the stochastic process. Nevertheless, the
stationarity result closes an important conceptual gap. It explains why
score-based drift, together with Brownian noise, provides a principled route
toward sampling from distributions of interest.

If the Langevin process is allowed to evolve long enough under suitable
conditions, its distribution may approach the stationary law \(p^\ast\). At the
density level, this means that probability mass progressively takes on the shape
of the desired target distribution. At the particle level, it means that
repeatedly updating samples through Langevin dynamics can produce particles
whose statistical behaviour resembles draws from the target law.

Thus the density-level and particle-level viewpoints are consistent with one
another. The Fokker--Planck equation explains how the distribution evolves,
while the Langevin update explains how individual samples move. The score links
the two: it acts locally as a probabilistic drift field, and globally as the
ingredient that makes the target density stationary.

\section{Learning the Score: Fisher Divergence and Score Matching}

The previous section showed why the score function is important for sampling.
If the score of a target density \(p^\ast(x)\) is available, then it can be used
as the drift field in Langevin dynamics,
\[
dX_t
=
\nabla_x \log p^\ast(X_t)\,dt
+
\sqrt{2}\,dW_t,
\]
whose stationary distribution is \(p^\ast\). This gives a principled connection
between a probability density and a stochastic sampling dynamics.

In generative modelling, however, the target density is not usually available
in closed form. We are given samples from an unknown data-generating
distribution, but not the density itself. Let this unknown distribution be
denoted by \(p_{\text{data}}(x)\). The central learning problem is therefore to
approximate its score
\[
s_{\text{data}}(x)
=
\nabla_x \log p_{\text{data}}(x).
\]
If we can learn this vector field from data, then we can use it inside
Langevin-type sampling dynamics.

We model the score using a parameterised function \(s_\theta(x)\), typically a
neural network with parameters \(\theta\). A natural objective is to minimise
the discrepancy between the learned score and the true data score under the data
distribution. This leads to the \emph{Fisher divergence}:
\[
\mathcal{J}(\theta)
=
\mathbb{E}_{x\sim p_{\text{data}}}
\left[
\left\|
s_\theta(x)-s_{\text{data}}(x)
\right\|_2^2
\right].
\]
Equivalently, the learning problem is $\min_\theta \mathcal{J}(\theta).$ Here the divergence is defined between vector fields, while optimisation is
performed with respect to the parameters \(\theta\). If
\(\mathcal{J}(\theta)=0\), then
\[
s_\theta(x)=s_{\text{data}}(x)
\]
almost everywhere under \(p_{\text{data}}\). In that case, the learned vector
field matches the local geometry of the data distribution.

There is, however, an immediate difficulty. The objective appears to require the
true score \(s_{\text{data}}(x)\), which depends on the unknown density
\(p_{\text{data}}(x)\). Score matching~\cite{hyvarinen2005estimation} resolves this difficulty by rewriting the
objective so that the unknown score no longer appears explicitly.

\subsection{Expanding the Fisher Divergence}

We begin by expanding the squared norm:
\[
\begin{aligned}
\mathcal{J}(\theta)
&=
\mathbb{E}_{p_{\text{data}}}
\left[
\bigl(s_\theta(x)-s_{\text{data}}(x)\bigr)^\top
\bigl(s_\theta(x)-s_{\text{data}}(x)\bigr)
\right] \\[4pt]
&=
\mathbb{E}_{p_{\text{data}}}
\left[
\|s_\theta(x)\|_2^2
-
2s_\theta(x)^\top s_{\text{data}}(x)
+
\|s_{\text{data}}(x)\|_2^2
\right].
\end{aligned}
\]
The final term,
\[
\mathbb{E}_{p_{\text{data}}}
\left[
\|s_{\text{data}}(x)\|_2^2
\right],
\]
does not depend on \(\theta\), and therefore does not affect the optimisation.
Thus minimising \(\mathcal{J}(\theta)\) is equivalent to minimising
\[
\tilde{\mathcal{J}}(\theta)
=
\mathbb{E}_{p_{\text{data}}}
\left[
\|s_\theta(x)\|_2^2
\right]
-
2
\mathbb{E}_{p_{\text{data}}}
\left[
s_\theta(x)^\top s_{\text{data}}(x)
\right].
\]

The first term is directly computable from samples, because it only requires
evaluating the model \(s_\theta(x)\) at data points. The challenge lies in the
cross term
\[
\mathbb{E}_{p_{\text{data}}}
\left[
s_\theta(x)^\top s_{\text{data}}(x)
\right],
\]
because it still involves the unknown true score.

The key idea of score matching is to rewrite this cross term using integration
by parts. We first do this carefully in one dimension, where the calculation is
most transparent, and then extend the result to multiple dimensions.

\subsection{The One-Dimensional Case}

To make the main idea clear, suppose a one-dimensional scenario, with the true data score
\[
s_{\text{data}}(x)
=
\frac{d}{dx}\log p_{\text{data}}(x)
=
\frac{1}{p_{\text{data}}(x)}
\frac{d}{dx}p_{\text{data}}(x).
\]
Substituting this into the cross term gives
\[
\begin{aligned}
\mathbb{E}_{p_{\text{data}}}
\left[
s_\theta(x)s_{\text{data}}(x)
\right]
&=
\int
s_\theta(x)
\frac{d}{dx}\log p_{\text{data}}(x)
p_{\text{data}}(x)
\,dx \\[4pt]
&=
\int
s_\theta(x)
\frac{1}{p_{\text{data}}(x)}
\frac{d}{dx}p_{\text{data}}(x)
p_{\text{data}}(x)
\,dx \\[4pt]
&=
\int
s_\theta(x)
\frac{d}{dx}p_{\text{data}}(x)
\,dx.
\end{aligned}
\]

At this stage, the derivative acts on the unknown density
\(p_{\text{data}}(x)\). Our goal is to transfer this derivative onto the model
\(s_\theta(x)\) instead, using  the integration by parts.

\begin{tcolorbox}[
  colback=blue!4,
  colframe=blue!50!black,
  title={Integration by parts in one dimension},
  boxrule=0.6pt,
  arc=2mm,
  left=2mm,
  right=2mm,
  top=1mm,
  bottom=1mm
]
The product rule states that
\[
\frac{d}{dx}\bigl(u(x)v(x)\bigr)
=
u'(x)v(x)+u(x)v'(x).
\]
Integrating both sides gives
\[
\int u(x)v'(x)\,dx
=
\bigl[u(x)v(x)\bigr]_{-\infty}^{\infty}
-
\int u'(x)v(x)\,dx.
\]
This identity transfers a derivative from \(v\) to \(u\), at the cost of a
boundary term.
\end{tcolorbox}

We now apply this identity by choosing
\[
u(x)=s_\theta(x),
\qquad
v'(x)=\frac{d}{dx}p_{\text{data}}(x),
\qquad
v(x)=p_{\text{data}}(x).
\]
Then
\[
\int
s_\theta(x)
\frac{d}{dx}p_{\text{data}}(x)
\,dx
=
\bigl[
s_\theta(x)p_{\text{data}}(x)
\bigr]_{-\infty}^{\infty}
-
\int
\frac{d}{dx}s_\theta(x)
p_{\text{data}}(x)
\,dx.
\]

Under the standard assumption that \(p_{\text{data}}(x)\) decays sufficiently
fast as \(|x|\to\infty\), and that \(s_\theta(x)\) does not grow too rapidly,
the product \(s_\theta(x)p_{\text{data}}(x)\) vanishes at the boundary. Hence
the boundary contribution is zero. We therefore obtain
\[
\int
s_\theta(x)
\frac{d}{dx}p_{\text{data}}(x)
\,dx
=
-
\int
\frac{d}{dx}s_\theta(x)
p_{\text{data}}(x)
\,dx.
\]
Recognising the right-hand side as an expectation under \(p_{\text{data}}\), we
conclude that
\[
\mathbb{E}_{p_{\text{data}}}
\left[
s_\theta(x)s_{\text{data}}(x)
\right]
=
-
\mathbb{E}_{p_{\text{data}}}
\left[
\frac{d}{dx}s_\theta(x)
\right].
\]

The derivative has therefore been transferred from the unknown
density \(p_{\text{data}}(x)\) to the model \(s_\theta(x)\), requiring no longer explicit knowledge of the true data score. Substituting this result into the reduced Fisher divergence gives
\[
\begin{aligned}
\tilde{\mathcal{J}}(\theta)
&=
\mathbb{E}_{p_{\text{data}}}
\left[
s_\theta(x)^2
\right]
-
2
\mathbb{E}_{p_{\text{data}}}
\left[
s_\theta(x)s_{\text{data}}(x)
\right] \\[4pt]
&=
\mathbb{E}_{p_{\text{data}}}
\left[
s_\theta(x)^2
\right]
+
2
\mathbb{E}_{p_{\text{data}}}
\left[
\frac{d}{dx}s_\theta(x)
\right].
\end{aligned}
\]

Since multiplying an objective by a positive constant does not change its
minimiser, it is common to write the one-dimensional score matching objective as
\[
\boxed{
\mathcal{L}_{\text{SM}}(\theta)
=
\mathbb{E}_{p_{\text{data}}}
\left[
\frac{1}{2}s_\theta(x)^2
+
\frac{d}{dx}s_\theta(x)
\right].
}
\]

\subsection{Extension to Multiple Dimensions}

We now extend the same argument to the general case
\(x\in\mathbb{R}^d\). The true data score is
\[
s_{\text{data}}(x)
=
\nabla_x\log p_{\text{data}}(x),
\]
and the cross term in the Fisher divergence becomes
\[
\begin{aligned}
\mathbb{E}_{p_{\text{data}}}
\left[
s_\theta(x)^\top s_{\text{data}}(x)
\right]
&=
\int
s_\theta(x)^\top
\nabla_x\log p_{\text{data}}(x)
p_{\text{data}}(x)
\,dx \\[4pt]
&=
\int
s_\theta(x)^\top
\nabla_x p_{\text{data}}(x)
\,dx.
\end{aligned}
\]

The role played by the ordinary derivative in one dimension is now played by
the divergence. For a vector field
\[
v(x)=
\begin{bmatrix}
v_1(x)\\
\vdots\\
v_d(x)
\end{bmatrix},
\]
the divergence is
\[
\nabla_x\cdot v(x)
=
\sum_{i=1}^d
\frac{\partial v_i(x)}{\partial x_i}.
\]
As introduced in the previous chapter, divergence measures the local net
outflow of a vector field. Here it appears because the model score
\(s_\theta(x)\) is itself a vector field.

The multidimensional analogue of integration by parts follows from the
divergence theorem. The relevant product rule is
\[
\nabla_x\cdot
\bigl(p_{\text{data}}(x)s_\theta(x)\bigr)
=
s_\theta(x)^\top\nabla_x p_{\text{data}}(x)
+
p_{\text{data}}(x)\nabla_x\cdot s_\theta(x).
\]
Rearranging gives
\[
s_\theta(x)^\top\nabla_x p_{\text{data}}(x)
=
\nabla_x\cdot
\bigl(p_{\text{data}}(x)s_\theta(x)\bigr)
-
p_{\text{data}}(x)\nabla_x\cdot s_\theta(x).
\]
Integrating over \(\mathbb{R}^d\), and assuming suitable boundary conditions so
that the divergence term contributes no boundary flux, we obtain
\[
\int
s_\theta(x)^\top\nabla_x p_{\text{data}}(x)
\,dx
=
-
\int
p_{\text{data}}(x)
\nabla_x\cdot s_\theta(x)
\,dx.
\]
Therefore,
\[
\boxed{
\mathbb{E}_{p_{\text{data}}}
\left[
s_\theta(x)^\top s_{\text{data}}(x)
\right]
=
-
\mathbb{E}_{p_{\text{data}}}
\left[
\nabla_x\cdot s_\theta(x)
\right].
}
\]

This is the direct multidimensional counterpart of the one-dimensional result
\[
\mathbb{E}_{p_{\text{data}}}
\left[
s_\theta(x)s_{\text{data}}(x)
\right]
=
-
\mathbb{E}_{p_{\text{data}}}
\left[
\frac{d}{dx}s_\theta(x)
\right].
\]
The only structural change is that the derivative of the scalar score is
replaced by the divergence of the vector-valued score field.

Substituting the cross-term identity into the reduced Fisher divergence gives
\[
\tilde{\mathcal{J}}(\theta)
=
\mathbb{E}_{p_{\text{data}}}
\left[
\|s_\theta(x)\|_2^2
+
2\nabla_x\cdot s_\theta(x)
\right].
\]
Equivalently, after multiplying by \(1/2\), the general score matching objective
is
\[
\boxed{
\mathcal{L}_{\text{SM}}(\theta)
=
\mathbb{E}_{p_{\text{data}}}
\left[
\frac{1}{2}
\|s_\theta(x)\|_2^2
+
\nabla_x\cdot s_\theta(x)
\right].
}
\]

This is the classical score matching objective. It was derived from the Fisher
divergence between the learned score and the true data score, but the final
expression no longer contains \(s_{\text{data}}(x)\) explicitly. 
Once \(s_\theta(x)\) approximates \(s_{\text{data}}(x)\), it can be inserted into the
Langevin dynamics introduced earlier. Under suitable conditions, this learned
dynamics can then be used to generate samples whose distribution approximates
the data distribution.

\section{Stabilising Learning: Denoising Score Matching via Corruption}

In the previous section, we derived the score matching objective, which allows a
score function \(s_\theta(x)\) to be learned from data without explicit
knowledge of the data density. While theoretically elegant, direct score
matching can be difficult to apply in high-dimensional generative modelling.

The core difficulty stems from the geometry of real data distributions.
Empirical data often concentrate on or near low-dimensional manifolds embedded
in high-dimensional space. As a result, the data density may be sharply
concentrated, irregular, or poorly behaved away from the data support. Directly
learning the score of such a distribution can therefore be unstable, because the
induced vector field may behave erratically in regions with little or no data
coverage.

A natural remedy is to introduce noise. Gaussian corruption turns the empirical
or sharply concentrated data distribution into a smoothed distribution whose score field is much better
behaved. Instead of learning the score of the original data
distribution directly, we learn the score of a noisy version of it. This idea is
the basis of \emph{denoising score matching~\cite{vincent2011connection}}.

\subsection{Gaussian Corruption and Smoothed Scores}

We begin by perturbing the data with additive Gaussian noise. Given a clean
sample
\[
x\sim p_{\text{data}}(x),
\]
we construct a noisy observation
\[
\tilde{x}
=
x+\varepsilon,
\qquad
\varepsilon\sim\mathcal{N}(0,\sigma^2 I).
\]
Equivalently, the conditional distribution of the corrupted sample given the
clean sample is
\[
p(\tilde{x}\mid x)
=
\mathcal{N}(\tilde{x};x,\sigma^2 I).
\]

The marginal distribution of \(\tilde{x}\) is obtained by integrating out the
clean variable \(x\):
\[
p_\sigma(\tilde{x})
=
\int
p_{\text{data}}(x)
p(\tilde{x}\mid x)
\,dx.
\]
This is a convolution of the original data distribution with a Gaussian kernel.
The effect is smoothing: probability mass is spread into nearby regions, sharp
structures are softened, and areas that previously had negligible density may
receive nonzero probability under the corrupted distribution.

Rather than learning the score of the original data distribution, we now aim to
learn the score of this smoothed density:
\[
s_\sigma(\tilde{x})
=
\nabla_{\tilde{x}}\log p_\sigma(\tilde{x}).
\]

We next examine the structure of this noisy score. By the logarithmic gradient
identity,
\[
\nabla_{\tilde{x}}\log p_\sigma(\tilde{x})
=
\frac{1}{p_\sigma(\tilde{x})}
\nabla_{\tilde{x}}p_\sigma(\tilde{x}).
\]
The gradient of the smoothed density is
\[
\begin{aligned}
\nabla_{\tilde{x}}p_\sigma(\tilde{x})
&=
\nabla_{\tilde{x}}
\int
p_{\text{data}}(x)
p(\tilde{x}\mid x)
\,dx \\[4pt]
&=
\int
p_{\text{data}}(x)
\nabla_{\tilde{x}}p(\tilde{x}\mid x)
\,dx,
\end{aligned}
\]
where we differentiate under the integral sign under standard regularity
conditions. Therefore,
\[
s_\sigma(\tilde{x})
=
\frac{
\int
p_{\text{data}}(x)
\nabla_{\tilde{x}}p(\tilde{x}\mid x)
\,dx
}{
\int
p_{\text{data}}(x)
p(\tilde{x}\mid x)
\,dx
}.
\]

This expression has a useful probabilistic interpretation. By Bayes' rule, the
posterior distribution of the clean variable given the corrupted observation is
\[
p(x\mid\tilde{x})
=
\frac{
p_{\text{data}}(x)p(\tilde{x}\mid x)
}{
p_\sigma(\tilde{x})
}.
\]
Using this posterior, the noisy score can be written as
\[
s_\sigma(\tilde{x})
=
\int
p(x\mid\tilde{x})
\nabla_{\tilde{x}}\log p(\tilde{x}\mid x)
\,dx,
\]
or equivalently,
\[
s_\sigma(\tilde{x})
=
\mathbb{E}_{x\sim p(x\mid\tilde{x})}
\left[
\nabla_{\tilde{x}}\log p(\tilde{x}\mid x)
\right].
\]

For Gaussian corruption,
\[
p(\tilde{x}\mid x)
=
\mathcal{N}(\tilde{x};x,\sigma^2 I).
\]
Its logarithm is, up to constants independent of \(\tilde{x}\),
\[
\log p(\tilde{x}\mid x)
=
-\frac{1}{2\sigma^2}
\|\tilde{x}-x\|_2^2
+\text{constant}.
\]
Hence
\[
\nabla_{\tilde{x}}\log p(\tilde{x}\mid x)
=
-\frac{1}{\sigma^2}(\tilde{x}-x)
=
\frac{1}{\sigma^2}(x-\tilde{x}).
\]
Substituting this into the conditional expectation gives
\[
\boxed{
s_\sigma(\tilde{x})
=
\frac{1}{\sigma^2}
\mathbb{E}_{x\sim p(x\mid\tilde{x})}
\left[
x-\tilde{x}
\right].
}
\]

This identity gives an intuitive interpretation of the smoothed score. At a
corrupted point \(\tilde{x}\), the score points toward the conditional average
clean point that could have produced it. In other words, the noisy score gives a
local denoising direction: it tells us how \(\tilde{x}\) should be shifted, on
average, in order to move back toward the data distribution.

\subsection{The Intractability of the Noisy Score}

The identity above is conceptually very useful, but it does not yet give us a
directly computable training target. The forward corruption process is simple:
we can sample
\[
x\sim p_{\text{data}}(x),
\qquad
\tilde{x}=x+\varepsilon,
\qquad
\varepsilon\sim\mathcal{N}(0,\sigma^2 I).
\]
Thus we can generate pairs \((x,\tilde{x})\) without difficulty.

The problem is that the noisy score involves an expectation under the posterior
distribution \(p(x\mid\tilde{x})\):
\[
s_\sigma(\tilde{x})
=
\frac{1}{\sigma^2}
\mathbb{E}_{x\sim p(x\mid\tilde{x})}
\left[
x-\tilde{x}
\right].
\]
By Bayes' rule,
\[
p(x\mid\tilde{x})
=
\frac{
p(\tilde{x}\mid x)p_{\text{data}}(x)
}{
p_\sigma(\tilde{x})
},
\]
where
\[
p_\sigma(\tilde{x})
=
\int
p_{\text{data}}(x)p(\tilde{x}\mid x)\,dx.
\]
Although the likelihood \(p(\tilde{x}\mid x)\) is known, the data distribution
\(p_{\text{data}}(x)\) is not available in closed form. We only have samples
from it. Therefore the marginal density \(p_\sigma(\tilde{x})\) and the
posterior \(p(x\mid\tilde{x})\) are not directly available.

This prevents us from directly minimising the noisy Fisher divergence
\[
\mathbb{E}_{\tilde{x}\sim p_\sigma}
\left[
\left\|
s_\theta(\tilde{x})
-
s_\sigma(\tilde{x})
\right\|_2^2
\right],
\]
because evaluating this objective would require explicit access to the true
noisy score \(s_\sigma(\tilde{x})\). Thus, although the forward corruption
process is tractable, the corresponding score remains inaccessible in closed
form.

We therefore face a familiar situation: we have a well-defined score estimation
problem, but its target cannot be computed directly. The crucial question is
whether the known corruption process can be used to avoid evaluating the
posterior.

\subsection{Denoising Score Matching as Regression}

The key observation is that the corruption process defines a tractable joint
distribution over \((x,\tilde{x})\):
\[
p_{\text{data}}(x)p(\tilde{x}\mid x).
\]
We may not know the marginal density \(p_\sigma(\tilde{x})\) or the posterior
\(p(x\mid\tilde{x})\) explicitly, but we can sample from this joint distribution
exactly: draw a clean data point \(x\), add Gaussian noise, and obtain
\(\tilde{x}\).

Denoising score matching exploits this joint sampling procedure through a basic
property of least-squares regression.

\begin{tcolorbox}[
  colback=blue!4,
  colframe=blue!50!black,
  title={Least-squares regression identity},
  boxrule=0.6pt,
  arc=2mm,
  left=2mm,
  right=2mm,
  top=1mm,
  bottom=1mm
]
For random variables \((U,Y)\), the function minimising
\[
\mathbb{E}
\left[
\|f(U)-Y\|_2^2
\right]
\]
is
\[
f^\star(U)=\mathbb{E}[Y\mid U].
\]
Thus, under squared error loss, the best prediction is the conditional mean of
the target given the input.
\end{tcolorbox}

We now apply this identity to the denoising setting. Let the input variable be
the corrupted observation
\[
U=\tilde{x},
\]
and let the regression target be the scaled residual
\[
Y=
\frac{1}{\sigma^2}(x-\tilde{x}).
\]
Now consider the objective
\[
\boxed{
\mathcal{L}_{\text{DSM}}(\theta)
=
\mathbb{E}_{x\sim p_{\text{data}},\,
\tilde{x}\sim p(\tilde{x}\mid x)}
\left[
\left\|
s_\theta(\tilde{x})
-
\frac{1}{\sigma^2}(x-\tilde{x})
\right\|_2^2
\right].
}
\]

By the least-squares regression identity, the optimal function satisfies
\[
s_\theta^\star(\tilde{x})
=
\mathbb{E}
\left[
\frac{1}{\sigma^2}(x-\tilde{x})
\;\middle|\;
\tilde{x}
\right].
\]
But this conditional expectation is exactly the noisy score derived earlier:
\[
s_\sigma(\tilde{x})
=
\frac{1}{\sigma^2}
\mathbb{E}
\left[
x-\tilde{x}
\;\middle|\;
\tilde{x}
\right].
\]
Therefore,
\[
\boxed{
s_\theta^\star(\tilde{x})=s_\sigma(\tilde{x}).
}
\]

This is the central insight of denoising score matching. Individual training
pairs provide tractable conditional targets, while squared-error regression
averages those targets conditionally on \(\tilde{x}\), recovering the score of the
smoothed density. Figure~\ref{fig:score-dsm-regression} makes this averaging
geometry explicit.

\input{figures/ch06/fig_ch06_03_dsm_regression}

Equivalently, since
\[
\tilde{x}=x+\varepsilon,
\]
we have
\[
x-\tilde{x}=-\varepsilon.
\]
Thus the denoising score matching objective may also be written as
\[
\mathcal{L}_{\text{DSM}}(\theta)
=
\mathbb{E}_{x\sim p_{\text{data}},\,
\varepsilon\sim\mathcal{N}(0,\sigma^2 I)}
\left[
\left\|
s_\theta(x+\varepsilon)
+
\frac{\varepsilon}{\sigma^2}
\right\|_2^2
\right].
\]
This form makes the learning target especially concrete: the model is trained
to predict the negative noise direction, scaled by \(1/\sigma^2\).

Denoising score matching therefore replaces an intractable score estimation
problem with a supervised regression problem defined entirely through a known
corruption process. The objective depends only on clean data samples and
Gaussian noise. It does not require evaluating \(p_{\text{data}}(x)\),
\(p_\sigma(\tilde{x})\), or the posterior \(p(x\mid\tilde{x})\).

It is important to note that, for a fixed \(\sigma\), the model learns the score
of the corrupted density \(p_\sigma\), not directly the score of
\(p_{\text{data}}\). The noise level therefore controls the distribution whose
geometry is being learned.
However, learning the score at a single noise level raises a further question:
how should the noise scale \(\sigma\) be chosen?

\section{From Multi-Scale Score Learning to Continuous-Time Diffusion}

So far, this chapter has developed score-based generative modelling from the
bottom up. We first introduced the score as a probabilistic drift field, then
showed how Langevin dynamics uses the score for sampling, and then derived
score matching and denoising score matching as ways of learning score fields
from data.

We now make the final conceptual transition: from learning the score at one
noise level to learning scores across a whole path of noisy distributions, a
multi-scale viewpoint that became central in modern score-based generative
modelling~\cite{song2019generative,song2021scorebased}.
This is the point at which score-based learning begins to connect directly with
diffusion models.

This connection should be understood carefully. In the earlier DDPM chapter, we
introduced diffusion models from a discrete latent-variable perspective: a
forward noising Markov chain, a learned reverse chain, and a variational
training objective. Here we approach the same broad idea from a different
direction. Starting from score learning, we ask how a family of noisy score
fields can be used to move samples from noise back to data. This leads first to
multi-scale denoising score matching, then to annealed Langevin dynamics, and
finally to a continuous-time diffusion process.

Thus, this section does not repeat the DDPM derivation. Instead, it shows how
the score-based viewpoint provides a continuous-time route to diffusion:
\[
\begin{aligned}
\text{single noisy score}
&\;\longrightarrow\;
\text{many noisy scores}
\\
&\;\longrightarrow\;
\text{annealed reverse sampling}
\;\longrightarrow\;
\text{continuous-time score-guided diffusion}.
\end{aligned}
\]

\subsection{Multi-Scale Denoising Score Matching}

The previous section showed that, for a fixed noise level \(\sigma\), denoising
score matching learns the score of a smoothed data distribution \(p_\sigma\).
We corrupt data by
\[
\tilde{x}=x+\varepsilon,
\qquad
\varepsilon\sim\mathcal{N}(0,\sigma^2I),
\]
and train
\[
s_\theta(\tilde{x},\sigma)
\approx
\nabla_{\tilde{x}}\log p_\sigma(\tilde{x}).
\]
The corresponding denoising target is
\[
\frac{1}{\sigma^2}(x-\tilde{x})
=
-\frac{\varepsilon}{\sigma^2}.
\]

However, a single noise level gives only one smoothed view of the data
distribution. If \(\sigma\) is large, the density is smooth and the score field
provides stable global guidance toward the broad data region, but fine details
are blurred away. If \(\sigma\) is small, the density is closer to the original
data distribution and the score field captures sharper local detail, but it may
be difficult to estimate reliably away from the data support. Thus one noise
level cannot simultaneously provide robust global guidance and precise local
refinement.

Figure~\ref{fig:score-multiscale-fields} shows this effect for the same
multimodal density across three Gaussian smoothing scales: large noise reveals
coarse global geometry, whereas small noise recovers sharper local structure.

\input{figures/ch06/fig_ch06_04_multiscale_scores}

This suggests that the noise scale should not be chosen once and for all.
Instead, we should learn a family of score fields across a sequence of noise
levels.

Let
\[
\sigma_1>\sigma_2>\cdots>\sigma_K>0
\]
be a decreasing sequence of noise scales. For each \(\sigma_k\), define the
corrupted density
\[
p_{\sigma_k}(\tilde{x})
=
\int
p_{\text{data}}(x)
\mathcal{N}(\tilde{x};x,\sigma_k^2I)
\,dx.
\]
Instead of learning one score field, we now train a noise-conditional model
\[
s_\theta(\tilde{x},\sigma_k)
\approx
\nabla_{\tilde{x}}\log p_{\sigma_k}(\tilde{x}).
\]

The denoising score matching objective extends directly from the single-scale
case:
\[
\boxed{
\mathcal{L}_{\text{MS-DSM}}(\theta)
=
\mathbb{E}_{k}
\mathbb{E}_{x\sim p_{\text{data}},\,
\tilde{x}\sim\mathcal{N}(x,\sigma_k^2I)}
\left[
\left\|
s_\theta(\tilde{x},\sigma_k)
-
\frac{1}{\sigma_k^2}(x-\tilde{x})
\right\|_2^2
\right].
}
\]
Equivalently, writing \(\tilde{x}=x+\varepsilon\) with
\(\varepsilon\sim\mathcal{N}(0,\sigma_k^2I)\), this becomes
\[
\mathcal{L}_{\text{MS-DSM}}(\theta)
=
\mathbb{E}_{k}
\mathbb{E}_{x\sim p_{\text{data}},\,
\varepsilon\sim\mathcal{N}(0,\sigma_k^2I)}
\left[
\left\|
s_\theta(x+\varepsilon,\sigma_k)
+
\frac{\varepsilon}{\sigma_k^2}
\right\|_2^2
\right].
\]

For each fixed noise level, the same regression argument as before applies. The
optimal predictor recovers the score of the corresponding smoothed density:
\[
s_\theta^\star(\tilde{x},\sigma_k)
=
\nabla_{\tilde{x}}\log p_{\sigma_k}(\tilde{x}).
\]
The difference is that the model now learns a family of score fields rather than
a single one. Instead of modelling one density, we therefore model a hierarchy
of smoothed densities, ranging from highly noisy and simple to lightly noisy and
detailed.

\subsection{Annealed Langevin Dynamics}

Once scores have been learned at multiple noise levels, they can be used for
sampling. This leads to \emph{annealed Langevin dynamics}, which applies
Langevin updates from large noise scales to small noise scales.

Suppose
\[
s_\theta(x,\sigma_k)
\approx
\nabla_x\log p_{\sigma_k}(x),
\qquad
k=1,\dots,K,
\]
with
\[
\sigma_1>\sigma_2>\cdots>\sigma_K.
\]
We initialise from a broad noise distribution, for example
\[
x^{(0)}\sim\mathcal{N}(0,\sigma_1^2I),
\]
and then, for each scale \(\sigma_k\), perform several updates of the form
\[
\boxed{
x
\leftarrow
x
+
\epsilon_k s_\theta(x,\sigma_k)
+
\sqrt{2\epsilon_k}\,\xi,
\qquad
\xi\sim\mathcal{N}(0,I).
}
\]
After several steps at scale \(\sigma_k\), the procedure moves to the next
smaller scale \(\sigma_{k+1}\).

The logic is simple. At large noise levels, the score gives coarse guidance
toward the broad data region. At smaller noise levels, the score gives sharper
local corrections. Sampling therefore proceeds as progressive refinement:
\[
\text{random noise}
\;\longrightarrow\;
\text{coarse data region}
\;\longrightarrow\;
\text{local structure}
\;\longrightarrow\;
\text{fine detail}.
\]

Annealed Langevin dynamics turns multi-scale score learning into an explicit
generative procedure. It also suggests the next conceptual step. If the number
of noise levels becomes very large and the gap between neighbouring levels
becomes very small, this discrete refinement process begins to resemble a
continuous-time diffusion process.

Annealed Langevin dynamics is already a generative procedure, but it is not yet
a diffusion model in the DDPM sense. It uses a collection of noisy densities
\(p_{\sigma_k}\) and applies Langevin refinement at each scale. A diffusion
model organises the intermediate noisy distributions as the marginals of a
single forward noising process. In DDPMs, this process is a discrete Markov
chain; in the continuous-time score-based view, it is described by a stochastic
differential equation.
\subsection{From Discrete Noise Levels to a Continuous Diffusion Process}

The multi-scale construction uses finitely many noise levels to approximate a
smooth progression from data to noise. When the number of levels becomes large
and neighbouring levels are close together, it is natural to replace the
discrete index \(\sigma_k\) by a continuous time variable \(t\). Instead of
thinking of separate smoothed densities
\[
p_{\sigma_1},p_{\sigma_2},\dots,p_{\sigma_K},
\]
we consider a continuous family of densities
\[
p_t(x),
\qquad
0\leq t\leq T.
\]
This continuous-time description should be read as the analogue of the discrete
noising path introduced in the DDPM chapter: the finite sequence of intermediate
noisy variables is replaced by a continuously indexed family of noisy
distributions.

At one end, \(p_0\) is the data distribution, or an idealised version of it. At
the other end, \(p_T\) is designed to be a simple noise-like distribution, often
approximately Gaussian.

The score model then changes from a noise-conditional field
\[
s_\theta(x,\sigma_k)
\approx
\nabla_x\log p_{\sigma_k}(x)
\]
to a time-dependent field
\[
s_\theta(x,t)
\approx
\nabla_x\log p_t(x).
\]
Thus the multi-scale score-learning picture becomes
\[
\text{finite noise scales}
\quad\longrightarrow\quad
\text{continuous family of noisy densities}
\quad\longrightarrow\quad
\text{a forward diffusion process}.
\]

This is the point at which multi-scale score learning becomes tied to a
diffusion model. Learning scores at many noise levels gives us a family of
vector fields, but by itself it does not yet specify a diffusion model. A
diffusion model is obtained when the noisy densities are organised as the time
marginals of a single forward noising process. In other words, there must be a
stochastic dynamics whose evolving distribution is \(p_t\). The learned score
\(s_\theta(x,t)\) is then used not merely as an isolated denoising field, but as
the correction term needed to run that noising process backward.

A continuous noising process can be described by a stochastic differential
equation of the form
\[
\boxed{
dX_t
=
f(X_t,t)\,dt
+
g(t)\,dW_t.
}
\]
The term \(f(X_t,t)\,dt\) is the deterministic drift, while
\(g(t)\,dW_t\) injects Brownian noise. The function \(g(t)\) controls the
strength of the noise at time \(t\).

This is exactly the drift-plus-noise structure developed in the previous
chapter. If \(p_t(x)\) denotes the density of \(X_t\), then its evolution is
governed by the Fokker--Planck equation
\[
\partial_t p_t(x)
=
-\nabla_x\cdot\bigl(p_t(x)f(x,t)\bigr)
+
\frac{1}{2}g(t)^2\Delta p_t(x),
\]
for the scalar, isotropic diffusion coefficient \(g(t)\). The drift term
transports probability mass through space, while the diffusion term spreads and
smooths the density.

In generative modelling, the forward process is not used to generate new data
directly. Instead, its role is to define a controlled path from data to noise:
\[
p_0(x)
\approx
p_{\text{data}}(x),
\qquad
p_T(x)
\approx
\text{simple noise distribution}.
\]
The generative problem is then to run this process backward: start from noise
and recover a structured sample.

\subsection{Reverse-Time Dynamics and the Role of the Score}

The forward diffusion process is easy to simulate. Starting from a data sample,
we add noise according to the chosen stochastic differential equation and obtain
increasingly noisy states. But generation requires the opposite direction. We
want to start from a simple noise sample and move backward toward the data
distribution.

At first sight, reversing a noisy process may seem impossible, because diffusion
appears to destroy information. However, under suitable regularity conditions,
the reverse-time evolution of a diffusion process can also be described by a
stochastic differential equation. The crucial fact is that the reverse drift
depends on the score of the intermediate density \(p_t\).

For the forward process
\[
dX_t
=
f(X_t,t)\,dt
+
g(t)\,dW_t,
\]
in the scalar-diffusion setting considered here, the reverse-time dynamics has
the form
\[
\boxed{
dX_t
=
\left[
f(X_t,t)
-
g(t)^2\nabla_x\log p_t(X_t)
\right]dt
+
g(t)\,d\bar{W}_t.
}
\]
Here the equation is written in the conventional reverse-time notation: the
process is simulated from \(t=T\) down to \(t=0\), so the time increment is
understood in the reverse direction. The Brownian motion \(\bar{W}_t\) denotes
the noise associated with this reversed process.

This equation is the point where the learned score becomes operational for
generation. The forward SDE defines the family of noisy densities \(p_t\), while
the reverse drift requires the score-dependent correction
\[
-g(t)^2\nabla_x\log p_t(X_t).
\]
Since the true score is unknown, a time-conditioned model is learned:
\[
s_\theta(x,t)
\approx
\nabla_x\log p_t(x),
\qquad
0\leq t\leq T.
\]
Figure~\ref{fig:score-forward-reverse-sde} summarises how these pieces fit
together.

\input{figures/ch06/fig_ch06_05_forward_reverse_sde}

The model therefore learns the scores of the intermediate noisy distributions,
not only the score of the data distribution. These local fields supply the
corrections needed to travel from the simple terminal noise law back toward
structured data.

Once these scores are known or well approximated, generation can be carried out
by simulating the reverse-time dynamics. One begins with
\[
X_T\sim p_T,
\]
where \(p_T\) is a simple noise distribution, and then evolves backward from
\(T\) to \(0\) using the learned score \(s_\theta(x,t)\). The final state is then
treated as an approximate sample from the data distribution.

This completes the conceptual transition from score learning to score-based
diffusion. Denoising score matching explains how to learn scores of noisy
distributions. Multi-scale score learning extends this across many noise levels.
The forward diffusion process organises these noisy distributions into a single
time-indexed path, and the reverse-time SDE shows how learned scores can be used
to travel back along that path.

Thus, a diffusion model is not simply a score estimator, and score matching
alone is not yet a complete generative model. Rather, the forward diffusion
process defines a structured path from data to noise, while score learning
provides the local information needed to reverse that path. Generation is
obtained by combining the two: start from a simple noise distribution and follow
a learned, score-guided reverse dynamics back toward the data distribution.
\section{Discrete Diffusion Models Through the Score-Based Lens}

The discussion above developed the continuous-time score-based view of
diffusion. We now connect this back to the discrete DDPM formulation introduced
earlier in the book. The purpose is not to repeat the latent-variable
derivation, but to reinterpret one familiar part of DDPMs through the
score-based viewpoint: the noise-prediction objective.

In the DDPM chapter, diffusion models were introduced through a forward noising
Markov chain, a learned reverse Markov chain, and a variational training
objective that simplified to a noise-prediction loss. Here we show why that
noise-prediction loss can also be understood as carrying score-like information
about the intermediate noisy distributions.

\subsection{Noise Prediction and Conditional Scores}

Recall that the DDPM forward process gradually corrupts a clean data sample
\(x_0\) by adding Gaussian noise. A key property of this process is that the
noisy variable \(x_t\) can be sampled directly from \(x_0\) as
\[
x_t
=
\sqrt{\bar{\alpha}_t}\,x_0
+
\sqrt{1-\bar{\alpha}_t}\,\varepsilon,
\qquad
\varepsilon\sim\mathcal{N}(0,I),
\]
where \(\bar{\alpha}_t\) controls the remaining signal strength at time \(t\).
Equivalently,
\[
q(x_t\mid x_0)
=
\mathcal{N}
\left(
x_t;
\sqrt{\bar{\alpha}_t}\,x_0,
(1-\bar{\alpha}_t)I
\right).
\]

From the score-based viewpoint, the relevant object is the gradient of a
log-density with respect to the noisy variable \(x_t\). For the conditional
Gaussian \(q(x_t\mid x_0)\), we have
\[
\log q(x_t\mid x_0)
=
-\frac{1}{2(1-\bar{\alpha}_t)}
\left\|
x_t-\sqrt{\bar{\alpha}_t}x_0
\right\|_2^2
+
\text{constant}.
\]
Differentiating with respect to \(x_t\) gives
\[
\nabla_{x_t}\log q(x_t\mid x_0)
=
-\frac{1}{1-\bar{\alpha}_t}
\left(
x_t-\sqrt{\bar{\alpha}_t}x_0
\right).
\]
Using the reparameterisation
\[
x_t-\sqrt{\bar{\alpha}_t}x_0
=
\sqrt{1-\bar{\alpha}_t}\,\varepsilon,
\]
this becomes
\[
\boxed{
\nabla_{x_t}\log q(x_t\mid x_0)
=
-\frac{1}{\sqrt{1-\bar{\alpha}_t}}\,
\varepsilon.
}
\]

This identity explains why noise prediction and score estimation are closely
connected. If a model predicts the added noise \(\varepsilon\), then, up to the
known scaling factor
\[
-\frac{1}{\sqrt{1-\bar{\alpha}_t}},
\]
it also predicts the conditional score of the noisy Gaussian distribution
\(q(x_t\mid x_0)\).

It is important to be precise about what this means. The expression above is the
conditional score, where the clean sample \(x_0\) is treated as known. The score
needed by the reverse process is the marginal score
\[
\nabla_{x_t}\log q_t(x_t),
\]
where \(q_t(x_t)\) is the full distribution of noisy samples at time \(t\). As
in denoising score matching, the marginal score is obtained by averaging over
the possible clean samples \(x_0\) that could have produced the same noisy
observation \(x_t\).

Thus DDPM noise prediction provides a practical regression route to the same
kind of information required by score-based models: the local direction that
tells a noisy sample how it should be corrected in order to become more
data-like.

\subsection{A Unified View of Score-Based Diffusion}

We can now place the different viewpoints side by side.

From the DDPM latent-variable perspective, a diffusion model is built from a
discrete forward noising process
\[
x_0 \rightarrow x_1 \rightarrow \cdots \rightarrow x_T,
\]
together with a learned reverse process
\[
x_T \rightarrow x_{T-1} \rightarrow \cdots \rightarrow x_0.
\]
Training can be expressed through a variational objective, and in the standard
parameterisation this leads to a neural network that predicts the noise added at
each timestep.

From the score-based perspective, the same sequence of noisy variables defines a
family of intermediate noisy distributions
\[
q_t(x_t),
\qquad
t=0,\dots,T.
\]
At each timestep, the reverse process needs local information about how to move
a noisy sample toward regions of higher probability under the corresponding
intermediate distribution. This local information is encoded by the score
\[
\nabla_{x_t}\log q_t(x_t).
\]

The two views are therefore complementary. The DDPM view emphasises the
probabilistic latent-variable structure and the discrete reverse Markov chain.
The score-based view emphasises the geometry of the intermediate noisy
densities and the local denoising directions needed to reverse the corruption.

In continuous time, this geometric picture becomes especially clear. The
discrete family
\[
q_0,q_1,\dots,q_T
\]
is replaced by a continuous path of densities
\[
p_t(x),
\qquad
0\leq t\leq T,
\]
and the reverse dynamics explicitly contains the score
\[
\nabla_x\log p_t(x).
\]
Thus the score-based continuous-time formulation can be viewed as a smooth
analogue of the same denoising principle that appears in DDPMs.

The main conceptual message is not that all formulations are identical in their
details. DDPMs, annealed Langevin dynamics, and continuous-time score-based
diffusion organise the generative process in different ways. But they share the
same central idea:
\[
\text{learn how noisy samples should be locally corrected,}
\]
and use those learned corrections to transform noise into data.

In this sense, diffusion models are structured forms of time-dependent
denoising. The forward process creates a hierarchy of noisy distributions. The
learning problem estimates the information needed to move backward through that
hierarchy. The sampling procedure starts from simple noise and repeatedly
applies learned corrections until a data-like sample is obtained.

This completes the connection between the discrete DDPM viewpoint and the
score-based viewpoint developed in this chapter. The DDPM chapter showed how
diffusion models can be derived as latent-variable generative models. The
present chapter shows why their denoising behaviour can also be understood
geometrically: they learn score-like information across noise levels and use it
to guide the reverse transformation from noise to data.

\section*{Summary and References}
\addcontentsline{toc}{section}{Summary and References}

This chapter developed the score-based view of generative modelling and used it
to build a continuous-time interpretation of diffusion. Whereas the earlier
DDPM chapter introduced diffusion through discrete latent variables and
variational learning, this chapter approached the same denoising idea through
local probability geometry.

The central object was the score function, $\nabla_x \log p(x),$
which turns a density into a vector field. We first interpreted the score as a
probabilistic drift direction and showed how, when combined with Brownian noise,
it leads to Langevin dynamics. The stationarity calculation using the
Fokker--Planck equation explained why the score of a target density can support
a sampling dynamics whose stationary distribution is that target.

We then turned to the learning problem. Since the true data density is unknown,
its score cannot be evaluated directly. Score matching addresses this by
minimising a Fisher-divergence objective and using integration by parts to
remove the unknown data score from the training loss. Denoising score matching
then made this idea more practical by corrupting data with Gaussian noise and
learning the score of the smoothed density through a regression problem. In this
view, the model learns how a noisy sample should move back toward the clean data
distribution.

The chapter then moved from a single noise level to a family of noisy
distributions. Multi-scale score learning provides coarse global guidance at
large noise levels and fine local refinement at small noise levels. Annealed
Langevin dynamics turns these score fields into a sampling procedure, while the
continuous-time formulation organises the noisy distributions as the marginals
of a forward stochastic differential equation. The corresponding reverse-time
dynamics shows why the time-dependent score $\nabla_x \log p_t(x)$ is the key correction term needed to transform noise back into data.

Finally, we connected this score-based view back to DDPMs. The DDPM
noise-prediction objective can be understood as learning score-like information
about noisy intermediate distributions. Thus the DDPM and score-based views are
not competing explanations, but complementary ones: DDPMs emphasise the
discrete latent-variable construction, while score-based diffusion emphasises
the geometry and dynamics of reversing noise.

\bigskip

The development of score-based diffusion draws on several important strands of
work. The score matching principle was introduced by
Hyv\"arinen~\cite{hyvarinen2005estimation}, who showed how models could be estimated
by matching gradients of log-densities without requiring the normalising
constant of the density. Vincent~\cite{vincent2011connection} later connected
score matching to denoising, showing that denoising objectives can recover
score information for corrupted data distributions. These ideas form the
foundation for the score learning and denoising score matching derivations used
in this chapter.

Score-based generative modelling was brought into the modern deep generative
modelling setting by Song and Ermon~\cite{song2019generative}, who learned
scores across multiple Gaussian noise levels and used annealed Langevin dynamics
for sampling. This work made explicit the importance of multi-scale score
fields, especially when data concentrate near lower-dimensional structures.
The continuous-time formulation was then unified and extended by
Song et al.~\cite{song2021scorebased}, who described score-based generative
modelling through stochastic differential equations, reverse-time SDEs, and
related sampling procedures. Together with the DDPM formulation of
Ho et al.~\cite{ho2020denoising}, these works provide the main conceptual
foundation for the modern understanding of diffusion models as both
probabilistic denoising models and score-guided generative dynamics.

%% file: figures/ch06/chapter6_score_at_a_glance.tex
\begin{center}
\includegraphics[width=0.95\textwidth]{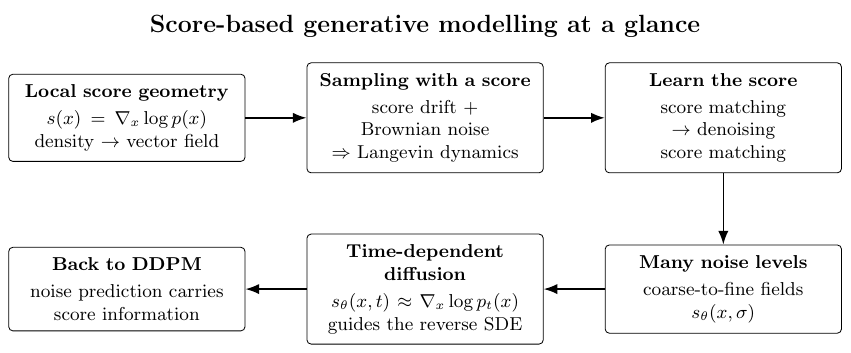}
\end{center}

%% file: figures/ch06/fig_ch06_01_score_field_geometry.tex
\begin{figure}[H]
\centering
\includegraphics[width=0.98\textwidth]{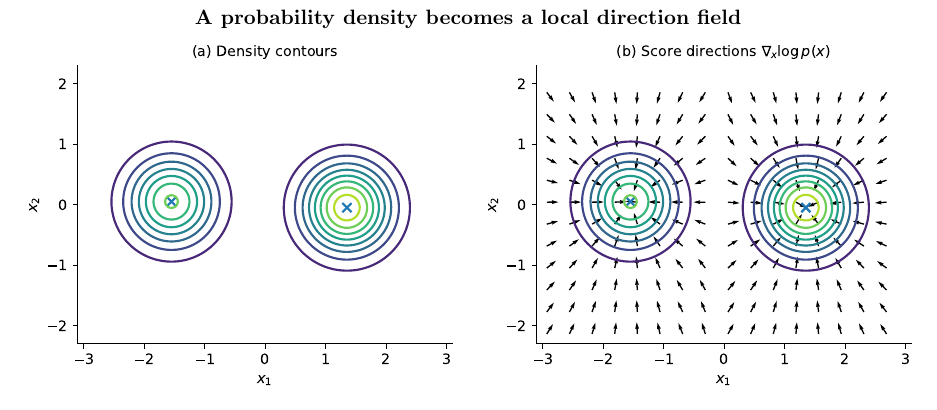}
\caption{\textbf{The score turns a multimodal density into a local direction
field.}
The left panel shows density contours for an exact two-component Gaussian
mixture, while the right shows the corresponding score directions
$\nabla_x\log p(x)$. The arrows point toward increasing log-density and bend
toward the locally more influential mode. Arrow lengths are normalised for
legibility, so the figure emphasises direction rather than magnitude.}
\label{fig:score-field-geometry}
\end{figure}

%% file: figures/ch06/fig_ch06_03_dsm_regression.tex
\begin{figure}[H]
\centering
\includegraphics[width=0.80\textwidth]{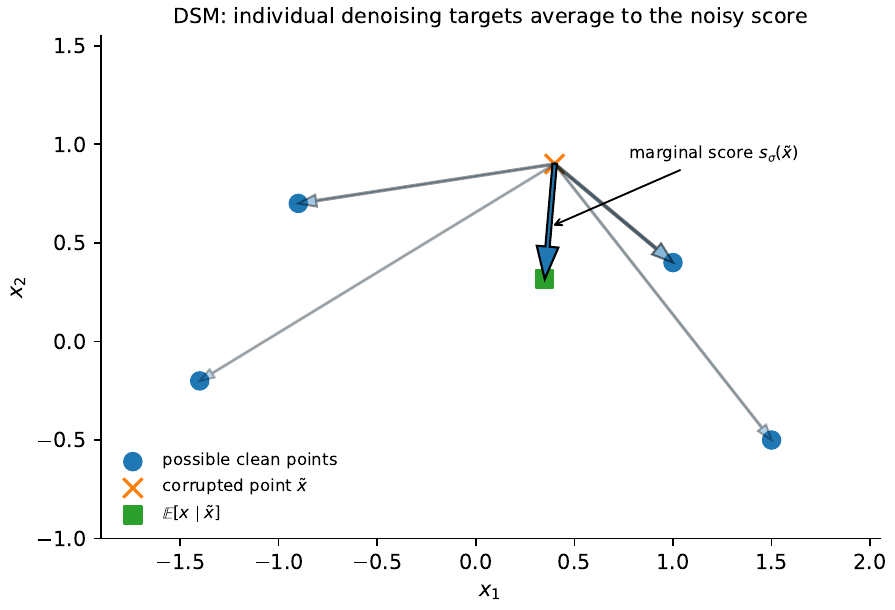}
\caption{\textbf{Denoising score matching as least-squares regression.}
In this finite illustration, the thin arrows show the tractable conditional
targets $(x-\tilde{x})/\sigma^2$ for possible clean points, with visual strength
reflecting their posterior responsibility. Their posterior-weighted average,
shown by the thick arrow, terminates at $\mathbb{E}[x\mid\tilde{x}]$ and equals
the marginal score $\nabla_{\tilde{x}}\log p_\sigma(\tilde{x})$. Thus
squared-error regression recovers the noisy-distribution score from individual
denoising targets.}
\label{fig:score-dsm-regression}
\end{figure}

%% file: figures/ch06/fig_ch06_04_multiscale_scores.tex
\begin{figure}[H]
\centering
\begin{minipage}[t]{0.43\textwidth}
  \vspace{0pt}\centering
  \includegraphics[width=\linewidth]{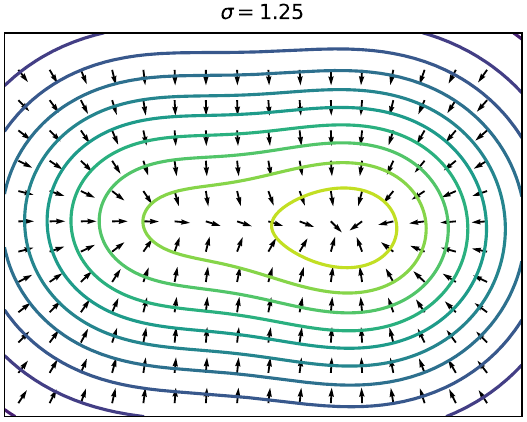}\\[-0.2em]
  {\small large noise: coarse geometry}
\end{minipage}\hfill
\begin{minipage}[t]{0.43\textwidth}
  \vspace{0pt}\centering
  \includegraphics[width=\linewidth]{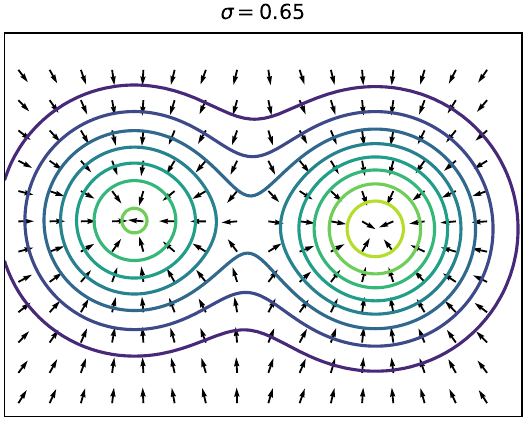}\\[-0.2em]
  {\small intermediate scale}
\end{minipage}

\vspace{0.8em}

\begin{minipage}[t]{0.43\textwidth}
  \vspace{0pt}\centering
  \includegraphics[width=\linewidth]{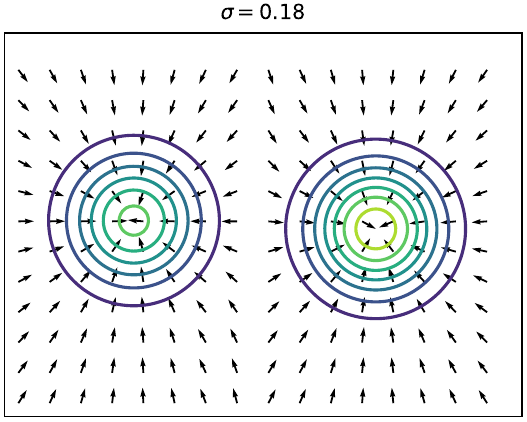}\\[-0.2em]
  {\small small noise: local detail}
\end{minipage}\hfill
\begin{minipage}[t]{0.43\textwidth}
  \vspace{0pt}
  \caption{\textbf{Different noise levels define different score fields.}
  The same two-component distribution is smoothed with Gaussian noise at three
  scales. Large noise reveals coarse global geometry, while decreasing the noise
  level exposes separate modes and sharper local structure. Multi-scale score
  models therefore condition on the noise level rather than sharing one score
  field across all scales. Arrow lengths are normalised for legibility.}
  \label{fig:score-multiscale-fields}
\end{minipage}
\end{figure}

%% file: figures/ch06/fig_ch06_05_forward_reverse_sde.tex
\begin{figure}[H]
\centering
\begin{tikzpicture}[
  >=Latex,
  every node/.style={font=\small},
  state/.style={
    draw, rounded corners=2pt, align=center, inner sep=5pt,
    text width=2.20cm, minimum height=0.82cm
  },
  eqbox/.style={
    draw, rounded corners=2pt, align=center, inner sep=6pt,
    text width=10.8cm, minimum height=1.12cm
  },
  flow/.style={->,thick},
  reverse/.style={->,thick,dashed}
]
\node[font=\normalsize\bfseries] at (6.1,3.25)
  {The score supplies the correction needed to reverse diffusion};

% Forward row
\node[state] (f0) at (0,1.75) {$p_0$\\data};
\node[state] (ft) at (6.1,1.75) {$p_t$\\intermediate};
\node[state] (fT) at (12.2,1.75) {$p_T$\\simple noise};

\draw[flow] (f0) -- node[above] {\scriptsize forward noising} (ft);
\draw[flow] (ft) -- (fT);

\node[eqbox] (forward) at (6.1,0.15) {
\textbf{Forward SDE}\\[3pt]
$\displaystyle
dX_t=f(X_t,t)\,dt+g(t)\,dW_t
$
\qquad
\textbf{learn along its marginals: }
$\displaystyle
s_\theta(x,t)\approx\nabla_x\log p_t(x)
$
};

% Reverse row
\node[state] (r0) at (0,-1.65) {$p_0$\\data};
\node[state] (rt) at (6.1,-1.65) {$p_t$\\intermediate};
\node[state] (rT) at (12.2,-1.65) {$p_T$\\simple noise};

\draw[reverse] (rT) -- node[above] {\scriptsize reverse-time generation} (rt);
\draw[reverse] (rt) -- (r0);

\node[eqbox] (reversebox) at (6.1,-3.45) {
\textbf{Reverse-time SDE, simulated from $T$ down to $0$}\\[3pt]
$\displaystyle
dX_t=
\left[
f(X_t,t)-g(t)^2s_\theta(X_t,t)
\right]dt
+
g(t)\,d\bar W_t
$
};

\draw[flow] (forward) -- (reversebox);

\end{tikzpicture}
\caption{\textbf{The score supplies the correction needed to reverse
 diffusion.}
The forward SDE defines the time-indexed marginals $p_t$, while a
time-conditioned model learns $s_\theta(x,t)\approx\nabla_x\log p_t(x)$. For
the scalar, state-independent diffusion coefficient used here, the reverse
drift differs from the forward drift by $-g(t)^2\nabla_x\log p_t(x)$.
Replacing the unknown score by $s_\theta(x,t)$ gives the practical reverse-time
SDE, simulated from $T$ down to $0$.}
\label{fig:score-forward-reverse-sde}
\end{figure}

%% file: chapters/cht_8_9_merged_new.tex
\chapter{Exact Density Models: Normalising Flows and Autoregressive Factorisations}
\section*{Overview}
\addcontentsline{toc}{section}{Overview}

The previous chapters developed several major routes into modern generative
modelling. Variational autoencoders introduced latent-variable modelling and
approximate inference. Denoising diffusion probabilistic models shifted the
perspective to stochastic noising and denoising processes. The continuous-time
and score-based chapters then developed this further through density evolution,
stochastic differential equations, and learned score fields.

This chapter turns to a different paradigm: \emph{exact density modelling}.
Here, the aim is to construct generative models for which the probability
density of an observed data point can be evaluated directly by design. This
changes the modelling philosophy. Instead of relying on an intractable latent
marginal, or learning to reverse a stochastic corruption process, we impose
mathematical structure that keeps likelihood evaluation tractable from the
start.

The first route is provided by \textbf{normalising flows}. A normalising flow
starts from a simple base distribution, such as a standard Gaussian, and applies
a sequence of differentiable, invertible transformations. The
change-of-variables formula tells us exactly how density changes under each
transformation by accounting for local stretching or compression through the
Jacobian determinant. As long as the transformations are invertible and their
Jacobian determinants are tractable, both likelihood evaluation and sampling are
exact model operations.

The second route is provided by \textbf{autoregressive models}. Rather than
transforming a base distribution through invertible maps, autoregressive models
construct the joint density directly using the chain rule of probability. A
high-dimensional data point is decomposed into an ordered sequence of
components, and each component is modelled conditionally on those that came
before it. This gives exact likelihoods without latent variables, variational
bounds, or Jacobian determinants, but sampling usually becomes sequential.

These two approaches remain important in modern generative modelling because
they show how far exact likelihood-based modelling can be pushed. Normalising
flows provide a clean framework for invertible density estimation, uncertainty
modelling, and tractable transformations of probability distributions.
Autoregressive models, meanwhile, underpin many influential sequence, audio,
language, and pixel-level image generation systems. Even when newer systems combine several modelling ideas, the
principles developed in this chapter---change of variables, tractable
Jacobians, conditional factorisation, and ordered generation---continue to
reappear as core building blocks.

\section*{Concept Map}
\addcontentsline{toc}{section}{Concept Map}

The chapter develops exact density modelling as a contrasting paradigm to the
latent-variable, diffusion, and score-based approaches introduced earlier. The
central question is how flexible high-dimensional generative models can retain
exact likelihood evaluation.

\begin{itemize}

    \item[\(\rightarrow\)] \textbf{We first introduce normalising flows through the
    change-of-variables formula.}  
    A normalising flow maps a simple base variable \(z\) to a data variable
    \(x=f(z)\) using differentiable, invertible transformations. The
    change-of-variables formula tracks the resulting density change through the
    Jacobian determinant.

    \item[\(\rightarrow\)] \textbf{We then study planar flows as a first nonlinear
    flow layer.}  
    Planar flows introduce nonlinearity through a residual rank-one
    perturbation. This keeps the Jacobian determinant analytically tractable,
    while also illustrating the importance of invertibility constraints.

    \item[\(\rightarrow\)] \textbf{Next, we move to affine coupling layers as a
    practical flow construction.}  
    Affine coupling layers transform only part of the input while conditioning
    on the rest. This creates closed-form inverses and efficient
    log-determinant computation, making them useful for practical flow models.

    \item[\(\rightarrow\)] \textbf{We then compose flow layers into full generative
    models.}  
    Stacking many invertible layers allows simple local deformations to become
    an expressive global transformation. Likelihood evaluation maps data back to
    the base space, while sampling maps base noise forward to data space.

    \item[\(\rightarrow\)] \textbf{Next, we introduce autoregressive factorisation as
    a second route to exact likelihoods.}  
    Autoregressive models use the chain rule $p(x)=\prod_{i=1}^d p(x_i\mid x_{<i})$
    to construct exact likelihoods without latent variables, variational bounds,
    or Jacobian determinants.

    \item[\(\rightarrow\)] \textbf{Finally, we explain conditional parameterisation,
    training, sampling, and masking.}  
    A shared model can parameterise all conditional distributions. During
    training, the full data vector is observed, so likelihood terms can often be
    evaluated in parallel. During sampling, components are generated
    sequentially. Masking enforces the ordered information flow that makes the
    autoregressive factorisation tractable.

\end{itemize}

The two routes developed in this chapter, and the structural principle that makes
both of them tractable, are summarised in the visual overview below.

\input{figures/ch07/chapter7_exact_density_at_a_glance_book.tex}

\section{Normalising Flows: Exact Density via Invertible Transformations}

Normalising flows provide a direct, fully probabilistic route to generative modelling by constructing an explicit mapping between a simple reference distribution and the data distribution, while retaining exact likelihood evaluation throughout~\cite{rezende2015variational}. The term \emph{normalising flow} refers to this broader family of invertible
density models, rather than to one specific transformation; planar flows,
affine coupling layers, and related architectures are different ways of
designing such transformations so that they remain expressive, invertible, and
computationally tractable.

The central mathematical tool underlying normalising flows is the \emph{change-of-variables formula}. This formula precisely characterises how probability density transforms under a differentiable, invertible mapping. Rather than introducing latent variables or auxiliary stochastic processes, normalising flows rely solely on deterministic transformations whose effect on density can be computed exactly. By chaining together many such transformations, a simple base distribution—typically a standard Gaussian—can be gradually reshaped into a highly structured data distribution.

From a geometric perspective, each transformation acts as a controlled deformation of space, locally stretching or compressing volumes in a way that is explicitly accounted for by the Jacobian determinant. As long as each transformation is invertible and its Jacobian determinant remains tractable, the resulting model supports exact likelihood evaluation and exact sampling. This combination of expressiveness and mathematical transparency distinguishes normalising flows from other generative modelling paradigms and makes them a natural starting point for our study of exact density models.

\subsection{Revisiting the Change-of-Variables Formula}

Although the change-of-variables formula was introduced in Chapter~5 from the
more general viewpoint of deterministic density transport, it is worth revisiting
it here because it is the modelling
principle of normalising flows. 

Suppose \(X\) has a known density \(p_X(x)\), and we apply a differentiable,
invertible function $Y=f(X).$
Our goal is to determine the density of \(Y\), denoted \(p_Y(y)\). Intuitively, the density of \(Y\) needs to reflect two things.

First, it must account for \emph{where probability mass came from}. Since
\(y=f(x)\), the likelihood of observing \(y\) depends on how likely its
corresponding pre-image $x=f^{-1}(y)$
was under the original distribution.

Second, it must account for \emph{how the transformation stretches or compresses
space}. A transformation that spreads points further apart should reduce the
density, while one that compresses points should increase it. Density is always
probability per unit length, area, or volume, so local spacing matters.

In one dimension, these ideas combine in the change-of-variables formula
\[
\boxed{
p_Y(y)
=
p_X(f^{-1}(y))
\left|
\frac{d f^{-1}(y)}{dy}
\right|.
}
\]
The first term tells us which value of \(x\) generated \(y\). The second term is
a scaling factor that accounts for how stretched or compressed the neighbourhood
around \(x\) becomes when mapped to \(y\).

\paragraph{A simple geometric example.}

Let $X\sim \mathrm{Uniform}(0,1),
\qquad
Y=f(X)=2X.$

This transformation doubles every point and stretches the interval \([0,1]\) to
\([0,2]\). The inverse is
\[
x=f^{-1}(y)=\frac{y}{2},
\]
so
\[
\frac{d f^{-1}(y)}{dy}
=
\frac{1}{2}.
\]
Applying the formula gives
\[
p_Y(y)
=
p_X\!\left(\frac{y}{2}\right)\frac{1}{2}.
\]
Within the new domain \([0,2]\), the original density satisfies
\(p_X(y/2)=1\), and therefore
\[
p_Y(y)=\frac{1}{2},
\qquad 0\leq y\leq 2.
\]
This matches intuition exactly: the same total probability now fills an
interval twice as wide, so its density is halved. This simple logic scales to higher dimensions. 

\paragraph{The higher-dimensional case: from stretching lengths to stretching volumes.}

In multiple dimensions, the derivative is replaced by a volume-scaling
determinant. For the simple linear map
\[
f(x)=Ax,
\qquad
A=
\begin{bmatrix}
2 & 0\\
0 & 3
\end{bmatrix},
\]
local area is multiplied by $|\det A|=6$, and therefore
\[
p_Y(y)=p_X(A^{-1}y)\frac{1}{|\det A|}.
\]
The same probability mass occupies a larger local volume, so the density falls
by the reciprocal factor.

\paragraph{The Jacobian as a local linear lens.}

To generalise beyond linear transformations, we use the \emph{Jacobian matrix}.
For a differentiable function
\[
f:\mathbb{R}^d\to\mathbb{R}^d,
\]
the Jacobian is defined by
\[
[J_f(x)]_{ij}
=
\frac{\partial f_i(x)}{\partial x_j}.
\]
The Jacobian can be viewed as a local linear lens. Around a point \(x\), a small
displacement \(\delta x\) is transformed approximately as
\[
f(x+\delta x)
\approx
f(x)+J_f(x)\delta x.
\]
Thus \(J_f(x)\) tells us how an infinitesimally small neighbourhood around
\(x\) is distorted when mapped to \(y=f(x)\). Its determinant quantifies how
much the corresponding local volume expands or contracts.

For a general differentiable, invertible transformation \(y=f(x)\), the
multivariate change-of-variables formula becomes
\[
\boxed{
p_Y(y)
=
p_X(f^{-1}(y))
\left|
\det J_{f^{-1}}(y)
\right|=
p_X(x)
\frac{1}{|\det J_f(x)|},
\qquad y=f(x).
}
\]

This is the mathematical heart of normalising flows:  
\emph{if each transformation is invertible and its Jacobian determinant is tractable, we can express the density of the transformed variable exactly, even when the overall mapping becomes highly expressive.}
Figure~\ref{fig:flow-change-variable-geometry} shows the geometric content of
this formula: a fixed amount of probability mass occupies a locally stretched
or compressed region, and the density changes by the reciprocal volume factor.

\input{figures/ch07/fig_ch07_01_change_of_variables.tex}

\begin{tcolorbox}[
  colback=blue!4,
  colframe=blue!50!black,
  title={Why standard normalising flows preserve dimension},
  boxrule=0.6pt,
  arc=2mm,
  left=2mm,
  right=2mm,
  top=1mm,
  bottom=1mm
]
Standard normalising flows use differentiable, invertible transformations of
the form
\[
f:\mathbb{R}^d\to\mathbb{R}^d.
\]
This equal dimensionality is required by the standard
change-of-variables formula: the Jacobian must be square so that
\(\det J_f(x)\) is defined and can measure local volume change. If the map sends \(\mathbb{R}^d\) to a lower-dimensional space, information is
collapsed and different inputs may map to the same output. If it sends
\(\mathbb{R}^d\) into a higher-dimensional space, the image lies on a
lower-dimensional manifold and does not define an ordinary full-dimensional
density without extra structure. Thus, standard normalising flows preserve dimension. Without this
full-dimensional invertibility, probability mass cannot be traced exactly using
the change-of-variables formula.
\end{tcolorbox}

\subsection{Planar Flows}

The examples above were intentionally simple. They showed how the
change-of-variables formula tracks density under linear stretching and volume
scaling. However, linear transformations have limited expressive power. They can
stretch, rotate, or shear space, but cannot bend or warp it in rich
nonlinear ways. To approximate complex distributions, we need transformations
that introduce carefully controlled nonlinearity while preserving the two core
requirements of normalising flows: invertibility and a tractable Jacobian
determinant.

Planar flows provide a classic example of this trade-off, originally introduced
as part of the normalising-flow framework of Rezende and
Mohamed~\cite{rezende2015variational}. A foundational
nonlinear flow layer is defined as
\[
f(x)
=
x+u\,h(w^\top x+b),
\]
where \(x\in\mathbb{R}^d\), \(u,w\in\mathbb{R}^d\), \(b\in\mathbb{R}\), and
\[
h:\mathbb{R}\to\mathbb{R}
\]
is a smooth activation function such as \(\tanh\) or \texttt{sigmoid}. Here
\(x\) denotes the input to this individual flow layer; in a stacked flow it
would usually be one of the intermediate variables.

The term
\[
w^\top x+b
\]
acts as an affine projection of the input, reducing the high-dimensional vector
\(x\) to a single scalar. On its own, this projection would contribute only a
linear quantity. To introduce curvature and richer behaviour, we pass this
scalar through a nonlinear activation \(h(\cdot)\). Functions such as
\(\tanh\) and \texttt{sigmoid} squash their inputs into bounded ranges, so the
output of \(h\) can be interpreted as a soft gating value that controls how
strongly the transformation should modify the input.

Multiplying this gated scalar by the vector \(u\) produces a directionally
scaled perturbation. The deformation always occurs along the direction of
\(u\), but its magnitude depends nonlinearly on the projection \(w^\top x+b\).
Adding this perturbation back to \(x\) ensures that the output remains in the
full \(d\)-dimensional space while introducing a flexible nonlinear adjustment.

\begin{tcolorbox}[
  colback=blue!4,
  colframe=blue!50!black,
  title={Residual structure and motivation},
  boxrule=0.6pt,
  arc=2mm,
  left=2mm,
  right=2mm,
  top=1mm,
  bottom=1mm
]
Planar flows follow the spirit of residual connections in deep learning:
\[
f(x)=x+\text{learned perturbation}.
\]
This preserves the identity mapping while adding a controlled deformation.

If we used only the nonlinear term
\[
x\mapsto u\,h(w^\top x+b),
\]
the output would collapse onto the one-dimensional direction of \(u\), since
\(h(\cdot)\) produces a scalar. This would destroy the full-dimensional
structure needed for a normalising flow.

By adding the perturbation back to \(x\), the planar flow keeps the full
\(d\)-dimensional space accessible while introducing targeted nonlinear
distortions. 
\end{tcolorbox}

\paragraph{Jacobian derivation.}

To compute how densities transform under a planar flow, we derive the Jacobian
of
\[
f(x)=x+u\,h(w^\top x+b).
\]
For convenience, let
\[
a=w^\top x+b,
\qquad
f(x)=x+u\,h(a).
\]
Differentiating with respect to \(x\), and applying the chain rule, gives
\[
\frac{\partial f(x)}{\partial x}
=
I+u\,h'(a)\frac{\partial a}{\partial x}.
\]
Since $\frac{\partial a}{\partial x}=w^\top,$ the Jacobian takes the compact form
\[
J_f(x)
=
I+u\,w^\top h'(a).
\]
Equivalently,
\[
J_f(x)
=
I+u\psi(x)^\top,
\qquad
\psi(x):=h'(w^\top x+b)w.
\]

This structure is extremely convenient. The Jacobian is a perturbation of the
identity by a rank-one matrix. Instead of computing a full \(d\times d\)
determinant, which would usually cost \(O(d^3)\), the determinant reduces to a
single scalar expression. Using the matrix determinant lemma,
\[
\det(I+uv^\top)=1+v^\top u,
\]
we obtain
\[
\det J_f(x)
=
1+\psi(x)^\top u.
\]
Substituting \(\psi(x)=h'(w^\top x+b)w\), this becomes
\[
\boxed{
\det J_f(x)
=
1+h'(w^\top x+b)\,w^\top u.
}
\]
Therefore the log-determinant contribution of a planar flow is
\[
\log|\det J_f(x)|
=
\log\left|
1+h'(w^\top x+b)\,w^\top u
\right|.
\]
This closed form makes planar flows tractable for likelihood-based training.

\begin{tcolorbox}[
  colback=blue!4,
  colframe=blue!50!black,
  title={Understanding the rank-one update},
  boxrule=0.6pt,
  arc=2mm,
  left=2mm,
  right=2mm,
  top=1mm,
  bottom=1mm
]
The matrix \(u w^\top\) is an outer product of two vectors in
\(\mathbb{R}^d\), and such a matrix has rank one.

To see why, apply it to any vector \(v\):
\[
u w^\top v
=
u\,(w^\top v).
\]
The scalar \(w^\top v\) changes with \(v\), but the output is always a scalar
multiple of \(u\). Thus the matrix has only one independent output direction.

This is why the determinant is simple: the planar flow does not attempt an
arbitrary nonlinear deformation of all dimensions at once. It introduces a
controlled one-directional deformation whose volume change can be computed
analytically.
\end{tcolorbox}

Figure~\ref{fig:planar-flow-rank-one} shows the corresponding geometry: the
amount of deformation changes across space, but the residual displacement
remains parallel to the single direction $u$.

\input{figures/ch07/fig_ch07_02_planar_flow.tex}

\paragraph{Invertibility condition.}

Besides a tractable Jacobian determinant, a flow layer must also be invertible.
For the planar flow
\[
f(x)=x+u\,h(w^\top x+b),
\]
the inverse is not generally available in closed form, so invertibility is
usually controlled through a parameter constraint.

From the Jacobian derivation,
\[
\det J_f(x)
=
1+h'(w^\top x+b)\,w^\top u.
\]
If this determinant becomes zero, the transformation locally collapses volume
and cannot be locally inverted. A common sufficient condition is therefore to
keep the determinant positive for all \(x\).

For activations such as \(\tanh\), whose derivative satisfies
\[
0\leq h'(a)\leq 1,
\]
it is sufficient to require
\[
w^\top u>-1.
\]
Then
\[
1+h'(a)\,w^\top u>0
\]
for all \(a\), so the planar flow avoids local volume collapse. In practice,
this condition can be enforced by reparameterising \(u\).

Thus planar flows illustrate a central design trade-off: they introduce
nonlinearity through a simple residual perturbation while keeping the Jacobian
determinant analytically tractable and invertibility controllable through a
scalar condition.

\section{Normalising Flows in Practice: Coupling Layers and Composition}

The previous section introduced normalising flows as exact density models based
on invertible transformations. Planar flows gave a simple example of a nonlinear
flow layer whose Jacobian determinant remains tractable, but they also revealed
a practical limitation: the inverse is not generally available in closed form.

We now move from this pedagogical construction to one of the central design
principles behind practical flow architectures: \emph{coupling}. Coupling layers
preserve exact invertibility by transforming only part of the input at a time,
while using the remaining part as a conditioning signal. This creates a
triangular Jacobian, gives a closed-form inverse, and allows the layer to be
stacked efficiently.

The section then shows how such layers are composed into a full generative
model. This composition is what turns simple local transformations into an
expressive density model, while preserving the two defining operations of
normalising flows: exact likelihood evaluation and exact sampling from the
learned model.

\subsection{Affine Coupling Layers and Triangular Jacobians}

Affine coupling layers, introduced in coupling-based flow models such as
NICE and RealNVP~\cite{dinh2014nice,dinh2017density}, provide one of the most
practical ways to build expressive flows with tractable inverses and
log-determinants.

An affine coupling layer begins by splitting the input vector into two parts.
Let \(x\in\mathbb{R}^d\) be the input to a flow layer, and write
\[
x=(x_1,x_2),
\]
where \(x_1\in\mathbb{R}^{d_1}\), \(x_2\in\mathbb{R}^{d_2}\), and
\(d_1+d_2=d\). The affine coupling transformation is defined by
\[
\begin{aligned}
y_1 &= x_1,\\
y_2 &= x_2\odot \exp(s(x_1)) + t(x_1),
\end{aligned}
\]
where \(s(\cdot)\) and \(t(\cdot)\) are functions, usually neural networks, and
\(\odot\) denotes elementwise multiplication.

The first part of the input is left unchanged. The second part is transformed
by an elementwise affine map whose scale and shift are predicted from the first
part. Thus \(x_1\) acts as a context variable, while \(x_2\) is scaled and
translated according to that context.

Although the transformation is affine in \(x_2\), the functions \(s(x_1)\) and
\(t(x_1)\) may be highly nonlinear. This allows the overall transformation to
be expressive while preserving a simple algebraic structure. The layer is
therefore deliberately asymmetric: it sacrifices full simultaneous mixing
inside one layer in exchange for exact inversion and an efficient determinant.

\paragraph{Jacobian determinant.}

To compute the Jacobian, write
\[
y=
\begin{bmatrix}
y_1\\
y_2
\end{bmatrix}
=
\begin{bmatrix}
x_1\\
x_2\odot \exp(s(x_1))+t(x_1)
\end{bmatrix}.
\]
The Jacobian has block form
\[
J
=
\frac{\partial y}{\partial x}
=
\begin{bmatrix}
\frac{\partial y_1}{\partial x_1}
&
\frac{\partial y_1}{\partial x_2}
\\[4pt]
\frac{\partial y_2}{\partial x_1}
&
\frac{\partial y_2}{\partial x_2}
\end{bmatrix}.
\]

Since \(y_1=x_1\), we have
\[
\frac{\partial y_1}{\partial x_1}=I,
\qquad
\frac{\partial y_1}{\partial x_2}=0.
\]
The lower-left block \(\frac{\partial y_2}{\partial x_1}\) may be complicated,
because it contains derivatives of the neural networks \(s(\cdot)\) and
\(t(\cdot)\). Crucially, however, this block will not affect the determinant.

For the lower-right block, note that each component of \(y_2\) is
\[
y_{2,i}
=
x_{2,i}\exp(s_i(x_1))+t_i(x_1).
\]
Differentiating with respect to \(x_{2,j}\), while holding \(x_1\) fixed, gives
\[
\frac{\partial y_{2,i}}{\partial x_{2,j}}
=
\begin{cases}
\exp(s_i(x_1)), & i=j,\\
0, & i\neq j.
\end{cases}
\]
Therefore
\[
\frac{\partial y_2}{\partial x_2}
=
\operatorname{diag}\!\big(\exp(s(x_1))\big).
\]

Putting the blocks together,
\[
J
=
\begin{bmatrix}
I & 0\\
* & \operatorname{diag}\!\big(\exp(s(x_1))\big)
\end{bmatrix}.
\]
This is a block lower-triangular matrix. For any triangular or block triangular
matrix, the determinant is the product of the determinants of the diagonal
blocks. Hence
\[
\det J
=
\det(I)\,
\det\!\left(
\operatorname{diag}\!\big(\exp(s(x_1))\big)
\right).
\]
Since \(\det(I)=1\), and the determinant of a diagonal matrix is the product of
its diagonal entries,
\[
\det J
=
\prod_i \exp(s_i(x_1)).
\]
Taking logarithms gives the especially simple expression
\[
\boxed{
\log|\det J|
=
\sum_i s_i(x_1).
}
\]

This is the main tractability property of affine coupling layers. The dense and
possibly complicated block \(\frac{\partial y_2}{\partial x_1}\) does not need
to be computed for the determinant. The entire volume change is controlled by
the scale output \(s(x_1)\). Thus, despite using nonlinear neural networks
inside the layer, the log-determinant can be evaluated in linear time with
respect to the transformed dimension.

The triangular block structure is the key computational simplification: the
off-diagonal block may be complicated, but it does not enter the determinant.
This is an example of a recurring principle in exact density modelling: nonlinear
dependencies are arranged so that the high-dimensional probability calculation
still factorises into simple terms.

\paragraph{Invertibility.}

The same asymmetric structure that gives a triangular Jacobian also gives a
closed-form inverse. From the forward transformation,
\[
\begin{aligned}
y_1 &= x_1,\\
y_2 &= x_2\odot \exp(s(x_1)) + t(x_1),
\end{aligned}
\]
we immediately recover
\[
x_1=y_1.
\]
Substituting this into the second equation gives
\[
y_2
=
x_2\odot \exp(s(y_1)) + t(y_1).
\]
Since \(\exp(s(y_1))\) is strictly positive componentwise, this elementwise
affine transformation can be inverted:
\[
x_2
=
\bigl(y_2-t(y_1)\bigr)\odot \exp(-s(y_1)).
\]
Thus the inverse is
\[
\boxed{
\begin{aligned}
x_1 &= y_1,\\
x_2 &= \bigl(y_2-t(y_1)\bigr)\odot \exp(-s(y_1)).
\end{aligned}
}
\]

No root finding or iterative inversion is needed. The inverse has essentially
the same computational cost as the forward transformation. This makes affine
coupling layers especially practical for large-scale models, where sampling and
likelihood evaluation may require repeated application of many flow
transformations. Figure~\ref{fig:affine-coupling-structure} summarises the
asymmetric information flow that makes both properties possible: one subset is
passed through unchanged while also controlling the transformation of the other.

\input{figures/ch07/fig_ch07_03_affine_coupling.tex}

\subsection{Stacking Flows: Likelihood Evaluation and Sampling}

A single flow layer can only produce a limited deformation of space. Planar
flows perturb the identity in one learned direction, while affine coupling
layers transform only part of the input at a time. The expressive power of
normalising flows therefore comes from \emph{composition}: many simple,
invertible transformations are stacked so that their small deformations
accumulate into a rich global mapping.

Let
\[
f_1,f_2,\dots,f_K
\]
be a sequence of differentiable and invertible transformations. Starting from a
base variable
\[
z_0\sim p_Z(z_0),
\]
we define intermediate variables
\[
z_k=f_k(z_{k-1}),
\qquad k=1,\dots,K,
\]
and the final output
\[
x=z_K.
\]
Equivalently,
\[
x
=
f_K\circ f_{K-1}\circ \cdots \circ f_1(z_0).
\]
Because each layer is invertible, the full composition is also invertible. This
means we can move in both directions:
\[
z_0
\xrightarrow{f_1}
z_1
\xrightarrow{f_2}
\cdots
\xrightarrow{f_K}
z_K=x,
\]
or
\[
x=z_K
\xrightarrow{f_K^{-1}}
z_{K-1}
\xrightarrow{f_{K-1}^{-1}}
\cdots
\xrightarrow{f_1^{-1}}
z_0.
\]

This two-way structure gives the two basic operations of a flow model:
likelihood evaluation and sampling.

\paragraph{Likelihood evaluation: data to base space.}

During likelihood-based training, we are given a data point \(x\). Under the
convention above, \(x\) is the final variable \(z_K\). To evaluate its density,
we map it backwards through the inverse transformations:
\[
x=z_K
\xrightarrow{f_K^{-1}}
z_{K-1}
\xrightarrow{}
\cdots
\xrightarrow{f_1^{-1}}
z_0.
\]
If the model is well fitted, the recovered base-space variable \(z_0\) should
look likely under the simple base density \(p_Z\).

Applying the change-of-variables formula one layer at a time gives
\[
p_{Z_k}(z_k)
=
p_{Z_{k-1}}(z_{k-1})
\left|
\det J_{f_k}(z_{k-1})
\right|^{-1}.
\]
Repeating this recursion from \(k=1\) to \(K\), and using \(z_K=x\), yields
\[
\boxed{
p_X(x)
=
p_Z(z_0)
\prod_{k=1}^K
\left|
\det J_{f_k}(z_{k-1})
\right|^{-1}.
}
\]
Taking logarithms gives
\[
\boxed{
\log p_X(x)
=
\log p_Z(z_0)
-
\sum_{k=1}^K
\log
\left|
\det J_{f_k}(z_{k-1})
\right|.
}
\]

This expression has a direct interpretation. The base-density term
\(\log p_Z(z_0)\) measures how plausible the inverse-transformed data point is
under the simple reference distribution. The log-determinant terms correct for
all volume changes introduced by the sequence of transformations. Each layer
contributes one additive volume-correction term to the final log-likelihood.

\begin{tcolorbox}[
  colback=blue!3,
  colframe=blue!50!black,
  title={How the product form arises},
  boxrule=0.6pt,
  arc=2mm,
  left=2mm,
  right=2mm,
  top=1mm,
  bottom=1mm
]
For one flow layer \(z_k=f_k(z_{k-1})\), the change-of-variables formula gives
\[
p_{Z_k}(z_k)
=
p_{Z_{k-1}}(z_{k-1})
\left|\det J_{f_k}(z_{k-1})\right|^{-1}.
\]
Applying this relation repeatedly gives
\[
\begin{aligned}
p_{Z_2}(z_2)
&=
p_{Z_0}(z_0)
\left|\det J_{f_1}(z_0)\right|^{-1}
\left|\det J_{f_2}(z_1)\right|^{-1},\\
&\ \vdots\\
p_{Z_K}(z_K)
&=
p_{Z_0}(z_0)
\prod_{k=1}^K
\left|\det J_{f_k}(z_{k-1})\right|^{-1}.
\end{aligned}
\]
Since \(z_K=x\), this gives the likelihood of the observed data point. 
\end{tcolorbox}

\paragraph{Sampling: base space to data space.}

Sampling uses the same transformations in the opposite direction. We first draw
a point from the base distribution,
\[
z_0\sim p_Z(z_0),
\]
and then apply the learned transformations:
\[
z_0
\xrightarrow{f_1}
z_1
\xrightarrow{f_2}
\cdots
\xrightarrow{f_K}
z_K=x.
\]
The resulting \(x\) is a sample from the learned model distribution.

This procedure is exact with respect to the model. No Markov chain, variational
approximation, or numerical integration is required. Once the transformations
have been learned, generation consists simply of sampling from the base
distribution and applying a deterministic sequence of invertible maps.

This also clarifies the contrast with diffusion models. Diffusion models
generate by simulating a learned reverse stochastic process from noise to data,
whereas normalising flows apply a learned deterministic invertible transformation.

\paragraph{Composition, mixing, and expressiveness.}

The composition of many flow layers makes the model expressive, but composition
alone is not enough. The layers must also be arranged so that information can
mix across dimensions.  If the same split were used
in every layer, some variables would repeatedly act as conditioning variables
without being transformed themselves.

To address this, practical flow models interleave nonlinear flow layers with
simple invertible mixing operations. These may be permutations,
\[
z\mapsto Pz,
\]
or more general invertible linear transformations,
\[
z\mapsto Az,
\]
where \(P\) is a permutation matrix and \(A\) is an invertible matrix with a
tractable determinant.

The purpose of these mixing operations is to change which coordinates interact
in later layers. A variable left unchanged in one coupling layer may be moved
into the transformed subset in the next. Geometrically, each mixing operation
changes the coordinate system in which the next nonlinear deformation is
applied.

Thus, a modern flow model is typically built from a repeated pattern:
\[
\text{mix coordinates}
\quad\longrightarrow\quad
\text{apply tractable nonlinear deformation}
\quad\longrightarrow\quad
\text{mix again}.
\]
This repeated alternation allows local, structured transformations to combine
into a flexible global mapping while preserving exact invertibility and
tractable likelihood evaluation.

Architectures such as RealNVP and Glow are built around this principle. They
differ in the details of their coupling transformations, scale parameterisation,
and coordinate mixing, but their underlying mathematical structure is the same:
compose simple invertible layers, keep track of each layer's volume change, and
train the whole model by exact maximum likelihood.

\section{Autoregressive Factorisation as a Generative Model}

Normalising flows achieve exact density modelling by constructing an invertible
map between a simple base distribution and the data distribution. Their
tractability comes from the change-of-variables formula: if the transformation
is invertible and its Jacobian determinant is tractable, the likelihood can be
evaluated exactly.

Autoregressive models provide a second route to exact density modelling. Instead
of transforming a base random variable into data, they construct the joint
density directly by factorising it into a sequence of conditional densities.
Here, exact likelihoods come from the chain rule of probability. This shifts where structure is imposed. In normalising flows, structure is
imposed on the transformation. In autoregressive models, it is imposed on the
dependencies among the components of a data point. By choosing an order and
allowing each component to depend only on previous components, a complex joint
density becomes a product of simpler conditional terms.

Thus, a high-dimensional data point is viewed as a sequence of components
revealed one at a time. Each component is modelled conditionally on those that
came before it, and the resulting conditional pieces define a valid joint
distribution with an exactly computable likelihood. Autoregressive models therefore form a complete generative modelling framework
without latent variables, variational lower bounds, or global invertible
transformations. Their price is different: they impose an explicit ordering, and
sampling becomes sequential.

\subsection{The Chain Rule as an Exact Density Model}

Let
\[
x=(x_1,x_2,\dots,x_d)
\]
be a \(d\)-dimensional random vector. The chain rule of probability follows
from repeated application of the identity
\[
p(a,b)=p(a\mid b)p(b).
\]
Starting from the full joint density, we can first separate out the final
variable:
\[
p(x_1,\dots,x_d)
=
p(x_d\mid x_1,\dots,x_{d-1})p(x_1,\dots,x_{d-1}).
\]
Applying the same identity recursively to the remaining joint term gives
\[
\begin{aligned}
p(x_1,\dots,x_d)
&=
p(x_d\mid x_1,\dots,x_{d-1})
p(x_{d-1}\mid x_1,\dots,x_{d-2})
\cdots
p(x_2\mid x_1)p(x_1).
\end{aligned}
\]
Reordering the factors yields the standard autoregressive factorisation:
\[
\boxed{
p(x_1,\dots,x_d)
=
\prod_{i=1}^d p(x_i\mid x_{<i}),
}
\]
where
\[
x_{<i}:=(x_1,\dots,x_{i-1}),
\]
and for \(i=1\), the term \(p(x_1\mid x_{<1})\) is simply interpreted as
\(p(x_1)\).

This identity is exact. It makes no approximation and imposes no modelling
assumption by itself. Any joint distribution can be decomposed in this way once
an ordering of the variables has been chosen.

The modelling step comes from parameterising the conditional densities:
\[
p_\theta(x)
=
\prod_{i=1}^d p_\theta(x_i\mid x_{<i}).
\]
If each conditional density is normalised, then the product defines a properly
normalised joint density. This is why autoregressive models give exact
likelihoods by construction.

\begin{tcolorbox}[
  colback=blue!4,
  colframe=blue!50!black,
  title={Why the autoregressive product is a valid density},
  boxrule=0.6pt,
  arc=2mm,
  left=2mm,
  right=2mm,
  top=1mm,
  bottom=1mm
]
Suppose each conditional density \(p_\theta(x_i\mid x_{<i})\) is normalised:
\[
\int p_\theta(x_i\mid x_{<i})\,dx_i=1.
\]
Then the joint density
\[
p_\theta(x_1,\dots,x_d)
=
\prod_{i=1}^d p_\theta(x_i\mid x_{<i})
\]
is also normalised. Integrating from \(x_d\) backwards,
\[
\int p_\theta(x_d\mid x_{<d})\,dx_d=1,
\]
then
\[
\int p_\theta(x_{d-1}\mid x_{<d-1})\,dx_{d-1}=1,
\]
and so on until only \(p_\theta(x_1)\) remains, which also integrates to one.

Thus the autoregressive product is not merely a convenient factorisation; it
defines a valid probability density as long as each conditional term is valid.
\end{tcolorbox}

\subsection{Conditional Parameterisation and Generation}

The chain-rule factorisation becomes a learnable generative model once we
specify how each conditional density is represented. In general, we write
\[
p_\theta(x_i\mid x_{<i}),
\qquad i=1,\dots,d,
\]
where \(\theta\) denotes model parameters.

An important point is that these conditionals are usually not represented by
\(d\) unrelated models. Instead, a single parameterised function, often a neural
network, is used to output the parameters of all conditional distributions while
respecting the autoregressive ordering. The same parameters are reused across
the factorisation; what changes from one conditional to the next is the
conditioning context.

For continuous variables, a common simple choice is a Gaussian conditional:
\[
p_\theta(x_i\mid x_{<i})
=
\mathcal{N}\!\left(
x_i;
\mu_i(x_{<i}),
\sigma_i^2(x_{<i})
\right),
\]
where the mean and variance are functions of the previous variables. More
flexible conditionals may use mixtures of Gaussians, discretised logistic
mixtures, or other expressive one-dimensional distributions.

For discrete variables, the model may output probabilities for a categorical
distribution:
\[
p_\theta(x_i\mid x_{<i})
=
\operatorname{Categorical}\!\left(\pi_i(x_{<i})\right).
\]
In both cases, the model predicts a full conditional distribution, not merely a
point estimate.

\begin{tcolorbox}[
  colback=blue!4,
  colframe=blue!50!black,
  title={A three-variable autoregressive example},
  boxrule=0.6pt,
  arc=2mm,
  left=2mm,
  right=2mm,
  top=1mm,
  bottom=1mm
]
For $x=(x_1,x_2,x_3),$ the chain rule gives
\[
p_\theta(x_1,x_2,x_3)
=
p_\theta(x_1)\,
p_\theta(x_2\mid x_1)\,
p_\theta(x_3\mid x_1,x_2).
\]

If Gaussian conditionals are used, then each conditional can be represented by a
mean and variance. A shared parameter-generating function \(f_\theta\) may
produce these parameters:
\[
(\mu_1,\sigma_1^2)=f_\theta(\varnothing),
\]
\[
(\mu_2,\sigma_2^2)=f_\theta(x_1),
\]
\[
(\mu_3,\sigma_3^2)=f_\theta(x_1,x_2).
\]

Here \(f_\theta(\varnothing)\) means that the first conditional has no previous
variables to condition on, so it produces the unconditional parameters for
\(p_\theta(x_1)\). Then \(f_\theta(x_1)\) uses the available value of \(x_1\) to
produce the parameters for \(p_\theta(x_2\mid x_1)\), and
\(f_\theta(x_1,x_2)\) uses both previous values to produce the parameters for
\(p_\theta(x_3\mid x_1,x_2)\).

Thus, for example, 
\[
p_\theta(x_2\mid x_1)
=
\mathcal{N}\!\left(x_2;\mu_2,\sigma_2^2\right),
\qquad
(\mu_2,\sigma_2^2)=f_\theta(x_1),
\]

and similarly
\[
p_\theta(x_3\mid x_1,x_2)
=
\mathcal{N}\!\left(x_3;\mu_3,\sigma_3^2\right),
\qquad
(\mu_3,\sigma_3^2)=f_\theta(x_1,x_2).
\]
\end{tcolorbox}

The same model parameters \(\theta\) are reused throughout. What changes from
one conditional distribution to the next is not the model itself, but the amount
of conditioning information available at that step. This resolves a common
misconception: the autoregressive factorisation lists a sequence of conditional
distributions, but this does not mean that we train a separate unrelated model
for each factor. In modern autoregressive models, the conditionals are usually
represented together by one shared architecture with restricted information
flow.

The same factorisation also gives a direct sampling procedure. Given
\[
p_\theta(x)
=
\prod_{i=1}^d p_\theta(x_i\mid x_{<i}),
\]
we may generate a sample sequentially:
\[
x_1\sim p_\theta(x_1),
\]
then
\[
x_2\sim p_\theta(x_2\mid x_1),
\]
then
\[
x_3\sim p_\theta(x_3\mid x_1,x_2),
\]
and so on until
\[
x_d\sim p_\theta(x_d\mid x_1,\dots,x_{d-1}).
\]
After all components have been generated, the resulting vector
\[
x=(x_1,\dots,x_d)
\]
is a sample from the autoregressive model.

This procedure should be read as a full generative mechanism. The model does
not first sample a latent variable and then decode it. Nor does it run a
stochastic denoising process. It directly constructs the data point one
component at a time, using previously generated components as context for later
ones.

\subsection{Parallel Training, Sequential Sampling, and Masking}

Autoregressive models have an important computational asymmetry. Likelihood
evaluation during training can often be parallelised, while sampling is
inherently sequential.

Given an observed data point
\[
x=(x_1,\dots,x_d),
\]
the log-likelihood is
\[
\log p_\theta(x)
=
\sum_{i=1}^d
\log p_\theta(x_i\mid x_{<i}).
\]
During training, all components of \(x\) are known. Therefore, every
conditioning context \(x_{<i}\) can be read directly from the observed data.
No sampling is required. The model evaluates each conditional term using the
ground-truth previous components.

This is why autoregressive likelihood training is exact and stable. The model
is not asked to generate \(x_1\), then use the generated value to evaluate
\(x_2\), and so on. Instead, it evaluates all conditional probabilities against
the observed data:
\[
\log p_\theta(x_1),
\quad
\log p_\theta(x_2\mid x_1),
\quad
\dots,
\quad
\log p_\theta(x_d\mid x_{<d}).
\]
Depending on the architecture, many or all of these conditional terms can be
computed in parallel.

Sampling is different. When generating a new sample, future components are
unknown because they have not yet been produced. We must therefore proceed in
order:
\[
x_1\sim p_\theta(x_1),
\]
\[
x_2\sim p_\theta(x_2\mid x_1),
\]
\[
\cdots
\]
\[
x_d\sim p_\theta(x_d\mid x_{<d}).
\]
Each newly sampled component becomes part of the conditioning context for the
next one. This makes sampling sequential by construction.

To implement the autoregressive factorisation in a neural network, we must also
ensure that the parameters of the \(i\)-th conditional distribution depend only
on \(x_{<i}\), and not on \(x_i\) itself or on future variables. This restriction
is commonly enforced through \emph{masking}. A mask blocks forbidden connections
inside the model, ensuring that information flows only from earlier variables
to later variables.

Thus, even if the network processes the whole vector \(x\) at once during
training, the output parameters for \(x_i\) cannot depend on
\(x_i,x_{i+1},\dots,x_d\). The computation may be parallel, but the dependency
structure remains ordered.

Abstractly, masking imposes a lower-triangular dependency pattern. If we define
a dependency matrix \(D\) whose entry \(D_{ij}\) indicates whether the
conditional for \(x_i\) is allowed to depend on \(x_j\), then autoregressive
ordering requires
\[
D_{ij}=0
\qquad
\text{whenever}
\qquad
j\geq i.
\]
For example,
\[
D
=
\begin{pmatrix}
0      & 0      & 0      & \cdots & 0 \\
\ast   & 0      & 0      & \cdots & 0 \\
\ast   & \ast   & 0      & \cdots & 0 \\
\vdots & \vdots & \vdots & \ddots & 0 \\
\ast   & \ast   & \ast   & \cdots & 0
\end{pmatrix},
\]
where \(\ast\) denotes a permitted dependency.

This triangular dependency is a key reason the model remains tractable in
practice. It avoids cyclic relationships: the distribution of \(x_i\) may depend
on earlier components, but earlier components do not simultaneously depend on
\(x_i\). As a result, the high-dimensional joint likelihood decomposes into a
sum of conditional log-likelihoods.

In this sense, triangular dependency in autoregressive models plays a role
analogous to triangular Jacobians in coupling-based normalising flows. In both
cases, dependencies are organised in one direction so that a high-dimensional
probability calculation factorises into simple terms.

With this principle in place, many well-known architectures can be understood
at a structural level. MADE uses masking to enforce autoregressive structure in
feedforward networks~\cite{germain2015made}; PixelRNN and PixelCNN apply
autoregressive factorisation to images~\cite{oord2016pixel}; and WaveNet uses
causal convolutions for raw audio generation~\cite{oord2016wavenet}.

Autoregressive models therefore provide a second major route to exact density
modelling. Normalising flows achieve exact likelihoods through invertible
transformations and tractable Jacobian determinants. Autoregressive models
achieve exact likelihoods through ordered conditional factorisation. Both show
how high-dimensional probability models can remain computable when their
dependencies are carefully structured.

\section*{Summary and References}
\addcontentsline{toc}{section}{Summary and References}

This chapter introduced exact density modelling as a complementary paradigm to
the latent-variable and diffusion-based models developed earlier, where
likelihoods are evaluated directly without variational bounds,
stochastic noising processes, or learned score fields.

We first studied normalising flows. Their central idea is to transform a simple
base distribution into a complex data distribution using differentiable,
invertible mappings. The change-of-variables formula tracks how local volume
expansion or compression changes probability density. This led to the main
design requirements of flow layers: invertibility, dimension preservation, and
tractable Jacobian determinants.

We then examined representative flow constructions. Planar flows showed how a
simple nonlinear rank-one perturbation can keep the Jacobian determinant
tractable. Affine coupling layers introduced a more practical mechanism: by
transforming only part of the input while conditioning on the rest, they provide
closed-form inverses and efficient log-determinant computation. Stacking such
layers, together with coordinate mixing operations, allows simple local
transformations to form expressive generative models with exact likelihood
evaluation and exact model sampling.

The second half of the chapter introduced autoregressive models as another
route to exact likelihoods. Rather than transforming densities through
invertible maps, autoregressive models use the chain rule of probability to
factorise a joint density into a product of conditional distributions.
This gives exact likelihoods without latent variables, variational bounds, or
Jacobian determinants. The price is that an ordering must be imposed, and
sampling is usually sequential. We also saw how masking enforces the required
ordered information flow in neural architectures, allowing likelihood terms to
be evaluated efficiently during training while preserving the autoregressive
factorisation.

Across both model families, the central message is that exact likelihoods are
made possible by structural constraints. In normalising flows, these constraints
take the form of invertibility and tractable Jacobian determinants. In
autoregressive models, they take the form of ordered conditional dependencies.
Both approaches show how high-dimensional probability models can remain
computable when their transformations or dependencies are carefully organised.

\bigskip

Normalising flows were introduced as a general framework for constructing
flexible distributions through sequences of invertible transformations by
Rezende and Mohamed~\cite{rezende2015variational}. Important practical flow
architectures include NICE and RealNVP by Dinh et
al.~\cite{dinh2014nice,dinh2017density}, which developed coupling-based
transformations with tractable Jacobians, and Glow by Kingma and
Dhariwal~\cite{kingma2018glow}, which extended these ideas using invertible
 \(1\times1\) convolutions and scalable flow design. 
A broader review of normalising flows
and their role in density estimation is provided by  Papamakarios et
al.~\cite{papamakarios2021normalizing}.

Autoregressive neural density estimation has also played a major role in modern
generative modelling. MADE by Germain et al.~\cite{germain2015made} showed how
masking can enforce autoregressive structure in feedforward neural networks.
PixelRNN and PixelCNN by van den Oord et al.~\cite{oord2016pixel} developed
autoregressive image models, while WaveNet by van den Oord et
al.~\cite{oord2016wavenet} applied causal autoregressive modelling to raw audio.
Together, these works show the continuing importance of exact likelihood-based
modelling, both through invertible transformations and through ordered
conditional factorisations.

%% file: figures/ch07/chapter7_exact_density_at_a_glance_book.tex
\begin{center}
\includegraphics[width=0.96\textwidth]{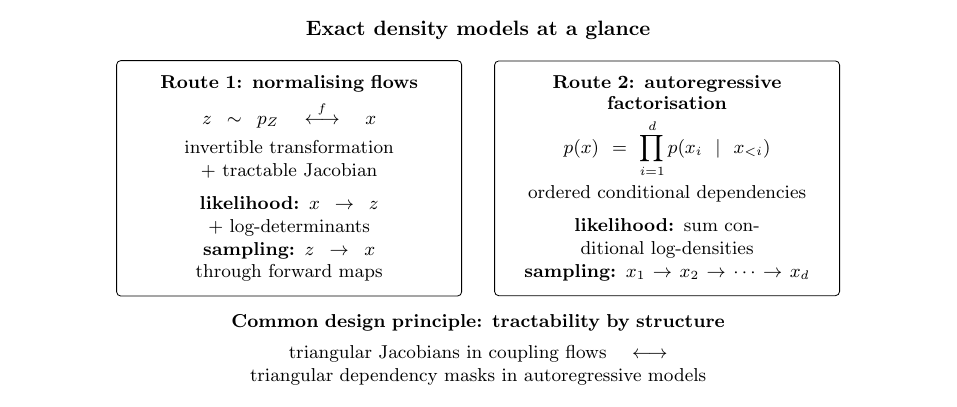}
\end{center}

%% file: figures/ch07/fig_ch07_01_change_of_variables.tex
\begin{figure}[H]
\centering
\includegraphics[width=0.88\textwidth]{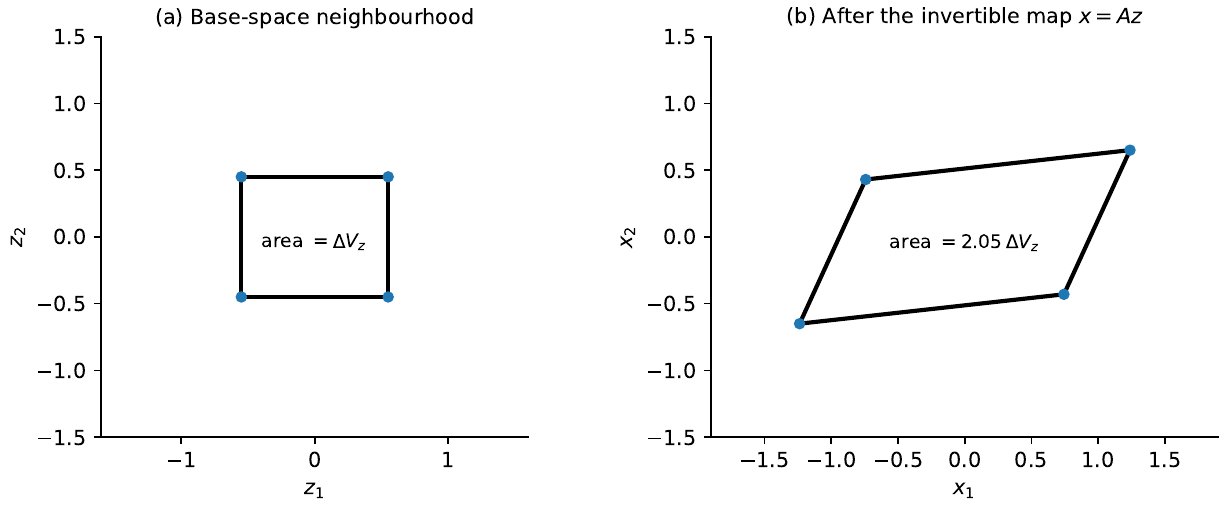}
\caption{\textbf{Change of variables corrects density for local volume change.}
The same probability mass occupies a different local area after an invertible
transformation. In the linear example, the determinant gives the area scaling;
for a nonlinear flow, the Jacobian gives the corresponding local scaling point
by point. This is the geometric origin of the log-determinant correction.}
\label{fig:flow-change-variable-geometry}
\end{figure}

%% file: figures/ch07/fig_ch07_02_planar_flow.tex
\begin{figure}[H]
\centering
\includegraphics[width=0.92\textwidth]{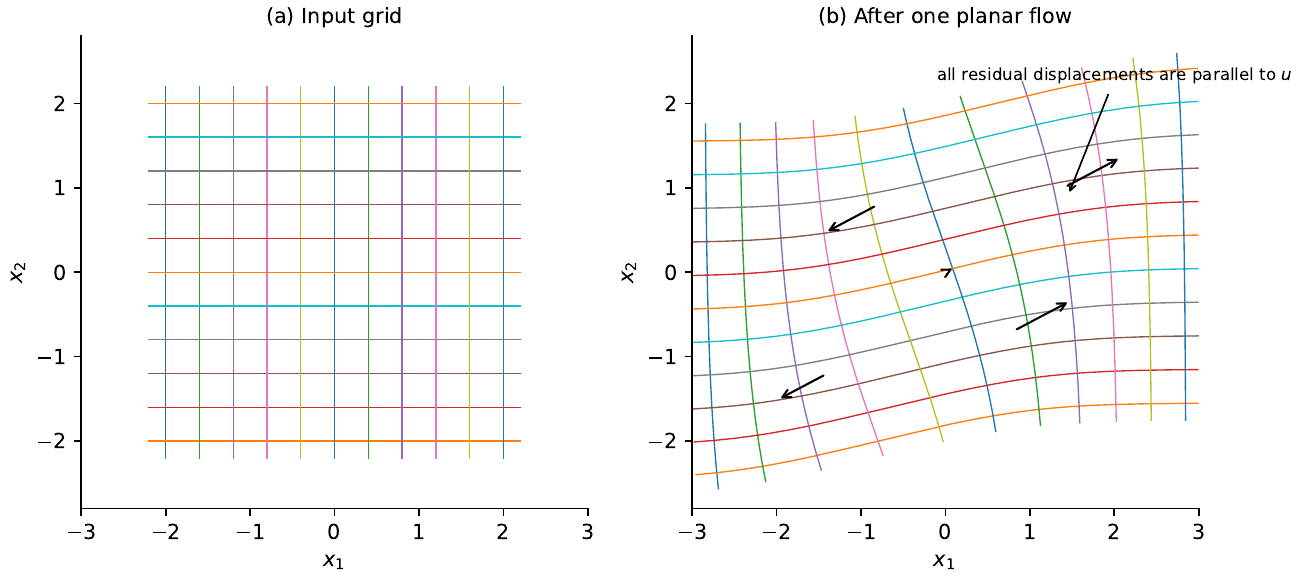}
\caption{\textbf{A planar flow is nonlinear but directionally constrained.}
The displacement varies with location through the scalar nonlinear response,
while every residual displacement remains parallel to the same vector $u$.
This is the geometric consequence of the rank-one structure derived in the text.}
\label{fig:planar-flow-rank-one}
\end{figure}

%% file: figures/ch07/fig_ch07_03_affine_coupling.tex
\begin{figure}[H]
\centering
\begin{tikzpicture}[
  >=Latex,
  every node/.style={font=\small},
  state/.style={draw,rounded corners=2pt,align=center,inner sep=5pt,text width=2.35cm,minimum height=.82cm},
  note/.style={draw,rounded corners=2pt,align=center,inner sep=5pt,text width=4.45cm,minimum height=.82cm},
  flow/.style={->,thick}
]
\node[font=\normalsize\bfseries] at (6.0,2.75) {Affine coupling: one-way dependence creates tractability};

\node[state] (x1) at (.8,1.20) {$x_1$\\context};
\node[state] (x2) at (.8,-.55) {$x_2$\\transformed};
\node[state] (st) at (4.0,.30) {scale and shift\\from $x_1$};
\node[state] (y1) at (7.35,1.20) {$y_1$\\unchanged};
\node[state] (y2) at (7.35,-.55) {$y_2$\\transformed};

\draw[flow] (x1) -- (st);
\draw[flow] (x1) -- (y1);
\draw[flow] (st) -- (y2);
\draw[flow] (x2) -- (y2);

\node[note] (tri) at (3.0,-2.45) {one-way dependency\\$\Rightarrow$ block-triangular Jacobian};
\node[note] (inv) at (8.7,-2.45) {unchanged branch\\$\Rightarrow$ direct recovery in the inverse};
\end{tikzpicture}
\caption{\textbf{Asymmetric information flow in affine coupling.}
One subset passes through unchanged while also determining how the other subset
is transformed. This one-way dependence gives both the triangular Jacobian and
the closed-form inverse.}
\label{fig:affine-coupling-structure}
\end{figure}

%% file: chapters/cht10_new.tex
\chapter{Beyond Likelihoods: Generative Adversarial Networks and Energy-Based Models}

%------------------------------------------------
% Overview
%------------------------------------------------
\section*{Overview}
\addcontentsline{toc}{section}{Overview}
Throughout this book, we have studied generative models that represent complex
data distributions in different mathematical ways. PPCA and VAEs introduced
latent-variable modelling and approximate inference. Diffusion models extended
this perspective into sequential noising and denoising processes, while
score-based models reinterpreted denoising through learned score fields and
stochastic dynamics. Normalising flows and autoregressive models showed how
exact likelihoods can be preserved through carefully structured transformations
or factorisations.

This final chapter turns to a different question: what happens when tractable
likelihoods are no longer the central object? In many important generative
models, the data distribution is not represented through an explicitly
normalised density that can be evaluated pointwise. Instead, learning proceeds
through comparison, geometry, or unnormalised scalar landscapes.

Generative Adversarial Networks (GANs) provide the first example of this shift. A
generator learns to produce samples that are difficult to distinguish from real
data, while a discriminator supplies the comparison signal. Wasserstein GANs
refine this adversarial idea by replacing classification-based comparison with
a geometric notion of distributional distance based on optimal transport.

Energy-Based Models offer a second route beyond explicit likelihoods. Rather
than defining a normalised density directly, they assign each configuration a
scalar energy. Low energy marks plausible regions of data space, while high
energy discourages implausible configurations. Learning then becomes the task of
shaping an energy landscape rather than directly maximising a tractable
likelihood.

The purpose of this chapter is to complete the conceptual landscape developed
throughout the book. GANs, Wasserstein GANs, and Energy-Based Models show that
generative learning can proceed even when normalised probabilities are
unavailable: by comparing samples, measuring geometric discrepancy, or learning
landscapes whose gradients guide movement through data space.

\section*{Concept Map}
\addcontentsline{toc}{section}{Concept Map}

This chapter studies generative models that do not require tractable likelihood
evaluation. The central theme is how learning can proceed when probabilities are
implicit, unnormalised, or replaced by comparisons.

\begin{itemize}

    \item[\(\rightarrow\)] \textbf{We first introduce GANs as learning by comparison.}  
    A generator maps latent noise to data-like samples, while a discriminator
    learns to distinguish generated samples from real data. The generator learns
    by trying to make this comparison fail.

    \item[\(\rightarrow\)] \textbf{We then interpret the adversarial game
    distributionally.}  
    For a fixed generator, the optimal discriminator estimates a local density
    ratio between the data distribution and the generator distribution.
    Substituting this discriminator back into the GAN objective reveals the
    Jensen--Shannon divergence.

    \item[\(\rightarrow\)] \textbf{Next, we examine GAN optimisation and its
    limitations.}  
    The original minimax generator update can produce weak gradients when the
    discriminator is confident. The non-saturating loss improves the generator's
    gradients, but mode collapse and overlap-based comparison remain important
    limitations.

    \item[\(\rightarrow\)] \textbf{We then move from overlap-based comparison to
    Wasserstein geometry.}  
    Wasserstein GANs replace classification-based discrepancy with the
    \(1\)-Wasserstein distance, which measures how far probability mass must be
    transported. This leads to critic functions, Lipschitz constraints, and
    geometry-aware training.

    \item[\(\rightarrow\)] \textbf{The chapter then turns to energy-based models
    as scalar landscapes.}  
    An energy-based model assigns each configuration a scalar energy
    \(E_\theta(x)\). Low energy marks plausible regions of data space, while high
    energy discourages implausible configurations. The model usually defines an
    unnormalised density rather than a tractable likelihood.

    \item[\(\rightarrow\)] \textbf{We next derive the contrastive structure of
    EBM learning.}  
    Maximum likelihood for energy-based models lowers energy on data samples and
    raises energy on samples from the model distribution. This exposes the
    central difficulty: learning requires sampling from the current model.

        \item[\(\rightarrow\)] \textbf{Finally, we connect energies, scores, and
    diffusion.}  
    In unnormalised energy models, low energy corresponds to high probability,
    and the score becomes the negative energy gradient. Diffusion models can
    therefore be viewed as learning energy-gradient-like directions across
    multiple noise levels, providing a stable multi-scale route beyond explicit
    likelihoods.

\end{itemize}

\noindent The three beyond-likelihood routes developed in this chapter are summarised visually below.
\input{figures/ch08/chapter8_beyond_likelihoods_at_a_glance}

\section{Generative Adversarial Networks: Learning by Comparison}

Many of the generative models introduced so far rely on explicit probability
distributions. Whether through exact likelihoods or carefully constructed
approximations, learning has often been framed as the problem of assigning high
probability to observed data. This provides a principled foundation, but it also
imposes strong structural constraints on how models are designed and trained.

This naturally raises a simpler question:
\[
\textit{do we need to compute probabilities at all in order to generate
realistic data?}
\]
If the goal is to produce samples that look like they came from the data
distribution, it may be enough to learn by comparison. Rather than asking how
likely a sample is, we can ask whether it is indistinguishable from real data.
Learning then proceeds by matching samples, not by evaluating densities.

Generative Adversarial Networks (GANs)~\cite{goodfellow2014generative} are built around this idea. They define
a generative model through its ability to produce convincing samples, without
ever specifying or evaluating a probability density. Training is driven by
comparison: generated samples are pitted against real data, and the model
improves by learning to make the two increasingly difficult to tell apart.

\subsection{The Generator as an Implicit Sampling Mechanism}

A GAN begins with a \emph{generator}, a parameterised function
\[
x = G_\theta(z),
\qquad z \sim p(z),
\]
where \(z\) is a latent variable drawn from a simple fixed distribution, such as
a standard Gaussian or a uniform distribution. The generator \(G_\theta\), often
implemented as a neural network, maps this simple source of randomness into the
data space. 

The mapping
\[
z \longmapsto G_\theta(z)
\]
transforms simple noise into structured, data-like samples. By sampling
\(z\sim p(z)\) and passing it through \(G_\theta\), the generator induces a
distribution over \(x\), denoted \(p_\theta(x)\). Crucially, this distribution
is defined only \emph{implicitly}: we can draw samples from it, but we cannot
usually evaluate its density at a given point.

This distinguishes GANs from the likelihood-based models studied earlier.
Unlike VAEs, the latent variable \(z\) is not inferred from data and no
posterior distribution is learned. Unlike diffusion models, \(z\) does not arise
from a forward noising process. Unlike normalising flows or autoregressive
models, the generator need not be invertible, and no Jacobian determinant or
explicit normalisation of \(p_\theta(x)\) is required. The generator therefore
behaves like a deterministic simulator: it specifies \emph{how to generate
data}, but not \emph{how likely a given data point is}.

Once training is complete, sampling from a GAN is straightforward. We draw $z\sim p(z)$ from the latent noise distribution and compute $x=G_\theta(z).$ The discriminator is not needed at generation time; it serves only as a training
signal that shapes the generator. Thus GANs provide direct sampling, but not
direct likelihood evaluation.
\subsection{The Discriminator and the Adversarial Game}

If a model produces only samples, a fundamental question immediately arises:
\[
\textit{how can we tell whether those samples are good?}
\]
In likelihood-based models, this question is answered by probability: samples
are good if they receive high likelihood under the model. In GANs, likelihoods
are unavailable. There is no tractable function \(p_\theta(x)\) to evaluate or
maximise.

To resolve this, GANs introduce a second learned component: the
\emph{discriminator}. Rather than measuring probability, the discriminator
provides a data-driven comparison between two sources of samples:
\[
x\sim p_{\text{data}}(x),
\qquad
x=G_\theta(z),\quad z\sim p(z).
\]
The first source consists of real data samples; the second consists of generated
samples.

Formally, the discriminator is a function
\[
D_\phi:\mathbb{R}^d\to(0,1),
\]
parameterised by \(\phi\). For an input \(x\), the value \(D_\phi(x)\) is
interpreted as the discriminator's confidence that \(x\) came from the real data
distribution rather than from the generator. Thus the discriminator is trained
as a binary classifier:
\[
D_\phi(x)\approx 1
\quad\text{for real data,}
\qquad
D_\phi(G_\theta(z))\approx 0
\quad\text{for generated data.}
\]

The discriminator is not an end in itself. It does not define a probability
density, nor is it the final object we want to use after training. Its role is
to provide a moving, adaptive notion of discrepancy between real and generated
samples. The generator learns only through this feedback, adjusting its
parameters so that generated samples become harder to distinguish from real
ones.

This interaction is formalised by the minimax objective
\[
\boxed{
\min_\theta \max_\phi
\;\;
\mathbb{E}_{x \sim p_{\text{data}}}
\big[\log D_\phi(x)\big]
+
\mathbb{E}_{z \sim p(z)}
\big[\log (1 - D_\phi(G_\theta(z)))\big].
}
\]
The discriminator tries to maximise this objective by assigning high scores to
real samples and low scores to generated samples. The generator tries to
minimise it by producing samples for which \(D_\phi(G_\theta(z))\) becomes large,
so that the discriminator struggles to classify them as fake.

\begin{tcolorbox}[
  colback=blue!4,
  colframe=blue!50!black,
  title={How to read the minimax objective},
  boxrule=0.6pt,
  arc=2mm,
  left=2mm,
  right=2mm,
  top=1mm,
  bottom=1mm
]
The expression
\[
\min_\theta \max_\phi \; \mathcal{L}(\theta,\phi)
\]
should be read as a two-player game.

For fixed generator parameters \(\theta\), the inner maximisation asks:
\[
\textit{what is the best discriminator against the current generator?}
\]
This trains \(D_\phi\) to separate real samples from generated ones.

Then, with the discriminator treated as fixed, the outer minimisation asks:
\[
\textit{how should the generator change to make real and generated samples harder to tell apart?}
\]
In practice, these two steps are interleaved: we alternate between improving the
discriminator and improving the generator.
\end{tcolorbox}

Although the objective resembles a classification loss, its role is
fundamentally generative. The discriminator defines a learned comparison between
two distributions, while the generator reshapes its output distribution so that
this comparison gradually disappears.

\subsection{The Optimal Discriminator}

To understand adversarial training at the distributional level, consider an
idealised setting in which the generator \(G_\theta\) is fixed. Sampling
\(z\sim p(z)\) and mapping \(x=G_\theta(z)\) induces an implicit generator
distribution \(p_\theta(x)\). With the generator fixed, the discriminator's
task reduces to binary classification between
\[
p_{\text{data}}(x)
\qquad\text{and}\qquad
p_\theta(x).
\]

The discriminator objective can then be written as
\[
\mathbb{E}_{x \sim p_{\text{data}}}\!\big[\log D(x)\big]
+
\mathbb{E}_{x \sim p_\theta}\!\big[\log(1 - D(x))\big].
\]
Writing these expectations as integrals gives
\[
\int
p_{\text{data}}(x)\log D(x)\,dx
+
\int
p_\theta(x)\log(1-D(x))\,dx.
\]
Because this objective decomposes over \(x\), the optimal discriminator can be
derived pointwise. For each fixed \(x\), we maximise
\[
p_{\text{data}}(x)\log D(x)
+
p_\theta(x)\log(1-D(x))
\]
with respect to the scalar \(D(x)\).

Differentiating with respect to \(D(x)\) gives
\[
\frac{p_{\text{data}}(x)}{D(x)}
-
\frac{p_\theta(x)}{1-D(x)}.
\]
Setting this derivative to zero yields
\[
\frac{p_{\text{data}}(x)}{D(x)}
=
\frac{p_\theta(x)}{1-D(x)}.
\]
Solving for \(D(x)\), we obtain the optimal discriminator
\[
\boxed{
D^*(x)
=
\frac{p_{\text{data}}(x)}
{p_{\text{data}}(x)+p_\theta(x)}.
}
\]

This formula gives an important interpretation: the optimal discriminator is a
local density-ratio object, even though neither density is explicitly available
to the model. Fig.~\ref{fig:gan-optimal-discriminator-geometry} shows this
point directly: its output moves toward one where the data density dominates,
toward zero where the generator density dominates, and equals one half where
the two densities are equal.

\input{figures/ch08/fig_ch08_01_optimal_discriminator}

\subsection{From the Optimal Discriminator to Jensen--Shannon Divergence}

Having derived the optimal discriminator for a fixed generator, we can ask what
the adversarial game represents at the level of entire distributions. To do
this, we write the discriminator objective as a \emph{value function}, which
scores a generator--discriminator pair by how well the discriminator separates
real samples from generated ones.

We define
\[
V(G_\theta,D)
=
\mathbb{E}_{x \sim p_{\text{data}}}\!\big[\log D(x)\big]
+
\mathbb{E}_{x \sim p_\theta}\!\big[\log(1-D(x))\big].
\]
For fixed \(G_\theta\), maximising \(V(G_\theta,D)\) over \(D\) gives the best
discriminator against that generator. If we then evaluate the value function at
this best discriminator, we can identify the distributional discrepancy that
the generator is implicitly minimising.

Using
\[
D^*(x)
=
\frac{p_{\text{data}}(x)}
{p_{\text{data}}(x)+p_\theta(x)},
\]
after algebraic simplification, we obtain
\[
V(G_\theta,D^*)
=
-\log 4
+
2\,\mathrm{JS}\!\left(p_{\text{data}}\,\|\,p_\theta\right),
\]
where \(\mathrm{JS}(\cdot\|\cdot)\) denotes the Jensen--Shannon divergence.

The constant term \(-\log 4\) does not depend on the generator, and therefore
does not affect the optimisation. Thus, in the idealised setting where the
discriminator is optimal, GAN training corresponds to minimising the
Jensen--Shannon divergence between the data distribution and the generator
distribution.

The expression above introduces a new divergence between probability
distributions: the \emph{Jensen--Shannon divergence}. To understand it, it is
helpful to recall the Kullback--Leibler divergence, which appeared earlier in
the VAE chapter as part of the evidence lower bound. The KL divergence compares
two distributions by
\[
\mathrm{KL}(p\,\|\,q)
=
\int p(x)\log\frac{p(x)}{q(x)}\,dx.
\]
It measures how costly it is to use \(q\) as an approximation to \(p\). However,
KL divergence is asymmetric:
\[
\mathrm{KL}(p\,\|\,q)
\neq
\mathrm{KL}(q\,\|\,p),
\]
and it is sensitive to support mismatch. If \(q(x)=0\) in a region where
\(p(x)>0\), then \(\mathrm{KL}(p\,\|\,q)\) becomes infinite.

The Jensen--Shannon divergence modifies this comparison by comparing both
distributions to a shared intermediate distribution. Given two distributions
\(p\) and \(q\), define their mixture
\[
m(x)
=
\frac{1}{2}\big(p(x)+q(x)\big).
\]
This mixture can be understood as the distribution obtained by first choosing
between \(p\) and \(q\) with equal probability, and then sampling from the chosen
distribution.

The Jensen--Shannon divergence is then defined as
\[
\boxed{
\mathrm{JS}(p\,\|\,q)
=
\frac{1}{2}\mathrm{KL}(p\,\|\,m)
+
\frac{1}{2}\mathrm{KL}(q\,\|\,m).
}
\]
Thus, rather than asking one distribution to explain the other directly, JS
divergence asks how far each distribution is from their shared mixture. This
makes it symmetric:
\[
\mathrm{JS}(p\,\|\,q)
=
\mathrm{JS}(q\,\|\,p),
\]
and bounded:
\[
0
\leq
\mathrm{JS}(p\,\|\,q)
\leq
\log 2.
\]

This is precisely the divergence that appears when the optimal discriminator is
substituted back into the original GAN value function. In the idealised setting
where the discriminator is optimal, the generator is therefore not maximising
likelihood. Instead, it is adjusting its implicit distribution \(p_\theta\) so
as to reduce
\[
\mathrm{JS}\!\left(p_{\text{data}}\,\|\,p_\theta\right).
\]

This reveals the distributional meaning of adversarial training. GANs do not
construct an explicit density; they perform \emph{distribution matching through
comparison}. The discriminator estimates how distinguishable the real and
generated distributions are, and the generator learns by trying to reduce that
distinguishability.

The global optimum is achieved when
\[
p_\theta(x)=p_{\text{data}}(x).
\]
In that case,
\[
D^*(x)=\frac{1}{2}
\]
everywhere, because real and generated samples are indistinguishable.

\section{Geometric Optimisation and Wasserstein GANs}

The preceding analysis showed that, under an optimal discriminator, the original
GAN objective corresponds to minimising the Jensen--Shannon divergence between
the data distribution and the generator's implicit distribution. This gives a
clean theoretical interpretation, but also reveals a limitation: when
\(p_{\text{data}}\) and \(p_\theta\) have little or no overlap, the JS divergence
can saturate, allowing the discriminator to separate real and generated samples
almost perfectly while giving the generator little guidance on how to move
toward the data distribution. This motivates the key shift:
from overlap-based comparison to geometric distance. Wasserstein GANs~\cite{arjovsky2017wasserstein} replace
the question of whether two distributions are distinguishable with the question
of how far probability mass must be moved to turn one distribution into the
other. 

\subsection{Generator Optimisation and the Non-Saturating Loss}

Before moving to Wasserstein GANs, we first discuss a common practical
workaround within the original GAN framework: changing the generator's loss to
improve the gradient signal during training.
The original GAN objective is written as the minimax problem
\[
\min_\theta \max_\phi
\;
\mathbb{E}_{x \sim p_{\text{data}}}\big[\log D_\phi(x)\big]
+
\mathbb{E}_{z \sim p(z)}\big[\log (1 - D_\phi(G_\theta(z)))\big].
\]
For the discriminator, the interpretation is straightforward: it is rewarded
for assigning high values to real samples and low values to generated samples.
For the generator, the situation is more subtle.

If we fix the discriminator and minimise the same value function with respect
to the generator parameters \(\theta\), the only term that depends on
\(\theta\) is
\[
\mathbb{E}_{z \sim p(z)}
\big[\log (1 - D_\phi(G_\theta(z)))\big].
\]
This objective asks the generator to reduce the discriminator's confidence that
generated samples are fake.

The difficulty appears early in training. At this stage, the discriminator can
often identify generated samples confidently:
\[
D_\phi(G_\theta(z))\approx 0.
\]
Then
\[
1-D_\phi(G_\theta(z))\approx 1,
\qquad
\log(1-D_\phi(G_\theta(z)))\approx 0.
\]
The generator objective is therefore close to flat in this region, and the
gradient passed back to the generator can become very small even though its
samples are still poor.

For this reason, GANs are commonly trained with a different generator objective,
called the \emph{non-saturating loss}:
\[
\min_\theta
\;
\mathbb{E}_{z \sim p(z)}
\big[
-\log D_\phi(G_\theta(z))
\big].
\]
This loss is not obtained by algebraically simplifying the previous expression.
Instead, it changes the generator's goal from reducing the discriminator's
``fake'' confidence to increasing its ``real'' confidence.

The two objectives are different, but they have the same desired equilibrium:
generated samples should become indistinguishable from real data. Their
difference lies in the gradient behaviour. When
\[
D_\phi(G_\theta(z))\approx 0,
\]
the non-saturating term
\[
-\log D_\phi(G_\theta(z))
\]
is large, so the generator receives a much stronger learning signal. This is why
the non-saturating loss is commonly preferred in practice, even though the
original minimax objective provides the cleaner theoretical link to
Jensen--Shannon divergence.
This optimisation behaviour helps explain both the strength and the weakness of
GANs. On the positive side, the discriminator penalises features that make
generated samples look unrealistic, so adversarial training can encourage sharp,
visually convincing outputs rather than blurry averages. On the negative side, the mechanism can give rise to \emph{mode collapse}.
Since the generator is trained to fool the discriminator, it may model only a
subset of the data distribution convincingly. If that subset is enough to
confuse the discriminator, other modes may be ignored.

Likelihood-based models are penalised when they fail to cover observed modes,
because likelihood depends on assigning probability mass to the data. GANs, by
contrast, optimise distinguishability rather than explicit probability coverage. Thus the non-saturating loss improves the generator's gradients, but it does not
change the underlying geometry of the adversarial comparison. When two
distributions have little overlap, a classification-based discrepancy may still
say mainly that they are distinguishable, not how one should move toward the
other. This motivates a more geometric notion of distributional discrepancy.

\subsection{From Overlap to Wasserstein Geometry}

To develop this idea, it is helpful to think of probability distributions
geometrically. Rather than viewing a distribution as a static density, imagine
its probability mass as something that can be moved through space. If
transforming one distribution into another requires only small displacements of
mass, the distributions should be considered close. If large amounts of mass
must be transported over long distances, they should be considered far apart.

The Wasserstein distance formalises this intuition. In the form used by
Wasserstein GANs, the relevant object is the \(1\)-Wasserstein distance, also
called the Earth Mover's distance. It measures the minimal cost of transporting
probability mass from one distribution to another, with cost proportional to the
distance moved.

A key theoretical result, the Kantorovich--Rubinstein duality, gives a practical
way to express this distance. For two probability distributions \(p\) and \(q\)
on \(\mathbb{R}^d\), the \(1\)-Wasserstein distance can be written as
\[
\boxed{
W(p,q)
=
\sup_{\|f\|_{L} \le 1}
\left(
\mathbb{E}_{x \sim p}[f(x)]
-
\mathbb{E}_{x \sim q}[f(x)]
\right).
}
\]
Here, \(f\) is a real-valued scoring function, and the constraint
\[
\|f\|_{L}\leq 1
\]
means that \(f\) must be \(1\)-Lipschitz: its value cannot change faster than
unit rate with respect to distance in the input space.

The expression
\[
\mathbb{E}_{x \sim p}[f(x)]
-
\mathbb{E}_{x \sim q}[f(x)]
\]
measures how strongly the scoring function separates the two distributions on
average. The supremum means that we choose the scoring function that separates
them as much as possible, subject to the Lipschitz constraint.

This constraint is essential. Without it, the scoring function could assign
arbitrarily large values to samples from one distribution and arbitrarily small
values to samples from the other, making the comparison meaningless. The
Lipschitz constraint forces the score difference to reflect the geometry of the
data space: large score differences are only allowed when the corresponding
points are far apart.

Note this is different from the score in score-based diffusion models, where the
score means \(\nabla_x\log p(x)\), the gradient of a log-density; in WGANs, the critic output \(f_\phi(x)\) is a real-valued scalar used for
comparison, rather than a probability; the critic is constrained through its
Lipschitz behaviour, not through probabilistic normalisation.
\begin{tcolorbox}[
  colback=blue!4,
  colframe=blue!50!black,
  title={A worked example of the Wasserstein distance},
  boxrule=0.6pt,
  arc=2mm,
  left=2mm,
  right=2mm,
  top=1mm,
  bottom=1mm
]
Consider a one-dimensional example in which \(p\) places all its mass at
\(x=0\), while \(q\) places all its mass at \(x=10\). Sampling from \(p\) always
returns \(0\), and sampling from \(q\) always returns \(10\), so
\[
\mathbb{E}_{x\sim p}[f(x)] = f(0),
\qquad
\mathbb{E}_{x\sim q}[f(x)] = f(10),
\qquad
W(p,q)
=
\sup_{\|f\|_L\leq 1}
\big(f(0)-f(10)\big).
\]

The constraint \(\|f\|_L\leq 1\) means that
\[
|f(x_1)-f(x_2)|\leq |x_1-x_2|
\quad
\text{for all }x_1,x_2.
\]
Hence every admissible \(f\) satisfies
\[
f(0)-f(10)\leq |0-10|=10.
\]
This upper bound is achieved by \(f(x)=-x\), since
\[
f(0)-f(10)=0-(-10)=10.
\]
Therefore $W(p,q)=10,$ matching the transport intuition exactly: all probability mass must be moved a distance of \(10\).
\end{tcolorbox}

This example shows why the Wasserstein distance remains informative when
distributions do not overlap. An overlap-based divergence can detect that the
supports are disjoint, but may not reflect whether the distributions are one
unit apart or ten units apart. The Wasserstein distance, by contrast, directly
measures spatial separation. Fig.~\ref{fig:gan-js-wasserstein-geometry}
shows the same distinction for a continuous family of translated Gaussian
distributions: JS becomes nearly saturated as overlap disappears, whereas the
Wasserstein distance continues to track spatial separation.

\input{figures/ch08/fig_ch08_02_js_vs_wasserstein}

\subsection{The Wasserstein GAN Objective}

The dual characterisation of the Wasserstein distance leads directly to a
practical adversarial objective. If the generator induces the distribution
\(p_\theta\), then the Wasserstein distance between the data distribution and the
generator distribution can be written as
\[
W(p_{\text{data}},p_\theta)
=
\sup_{\|f\|_{L}\leq 1}
\left(
\mathbb{E}_{x\sim p_{\text{data}}}[f(x)]
-
\mathbb{E}_{x\sim p_\theta}[f(x)]
\right).
\]
Since samples from \(p_\theta\) are produced by
\[
x=G_\theta(z),
\qquad z\sim p(z),
\]
we can write
\[
\mathbb{E}_{x\sim p_\theta}[f(x)]
=
\mathbb{E}_{z\sim p(z)}[f(G_\theta(z))].
\]

Wasserstein GANs represent the scoring function \(f\) by a neural network
\(f_\phi\), and optimise
\[
\boxed{
\min_\theta \max_{\phi \in \mathcal{L}}
\;
\mathbb{E}_{x \sim p_{\text{data}}}[f_\phi(x)]
-
\mathbb{E}_{z \sim p(z)}[f_\phi(G_\theta(z))],
}
\]
where \(\mathcal{L}\) denotes the set of parameter values for which \(f_\phi\)
satisfies the required Lipschitz constraint.

The maximisation over \(\phi\) seeks a scoring function that separates real and
generated samples as strongly as possible while respecting the geometry of the
data space. The minimisation over \(\theta\) then updates the generator so that
this separation becomes smaller.

At this point, the terminology changes. In the original GAN, the sample-evaluating
network is called a \emph{discriminator}. Its output
\[
D_\phi(x)\in(0,1)
\]
is interpreted as the probability that \(x\) is real. In a Wasserstein GAN, the
corresponding network is called a \emph{critic}. It outputs a real-valued score
\[
f_\phi(x)\in\mathbb{R},
\]
not a probability. There is no sigmoid output, no binary classification
interpretation, and no requirement that the scores be normalised. Only
differences in critic values matter.

Seen in this way, the critic is a neural-network approximation to the optimal
scoring function in the Wasserstein distance. Its purpose is not to make a hard
real--fake decision, but to assign scores whose average difference reflects the
geometric separation between the two distributions. Real samples should receive
higher scores on average than generated samples, subject to the constraint that
scores cannot change arbitrarily fast across the data space. Since the
Wasserstein distance changes continuously as the generator distribution moves,
the generator can still receive useful gradients even when \(p_\theta\) and
\(p_{\text{data}}\) do not overlap, so optimisation is guided by geometric
proximity rather than overlap alone.
As in the original GAN, the critic is used only during training. Once the
generator has been learned, sampling is again performed directly by drawing
\(z\sim p(z)\) and returning \(G_\theta(z)\).

\subsection{Enforcing the Lipschitz Constraint}

The remaining challenge is practical: how can the Lipschitz constraint on the
critic be enforced during training? Recall that this constraint ensures the
critic's scores change at a controlled rate, so that score differences reflect
geometric distance rather than arbitrary scaling. Neural networks do not satisfy
this property automatically.

The original WGAN enforced the constraint through \emph{weight clipping}, which
restricts each parameter of the critic to a fixed interval. Although simple,
this is a crude way to control the Lipschitz constant. It can reduce the
capacity of the critic and may lead to unstable optimisation.

A more effective alternative is the \emph{gradient penalty}, introduced in
WGAN-GP~\cite{gulrajani2017improved}. Instead of directly
constraining the parameters, the gradient penalty constrains the critic's
input--output behaviour. The idea is to control how rapidly the critic's score
changes in response to small changes in its input.

A penalty term of the form
\[
\lambda
\,
\mathbb{E}_{\hat{x}}
\left[
\big(\|\nabla_{\hat{x}} f_\phi(\hat{x})\|_2-1\big)^2
\right]
\]
is added to the critic's training objective, where \(\hat{x}\) is sampled along
straight-line interpolations between real samples and generated samples.

The gradient
\[
\nabla_{\hat{x}} f_\phi(\hat{x})
\]
measures how sensitive the critic's score is to infinitesimal changes in the
input. Its Euclidean norm is therefore the local rate of change, or local slope,
of the critic at \(\hat{x}\). Penalising deviations of $\|\nabla_{\hat{x}} f_\phi(\hat{x})\|_2$ from one encourages the critic to change smoothly and at a controlled rate in
the regions most relevant for comparing real and generated samples.

The interpolation points \(\hat{x}\) lie between real and generated samples,
precisely where the critic must learn how to compare the two distributions.
Enforcing the constraint along these paths stabilises training while preserving
the geometric interpretation of the Wasserstein distance. The coefficient
\(\lambda\) controls the strength of the penalty: larger values enforce the
constraint more strongly, while smaller values allow greater flexibility.

By enforcing the Lipschitz constraint in this way, WGANs with gradient penalty
preserve the transport-based meaning of the Wasserstein distance while achieving
more reliable training dynamics. More broadly, they show that generative
learning need not rely on explicit likelihoods or normalised densities. It can
instead be guided by a learned real-valued function whose geometry pushes
generated samples toward the data distribution.

This provides a natural bridge to energy-based models. Like WGANs, energy-based
models use real-valued functions to compare, rank, or guide samples. But instead
of framing this through an adversarial game, they define distributions through
unnormalised energy landscapes. We turn to this perspective next.

\section{Energy-Based Models: Foundations and Learning}

The preceding sections considered generative models that learn by comparison.
In GANs, a discriminator provides a learned test for distinguishing real and
generated samples. In Wasserstein GANs, this comparison is replaced by a
geometric critic whose values reflect how far probability mass must be moved.
Both approaches avoid explicit likelihood evaluation and instead learn through
relative assessments of samples.

Energy-based models~\cite{lecun2006tutorial} take this idea one step further. Rather than defining a
generator, a discriminator, or an explicit likelihood, they begin with a single
scalar function over the data space: an \emph{energy function}. This function
assigns a value
\[
E_\theta(x)\in\mathbb{R}
\]
to each configuration \(x\). Lower energy indicates greater compatibility with
the structure of the data, while higher energy indicates implausibility.

The energy function is deliberately unconstrained. It need not be positive,
bounded, or normalised, and by itself it does not define a probability
distribution. At this stage, energy should therefore not be interpreted as a
probability. It is better understood as a measure of preference, cost, or
compatibility. Data-like configurations should lie in low-energy regions, while
unrealistic configurations should lie in high-energy regions.

This gives a geometric view of modelling. An energy-based model defines a
landscape over the data space:
\[
\text{low energy}
\quad\Longleftrightarrow\quad
\text{plausible configurations},
\qquad
\text{high energy}
\quad\Longleftrightarrow\quad
\text{implausible configurations}.
\]
Learning then means shaping this landscape so that observed data occupy valleys,
while regions away from the data are assigned higher energy.

\subsection{Energy as a Modelling Primitive}

The key modelling primitive in an energy-based model is the scalar energy
function
\[
E_\theta:\mathbb{R}^d\to\mathbb{R}.
\]
Unlike normalising flows or autoregressive models, this function does not need
to provide an exact density. Unlike VAEs, it does not require an explicit latent
posterior. Unlike GANs, it does not require a separate discriminator trained in
an adversarial game. The model is specified directly through a scalar landscape.

Only relative energy differences matter. If we add a constant \(c\) to the
energy function,
\[
\tilde{E}_\theta(x)=E_\theta(x)+c,
\]
then every configuration is shifted by the same amount. The ordering of points
from low to high energy is unchanged. Since the model's behaviour depends on
which configurations are preferred relative to others, this constant shift has
no meaningful effect.

This relative nature is important. An energy function is not trying to assign an
absolute probability value to every point. Instead, it encodes preferences:
which configurations should be encouraged, and which should be discouraged. In
this sense, energy-based modelling asks a different question from explicit
density modelling:
\[
\textit{where should the model place low-energy regions in the data space?}
\]
The answer is learned from data.

\subsection{From Energies to Unnormalised Distributions}

Although an energy function is not itself a probability distribution, it can be
connected to one through the Boltzmann form
\[
p_\theta(x)
=
\frac{1}{Z_\theta}
\exp\big(-E_\theta(x)\big),
\]
where
\[
Z_\theta
=
\int
\exp\big(-E_\theta(x)\big)\,dx
\]
is the normalising constant, or \emph{partition function}.

The negative exponential converts low energy into large positive weight, while
the partition function performs one global normalisation. Fig.~\ref{fig:ebm-energy-to-density}
shows this mapping from scalar energy to unnormalised weight and then to a
normalised density; it also makes clear why adding a constant to all energies
does not change the final density.

\input{figures/ch08/fig_ch08_03_energy_to_density}

The difficulty is that \(Z_\theta\) is usually intractable in high-dimensional
spaces. Computing it requires integrating over all possible configurations:
\[
Z_\theta
=
\int
\exp\big(-E_\theta(x)\big)\,dx.
\]
For images, audio, text representations, or other high-dimensional data, this
integral is generally impossible to evaluate exactly.

This is the central trade-off of energy-based modelling. By avoiding the
structural constraints required by exact density models, energy functions can be
very flexible. But this flexibility comes at the cost of normalisation: the
model usually defines an \emph{unnormalised} density
\[
p_\theta(x)
\propto
\exp\big(-E_\theta(x)\big),
\]
rather than a tractable normalised likelihood.

This distinguishes energy-based models from the exact density models studied in
the previous chapter. Normalising flows and autoregressive models are carefully
designed so that \(p_\theta(x)\) can be evaluated exactly. Energy-based models
instead place the emphasis on the energy landscape itself. Relative energy
differences and energy gradients can be useful even when the normalising
constant is unknown.

\subsection{Learning Energy Landscapes}

Learning in an energy-based model means shaping the energy function so that it
reflects the structure of the data. Suppose we observe samples
\[
x\sim p_{\text{data}}(x).
\]
A desirable energy function should assign low energy to these observed samples:
\[
E_\theta(x)
\quad\text{small for}\quad
x\sim p_{\text{data}}.
\]
But this condition alone is not enough. If we simply lowered the energy
everywhere, the model would not distinguish data-like configurations from
implausible ones. Learning must therefore be contrastive: it must lower the
energy on data samples while raising the energy elsewhere.

Informally, the desired behaviour is
\[
E_\theta(x)
\quad\text{low on data},
\qquad
E_\theta(x)
\quad\text{high away from data}.
\]
This contrastive structure is what makes energy-based learning meaningful. The
model must not merely reward observed data; it must also learn which alternative
configurations should be discouraged.

A useful way to see this is through the energy gradient
\[
\nabla_x E_\theta(x).
\]
This gradient describes how the energy changes locally in data space. Moving in
the direction
\[
-\nabla_x E_\theta(x)
\]
decreases energy, guiding a configuration toward regions that the model
considers more plausible. Thus the energy landscape defines not only a scalar
preference value, but also a local direction of improvement.

This connects energy-based models to the geometric viewpoint developed earlier
for GANs and score-based models. In a Wasserstein GAN, the critic assigns
real-valued scores whose differences guide the generator. In an energy-based
model, the energy function assigns scalar values whose gradients define how
configurations should move through the data space. The two ideas are related,
but not identical: a WGAN critic is trained to separate two distributions,
whereas an energy function is intended to define a landscape of plausibility.

\subsection{Maximum Likelihood and the Contrastive Gradient}

Although energy-based models are often discussed as unnormalised models, the
Boltzmann form still provides a principled learning objective. If
\[
p_\theta(x)
=
\frac{1}{Z_\theta}\exp(-E_\theta(x)),
\]
then maximum likelihood seeks to maximise
\[
\mathbb{E}_{x\sim p_{\text{data}}}
\big[\log p_\theta(x)\big].
\]
Substituting the energy-based density gives
\[
\log p_\theta(x)
=
-E_\theta(x)-\log Z_\theta,
\]
and therefore
\[
\mathbb{E}_{x\sim p_{\text{data}}}
\big[\log p_\theta(x)\big]
=
-\mathbb{E}_{x\sim p_{\text{data}}}
\big[E_\theta(x)\big]
-
\log Z_\theta.
\]

The first term has an intuitive effect: maximising it lowers the energy of data
samples. The second term prevents the trivial solution of lowering energy
everywhere, because the partition function increases when too much of the space
is assigned low energy. Thus maximum likelihood balances two forces:
\[
\text{lower energy on data}
\qquad
\text{and}
\qquad
\text{control energy elsewhere}.
\]

This becomes clearer by differentiating the objective with respect to
\(\theta\). The gradient of the data term is
\[
-\nabla_\theta
\mathbb{E}_{x\sim p_{\text{data}}}
\big[E_\theta(x)\big]
=
-
\mathbb{E}_{x\sim p_{\text{data}}}
\big[\nabla_\theta E_\theta(x)\big].
\]
For the partition function term, we use
\[
Z_\theta
=
\int
\exp(-E_\theta(x))\,dx.
\]
Differentiating gives
\[
\begin{aligned}
\nabla_\theta \log Z_\theta
&=
\frac{1}{Z_\theta}
\nabla_\theta
\int
\exp(-E_\theta(x))\,dx \\[4pt]
&=
\frac{1}{Z_\theta}
\int
\exp(-E_\theta(x))
\big(-\nabla_\theta E_\theta(x)\big)\,dx \\[4pt]
&=
-
\mathbb{E}_{x\sim p_\theta}
\big[\nabla_\theta E_\theta(x)\big].
\end{aligned}
\]
Since the log-likelihood contains \(-\log Z_\theta\), its contribution to the
gradient is
\[
-\nabla_\theta \log Z_\theta
=
\mathbb{E}_{x\sim p_\theta}
\big[\nabla_\theta E_\theta(x)\big].
\]

Combining the two terms, the maximum-likelihood gradient becomes
\[
\boxed{
\nabla_\theta
\mathbb{E}_{x\sim p_{\text{data}}}
\big[\log p_\theta(x)\big]
=
-
\mathbb{E}_{x\sim p_{\text{data}}}
\big[\nabla_\theta E_\theta(x)\big]
+
\mathbb{E}_{x\sim p_\theta}
\big[\nabla_\theta E_\theta(x)\big].
}
\]

This expression is central to energy-based learning. It has two parts:

\[
\text{data term}
\quad
-
\mathbb{E}_{p_{\text{data}}}
\big[\nabla_\theta E_\theta(x)\big],
\qquad
\text{model term}
\quad
+
\mathbb{E}_{p_\theta}
\big[\nabla_\theta E_\theta(x)\big].
\]

The data term lowers the energy of observed samples. The model term raises the
energy of samples drawn from the current model distribution. This is often
described as a contrast between a \emph{positive phase} and a \emph{negative
phase}: the positive phase makes data more preferred, while the negative phase
pushes back against configurations that the model itself currently assigns too
much probability mass.

This contrastive gradient exposes the central practical difficulty of
energy-based models. The expectation over the data distribution is easy to
approximate using the training set. The expectation over the model distribution
\[
\mathbb{E}_{x\sim p_\theta}
\big[\nabla_\theta E_\theta(x)\big]
\]
is much harder, because drawing samples from \(p_\theta(x)\propto
\exp(-E_\theta(x))\) is itself a nontrivial problem.

This creates a coupling between learning and sampling. To improve the energy
function, we need samples from the current model. But to draw good samples, we
need an energy landscape that has already been shaped well. This circular
dependence is one of the main reasons energy-based models are powerful but
difficult to train.

The next section develops this connection more explicitly. Since an
energy-based model does not provide a direct sampler, generation is usually
formulated as movement on the energy landscape. This leads naturally to
energy-gradient dynamics, Langevin sampling, and the connection between
energy-based modelling and the score-based diffusion methods studied earlier.

\section{Energy Gradients, Score Fields, and Diffusion}

The previous section showed that energy-based models define probability
distributions indirectly, through an unnormalised energy landscape
\[
p_\theta(x)
\propto
\exp(-E_\theta(x)).
\]
This gives great modelling flexibility, but it also creates a practical
difficulty: the model does not provide a direct sampling mechanism. To generate
samples, we must move through the energy landscape itself.

This is where energy gradients become central. The scalar energy
\(E_\theta(x)\) tells us how plausible a configuration is, while its gradient
\[
\nabla_x E_\theta(x)
\]
tells us how the energy changes locally in data space. Moving in the direction
\[
-\nabla_x E_\theta(x)
\]
decreases energy and therefore moves a configuration toward regions that the
model considers more plausible. In this sense, the negative energy gradient acts
as a local driving force.

This section connects that idea to the score-based methods developed earlier in
the book. The main point is simple but important: for a distribution of the form
\(p(x)\propto \exp(-E(x))\), the score is exactly the negative energy gradient,
\[
\nabla_x\log p(x)
=
-\nabla_x E(x).
\]
Thus energy-based models and score-based models are closely linked. They differ
less in the local direction they use, and more in how that direction is
parameterised, learned, and used for generation.

\subsection{Sampling as Energy-Guided Dynamics}

An energy-based model assigns low energy to plausible configurations, but it
does not directly tell us how to draw independent samples from the corresponding
distribution. Unlike a normalising flow, there is no invertible map from a base
distribution to data space. Unlike an autoregressive model, there is no
left-to-right product of conditionals that can be sampled sequentially. Instead,
sampling is usually formulated as an iterative dynamical process.

The basic idea is to initialise a configuration \(x_0\) and repeatedly update it
so that it moves toward lower-energy regions. A purely deterministic update
would take the form
\[
x_{t+1}
=
x_t
-
\eta\nabla_x E_\theta(x_t),
\]
where \(\eta>0\) is a step size. This resembles gradient descent, but now the
descent is performed in the data space rather than in the parameter space. The
configuration \(x_t\) itself is being moved toward lower energy.

However, deterministic descent alone is not sufficient for sampling. It may
collapse into a local minimum of the energy landscape and fail to explore
multiple plausible regions. To obtain a sampling procedure, randomness is added:
\[
\boxed{
x_{t+1}
=
x_t
-
\eta \nabla_x E_\theta(x_t)
+
\sqrt{2\eta}\,\xi_t,
\qquad
\xi_t\sim\mathcal{N}(0,I).
}
\]
This is the discrete Langevin-style update associated with the energy function.

The deterministic term
\[
-\eta\nabla_x E_\theta(x_t)
\]
moves the current configuration toward lower energy. The noise term
\[
\sqrt{2\eta}\,\xi_t
\]
injects randomness, preventing the dynamics from simply collapsing to a single
minimum and allowing it to explore the broader energy landscape. Under suitable conditions and with an appropriate continuous-time interpretation,
this gradient-plus-noise dynamics can have the energy-based distribution as its
stationary distribution.

This is the same Langevin idea studied earlier in the score-based chapter. The
only change is how the drift field is written. There, the drift was expressed
using the score \(\nabla_x\log p(x)\). Here, it is expressed using the negative
energy gradient \(-\nabla_xE_\theta(x)\). The next subsection makes this
connection explicit.

\subsection{Energy Gradients and Scores}

Recall that the score of a density \(p(x)\) is
\[
s(x)
=
\nabla_x\log p(x).
\]
For an energy-based model,
\[
p_\theta(x)
=
\frac{1}{Z_\theta}\exp(-E_\theta(x)).
\]
Taking logarithms gives
\[
\log p_\theta(x)
=
-E_\theta(x)-\log Z_\theta.
\]
Now differentiate with respect to \(x\). Since the partition function
\(Z_\theta\) depends on \(\theta\) but not on \(x\), we have
\[
\nabla_x\log Z_\theta=0.
\]
Therefore,
\[
\boxed{
\nabla_x\log p_\theta(x)
=
-\nabla_xE_\theta(x).
}
\]

This identity is the bridge between energy-based modelling and score-based
modelling. The score field of an energy-based distribution is exactly the
negative gradient of its energy function. Thus the local direction used to move
toward higher probability is the same as the direction used to move toward lower
energy.

The difference lies in how the object is represented. In a classical
energy-based model, we parameterise a scalar function \(E_\theta(x)\), and the
driving vector field is obtained by differentiating it:
\[
x
\quad\longmapsto\quad
-\nabla_xE_\theta(x).
\]
In a score-based model, we often parameterise the vector field directly:
\[
s_\theta(x)
\approx
\nabla_x\log p(x).
\]
Thus, an energy model starts from a scalar landscape and derives a vector field,
whereas a score model often learns the vector field itself.

This distinction is useful. A scalar energy automatically gives a conservative
gradient field: the vector field comes from the gradient of one underlying
function. A directly learned score network is more flexible as a vector field,
but it may not always be explicitly represented as the gradient of a scalar
energy unless additional structure is imposed. For the purposes of generative
modelling, however, the conceptual link remains strong: both approaches learn
local directions that guide samples toward regions of higher data density.

\subsection{Score Matching as Energy-Gradient Learning}

The previous section showed that maximum likelihood for energy-based models
leads to a contrastive gradient involving samples from the model distribution.
This creates a practical difficulty because sampling from the model is itself
hard. Score matching~\cite{hyvarinen2005estimation} provides a different route: instead of matching normalised
probabilities, it matches local score fields.

In the score-based chapter, we derived score matching in detail using the Fisher
divergence and integration by parts. Here we only recall the consequence needed
for the energy-based view. The goal is to learn a model score
\[
s_\theta(x)
\approx
\nabla_x\log p_{\text{data}}(x).
\]
For an energy-based model,
\[
s_\theta(x)
=
\nabla_x\log p_\theta(x)
=
-\nabla_xE_\theta(x).
\]
Thus score matching can be interpreted as learning an energy landscape whose
negative gradient agrees with the data score.

The classical score-matching objective can be written, up to constants
independent of \(\theta\):
\[
J_{\mathrm{SM}}(\theta)
=
\mathbb{E}_{x\sim p_{\text{data}}}
\left[
\frac{1}{2}\|s_\theta(x)\|_2^2
+
\nabla_x\cdot s_\theta(x)
\right].
\]
Substituting
\[
s_\theta(x)=-\nabla_xE_\theta(x)
\]
gives the energy-form objective
\[
\boxed{
J_{\mathrm{SM}}(\theta)
=
\mathbb{E}_{x\sim p_{\text{data}}}
\left[
\frac{1}{2}
\|\nabla_xE_\theta(x)\|_2^2
-
\nabla_x^2E_\theta(x)
\right],
}
\]
where
\[
\nabla_x^2E_\theta(x)
=
\sum_{j=1}^d
\frac{\partial^2E_\theta(x)}{\partial x_j^2}
\]
is the Laplacian of the energy function.

The important point is not the detailed derivation, which was already developed
earlier, but the change in what is being matched. Maximum likelihood compares
normalised probability distributions and produces a model expectation term.
Score matching compares local gradients of log-density and avoids the partition
function, because
\[
\nabla_x\log Z_\theta=0.
\]
In energy terms, it learns the shape of the energy landscape near data samples
through derivatives, rather than requiring exact evaluation of the normalised
density.

This is why score matching is so natural from an energy-based perspective. It
does not ask for the probability of a data point. It asks for the local
direction in which probability should increase, or equivalently the direction in
which energy should decrease.

\subsection{Diffusion as Multi-Scale Energy-Gradient Learning}

The relationship between energy-based models and score-based diffusion can now
be stated clearly. An energy-based model defines a scalar landscape
\(E_\theta(x)\). If this landscape is associated with the unnormalised density
\[
p_\theta(x)\propto \exp(-E_\theta(x)),
\]
then its score is
\[
\nabla_x\log p_\theta(x)
=
-\nabla_xE_\theta(x).
\]
Thus the local direction used in Langevin sampling is the same object that
score-based models try to learn directly: a direction pointing toward higher
probability, or equivalently toward lower energy.

From this perspective, score-based diffusion models~\cite{song2019generative,song2021scorebased}
can be viewed as learning time-dependent energy-gradient-like directions across
a hierarchy of noisy distributions.
The difference is therefore not the local geometry itself, but how that geometry
is represented and learned. Energy-based models parameterise a scalar energy and
derive a vector field by differentiation. Score-based models usually
parameterise the vector field directly. Diffusion models go one step further by
learning a time-dependent score field
\[
s_\theta(x,t)
\approx
\nabla_x\log p_t(x),
\]
where \(p_t(x)\) is the distribution obtained after corrupting the data with
noise up to time \(t\).

From an energy-based viewpoint, each noisy distribution \(p_t\) can be associated
with an implicit energy landscape
\[
E_t(x)
=
-\log p_t(x)+C_t,
\]
where \(C_t\) is an arbitrary constant. Since only gradients matter, this
constant has no effect, and
\[
\nabla_x\log p_t(x)
=
-\nabla_xE_t(x).
\]
Therefore, a diffusion model can be interpreted as learning
\[
s_\theta(x,t)
\approx
-\nabla_xE_t(x),
\]
the negative gradients of a whole family of energy landscapes indexed by noise
level.

This viewpoint is useful because the forward noising process creates a
multi-scale hierarchy of distributions. At large noise levels, \(p_t\) is smooth
and close to a simple distribution, so the corresponding energy landscape is
broad and regular. At small noise levels, \(p_t\) is closer to the data
distribution, so the energy landscape contains finer structure. Diffusion
training therefore learns the local geometry of this sequence:
\[
\text{coarse energy landscapes}
\quad\longrightarrow\quad
\text{fine data-level energy landscape}.
\]

In this sense, diffusion models may be viewed as a practical and stable form of
multi-scale energy-gradient learning. Instead of learning one sharply structured
energy landscape for the data distribution and sampling from it directly, they
spread the problem across many smoothed distributions. The high-noise levels
provide global guidance, while the low-noise levels provide detailed local
refinement.

This also explains why diffusion models avoid some of the traditional
difficulties of energy-based learning. Classical EBM maximum likelihood requires
samples from the current model distribution in order to estimate the negative
phase. Diffusion models instead create noisy training pairs through a known
forward corruption process and learn denoising directions across noise levels.
The result remains deeply connected to energy gradients, but the training
procedure is far more stable.

Energy-based models therefore provide the broader landscape view of this part of
generative modelling. They describe distributions through scalar compatibility
functions, while score-based diffusion can be seen as one particularly effective modern way to learn and use the corresponding energy-gradient directions across a
multi-scale path from noise to data.

%------------------------------------------------
% Summary and References
%------------------------------------------------
\section*{Summary and References}
\addcontentsline{toc}{section}{Summary and References}

This chapter moved beyond likelihood-based generative modelling and introduced
models that learn through comparison, geometry, and unnormalised landscapes.
GANs showed how a generator can be trained through adversarial comparison rather
than likelihood maximisation. The optimal discriminator connects the original
GAN objective to the Jensen--Shannon divergence, while practical training often
uses the non-saturating loss to provide stronger generator gradients.

Wasserstein GANs refined this adversarial view by replacing overlap-based
classification with a geometric distance between distributions. Through the
dual formulation, the \(1\)-Wasserstein distance can be
estimated using a real-valued Lipschitz critic. This critic does not output
probabilities; its score differences reflect how probability mass must move
through data space.

Energy-Based Models then provided a broader landscape view. An EBM assigns each
configuration a scalar energy \(E_\theta(x)\), with low energy corresponding to
plausible data and high energy to implausible configurations. Through $p_\theta(x)\propto \exp(-E_\theta(x)),$
the energy defines an unnormalised density. Learning and sampling are therefore
organised around energy differences and energy gradients rather than tractable
normalised likelihoods.

Finally, we connected EBMs to score-based diffusion through the identity $\nabla_x\log p_\theta(x)
=
-\nabla_xE_\theta(x).$
This shows that energy-based and score-based models share the same local
geometry: both use directions that guide samples toward higher-density,
lower-energy regions. Diffusion models extend this idea across a hierarchy of
noisy distributions, learning energy-gradient-like directions from coarse noise
levels to fine data-level structure.

\bigskip

Generative Adversarial Networks were introduced by Goodfellow et
al.~\cite{goodfellow2014generative}, who formulated generation as a minimax game
between a generator and a discriminator and connected the optimal discriminator
to the Jensen--Shannon divergence. Wasserstein GANs were introduced by Arjovsky
et al.~\cite{arjovsky2017wasserstein}, replacing classification-based
comparison with a transport-based distance. The gradient-penalty formulation was
developed by Gulrajani et al.~\cite{gulrajani2017improved}.

Energy-based modelling has a long history in machine learning and statistical
physics. LeCun et al.~\cite{lecun2006tutorial} provide a broad tutorial view of
learning as shaping energy landscapes. Classical examples include Hopfield
networks~\cite{hopfield1982neural} and Boltzmann
machines~\cite{hinton1985learning}. Score matching was introduced by
Hyv\"arinen~\cite{hyvarinen2005estimation} as a way to estimate unnormalised
statistical models without evaluating the partition function, and
Vincent~\cite{vincent2011connection} connected score matching to denoising.
The score-based diffusion chapter discussed Song and
Ermon~\cite{song2019generative} and Song et al.~\cite{song2021scorebased} in
detail; they are recalled here as they connect energy-gradient ideas to
multi-scale score learning and diffusion.

%% file: figures/ch08/chapter8_beyond_likelihoods_at_a_glance.tex
\begin{center}
\begin{tikzpicture}[
  >=Latex,
  every node/.style={font=\small},
  box/.style={draw,rounded corners=2pt,align=center,inner sep=6pt,
             text width=3.15cm,minimum height=1.02cm},
  flow/.style={->,thick}
]
\node[font=\normalsize\bfseries] at (6.0,2.65)
  {Beyond likelihoods at a glance};

\node[box] (gan) at (1.7,0.85) {
  \textbf{GAN comparison}\\[2pt]
  generator samples\\
  $\leftrightarrow$ discriminator\\
  feedback
};
\node[box] (wgan) at (6.0,0.85) {
  \textbf{WGAN geometry}\\[2pt]
  transport distance\\
  instead of\\
  classification
};
\node[box] (ebm) at (10.3,0.85) {
  \textbf{EBM landscapes}\\[2pt]
  energy assigns\\
  relative\\
  plausibility
};
\draw[flow] (gan) -- (wgan);
\draw[flow] (wgan) -- (ebm);

\node[box] (ratio) at (1.7,-2.15) {
  \textbf{Optimal}\\[-1pt]
  \textbf{discriminator}\\[2pt]
  local density-ratio\\
  information
};
\node[box] (energy) at (6.0,-2.15) {
  \textbf{Unnormalised density}\\[2pt]
  $p_\theta(x)\propto e^{-E_\theta(x)}$\\
  global\\
  partition function
};
\node[box] (score) at (10.3,-2.15) {
  \textbf{Back to scores}\\[2pt]
  $\nabla_x\log p_\theta(x)$\\
  $=-\nabla_xE_\theta(x)$
};
\draw[flow] (gan) -- (ratio);
\draw[flow] (ebm) -- (energy);
\draw[flow] (energy) -- (score);
\end{tikzpicture}
\end{center}

%% file: figures/ch08/fig_ch08_01_optimal_discriminator.tex
\begin{figure}[H]
\centering
\includegraphics[width=0.98\textwidth]{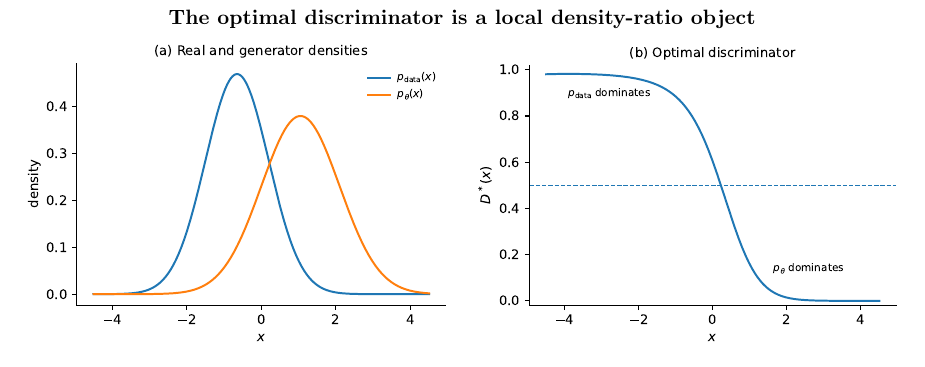}
\caption{\textbf{The optimal discriminator reflects local density balance.}
For a fixed generator,
$D^*(x)=p_{\mathrm{data}}(x)/(p_{\mathrm{data}}(x)+p_\theta(x))$.
It approaches one where the data density dominates, zero where the generator
density dominates, and equals $1/2$ where the two densities are equal.}
\label{fig:gan-optimal-discriminator-geometry}
\end{figure}

%% file: figures/ch08/fig_ch08_02_js_vs_wasserstein.tex
\begin{figure}[H]
\centering
\begin{minipage}[t]{0.48\textwidth}
  \vspace{0pt}\centering
  \includegraphics[width=\linewidth]{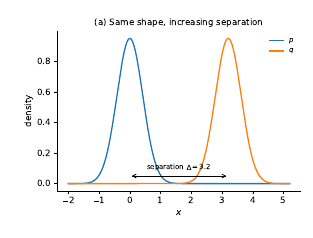}
\end{minipage}\hfill
\begin{minipage}[t]{0.48\textwidth}
  \vspace{0pt}\centering
  \includegraphics[width=\linewidth]{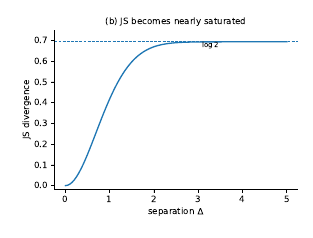}
\end{minipage}

\vspace{0.8em}

\begin{minipage}[t]{0.48\textwidth}
  \vspace{0pt}\centering
  \includegraphics[width=\linewidth]{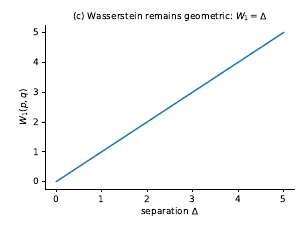}
\end{minipage}\hfill
\begin{minipage}[t]{0.48\textwidth}
  \vspace{0pt}
  \caption{\textbf{JS can saturate while Wasserstein remains geometric.}
  The panels compare equal-variance Gaussian distributions as their means move
  apart. As overlap disappears, the JS divergence approaches its upper bound
  $\log 2$ and becomes insensitive to further separation. For these translated
  distributions, the $1$-Wasserstein distance is $W_1=\Delta$, so it continues
  to measure how far probability mass must move.}
  \label{fig:gan-js-wasserstein-geometry}
\end{minipage}
\end{figure}

%% file: figures/ch08/fig_ch08_03_energy_to_density.tex
\begin{figure}[H]
\centering
\begin{minipage}[t]{0.48\textwidth}
  \vspace{0pt}\centering
  \includegraphics[width=\linewidth]{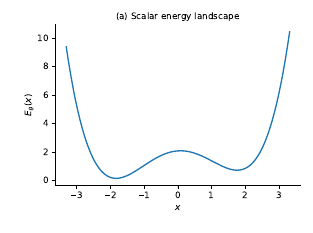}
\end{minipage}\hfill
\begin{minipage}[t]{0.48\textwidth}
  \vspace{0pt}\centering
  \includegraphics[width=\linewidth]{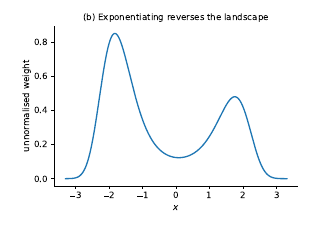}
\end{minipage}

\vspace{0.8em}

\begin{minipage}[t]{0.48\textwidth}
  \vspace{0pt}\centering
  \includegraphics[width=\linewidth]{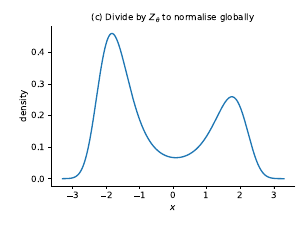}
\end{minipage}\hfill
\begin{minipage}[t]{0.48\textwidth}
  \vspace{0pt}
  \caption{\textbf{From energy to normalised density.}
  Low-energy regions receive large unnormalised weight $\exp(-E_\theta(x))$,
  and division by $Z_\theta$ turns this into a density without changing its
  shape. Adding a constant to all energies rescales the weights and $Z_\theta$
  by the same factor, leaving the normalised density unchanged.}
  \label{fig:ebm-energy-to-density}
\end{minipage}
\end{figure}

%% file: chapters/cht_closing_remarks.tex
\chapter*{Closing Remarks}
\addcontentsline{toc}{chapter}{Closing Remarks}

This book has followed a deliberately selective path through the mathematical
foundations of generative modelling. We began with linear representation:
matrices as transformations, PCA as variance-preserving projection, and
autoencoders as reconstruction-based compression. Probabilistic PCA then turned
this deterministic picture into a latent-variable generative model, introducing
the prior, likelihood, marginal likelihood, and posterior as central modelling
objects.

The Variational Autoencoder extended this latent-variable view into the
nonlinear setting, where exact marginal likelihoods and posteriors are no
longer available. This led naturally to the Evidence Lower Bound, variational
inference, approximate posteriors, and the reparameterisation trick. Diffusion
models then shifted the generative story from a single latent variable to a
whole trajectory of noisy latent states, turning generation into the learned
reverse of a gradual noising process.

The continuous-time and score-based chapters developed a second view of the
same broad problem. Instead of describing generation only as a discrete chain,
we studied motion, density evolution, Brownian noise, the Liouville equation,
the Fokker--Planck equation, Langevin dynamics, score matching, and reverse-time
diffusion. This showed that generative modelling can also be understood as
learning directions of motion through probability space.

The later chapters examined two complementary routes. Normalising flows and
autoregressive models showed how exact likelihoods can be preserved by design:
through invertible transformations with tractable Jacobians, or through ordered
conditional factorisation. GANs, Wasserstein GANs, and Energy-Based Models then
showed how generative learning can move beyond tractable likelihoods, relying
instead on comparison, geometric discrepancy, or scalar energy landscapes.

Taken together, these chapters present generative AI not as a collection of
unrelated architectures, but as a set of recurring mathematical ideas:
representation, transformation, marginalisation, approximation, noising and
denoising, density evolution, score fields, likelihood construction,
distributional comparison, and energy shaping. These ideas continue to reappear
in newer generative modelling frameworks.

For example, flow matching~\cite{lipman2023flow} and rectified
flows~\cite{liu2022rectified} can be viewed as further developments of the
continuous-time transport perspective: they learn vector fields that move
probability mass from a simple distribution toward the data distribution.
Consistency models~\cite{song2023consistency} can be viewed as another response
to the sampling-efficiency challenge in diffusion models, aiming to retain the
quality of iterative denoising while reducing the number of sampling steps.

The purpose of the book has therefore been to build a compact mathematical base
from which such later developments can be approached more confidently. Once the
reader understands latent variables, variational bounds, diffusion paths,
density evolution, score fields, exact likelihood models, adversarial
comparison, and energy landscapes, many modern extensions become easier to
place within a common conceptual map.

%% file: chapters/cht_appendix1.tex
\chapter{Gaussian Algebra and Completing the Square}
\label{app:completing-square}

Chapter 2 derived the PPCA marginal \(p(\mathbf{x})\) using a direct
linear-Gaussian argument.
That approach is the quickest and most transparent for the main flow of the
chapter.
There is, however, another derivation worth examining, because the same
algebraic structure reappears later in other Gaussian latent-variable models:
the method of \emph{completing the square}.

This appendix gathers a number of derivations that rely on explicit Gaussian
algebra, with particular emphasis on making precise the completing-the-square
identities used implicitly in the main text.
In the PPCA setting, the value of this approach is that it yields both key
Gaussian quantities in a single derivation:
\begin{itemize}
    \item the posterior \(p(\mathbf{z}\mid \mathbf{x})\),
    \item and, after integrating out \(\mathbf{z}\), the marginal \(p(\mathbf{x})\).
\end{itemize}

\paragraph{Completing the square.}
A Gaussian density is an exponential of a quadratic form.
So whenever the log-density of a variable can be written as a quadratic function
of that variable, we can try to rewrite it into the standard Gaussian shape.
Completing the square is precisely the algebraic tool that performs this
rewriting: it reorganises a quadratic expression into the form
\[
-\frac12 (\text{variable} - \text{mean})^\top
(\text{precision})
(\text{variable} - \text{mean})
+\text{constant},
\]
from which the Gaussian mean and covariance can be read off directly.

For resue, it is helpful to state the matrix form of the trick explicitly.
If \(A\) is symmetric positive definite, then
\[
-\frac12 \mathbf{x}^\top A \mathbf{x} + \mathbf{b}^\top \mathbf{x}
=
-\frac12 (\mathbf{x}-A^{-1}\mathbf{b})^\top A (\mathbf{x}-A^{-1}\mathbf{b})
+\frac12 \mathbf{b}^\top A^{-1}\mathbf{b}.
\]
This works because, if we expand the right-hand side, we obtain
\[
\begin{aligned}
&-\frac12 (\mathbf{x}-A^{-1}\mathbf{b})^\top A (\mathbf{x}-A^{-1}\mathbf{b})
+\frac12 \mathbf{b}^\top A^{-1}\mathbf{b} \\[4pt]
&\qquad=
-\frac12 \mathbf{x}^\top A \mathbf{x}
+\mathbf{b}^\top \mathbf{x}
-\frac12 \mathbf{b}^\top A^{-1}\mathbf{b}
+\frac12 \mathbf{b}^\top A^{-1}\mathbf{b} \\[4pt]
&\qquad=
-\frac12 \mathbf{x}^\top A \mathbf{x}+\mathbf{b}^\top \mathbf{x}.
\end{aligned}
\]
So the two expressions are exactly the same; the second is simply the first
rewritten in a Gaussian-centred form.
The vector \(A^{-1}\mathbf{b}\) is the centre of the quadratic, and \(A\) is the
precision matrix, whose inverse gives the covariance.

\paragraph{Step 1: Start from the joint density.}
Recall the PPCA model:
\[
p(\mathbf{z})=\mathcal{N}(\mathbf{z}\mid \mathbf{0},\mathbf{I}),
\qquad
p(\mathbf{x}\mid \mathbf{z})
=
\mathcal{N}(\mathbf{x}\mid \mathbf{Wz}+\boldsymbol{\mu},\alpha^2\mathbf{I}).
\]
Hence
\[
p(\mathbf{x},\mathbf{z})
=
p(\mathbf{x}\mid \mathbf{z})\,p(\mathbf{z}).
\]

Since Gaussian densities are exponentials of quadratic forms, it is convenient
to work with the log-joint:
\[
\log p(\mathbf{x},\mathbf{z})
=
-\frac{1}{2\alpha^2}\|\mathbf{x}-\boldsymbol{\mu}-\mathbf{Wz}\|^2
-\frac12 \|\mathbf{z}\|^2
+\text{const}.
\]

\paragraph{Step 2: Expand the quadratic in \(\mathbf{z}\).}
Expanding the first norm gives
\[
\|\mathbf{x}-\boldsymbol{\mu}-\mathbf{Wz}\|^2
=
(\mathbf{x}-\boldsymbol{\mu})^\top(\mathbf{x}-\boldsymbol{\mu})
-2\,\mathbf{z}^\top \mathbf{W}^\top(\mathbf{x}-\boldsymbol{\mu})
+\mathbf{z}^\top \mathbf{W}^\top\mathbf{W}\mathbf{z}.
\]
Substituting back,
\[
\log p(\mathbf{x},\mathbf{z})
=
-\frac{1}{2\alpha^2}(\mathbf{x}-\boldsymbol{\mu})^\top(\mathbf{x}-\boldsymbol{\mu})
+\frac{1}{\alpha^2}\mathbf{z}^\top\mathbf{W}^\top(\mathbf{x}-\boldsymbol{\mu})
-\frac12 \mathbf{z}^\top
\Bigl(\frac{1}{\alpha^2}\mathbf{W}^\top\mathbf{W}+\mathbf{I}\Bigr)\mathbf{z}
+\text{const}.
\]

Now regard \(\mathbf{x}\) as fixed.
Then the log-joint is a quadratic function of \(\mathbf{z}\), so it can be
rewritten as a Gaussian in \(\mathbf{z}\) by completing the square.

\paragraph{Step 3: Complete the square.}
Define
\[
\mathbf{M}
=
\mathbf{W}^\top\mathbf{W}+\alpha^2\mathbf{I},
\qquad
\mathbf{m}
=
\mathbf{M}^{-1}\mathbf{W}^\top(\mathbf{x}-\boldsymbol{\mu}).
\]
Since
\[
\frac{1}{\alpha^2}\mathbf{W}^\top\mathbf{W}+\mathbf{I}
=
\frac{1}{\alpha^2}\mathbf{M},
\]
the \(\mathbf{z}\)-dependent part of the log-joint becomes
\[
-\frac{1}{2\alpha^2}\mathbf{z}^\top\mathbf{M}\mathbf{z}
+
\frac{1}{\alpha^2}\mathbf{z}^\top\mathbf{W}^\top(\mathbf{x}-\boldsymbol{\mu}).
\]
This is now exactly in the form of the general identity above, with
\[
A=\frac{1}{\alpha^2}\mathbf{M},
\qquad
\mathbf{b}=\frac{1}{\alpha^2}\mathbf{W}^\top(\mathbf{x}-\boldsymbol{\mu}).
\]
Applying the identity gives
\[
-\frac{1}{2\alpha^2}\mathbf{z}^\top\mathbf{M}\mathbf{z}
+
\frac{1}{\alpha^2}\mathbf{z}^\top\mathbf{W}^\top(\mathbf{x}-\boldsymbol{\mu})
=
-\frac{1}{2\alpha^2}(\mathbf{z}-\mathbf{m})^\top\mathbf{M}(\mathbf{z}-\mathbf{m})
+\frac{1}{2\alpha^2}\mathbf{m}^\top\mathbf{M}\mathbf{m}.
\]
Substituting this back gives
\[
\log p(\mathbf{x},\mathbf{z})
=
-\frac{1}{2\alpha^2}(\mathbf{z}-\mathbf{m})^\top\mathbf{M}(\mathbf{z}-\mathbf{m})
-\frac{1}{2\alpha^2}(\mathbf{x}-\boldsymbol{\mu})^\top(\mathbf{x}-\boldsymbol{\mu})
+\frac{1}{2\alpha^2}\mathbf{m}^\top\mathbf{M}\mathbf{m}
+\text{const}.
\]

At this point, the Gaussian structure in \(\mathbf{z}\) is explicit.

\paragraph{Why this identifies a Gaussian posterior.}
Once \(\mathbf{x}\) is fixed, any terms independent of \(\mathbf{z}\) are part
of the normalising constant.
So the only \(\mathbf{z}\)-dependent term that matters is
\[
-\frac{1}{2\alpha^2}(\mathbf{z}-\mathbf{m})^\top\mathbf{M}(\mathbf{z}-\mathbf{m}),
\]
which is exactly the log-kernel of a Gaussian density in \(\mathbf{z}\).
Because \(\mathbf{M}\) is positive definite, this quadratic is concave in
\(\mathbf{z}\), so it indeed defines a proper Gaussian distribution.

\paragraph{Step 4: Read off the posterior \(p(\mathbf{z}\mid \mathbf{x})\).}
Therefore,
\[
p(\mathbf{z}\mid \mathbf{x})
=
\mathcal{N}\!\bigl(
\mathbf{z}\mid
\mathbf{m},
\alpha^2\mathbf{M}^{-1}
\bigr),
\]
that is,
\[
p(\mathbf{z}\mid \mathbf{x})
=
\mathcal{N}\!\bigl(
\mathbf{z}\mid
\mathbf{M}^{-1}\mathbf{W}^\top(\mathbf{x}-\boldsymbol{\mu}),
\alpha^2\mathbf{M}^{-1}
\bigr),
\qquad
\mathbf{M}=\mathbf{W}^\top\mathbf{W}+\alpha^2\mathbf{I}.
\]

So completing the square gives the posterior directly: its mean is the latent
estimate \(\mathbf{m}\), and its covariance is \(\alpha^2\mathbf{M}^{-1}\).

\paragraph{Step 5: Integrate out \(\mathbf{z}\) to obtain \(p(\mathbf{x})\).}
The marginal distribution is
\[
p(\mathbf{x})
=
\int p(\mathbf{x},\mathbf{z})\,d\mathbf{z}.
\]
From the completed-square form, the \(\mathbf{z}\)-dependent part is a Gaussian
density in \(\mathbf{z}\) centred at \(\mathbf{m}\).
Integrating over \(\mathbf{z}\) therefore removes that part, contributing only a
normalisation constant.
What remains is a quadratic function of \((\mathbf{x}-\boldsymbol{\mu})\), so
the marginal must also be Gaussian.

Carrying out the remaining algebra gives
\[
p(\mathbf{x})
=
\mathcal{N}\!\bigl(
\mathbf{x}\mid
\boldsymbol{\mu},
\mathbf{W}\mathbf{W}^\top+\alpha^2\mathbf{I}
\bigr).
\]

\bigskip
Thus, by completing the square once, we obtain both
\[
p(\mathbf{z}\mid \mathbf{x})
=
\mathcal{N}\!\bigl(
\mathbf{z}\mid
\mathbf{M}^{-1}\mathbf{W}^\top(\mathbf{x}-\boldsymbol{\mu}),
\alpha^2\mathbf{M}^{-1}
\bigr)
\]
and
\[
p(\mathbf{x})
=
\mathcal{N}\!\bigl(
\mathbf{x}\mid
\boldsymbol{\mu},
\mathbf{W}\mathbf{W}^\top+\alpha^2\mathbf{I}
\bigr).
\]

This derivation is not needed for the main conceptual flow of PPCA, but it is
useful because the same quadratic-completion trick will reappear later whenever
Gaussian latent-variable models are manipulated explicitly.

%% file: chapters/cht_appendix2.tex
\chapter{Diffusion Reverse-Posterior Derivations}
\label{app:diffusion-posterior}

In the Section 4.4, we stated that the true reverse posterior
\[
q(\mathbf{x}_{t-1}\mid \mathbf{x}_t,\mathbf{x}_0)
\]
is Gaussian, and gave its mean and variance in closed form.
The derivation, which we show here, follows from the same basic algebraic idea we used earlier in PPCA:
once all relevant terms are Gaussian, we can collect the quadratic terms in the
variable of interest and complete the square.

\subsection*{Step 1: Start from Bayes' rule}

From the forward process, we have
\[
q(\mathbf{x}_{t-1}\mid \mathbf{x}_t,\mathbf{x}_0)
=
\frac{
q(\mathbf{x}_t\mid \mathbf{x}_{t-1},\mathbf{x}_0)\,
q(\mathbf{x}_{t-1}\mid \mathbf{x}_0)
}{
q(\mathbf{x}_t\mid \mathbf{x}_0)
}.
\]
Since the denominator does not depend on \(\mathbf{x}_{t-1}\), it acts only as a
normalising constant when viewed as a function of \(\mathbf{x}_{t-1}\).
So, up to proportionality,
\[
q(\mathbf{x}_{t-1}\mid \mathbf{x}_t,\mathbf{x}_0)
\propto
q(\mathbf{x}_t\mid \mathbf{x}_{t-1},\mathbf{x}_0)\,
q(\mathbf{x}_{t-1}\mid \mathbf{x}_0).
\]

Because the forward process is Markovian,
\[
q(\mathbf{x}_t\mid \mathbf{x}_{t-1},\mathbf{x}_0)
=
q(\mathbf{x}_t\mid \mathbf{x}_{t-1}),
\]
so the two Gaussian factors are
\[
q(\mathbf{x}_t\mid \mathbf{x}_{t-1})
=
\mathcal{N}\!\bigl(
\mathbf{x}_t\mid
\sqrt{\alpha_t}\,\mathbf{x}_{t-1},
(1-\alpha_t)\mathbf{I}
\bigr),
\]
and
\[
q(\mathbf{x}_{t-1}\mid \mathbf{x}_0)
=
\mathcal{N}\!\bigl(
\mathbf{x}_{t-1}\mid
\sqrt{\bar{\alpha}_{t-1}}\,\mathbf{x}_0,
(1-\bar{\alpha}_{t-1})\mathbf{I}
\bigr).
\]

\subsection*{Step 2: Write the product in exponential form}

Ignoring normalising constants that do not depend on \(\mathbf{x}_{t-1}\), we have
\[
\begin{aligned}
q(\mathbf{x}_{t-1}\mid \mathbf{x}_t,\mathbf{x}_0)
&\propto
\exp\!\left(
-\frac{1}{2(1-\alpha_t)}
\left\|
\mathbf{x}_t-\sqrt{\alpha_t}\,\mathbf{x}_{t-1}
\right\|^2
\right)
\\[4pt]
&\qquad\times
\exp\!\left(
-\frac{1}{2(1-\bar{\alpha}_{t-1})}
\left\|
\mathbf{x}_{t-1}-\sqrt{\bar{\alpha}_{t-1}}\,\mathbf{x}_0
\right\|^2
\right).
\end{aligned}
\]
Combining the exponentials gives
\[
\begin{aligned}
q(\mathbf{x}_{t-1}\mid \mathbf{x}_t,\mathbf{x}_0)
\propto
\exp\!\Bigg(
&-\frac{1}{2(1-\alpha_t)}
\left\|
\mathbf{x}_t-\sqrt{\alpha_t}\,\mathbf{x}_{t-1}
\right\|^2
\\
&-\frac{1}{2(1-\bar{\alpha}_{t-1})}
\left\|
\mathbf{x}_{t-1}-\sqrt{\bar{\alpha}_{t-1}}\,\mathbf{x}_0
\right\|^2
\Bigg).
\end{aligned}
\]

\subsection*{Step 3: Expand the quadratic terms in \texorpdfstring{$\mathbf{x}_{t-1}$}{x\_\{t-1\}}}

We now expand the two squared norms, keeping only terms involving
\(\mathbf{x}_{t-1}\).

For the first term,
\[
\begin{aligned}
\left\|
\mathbf{x}_t-\sqrt{\alpha_t}\,\mathbf{x}_{t-1}
\right\|^2
&=
\mathbf{x}_t^\top\mathbf{x}_t
-2\sqrt{\alpha_t}\,\mathbf{x}_t^\top \mathbf{x}_{t-1}
+\alpha_t\,\mathbf{x}_{t-1}^\top \mathbf{x}_{t-1}.
\end{aligned}
\]
For the second term,
\[
\begin{aligned}
\left\|
\mathbf{x}_{t-1}-\sqrt{\bar{\alpha}_{t-1}}\,\mathbf{x}_0
\right\|^2
&=
\mathbf{x}_{t-1}^\top \mathbf{x}_{t-1}
-2\sqrt{\bar{\alpha}_{t-1}}\,\mathbf{x}_{t-1}^\top \mathbf{x}_0
+\bar{\alpha}_{t-1}\,\mathbf{x}_0^\top\mathbf{x}_0.
\end{aligned}
\]

Substituting these into the exponent and dropping terms independent of
\(\mathbf{x}_{t-1}\), we obtain
\[
\begin{aligned}
q(\mathbf{x}_{t-1}\mid \mathbf{x}_t,\mathbf{x}_0)
\propto
\exp\!\Bigg(
&-\frac{1}{2}
\left[
\frac{\alpha_t}{1-\alpha_t}
+
\frac{1}{1-\bar{\alpha}_{t-1}}
\right]
\mathbf{x}_{t-1}^\top \mathbf{x}_{t-1}
\\
&+
\left[
\frac{\sqrt{\alpha_t}}{1-\alpha_t}\,\mathbf{x}_t^\top
+
\frac{\sqrt{\bar{\alpha}_{t-1}}}{1-\bar{\alpha}_{t-1}}\,\mathbf{x}_0^\top
\right]
\mathbf{x}_{t-1}
\Bigg).
\end{aligned}
\]

It is now in the standard Gaussian quadratic form
\[
-\frac12 A\,\mathbf{x}_{t-1}^\top \mathbf{x}_{t-1}
+
\mathbf{b}^\top \mathbf{x}_{t-1},
\]
where
\[
A
=
\frac{\alpha_t}{1-\alpha_t}
+
\frac{1}{1-\bar{\alpha}_{t-1}},
\qquad
\mathbf{b}
=
\frac{\sqrt{\alpha_t}}{1-\alpha_t}\,\mathbf{x}_t
+
\frac{\sqrt{\bar{\alpha}_{t-1}}}{1-\bar{\alpha}_{t-1}}\,\mathbf{x}_0.
\]

\subsection*{Step 4: Complete the square}

Using the standard identity
\[
-\frac12 A\,\mathbf{x}^\top\mathbf{x}
+
\mathbf{b}^\top\mathbf{x}
=
-\frac12 A
\left\|
\mathbf{x}-A^{-1}\mathbf{b}
\right\|^2
+\text{constant},
\]
we see that the posterior is Gaussian with covariance coefficient
\[
\tilde{\beta}_t = A^{-1},
\]
and mean
\[
\tilde{\boldsymbol{\mu}}_t = A^{-1}\mathbf{b}.
\]

So we first simplify \(A\):
\[
\begin{aligned}
A
&=
\frac{\alpha_t}{1-\alpha_t}
+
\frac{1}{1-\bar{\alpha}_{t-1}}
\\[4pt]
&=
\frac{
\alpha_t(1-\bar{\alpha}_{t-1}) + (1-\alpha_t)
}{
(1-\alpha_t)(1-\bar{\alpha}_{t-1})
}.
\end{aligned}
\]
Since \(1-\alpha_t=\beta_t\) and \(\bar{\alpha}_t=\alpha_t\bar{\alpha}_{t-1}\),
the numerator becomes
\[
\alpha_t - \alpha_t\bar{\alpha}_{t-1} + 1 - \alpha_t
=
1-\bar{\alpha}_t.
\]
Hence
\[
A
=
\frac{
1-\bar{\alpha}_t
}{
\beta_t(1-\bar{\alpha}_{t-1})
},
\]
so
\[
\boxed{
\tilde{\beta}_t
=
A^{-1}
=
\frac{1-\bar{\alpha}_{t-1}}{1-\bar{\alpha}_t}\,\beta_t.
}
\]

Now compute the mean:
\[
\tilde{\boldsymbol{\mu}}_t
=
A^{-1}\mathbf{b}
=
\tilde{\beta}_t
\left(
\frac{\sqrt{\alpha_t}}{1-\alpha_t}\,\mathbf{x}_t
+
\frac{\sqrt{\bar{\alpha}_{t-1}}}{1-\bar{\alpha}_{t-1}}\,\mathbf{x}_0
\right).
\]
Substituting
\(
\tilde{\beta}_t
=
\frac{1-\bar{\alpha}_{t-1}}{1-\bar{\alpha}_t}\beta_t
\)
and \(1-\alpha_t=\beta_t\), we get
\[
\begin{aligned}
\tilde{\boldsymbol{\mu}}_t
&=
\frac{1-\bar{\alpha}_{t-1}}{1-\bar{\alpha}_t}\beta_t
\left(
\frac{\sqrt{\alpha_t}}{\beta_t}\,\mathbf{x}_t
+
\frac{\sqrt{\bar{\alpha}_{t-1}}}{1-\bar{\alpha}_{t-1}}\,\mathbf{x}_0
\right)
\\[6pt]
&=
\frac{\sqrt{\alpha_t}(1-\bar{\alpha}_{t-1})}{1-\bar{\alpha}_t}\,\mathbf{x}_t
+
\frac{\sqrt{\bar{\alpha}_{t-1}}\beta_t}{1-\bar{\alpha}_t}\,\mathbf{x}_0.
\end{aligned}
\]
So
\[
\boxed{
\tilde{\boldsymbol{\mu}}_t(\mathbf{x}_t,\mathbf{x}_0)
=
\frac{
\sqrt{\bar{\alpha}_{t-1}}\,\beta_t
}{
1-\bar{\alpha}_t
}\,\mathbf{x}_0
+
\frac{
\sqrt{\alpha_t}(1-\bar{\alpha}_{t-1})
}{
1-\bar{\alpha}_t
}\,\mathbf{x}_t.
}
\]

\subsection*{Step 5: Final result}

Putting everything together,
\[
\boxed{
q(\mathbf{x}_{t-1}\mid \mathbf{x}_t,\mathbf{x}_0)
=
\mathcal{N}\!\Bigl(
\mathbf{x}_{t-1}\mid
\tilde{\boldsymbol{\mu}}_t(\mathbf{x}_t,\mathbf{x}_0),
\tilde{\beta}_t\mathbf{I}
\Bigr),
}
\]
with
\[
\tilde{\boldsymbol{\mu}}_t(\mathbf{x}_t,\mathbf{x}_0)
=
\frac{
\sqrt{\bar{\alpha}_{t-1}}\,\beta_t
}{
1-\bar{\alpha}_t
}\,\mathbf{x}_0
+
\frac{
\sqrt{\alpha_t}(1-\bar{\alpha}_{t-1})
}{
1-\bar{\alpha}_t
}\,\mathbf{x}_t,
\]
and
\[
\tilde{\beta}_t
=
\frac{
1-\bar{\alpha}_{t-1}
}{
1-\bar{\alpha}_t
}\,\beta_t.
\]

This is the exact local reverse target that the learned denoising transition
\(p_\theta(\mathbf{x}_{t-1}\mid \mathbf{x}_t)\) is asked to match inside the
ELBO.